\documentclass[12pt,oneside, a4paper]{book}
\usepackage{etex}
\usepackage[table]{xcolor}
\usepackage[pdftex]{graphicx}
\usepackage{rotating}
\usepackage{epsfig}
\usepackage{epstopdf}
\usepackage{mathtools}
\usepackage{listings}
\usepackage{color}
\usepackage{svg}

\usepackage[T1]{fontenc}
\usepackage[utf8]{inputenc}
\usepackage{cmap}
\usepackage[english, croatian]{babel}
\usepackage{ae}
\usepackage[unicode]{hyperref}
\usepackage{mathptmx}
\usepackage{amscd}
\usepackage{amssymb}
\usepackage{xfrac}
\usepackage{amsmath}
\usepackage{amsfonts}
\usepackage{amsthm}

\usepackage{wrapfig}

\usepackage[left=2.5cm,right=2.5cm,top=2.5cm,bottom=2.5cm]{geometry}
\usepackage{setspace} 
\usepackage{fancyhdr} 
\usepackage{xspace}

\usepackage{ragged2e}

\usepackage[most]{tcolorbox}
\usepackage{tikz}
\usetikzlibrary{arrows,shapes,positioning,shadows,trees,mindmap}
\usepackage[edges]{forest}
\usetikzlibrary{arrows.meta}
\colorlet{linecol}{black!75}
\usetikzlibrary{backgrounds}
\usetikzlibrary{arrows,shapes}
\usetikzlibrary{tikzmark}
\usetikzlibrary{calc}
\usepackage{xkcdcolors} 

\usepackage{hhline}
\usepackage{caption}
\usepackage{subcaption}
\usepackage{cleveref}
\usepackage{enumerate}
\usepackage{delarray}
\usepackage{array}  
\usepackage{tabularx} 
\usepackage{multirow}  
\usepackage{booktabs}
\usepackage{pifont} 
\usepackage{wasysym}
\usepackage{subeqnarray}
\usepackage{pdflscape} 
\usepackage{enumitem} 
\usepackage{xspace}
\usepackage{adjustbox}
\setlist{nolistsep}   

\renewcommand{\arraystretch}{1.5} 

\newcommand{\jachess}{\textsc{JacHess}\xspace}

\DeclareMathOperator{\E}{\mathbb{E}}
\DeclareMathOperator{\trace}{Tr}

\DeclareMathOperator*{\argmax}{arg\,max}
\DeclareMathOperator*{\argmin}{arg\,min}
\DeclareMathOperator*{\softmax}{softmax}

\usepackage[square, numbers, sort]{natbib} 

\let\savenumberline\numberline
\def\numberline#1{\savenumberline{#1.}}

\makeatletter
\renewcommand*\l@chapter[2]{%
  \ifnum \c@tocdepth >\m@ne
  \addpenalty{-\@highpenalty}%
  \vskip 1.0em \@plus\p@
  \setlength\@tempdima{1.5em}%
  \begingroup
  \parindent \z@ \rightskip \@pnumwidth
  \parfillskip -\@pnumwidth
  \leavevmode \bfseries
  \advance\leftskip\@tempdima
  \hskip -\leftskip
  #1\nobreak\normalfont\leaders\hbox{$\m@th
    \mkern \@dotsep mu\hbox{.}\mkern \@dotsep
    mu$}\hfill\nobreak\hb@xt@\@pnumwidth{\hss #2}\par
  \penalty\@highpenalty
  \endgroup
  \fi}
\makeatother

\makeatletter
\renewcommand*\env@matrix[1][\arraystretch]{%
  \edef\arraystretch{#1}%
  \hskip -\arraycolsep
  \let\@ifnextchar\new@ifnextchar
  \array{*\c@MaxMatrixCols c}}
\makeatother

\AtBeginDocument{\addtocontents{toc}{\protect\thispagestyle{empty}}}
\AtBeginDocument{\addtocontents{lof}{\protect\thispagestyle{empty}}}
\AtBeginDocument{\addtocontents{lot}{\protect\thispagestyle{empty}}}

\graphicspath{{./images/}}

\newcommand{\concat}{%
  \mathbin{%
    \text{\textcircled{\scriptsize$\boldsymbol{;}$}}%
  }%
}

\definecolor{salmonpink}{RGB}{255,145,164}
\definecolor{turquoise}{RGB}{64,224,208}
\definecolor{lavender}{RGB}{220,208,255}
\definecolor{myblue}{RGB}{21,101,192}
\definecolor{myred}{RGB}{198,40,40}
\definecolor{mygreen}{RGB}{35,101,51}

\usetikzlibrary{shapes, arrows.meta, positioning, decorations.pathreplacing, calc}

\newenvironment{definition}[1][Definition]{%
    \vspace{1em}
    \noindent
    \begin{tcolorbox}[colback=blue!10!white, colframe=blue!50!black, title=#1, breakable] 
}{%
    \end{tcolorbox}
}

\newenvironment{example}[1][Example]{%
    \vspace{1em}
    \noindent
    \begin{tcolorbox}[colback=green!5!white, colframe=green!75!black, title=#1, breakable] 
}{%
    \end{tcolorbox}
}

\newenvironment{example-item}[1][Example]{%
    \noindent
    \begin{tcolorbox}[colback=green!5!white, colframe=green!75!black, title=#1, breakable] 
}{%
    \end{tcolorbox}
}

\newcommand{\nospacetext}[1]{\makebox[0pt][l]{#1}}

\newcommand{\trecb}[0]{\textsc{trec-2}}
\newcommand{\trec}[0]{\textsc{trec-6}}
\newcommand{\agnb}[0]{\textsc{agn-2}}
\newcommand{\agn}[0]{\textsc{agn-4}}
\newcommand{\subj}[0]{\textsc{subj}}

\newcommand{\bert}[0]{BERT}
\newcommand{\electra}[0]{ELECTRA}

\newcommand{\beast}[0]{\textsc{beast}}
\newcommand{\alsbi}[0]{\textsc{alsbi}}

\newcommand{\auc}[0]{\textsc{auc}}
\newcommand{\lcr}[0]{\textsc{lcr}}

\newcommand{\st}[0]{\textsc{st}}
\newcommand{\et}[0]{\textsc{et}}
\newcommand{\etb}[0]{\textsc{et$^\mathcal{B}$}}
\newcommand{\etbns}[0]{\textsc{et\nospacetext{$^\mathcal{B}$}}}
\newcommand{\etar}[0]{\textsc{eta}}
\newcommand{\etab}[0]{\textsc{eta$^\mathcal{B}$}}
\newcommand{\etabns}[0]{\textsc{eta\nospacetext{$^\mathcal{B}$}}}

\newcommand{\rnd}[0]{\textsc{rnd}}
\newcommand{\ent}[0]{\textsc{ent}}
\newcommand{\rg}[0]{\textsc{rg}}
\newcommand{\cs}[0]{\textsc{cs}}
\newcommand{\dal}[0]{\textsc{dal}}
\newcommand{\mc}[0]{\textsc{mc}}

\newcommand{\adapter}[0]{adapter}
\newcommand{\lora}[0]{LoRA}
\newcommand{\pt}[0]{prefix-tuning}
\newcommand{\uni}[0]{UniPELT}

\newcommand{\wilda}{\textsc{wilda}\xspace}
\newcommand{\sst}{\textsc{sst}\xspace}
\newcommand{\cola}{\textsc{cola}\xspace}
\newcommand{\rte}{\textsc{rte}\xspace}
\newcommand{\mrpc}{\textsc{mrpc}\xspace}
\newcommand{\qqp}{\textsc{qqp}\xspace}
\newcommand{\qnli}{\textsc{qnli}\xspace}
\newcommand{\mnli}{\textsc{mnli}\xspace}
\newcommand{\elmath}{\textsc{math}\xspace}
\newcommand{\misc}{\textsc{misc}\xspace}

\begin{document}


\frontmatter

\begin{titlepage}
  \fontsize{16pt}{20pt}\selectfont
  \fontfamily{phv}\fontseries{mc}\selectfont
  \newgeometry{left=3cm,right=3cm,top=3cm,bottom=2.5cm}
  \setlength{\intextsep}{0pt plus 0pt minus 0pt}

  \begin{center}
    \begin{figure}[ht!]
      \begin{center}
        \includegraphics[height=4.1184cm, width=5.94cm]{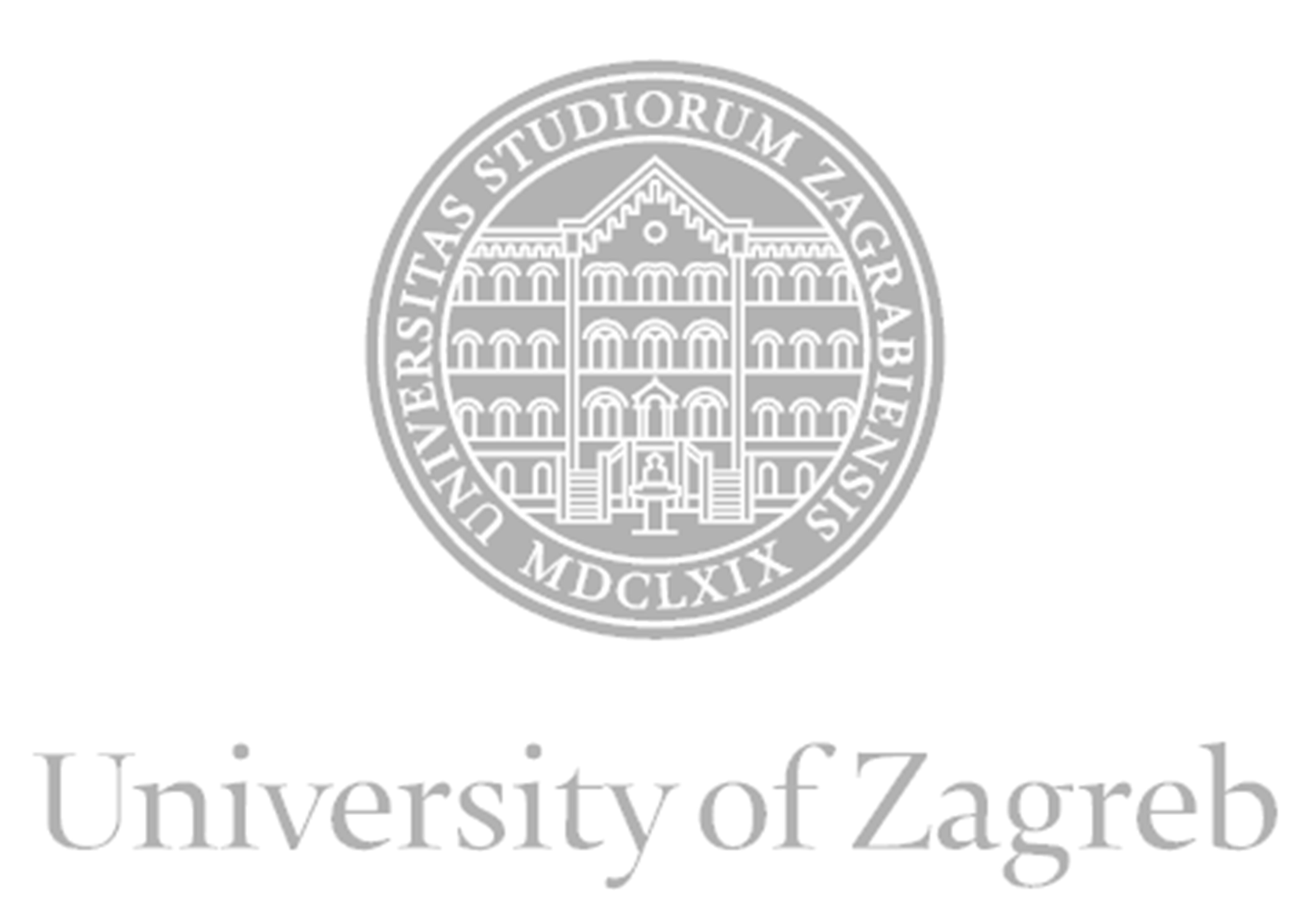}
      \end{center}
    \end{figure}
    \vspace{0cm}
    {FACULTY OF ELECTRICAL ENGINEERING AND COMPUTING} \\
    \vspace{3cm}
    Josip Jukić \\
    \vspace{2cm}
    {\fontsize{22pt}{22pt}\selectfont
\textbf{
IMPROVING DATA AND PARAMETER EFFICIENCY OF NEURAL LANGUAGE MODELS USING REPRESENTATION ANALYSIS}} \\
    \vspace{2cm}  
    DOCTORAL THESIS \\    
    \vfill{Zagreb, 2025}
  \end{center}
  \restoregeometry
\end{titlepage}

\begin{titlepage}
  \fontsize{16pt}{20pt}\selectfont
  \fontfamily{phv}\fontseries{mc}\selectfont
  \newgeometry{left=3cm,right=3cm,top=3cm,bottom=2.5cm}
  \setlength{\intextsep}{0pt plus 0pt minus 0pt}

  \begin{center}
    \begin{figure}[ht!]
      \begin{center}
        \includegraphics[height=4.1184cm, width=5.94cm]{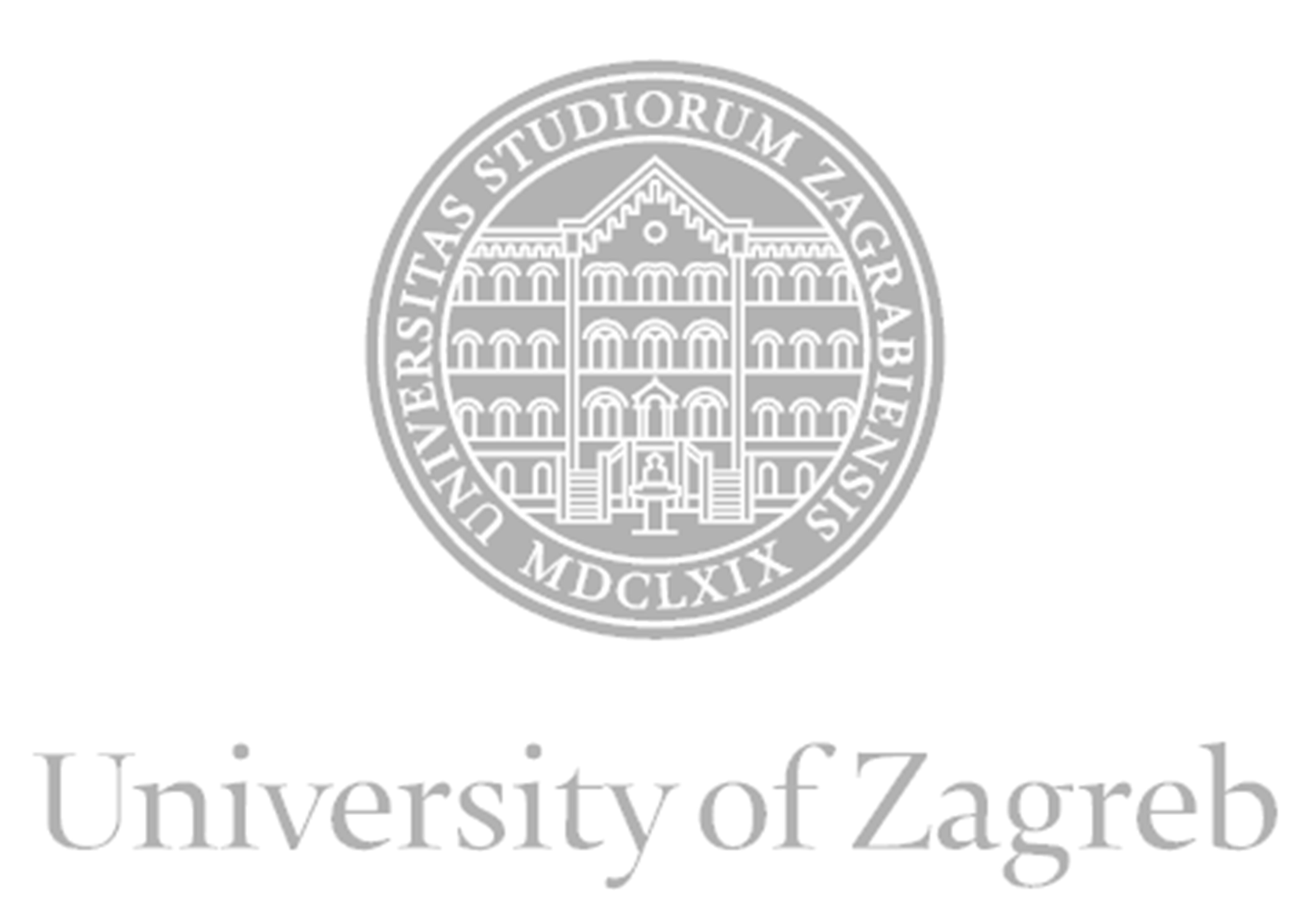}
      \end{center}
    \end{figure}		
    \vspace{0cm}
    {\fontsize{16pt}{16pt}{FACULTY OF ELECTRICAL ENGINEERING AND COMPUTING}} \\
    \vspace{3cm}
    Josip Jukić \\
    \vspace{2cm}
    {\fontsize{22pt}{22pt}\selectfont\textbf{
IMPROVING DATA AND PARAMETER EFFICIENCY OF NEURAL LANGUAGE MODELS USING REPRESENTATION ANALYSIS}} \\
    \vspace{2cm}   
    DOCTORAL THESIS \\  
    \vspace{5cm}   
    Supervisor: Professor Jan Šnajder, PhD \\
    \vfill{Zagreb, 2025}
  \end{center}
  \restoregeometry
\end{titlepage}

\begin{titlepage}
  \fontsize{16pt}{20pt}\selectfont
  \fontfamily{phv}\fontseries{mc}\selectfont
  \newgeometry{left=3cm,right=3cm,top=3cm,bottom=2.5cm}
  \setlength{\intextsep}{0pt plus 0pt minus 0pt}

  \begin{center}
    \begin{figure}[ht!]
      \begin{center}
        \includegraphics[height=4.1184cm, width=5.94cm]{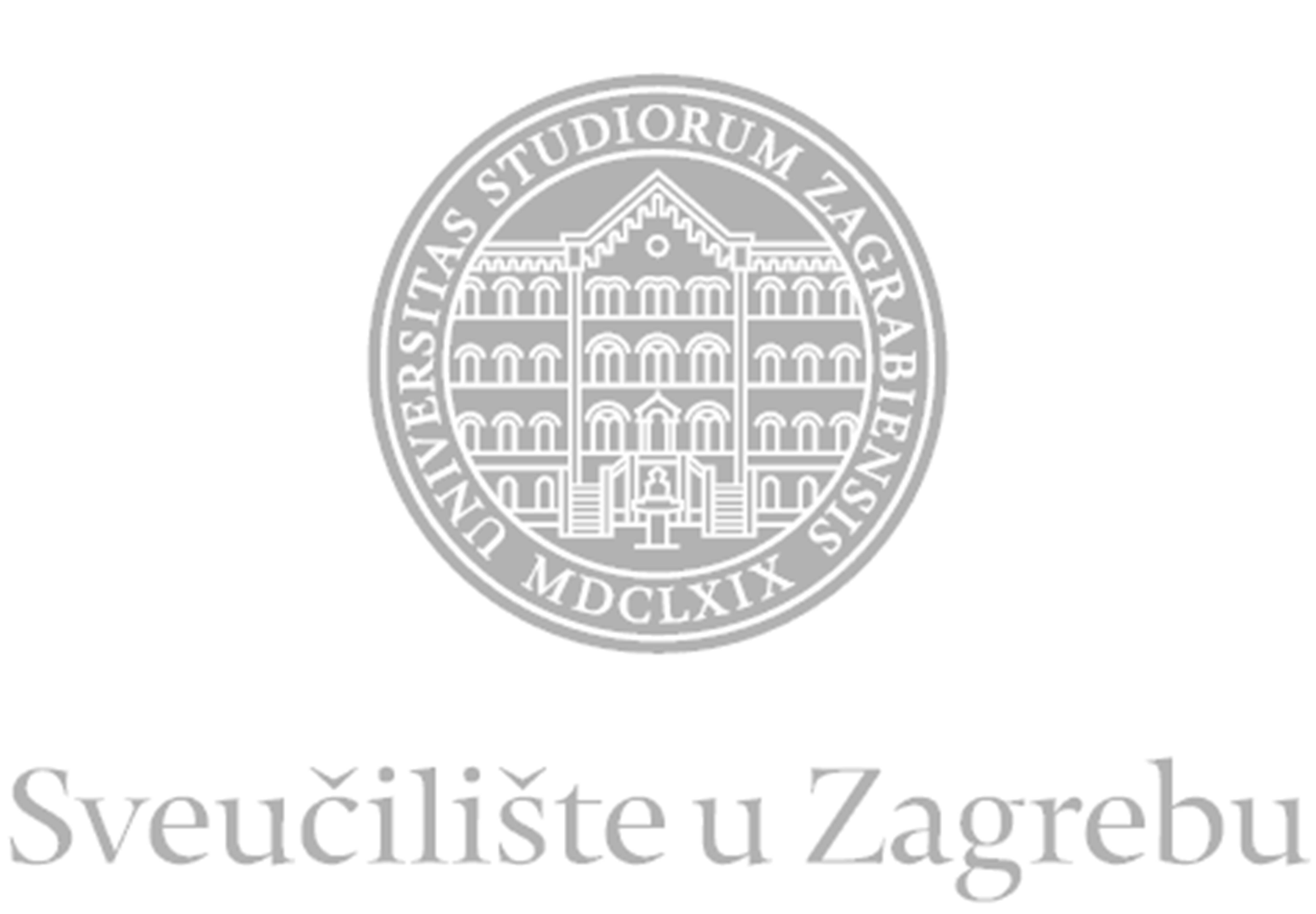}
      \end{center}
    \end{figure}		
    \vspace{0cm}
    {FAKULTET ELEKTROTEHNIKE I RAČUNARSTVA} \\
    \vspace{3cm}
    Josip Jukić \\
    \vspace{2cm}
    {\fontsize{22pt}{22pt}\selectfont\textbf{
POBOLJŠANJE PODATKOVNE I PARAMETARSKE UČINKOVITOSTI NEURONSKIH JEZIČNIH MODELA KORIŠTENJEM ANALIZE REPREZENTACIJA
}} \\
    \vspace{2cm}    
    DOKTORSKI RAD \\
    \vspace{5cm}    
	Mentor: prof. dr. sc. Jan Šnajder \\
    \vfill{Zagreb, 2025.}
  \end{center}
  \restoregeometry
\end{titlepage}

\begin{titlepage}
  \begin{minipage}{\dimexpr\textwidth-1cm}
    \vspace{3cm}
    The doctoral thesis has been made at the University of Zagreb,
    Faculty of Electrical Engineering and Computing, at the Department
    of Electronics, Microelectronics, Computer and Intelligent Systems,
    as part of the Text Analysis and Knowledge Engineering Laboratory (TakeLab).

    \vspace{1cm}
    Supervisor: Professor Jan Šnajder, PhD

    \vspace{1cm}
    The doctoral thesis consists of: 192 pages

    \vspace{1cm}
    Doctoral thesis num.: \line(1,0){0}
  \end{minipage}
\end{titlepage}

\thispagestyle{empty}

\section*{About the Supervisor}

Jan Šnajder has received his BSc, MSc, and PhD degrees in Computer Science from the University of Zagreb, Faculty of Electrical Engineering and Computing (FER), Zagreb, Croatia, in 2002, 2006, and 2010, respectively. From September 2002 he was working as a research assistant, from 2011 as Assistant Professor, from 2016 as Associate Professor, and from 2021 as Full Professor at the Department of Electronics, Microelectronics, Computer and Intelligent Systems at FER. He was a visiting researcher at the Institute for Computational Linguistics at the University of Heidelberg, the Institute for Natural Language Processing at the University of Stuttgart, the National Institute of Information and Communications Technology in Kyoto, and the University of Melbourne. He participated in a number of research and industry projects in the field of natural language processing and machine learning. He is the principal investigator on a HRZZ installation grant project and a HAMAG-BICRO proof-of-concept project, and a researcher on a UKF project. He has (co-)authored more than 100 papers in journals and conferences in natural language processing and information retrieval, and has been reviewing for major journals and conferences in the field. He is the lecturer in charge for six courses at FER and has supervised and co-supervised more than 100 BA and MA theses. He is a member of IEEE, ACM, ACL, the secretary of the Croatian Language Technologies Society, the co-founder and secretary of the Special Interest Group for Slavic NLP of the Association for Computational Linguistics (ACL SIGSLAV). He is a member of the Centre of Research Excellence for Data Science and Advanced Cooperative Systems and the associate editor of the Journal of Computing and Information Technology. He has been awarded the Silver Plaque ``Josip Lončar'' in 2010, the Croatian Science Foundation fellowship in 2012, the fellowship of the Japanese Society for the Promotion of Science in 2014, and the Endeavour Fellowship of the Australian Government in 2015.

\newpage
\section*{O mentoru}

Jan Šnajder diplomirao je, magistrirao i doktorirao u polju računarstva na Sveučilištu u Zagrebu, Fakultetu elektrotehnike i računarstva (FER), 2002., 2006.~odnosno 2010.~godine. Od 2002.~godine radio je kao znanstveni novak, od 2011.~godine kao docent, od 2016.~godine kao izvanredni profesor, a od 2021.~godine kao redoviti profesor na Zavodu za elektroniku, mikroelektroniku, računalne i inteligentne sustave FER-a. Usavršavao se na Institutu za računalnu lingvistiku Sveučilišta u Heidelbergu, Institutu za obradu prirodnog jezika Sveučilišta u Stuttgartu, Nacionalnom institutu za informacijske i komunikacijske tehnologije u Kyotu te Sveučilištu u Melbourneu. Sudjelovao je na nizu znanstvenih i stručnih projekata iz područja obrade prirodnog jezika i strojnog učenja. Voditelj je uspostavnog projekta HRZZ-a i projekta provjere koncepta HAMAG-BICRO-a te je istraživač na projektu UKF-a. Autor je ili suautor više od 100 znanstvenih radova u časopisima i zbornicima međunarodnih konferencija u području obrade prirodnog jezika i pretraživanja informacija te je bio recenzent za veći broj časopisa i konferencija iz tog područja. Nositelj je šest predmeta na FER-u te je bio mentor ili sumentor studentima na više od 100 preddiplomskih i diplomskih radova.

Član je stručnih udruga IEEE, ACM, ACL, tajnik Hrvatskoga društva za jezične tehnologije te suosnivač i tajnik posebne interesne skupine za obradu prirodnog jezika za slavenske jezike pri udruzi za računalnu lingvistiku (ACL SIGSLAV). Član je Znanstvenog centra izvrsnosti za znanost o podacima i kooperativne sustave te je pridruženi urednik časopisa Journal of Computing and Information Technology (CIT). Dobitnik je Srebrne plakete ``Josip Lončar'' 2010. godine, stipendije Hrvatske zaklade za znanost 2012.~godine, stipendije Japanskog društva za promicanje znanosti 2014.~godine te stipendije australske vlade Endeavour 2015.~godine.


\thispagestyle{empty}

\section*{Abstract}

This thesis addresses challenges related to data and parameter efficiency in neural language models, with a focus on representation analysis and the introduction of new optimization techniques. The first part examines the properties and dynamics of language representations within neural models, emphasizing their significance in enhancing robustness and generalization. It proposes innovative approaches based on representation smoothness, including regularization strategies that utilize Jacobian and Hessian matrices to stabilize training and mitigate sensitivity to input perturbations.
The second part focuses on methods to significantly enhance data and parameter efficiency by integrating active learning strategies with parameter-efficient fine-tuning, guided by insights from representation smoothness analysis. It presents smoothness-informed early-stopping techniques designed to eliminate the need for labeled validation sets and proposes innovative combinations of active learning and parameter-efficient fine-tuning to reduce labeling efforts and computational resources. Extensive experimental evaluations across various NLP tasks demonstrate that these combined approaches substantially outperform traditional methods in terms of performance, stability, and efficiency.
The third part explores weak supervision techniques enhanced by in-context learning to effectively utilize unlabeled data, further reducing dependence on extensive labeling. It shows that using in-context learning as a mechanism for weak supervision enables models to better generalize from limited labeled data by leveraging unlabeled examples more effectively during training. Comprehensive empirical evaluations confirm significant gains in model accuracy, adaptability, and robustness, especially in low-resource settings and dynamic data environments.
Overall, this thesis provides novel theoretical contributions and practical methodologies for the field of natural language processing, offering robust frameworks to efficiently train and adapt neural language models in diverse and resource-constrained scenarios.

\vspace{1cm}
\textbf{Keywords}: neural language models; representation analysis; data efficiency; parameter efficiency;  active learning; weak supervision; in-context learning

\selectlanguage{croatian}

\chapter*{Prošireni sažetak}

\section*{Poboljšanje podatkovne i parametarske učinkovitosti neuronskih jezičnih modela korištenjem analize reprezentacija}

Obrada prirodnog jezika (OPJ) jedno je od najdinamičnijih područja računarstva i umjetne inteligencije, usmjereno na razvoj računalnih metoda za razumijevanje i obradu tekstualnih podataka izraženih prirodnim jezikom. Posljednje desetljeće obilježila je eksplozija neuronskih modela, osobito neuronskih jezičnih modela (NJM), koji su značajno unaprijedili sposobnosti OPJ-a u zadatcima poput razumijevanja, prevođenja i generiranja teksta. Ključ ovog napretka leži u transformaciji teksta u računalno obradive reprezentacije jezika. Umjesto tradicionalnih simboličkih pristupa, moderni NJM-ovi oslanjaju se na raspodijeljene reprezentacije -- vektore koji učinkovito sažimaju semantičke odnose među riječima. Analiza tih reprezentacija ključna je za poboljšanje izvedbe NJM-ova i precizno identificiranje njihovih ograničenja, što omogućuje razvoj sofisticiranijih i učinkovitijih metoda obrade jezika. Duboko razumijevanje njihove strukture i svojstava nužno je za izgradnju stabilnih, prilagodljivih i robusnih modela.

Unatoč značajnim postignućima, suvremeni veliki jezični modeli suočavaju se s ozbiljnim izazovima, prvenstveno zbog iznimno visokih računalnih zahtjeva i potrebe za velikim količinama označenih podataka. Stalni rast broja parametara modela donosi visoke računalne troškove, čime se postavlja pitanje dugoročne održivosti. Osim toga, prikupljanje dovoljno kvalitetnih označenih podataka za specijalizirane zadatke često je složen, skup i vremenski zahtjevan proces. Sve to povećava potrebu za učinkovitim metodama prilagodbe NJM-ova u scenarijima s ograničenim resursima.

Kako bi se ti izazovi ublažili, ključno je istražiti strategije koje poboljšavaju učinkovitost NJM-ova, s naglaskom na bolje iskorištavanje dostupnih podataka. U tom kontekstu posebno se ističe aktivno učenje (AU) -- pristup koji smanjuje troškove označavanja tako što strateški bira najinformativnije primjere za ručno označavanje. Uz aktivno učenje, važnu ulogu ima i slabo nadziranje (engl.~\textit{weak supervision}), tehnika koja koristi neoznačene podatke generirajući pseudo-oznake, čime dodatno poboljšava učinkovitost modela.

U ovom radu detaljno se istražuje kako svojstva reprezentacija u NJM-ovima utječu na njihovu generalizaciju, stabilnost i prilagodljivost. Predložene metode oslanjaju se na analizu reprezentacija, aktivno učenje i parametarski učinkovite tehnike kako bi se poboljšala učinkovitost jezičnih modela. Ključni doprinosi rada uključuju analizu utjecaja svojstava reprezentacija na sposobnost modela da generalizira, razvoj metode za rano zaustavljanje učenja temeljene na analizi glatkoće reprezentacija, čime se eliminira potreba za označenim validacijskim skupovima, te integraciju parametarski učinkovitog finog podešavanja s aktivnim učenjem kako bi se omogućila prilagodba modela u uvjetima s ograničenim brojem označenih podataka. Nadalje, predložen je okvir temeljen na slabom nadziranju koji omogućuje stabilniju i učinkovitiju prilagodbu modela u scenarijima učenja iz konteksta. Ovi doprinosi zajedno čine sustavan pristup povećanju učinkovitosti i prilagodljivosti NJM-ova, omogućujući njihovu širu primjenu u okruženjima s ograničenim resursima.

\section*{I. dio}
Prvi dio ovoga rada sastoji se od tri poglavlja te istražuje prirodu i razvoj reprezentacija jezika u OPJ-u, naglašavajući njihovu ključnu ulogu u razumijevanju i poboljšanju modela. 
U ovome dijelu prvo je izložen razvoj reprezentacija jezika, od ranih simboličkih metoda do suvremenih kontekstualiziranih reprezentacija. Nakon toga, analiziraju se ključna svojstva reprezentacija koja utječu na robusnost i generalizaciju modela.
Kao doprinos prvog dijela, uvedena je nova metoda za regularizaciju, odnosno sprječavanje prenaučenosti modela, temeljena na zaglađivanju reprezentacija.

U drugome poglavlju (``\textit{Language Representation}'') opisan je razvoj tehnika reprezentacije jezika kroz povijest. Na početku se uvodi temeljna ideja o načinima predstavljanja riječi te se opisuju prva rješenja, poput simboličkih pristupa. Nakon toga obrađuje se distribucijska semantika, koja se temelji na supojavljivanju riječi i omogućuje otkrivanje latentnih semantičkih struktura. Predstavlja se značaj neuronskih distribuiranih reprezentacija, koje omogućuju učinkovito učenje vektorskih prikaza riječi i hvatanje složenih semantičkih odnosa. Posebna pozornost posvećena je naprednim metodama koje dinamički prilagođavaju značenje riječi ovisno o kontekstu. Ključnu ulogu pritom imaju modeli temeljeni na arhitekturi Transformatora (engl.~\textit{Transformer}), čija sposobnost kontekstualne prilagodbe značajno poboljšava razumijevanje jezika i generalizaciju modela.

U trećemu poglavlju (``\textit{Representation Properties}'') izlažu se ključna svojstva reprezentacija jezika koje formiraju NJM-ovi te se analizira njihov utjecaj na generalizaciju, stabilnost i robusnost modela. Poseban naglasak stavljen je na analizu karakteristika reprezentacijskog prostora, s obzirom na to da one određuju način na koji modeli obrađuju nove podatke, reagiraju na male perturbacije u ulazima te se prilagođavaju promjenama domene.
Središnji koncept poglavlja jest glatkoća reprezentacija -- svojstvo koje osigurava da male izmjene u ulaznim podacima ne uzrokuju velike promjene u izlazima modela.
Posebna pažnja posvećena je povezanosti između glatkoće reprezentacija i kalibracije modela, odnosno sposobnosti točne procjene vlastite nesigurnosti.
U završnome dijelu poglavlja uveden je koncept Besovljevih prostora, matematički okvir koji pruža mogućnost detaljnije analize reprezentacija kroz različite slojeve NJM-ova.

U četvrtome poglavlju (``\textit{Representation Regularization Based on Jacobian and Hessian Matrices}'') predstavljena je nova metoda za regularizaciju reprezentacija NJM-ova, pod nazivom \jachess{}. Metoda se temelji na minimiziranju normi Jacobijevih i Hesseovih matrica ulazno-izlaznih funkcija slojeva neuronske mreže,  što djeluje kao surogatni zadatak koji rezultira zaglađenijim reprezentacijama. Glavni cilj metode jest poboljšati generalizaciju i kalibraciju jezičnih modela povećanjem robusnosti modela. Metoda primjenjuje regularizaciju zaglađivanjem reprezentacija središnjih i izlaznih slojeva NJM-a.
Kako bi metoda bila računalno izvediva u visokodimenzionalnim prostorima NJM-ova, koristi se Hutchinsonov procjenitelj za izračun normi Jacobijevih i Hesseovih matrica. Predložene su strategije za određivanje faktora regularizacije koji kontroliraju jakost regularizacije na razini pojedinih slojeva, pri čemu se pokazalo da je optimalno rasporediti regularizacijske faktore proporcionalno postojećoj glatkoći reprezentacija prije finog podešavanja modela.
Empirijski je vrednovana učinkovitost metode \jachess{}-a za NJM-ove temeljene na arhitekturi Transformatora nizom eksperimenata na eksperimenata na skupu podataka GLUE-a. Rezultati su pokazali da \jachess{} značajno poboljšava robusnost NJM-ova, kako u slučaju kontinuiranih perturbacija nad vektorskim reprezentacijama riječi, tako i kod diskretnih korupcija, odnosno maskiranja pojedinih riječi. Metoda je pokazala izuzetne rezultate za generalizaciju na podatcima unutar distribucije, uz značajna poboljšanja u odnosu na druge srodne tehnike regularizacije.
Osim toga, predložena metoda ostvaruje poboljšanja i prilikom vrednovanja na neviđenim domenama podataka.
Dodatno, rezultati su pokazali kako metoda izravno doprinosi poboljšanju kalibracije modela. Grafička analiza kalibracije također je potvrdila superiornost \jachess{}-a u usklađivanju pouzdanosti modela s njegovom stvarnom točnošću.
Predložena metoda značajno doprinosi području regularizacije NJM-ova, omogućujući istovremeno poboljšanje generalizacije i kalibracije modela kroz promicanje glatkih reprezentacija. Osim što unapređuje temeljne sposobnosti modela, predstavlja vrijedan alat za daljnja istraživanja i praktične primjene NJM-ova, osobito u realnim i visokorizičnim okruženjima gdje su stabilnost i pouzdanost ključni zahtjevi.

\section*{II. dio}
Budući da upotreba prednaučenih NJM-ova često zahtijeva velike količine označenih podataka i značajne računalne resurse, njihova primjena u praksi može biti ograničena, osobito u okruženjima s oskudnim resursima. Kako bi se prevladali ti izazovi, drugi dio ovog rada, koji se sastoji od četiri poglavlja, usmjeren je na optimizaciju učinkovitosti podataka i parametara pri finom podešavanju NJM-ova.

Peto poglavlje (``\textit{Active Learning}'') uvodi aktivno učenje kao strategiju za povećanje učinkovitosti korištenja označenih podataka. AU omogućuje modelu da iterativno odabire najinformativnije primjere za označavanje, čime se znatno smanjuje potreba za velikim količinama ručno označenih primjera. Opisane su ključne metode i teorijske osnove AU-a, poput odabira primjera temeljenog na nesigurnosti modela, neslaganju modela, minimizaciji očekivane greške, te uzorkovanju temeljenom na raznolikosti odabranih primjera.

U šestome poglavlju (``\textit{Enhancing Active Learning in PLMs
through Representation Smoothness}'') predstavljena je metoda za poboljšanje AU-a u prednaučenim NJM-ovima korištenjem analize glatkoće reprezentacija temeljenih na teoriji Besovljevih prostora. Polazeći od uvida da glatkoća reprezentacija značajno utječe na generalizaciju i stabilnost NJM-ova tijekom finog podešavanja, razvijena je nova tehnika ranog zaustavljanja pod nazivom \beast{} (\textit{Besov early stopping}), koja eliminira potrebu za označenim validacijskim skupovima.
Eksperimenti su provedeni na više zadataka OPJ-a različite složenosti, uključujući klasifikaciju pitanja, subjektivnosti i tematike vijesti. Analizirani su temeljeni na arhitekturi Transformatora, uspoređujući različite strategije aktivnog učenja s predloženim pristupom temeljenim na glatkoći reprezentacija. Rezultati pokazuju da produljeno učenje u kombinaciji s predloženom metodom značajno povećava učinkovitost AU-a.
Predložena je i heuristika pod nazivom \alsbi{} (\textit{Active Learning Stopping by Besov Index}), koja koristi glatkoću aktivno odabranih primjera za automatsko određivanje optimalne točke zaustavljanja procesa AU-a.
Zaključno, poglavlje ističe kako iskorištavanje glatkoće reprezentacija može značajno unaprijediti stabilnost i efikasnost aktivnog učenja u PLM-ovima, pružajući praktična rješenja za probleme visoke cijene označavanja i ograničenosti resursa u OPJ-u.

Sedmo poglavlje (``\textit{Parameter-Efficient Learning}'') izlaže metode parametarski učinkovitog učenja, koje omogućuju prilagodbu prednaučenih NJM-ova uz minimalne promjene parametara, smanjujući računalne zahtjeve, a zadržavajući kvalitetnu izvedbu modela. Predstavljeni su različiti pristupi parametarski učinkovite prilagodbe temeljeni na modularnosti, koja omogućava fleksibilno kombiniranje komponenti modela za specifične zadatke. Izloženi su ključni pristupi parametarski učinkovitog finog podešavanja (PUFP), uključujući prilagodnike (engl.~\textit{adapters}), metode temeljene na dekompoziciji matrica parametara s niskim rangom te tehnike koje koriste manipulaciju tekstnim upitima. Zajednička značajka svih ovih metoda jest da se osvježava samo mali skup novih parametara, dok izvorni parametri modela ostaju zamrznuti, čime se postiže učinkovita prilagodba uz minimalne resursne zahtjeve.

U osmome poglavlju (``\textit{Integrating Active Learning with Parameter-Efficient Fine-Tuning}'') kao dio doprinosa istražuje se integracija AU-a s metodama PUFP-a radi povećanja efikasnosti prilagodbe prednaučenih NJM-ova u uvjetima ekstremno ograničenih resursa. U poglavlju je provedena sustavna eksperimentalna analiza metoda PUFP-a u kombinaciji s različitim strategijama AU-a na zadatcima klasifikacije teksta.
Rezultati pokazuju metode PUFP-a dosljedno nadmašuju tradicionalno fino podešavanje u scenarijima s ograničenim podatcima, osobito u kombinaciji s AU-om.
Dodatna analiza otkriva da metode PUFP-a selektivno preferiraju umjereno teške primjere, koji su ključni za učinkovito učenje, dok istovremeno zadržavaju reprezentacije bliže izvornom prednaučenom model. Rezultati jasno pokazuju prednosti integracije AU-a s metodama PUFP-a u stvaranju praktičnih i efikasnih modela za zadatke OPJ-a u uvjetima gdje su resursi oskudni.

\section*{III. dio}
Treći dio rada sastoji se od dva poglavlja te istražuje nov pristup prilagodbi velikih NJM-ova temeljen na učenju iz konteksta. Dok tradicionalne tehnike finog podešavanja izravno osvježavaju parametre modela, učenje u kontekstu omogućava modelima prilagodbu novim zadatcima bez eksplicitnog učenja. Unatoč prednostima, ovaj pristup suočava se s izazovima poput osjetljivosti na oblik tekstnih upita, odabir demonstracija i ograničenja dužine kontekstnog prozora.
U ovome dijelu uvodi se nov okvir za slabo nadziranje koji kombinira učenje u kontekstu s paradigmom ``učitelj-učenik''.

Deveto poglavlje (``\textit{In-Context Learning}'') uvodi koncept učenja iz konteksta, oblika prilagodbe velikih NJM-ova koji ne zahtijeva ažuriranje parametara, već modelu pruža nekoliko primjera (demonstracija) u sklopu samog ulaznog upita. Predstavljene su teorijske osnove učenja iz konteksta, uključujući koncept prepoznavanja latentnih zadataka ugrađenih u model tijekom primarnog učenja. Raspravlja se o mehanizmima kojima NJM-ovi koriste demonstracije za dinamičku prilagodbu reprezentacija te o faktorima koji utječu na učinkovitost ove prilagodbe.

Deseto poglavlje (``\textit{Weak Supervision for Disentangling Latent Shifts}'') uvodi nov pristup učenju unutar paradigme učenja iz konteksta, osmišljen kako bi prevladao izazove povezane s isprepletenošću latentnih reprezentacija demonstracija i upita kod velikih NJM-ova. Ovaj problem dovodi do nestabilnosti i otežava učinkovitu prilagodbu modela na nove zadatke. Predložena metoda, nazvana \wilda, koristi okvir ``učitelj-učenik'' za interno kodiranje znanja stečenog iz demonstracija unutar parametarski učinkovitih modula. 
Za razliku od postojećih metoda koje se temelje na manipulaciji mehanizama pažnje ili skrivenih stanja, \wilda iterativno prilagođava parametarski učinkovite module koristeći pseudo-oznake koje generira učiteljski model. Ovakav pristup omogućuje jasnije razdvajanje znanja iz demonstracija od obrade samog upita, čime se povećava stabilnost modela i poboljšava njegova sposobnost generalizacije.
Eksperimentalno vrednovanje pokazuje da \wilda dosljedno nadmašuje standardne metode učenja iz konteksta i postojeće pristupe koji se oslanjaju na manipulaciju latentnih reprezentacija, pokazujući značajna poboljšanja u stabilnosti i generalizaciji, unutar i izvan domena podataka za učenje. Također, kroz detaljnu analizu utvrđeno je da \wilda pokazuje karakteristike generalizacije od slabijeg k jačem (engl.~\textit{weak-to-strong generalization}), učinkovito koristeći mehanizme ispravljanja pseudo-oznaka i proširenja pokrivenosti.
Dodatno, \wilda omogućava učinkovito upravljanje velikim brojem demonstracija putem modularne arhitekture i tehnike aritmetike nad modulima, što dodatno pridonosi skalabilnosti i učinkovitosti prilagodbe NJM-ova. Ovi nalazi potvrđuju potencijal slabog nadziranja kao učinkovite metode za poboljšanje izvedbe i stabilnosti učenja iz konteksta u scenarijima s ograničenim resursima.

\section*{Rasprava i zaključak}
U pretposljednjemu poglavlju (``\textit{Discussion}'') analiziraju se implikacije dobivenih rezultata, pri čemu se razmatraju ključna ograničenja predloženih metoda.
Zaključno poglavlje (``\textit{Conclusion}'') nadovezuje se na prethodnu raspravu pružajući sažet pregled ključnih nalaza i doprinosa rada. Ističe se značaj predloženih metoda u kontekstu povećanja učinkovitosti korištenja podataka i parametara, poboljšanja stabilnosti učenja te unapređenja sposobnosti modela za generalizaciju. Također, predloženi su smjerovi budućih istraživanja koji bi mogli dodatno unaprijediti robusnost i praktičnu primjenjivost neuronskih jezičnih modela u stvarnim primjenama. Ovim završnim dijelom rada zaokružuje se cjelokupno istraživanje te otvaraju nove perspektive za daljnji razvoj.

\vspace{1cm}
\textbf{Ključne riječi}: neuronski jezični modeli; analiza reprezentacija; podatkovna učinkovitost; parametarska učinkovitost; aktivno učenje; slabo nadziranje; učenje u kontekstu

\selectlanguage{english}

\clearpage
\pagestyle{empty} 
\tableofcontents
\cleardoublepage 

\pagestyle{fancyplain} 

\mainmatter
\chapter{Introduction}
\label{ch:intro}

Natural language processing (NLP) is a prominent subfield of computer science and artificial intelligence (AI) designed to enable machines to process and interpret data encoded in natural language. The field experienced a dramatic surge in growth during the early 2010s, fueled by the neural revolution in AI, which profoundly influenced NLP. More recently, neural language models (NLMs) have taken NLP by storm and helped to advance the field with remarkable results in various aspects, such as language inference, generation, and translation. Contemporary language models commonly employ neural architectures, such as feedforward neural networks and their deep variants, to transform words, sentences, and text chunks into numeric representations. These representations are rooted in the \textit{distributional hypothesis} \cite{harris-1954-distributional}, which states that words appearing in similar contexts tend to have similar meanings. This notion was further popularized by Firth, who put it succinctly: ``A word is characterized by the company it keeps'' \cite{firth-1957-synopsis}. The distributional hypothesis serves as the foundation for \textit{distributional semantics}, which studies how linguistic meaning can be inferred from contextual usage patterns in large corpora \cite{turney-2010-vector}. To illustrate this concept, consider a simple sentence like:

\begin{center}
  \textit{``She used a \rule{1cm}{0.15mm} to cut the bread.''}
\end{center}

From the context of this sentence, we can infer that the missing word is some kind of a tool, most probably a ``knife''.  Distributional semantics leverages this as an idea to design supervised learning tasks from unstructured text in order to learn semantic representations of words from the contexts in which the word appears. These representations are commonly referred to as \textit{distributed representations} or \textit{embeddings} \cite{collobert-2008-unified}. In contrast to traditional symbolic representations, distributed representations are typically more compact and structured, embedding data into a low-dimensional, continuous space that captures semantic relationships more effectively.

While distributional semantics primarily focuses on the meaning of individual words, contemporary language models extend this idea to entire sequences, including phrases, sentences, and even paragraphs. This extension is grounded in the principle of semantic composition -- the idea that the meaning of a complex expression is derived from the meanings of its parts and the rules used to combine them \cite{frege-1956-thought}. NLMs implement this by leveraging deep learning architectures to process sequences of text, capturing both the individual meanings of words and their contextual interactions.

A major breakthrough in NLP has been the development of pre-trained language models (PLMs). These models, typically based on deep neural architectures, are trained in two distinct phases: \textit{pre-training} and \textit{fine-tuning}.
In the pre-training phase, PLMs undergo large-scale self-supervised learning on vast text corpora, where they learn to predict missing words (masked language modeling) or the next word in a sequence (causal language modeling). This phase enables the model to acquire broad linguistic knowledge, capturing syntax, semantics, and contextual relationships without requiring labeled data.
In the fine-tuning phase, the pre-trained model is further trained on labeled data specific to a particular NLP task. Fine-tuning allows the model to adapt its general language understanding to specialized applications with relatively little task-specific data.

Building on this foundation, the late 2010s witnessed the rise of large language models (LLMs) as an extension of PLMs, characterized by an unprecedented increase in both the number of parameters and the scale of pre-training data. This growth has driven remarkable advancements across various NLP tasks, enabling models to achieve impressive generalization. One of the key advantages of LLMs is their ability to perform \textbf{in-context learning} \cite{brown-etal-2020-language}. Unlike traditional supervised learning paradigms that require extensive fine-tuning on labeled datasets, in-context learning enables LLMs to infer task-specific patterns and generate relevant outputs based on a few provided examples.
However, despite their advances, LLMs face critical challenges, particularly their dependence on enormous datasets for effective training. State-of-the-art LLMs often require corpora containing billions of tokens, raising both practical and ethical concerns. The reliance on such large-scale data is increasingly unsustainable due to the significant computational costs, environmental impact, and financial burden of data collection and curation, and the fact that the amount of high-quality text data, while growing, remains inherently finite.

This unsustainable trajectory highlights the urgent need for \textbf{data-efficient} strategies in NLM training.. Empirical research has demonstrated that language model performance follows predictable power-law relationships with respect to model size, dataset size, and computational resources, a phenomenon known as \textit{LLM scaling laws} \cite{kaplan-2020-scaling, hoffmann-2022-chinchilla}. However, the current trend of scaling by consuming ever-larger datasets and computational resources is raising barriers to the democratization of AI research. For example, resource-constrained environments or smaller organizations often lack the financial and computational resources necessary to build or deploy such models. Similarly, underrepresented languages and specialized domains risk being left behind as the focus remains on widely available, high-volume datasets. These challenges underscore the pressing need for more efficient training paradigms that break the reliance on brute-force scaling while maintaining or improving model performance.

In the fine-tuning phase, a major bottleneck in data collection is the reliance on human-labeled annotations, which are labor-intensive, time-consuming, and expensive to produce at scale. Human annotators must label, verify, and refine the data to ensure quality, but this approach becomes increasingly infeasible as the demand for larger and more diverse datasets grows. The problem is further compounded in specialized domains, underrepresented languages, or niche applications, where large, high-quality datasets are often scarce or prohibitively expensive to create.

By prioritizing advancements in data efficiency, we can reduce the reliance on massive datasets and computational resources while maintaining high model performance. A key strategy in this pursuit is \textbf{active learning} \cite{settles-2009-active}, which enhances efficiency by selecting the most informative data points for labeling. Rather than passively consuming vast amounts of data, active learning enables models to focus on examples that maximize learning gains while reducing annotation costs.
Complementing this approach is \textbf{weak supervision} \cite{zhang-etal-2022-survey}, a paradigm that enables models to learn from imperfect, noisy, or indirect supervision signals, such as heuristics, distant supervision, or model-generated labels, rather than relying solely on manually annotated data. Within the framework of in-context learning, weak supervision can be implemented by prompting models with imperfect or automatically constructed demonstrations, allowing them to generalize from unlabeled data with minimal human input. This approach expands the training signal while reducing annotation costs, leading to improved generalization and efficiency, particularly in low-resource settings.

While improving data efficiency is crucial for making LLMs more sustainable, it is only one part of the equation. Another pressing challenge is \textbf{parameter efficiency}. Current LLMs at the high end of the spectrum contain hundreds of billions of parameters, driving up computational costs during both training and inference \cite{naveed-etal-2023-comprehensive}. Notably, some proprietary models with undisclosed architectures are speculated to be an order of magnitude larger, further exacerbating the resource demands. This immense scale necessitates substantial hardware resources, making LLMs inaccessible to many researchers and organizations and impractical for deployment in devices with limited computational capabilities. For state-of-the-art proprietary models, the cost of a single inference (producing a response to a single query) can reach hundreds of dollars, rendering their usability questionable in real-world applications where frequent or large-scale inference is required.

The substantial energy consumption of these models amplifies their sustainability challenges, emphasizing the importance of developing parameter-efficient solutions. Addressing parameter efficiency is important not only for ensuring the broader usability of LLMs but also for fostering innovation in scenarios where resources are constrained. Without significant advancements in this area, the long-term scalability and inclusivity of LLMs remain in jeopardy, limiting their potential to benefit diverse applications and communities.

Beyond simply increasing data and model size, a better understanding of how NLMs encode and organize information can lead to more efficient and effective models. \textbf{Representation analysis}, in particular, offers valuable insights into the internal mechanics of these models, helping to identify inefficiencies and enhance overall performance. One critical property is \textit{smoothness}, which ensures that similar inputs yield similar outputs, promoting stability and generalization. By analyzing and optimizing smoothness in vector spaces, inefficiencies in representation learning can be uncovered, reducing redundancy and improving expressiveness without adding unnecessary model complexity. Additionally, tracking how representations evolve throughout training enables more strategic data usage by prioritizing diverse and high-impact examples rather than relying on sheer dataset size. As models continue to scale, leveraging insights from representation analysis will be essential for developing neural language models that maintain strong performance while remaining computationally sustainable and widely accessible.

This thesis leverages representation analysis to enhance the data and parameter efficiency of NLMs, enabling robust performance with reduced reliance on large datasets and extensive computational resources. To achieve this, it integrates active learning, weak supervision, and parameter-efficient techniques within both fine-tuning and in-context learning paradigms. The findings show that these approaches effectively improve model efficiency, lowering computational costs and annotation requirements while preserving or enhancing generalization capabilities.

\section{Contribution}

This research introduces novel methods that leverage learned representations to improve the efficiency and adaptability of NLMs. The key original contributions of this work include:

\begin{itemize}
    \item An analysis of the effect of representation properties on generalization and uncertainty estimates of neural language models for natural language processing;
    
    \item An algorithm for early stopping prediction for language model fine-tuning based on representation smoothness analysis, without the need for labeled held-out data;
    
    \item A method for improving data and parameter efficiency of neural language models by combining parameter-efficient fine-tuning and active learning in low-resource scenarios;

    \item A weak supervision framework based on in-context learning that disentangles latent shifts introduced by demonstrations, enabling more stable and efficient few-shot adaptation.
\end{itemize}

Together, these contributions provide a systematic approach to improving the efficiency and adaptability of NLMs, addressing key challenges in generalization, computational cost, and data efficiency across both fine-tuning and in-context learning paradigms.

\section{Thesis Structure}
\label{sec:thesis_structure}

This thesis is structured into three parts. The first part of the thesis establishes a theoretical foundation by examining the evolution of language representations. \Cref{ch:background} provides an overview of language representations in NLP, tracing their development from symbolic approaches to distributed and contextualized embeddings. Together with \Cref{ch:rep-props}, which outlines key properties of representations, these chapters set the stage for the analysis in \Cref{ch:jachess}, which introduces techniques for studying the properties of learned representations in NLMs. It discusses regularization techniques such as Jacobian and Hessian norm regularization and presents a smoothness-informed regularization technique.

The second part of the thesis focuses on active learning and its integration with representation analysis and parameter-efficient methods. \Cref{ch:al} provides a theoretical and empirical exploration of AL, covering key query strategies and their applications in NLP. \Cref{ch:beast} builds on these insights, introducing a novel early stopping method for fine-tuning NLMs without requiring labeled validation data. This method leverages representation smoothness to improve model efficiency and generalization in AL settings, where models are iteratively updated with new labeled data. \Cref{ch:peft} then examines parameter-efficient fine-tuning techniques, including adapters, low-rank adaptation, and prompt tuning, highlighting their role in reducing computational costs while maintaining strong performance. \Cref{ch:al-peft} integrates these two efficiency-driven approaches, presenting a unified framework that combines active learning with parameter-efficient techniques. 

The third part of the thesis shifts focus to in-context learning and weak supervision. \Cref{ch:icl} introduces in-context learning and its limitations. Building on this, \Cref{ch:wilda} presents a framework that enables NLMs to internalize in-context knowledge through weak supervision while minimizing computational overhead with parameter-efficient techniques.

The final part of the thesis explores its implications, examines its limitations, and presents the conclusion in \Cref{ch:discussion,ch:conclusion}.

\part{Representation Analysis}
The first part of this thesis explores the nature and evolution of language representations in NLMs, emphasizing their central role in understanding and improving these models. Language representations serve as the foundation for how NLMs process text, influencing their ability to generalize, adapt, and perform across diverse tasks. A key challenge in NLP is constructing representations that effectively capture linguistic structure and meaning while remaining computationally efficient. Understanding these representations and their properties is crucial for both diagnosing model limitations and developing techniques to enhance learning dynamics, stability, and efficiency.  

\Cref{ch:background} provides an overview of language representations, tracing their development from early symbolic methods to contextualized embeddings in transformer-based models. \Cref{ch:rep-props} explores key properties that influence generalization and robustness, such as smoothness in representation spaces and model calibration. These concepts are essential for understanding how neural models learn and retain information.

A core contribution of this thesis in this domain is the introduction of a smoothness-informed regularization framework (\Cref{ch:jachess}). The proposed framework leverages Jacobian and Hessian regularization to improve generalization and robustness by enforcing smoothness constraints. This reduces sensitivity to perturbations and enhances model calibration, providing a principled approach to improving representation learning.

\chapter{Language Representations}
\label{ch:background}

NLP fundamentally deals with words, whereas traditional machine learning models operate on numerical data. This inherent gap necessitates translating linguistic information into a form that computational models can process effectively. Understanding how words are represented in NLP is crucial not only for appreciating the development of NLMs but also for identifying their strengths and limitations. The evolution of word representations, from early symbolic approaches to modern distributed embeddings, has closely followed advancements in computational techniques and the growing availability of large-scale datasets.

This chapter traces the historical progression of word representations in NLP, highlighting key milestones that have shaped the field. We examine the transition from symbolic methods to statistical approaches, the rise of distributed representations, and the advent of contextualized word embeddings. A deeper understanding of these representations provides critical insight into how NLMs capture meaning, structure, and relationships within language, forming the foundation for addressing the challenges of efficiency and generalization that this thesis aims to explore.

\section{Symbolic Representations of Words}

Early efforts to represent language in computational models relied on simple, straightforward techniques. These representations treated words as independent, isolated units, providing the foundation for the more advanced methods that followed.

\begin{definition}[Alphabet]
In formal language theory \cite{harrison-1978-introduction}, an \textit{alphabet}, denoted by $\Sigma$, is a finite set of symbols or letters. These symbols are the fundamental building blocks used to construct sequences of characters. From this alphabet $\Sigma$, we can generate all possible strings of finite length, including the \textit{empty string} $\epsilon$. The set of all such strings is denoted $\Sigma^*$ and is known as the \textit{Kleene star} of $\Sigma$.    
\end{definition}

In computational models of language, a concept closely related to the alphabet is the \textit{vocabulary},  which we denote by $V$. A vocabulary is essentially a finite collection of words that are used in specific language-processing tasks. While an alphabet $\Sigma$ typically represents individual characters or symbols, the vocabulary $V$ consists of \textit{tokens} -- strings or sequences of these symbols. Hence, a vocabulary $V$ can be thought of as a finite subset of $\Sigma^*$, where each element of $V$ is a word formed by concatenating one or more symbols from the alphabet $\Sigma$, i.e., $V \subset \Sigma^*$.
Thus, the vocabulary $V$ represents a finite selection of valid strings, often corresponding to words or subwords, from the infinite set of all possible strings over $\Sigma$.

\begin{figure}[]
    \centering
    \includegraphics[width=0.5\textwidth]{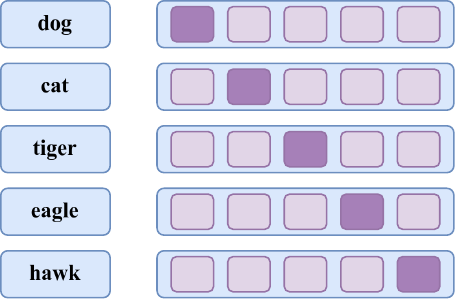}
    \caption[One-hot encoding]{Illustration of one-hot encoding for a vocabulary of five words. Each word is represented as a unique vector in a five-dimensional space, where dark cells indicate a value of $1$ and the light cells are $0$. However, all words are equidistant from one another. This uniform distance fails to capture semantic relationships between words, making one-hot encoding less favorable for representing word meanings in machine learning tasks.}
    \label{fig:one_hot}
\end{figure}

\begin{definition}[One-hot representation]
In symbolic representation schemes, tokens from the vocabulary $V$ were represented as atomic units or discrete symbols, where each word was treated independently of the others. One common method for representing words in these models was the \textit{one-hot} representation, where each word is assigned a unique identifier or index within the vocabulary. Given a vocabulary $V$ of size $N = |V|$, each token $t_i \in V$ is represented by a vector $\mathbf{v}_i \in \mathbb{R}^{N}$, where:
\[
\mathbf{v}_i = 
\left[
0, 0, \ldots, 1, \ldots, 0
\right] ,
\]
with the $1$ at the position corresponding to the index of a particular word in $ V $ (Figure \ref{fig:one_hot}).
\end{definition}

One-hot representation is widely used beyond NLP and serves as the default representation for categorical (multinomial) variables in machine learning classification models. While effective in scenarios where categories have no inherent structure or relationships, it becomes problematic in language modeling, where words exhibit rich semantic and syntactic dependencies.

Despite their simplicity and widespread use, one-hot representations suffer from several key limitations. The most fundamental issue is the \textit{lack of semantic relations} -- while words do exhibit similarities and connections in natural language, these relationships are not encoded in one-hot representations, where each word is treated as an independent entity. As a result, words with similar meanings or frequent co-occurrences remain equally distant in the representation space, preventing models from leveraging inherent linguistic structure and limiting their ability to generalize effectively.

Beyond semantic limitations, one-hot representations also pose serious \textit{scalability} challenges. As vocabulary size grows, the dimensionality of one-hot vectors increases proportionally, leading to inefficient computation and storage. Modern NLP applications, which rely on extensive textual data, struggle with this high-dimensional representation, making large-scale processing impractical. Furthermore, one-hot representations are inherently \textit{sparse}, with most dimensions in each vector containing zeros. This sparsity results in inefficient memory usage and increased computational overhead, as handling high-dimensional sparse vectors demands substantial resources without contributing meaningful information.

Despite these drawbacks, symbolic representations played a foundational role in early NLP systems. However, as the need for richer linguistic representations became apparent, their limitations motivated a shift toward more advanced word representation techniques. This transition paved the way for distributed embeddings and deep learning-based models, which offer a more expressive and efficient approach to encoding language.

\section{Distributional Semantics}

Moving beyond simple symbolic representations, we now explore \textit{distributional semantics}, a framework grounded in the distributional hypothesis \cite{harris-1954-distributional}. By analyzing co-occurrence patterns in large corpora, distributional semantics enables a data-driven, context-sensitive approach to meaning representation, forming the foundation of modern NLP models. This section examines its key components, from fundamental notions of semantic similarity to practical implementations and advancements such as latent semantic analysis.

\subsection{Co-occurrence Matrices}

Building on the principles of distributional semantics, co-occurrence matrices emerged as an early method for operationalizing the distributional hypothesis. These matrices provide a structured way to represent the linguistic data as a two-dimensional grid, where each row corresponds to a word, and each column represents its context.
Formally, the co-occurrences are represented as a matrix $ \mathbf{M} \in \mathbb{R}^{N \times N} $, where $ N = |V| $ is the size of the vocabulary.\footnote{The context need not encompass the entire vocabulary; early distributional semantic models (DSMs) often defined context using a more limited, task-specific subset of words. Moreover, DSMs do not necessarily have to use words as context elements -- contexts can also take the form of structured tuples \cite{pado-lapata-2007-dependency}. For example, in syntax-based DSMs, instead of simply considering words occurring nearby, the context might be a dependency relation paired with a headword.} Each entry $ M_{ij} $ in the matrix captures the frequency, or a function of frequency, with which token $ t_j $ occurs in the context of token $ t_i $. These co-occurrence counts form the basis for computing similarity measures, enabling the extraction of semantic relationships by leveraging shared contexts.

In the context of co-occurrence statistics, \textbf{semantic similarity} provides a valuable framework for comparing different contexts -- a fundamental step in understanding relationships between words and their meanings. Co-occurrence patterns, which reflect how often tokens appear together in a corpus, serve as the basis for quantifying these similarities. By leveraging these statistics, we can represent tokens as vectors that encode their distributional behavior, enabling the comparison of their contextual overlap.

Formally, these vectors exist within an \textit{embedding space}, which can be understood as a Euclidean vector space where each dimension corresponds to a basis vector associated with a contextual feature. In traditional co-occurrence-based models, these bases are often defined by context words, meaning that each word is positioned in the space according to how frequently it appears alongside others. This structured representation allows for measuring word similarity by comparing their vector positions in the embedding space.

One commonly used method to quantify the semantic similarity between tokens based on their co-occurrence statistics is \textbf{cosine similarity}. While there are other possible metrics, cosine similarity specifically measures the degree to which two context vectors align in the embedding space, offering an interpretable way to assess their co-occurrence-based relationships. There are also alternative measures, such as \textit{Jaccard similarity} and \textit{Euclidean distance}, which may be used depending on the application.

\begin{definition}[Cosine Similarity]
Let $t_i$ be a token in a vocabulary $V$, and let $C(t_i)$ represent the set of contexts in which $t_i$ appears within a corpus. According to the distributional hypothesis, ``the meaning of a word is derived from the company it keeps.'' This implies that tokens occurring in similar contexts tend to exhibit semantic similarity.

To express this similarity, we construct a \textbf{context vector} $\mathbf{c}(t_i)$ for each token $t_i$, where each dimension of $\mathbf{c}(t_i)$ corresponds to the frequency or weight of $t_i$ co-occurring with a specific context token. These vectors represent the co-occurrence patterns for each token.

Cosine similarity provides a way to measure the semantic similarity between two tokens $t_i$ and $t_j$ by comparing their respective co-occurrence vectors $\mathbf{c}(t_i)$ and $\mathbf{c}(t_j)$:
\[
\text{sim}(t_i, t_j) = \frac{\langle \mathbf{c}(t_i), \mathbf{c}(t_j) \rangle}{\|\mathbf{c}(t_i)\| \|\mathbf{c}(t_j)\|},
\]
where \( \langle \mathbf{c}(t_i), \mathbf{c}(t_j) \rangle \) represents the inner product of the vectors, while \( \|\mathbf{c}(t_i)\| \) and \( \|\mathbf{c}(t_j)\| \) are their magnitudes. This metric highlights the overlap between the co-occurrence patterns of \( t_i \) and \( t_j \), with higher values indicating greater semantic similarity.
\end{definition}

Despite their strengths, co-occurrence matrices pose significant challenges. Their size and sparsity make them computationally expensive to store and process, particularly for large vocabularies. Moreover, these matrices capture only direct co-occurrence patterns without accounting for higher-order or latent relationships between words. These limitations motivated further advancements in word representation techniques, such as dimensionality reduction and distributed embeddings, which sought to build on the strengths of co-occurrence matrices while addressing their weaknesses.

\subsection{Latent Semantic Analysis}

While co-occurrence matrices effectively capture first-order semantic relationships -- identifying words that appear in similar contexts -- they struggle to reveal higher-order connections between words that do not share direct co-occurrences. \textit{Latent semantic analysis} (LSA) \cite{deerwester-etal-1990-indexing, landauer-etal-1998-introduction} addresses this limitation by applying dimensionality reduction techniques based on linear algebra. Specifically, LSA utilizes \textit{singular value decomposition} (SVD) to project the high-dimensional co-occurrence matrix into a lower-dimensional space, capturing latent semantic structures. Unlike raw co-occurrence counts, which can already group synonyms such as ``house'' and ``home'' based on their shared first-order contexts, LSA uncovers deeper, second-order relationships. For example, it can associate ``house'' with ``mortgage'' or ``children'' with ``education'' even if these words do not co-occur directly but share common contextual patterns across multiple linguistic contexts. By reducing noise and emphasizing meaningful associations, LSA enhances the ability to infer conceptual similarities beyond direct co-occurrence.

\begin{definition}[SVD]
Let $\mathbf{M} \in \mathbb{R}^{N \times N}$ be the co-occurrence matrix, where $N=|V|$ is the size of the vocabulary. SVD decomposes $\mathbf{M}$ as:
\[
\mathbf{M} = \mathbf{U} \mathbf{D} \mathbf{V}^\top ,
\]
where $\mathbf{U}$ and $\mathbf{V}$ are orthogonal matrices, and $ \mathbf{D}$ is a diagonal matrix containing the singular values. By retaining only the top $k$ singular values, we obtain a reduced approximation:
\[
\mathbf{M}_k = \mathbf{U}_k \mathbf{D}_k \mathbf{V}_k^\top .
\]
This reduced matrix $\mathbf{M}_k$ captures the most critical semantic relationships while discarding noise and less significant information. Importantly, the dimensionality $k$ is typically much smaller than $N$, making computations more efficient.
\end{definition}

LSA represented a significant advancement in distributional semantic models (DSMs) by capturing higher-order semantic relationships -- associating words that are not only linked through direct co-occurrence but also through multiple layers of indirect contextual connections. While standard count-based models can already identify words that share first-order contexts, LSA extends this by revealing deeper associations that emerge from shared patterns across diverse contexts. This ability to infer semantic connections beyond direct co-occurrence enhances the expressiveness of word representations. Additionally, by reducing the dimensionality of co-occurrence matrices, LSA mitigates the computational challenges posed by their high dimensionality and sparsity, improving both efficiency and interpretability in semantic comparisons.

A key drawback of LSA is its reliance on linear algebra, as SVD is inherently linear. This restricts LSA’s ability to capture non-linear semantic relationships, which are often essential for modeling complex language patterns. Although LSA mitigates the problem of sparsity by reducing the dimensionality of co-occurrence matrices, it still produces \textit{context-independent} word embeddings, just like standard count-based DSMs. This means that a word is assigned the same representation regardless of its specific usage, preventing LSA from distinguishing between different senses of a word based on context. Furthermore, because LSA operates on a \textit{fixed vocabulary}, it cannot efficiently incorporate new words or adapt to evolving corpora without recomputing the entire co-occurrence matrix and performing SVD again, making it impractical for dynamic and continuously expanding datasets.

Despite these limitations, LSA's advancements laid the groundwork for subsequent developments in word representation, including \textit{neural embeddings} and \textit{contextualized models}, which build on its strengths while addressing its shortcomings.

\section{Neural Distributed Representations}
\label{sec:distributed_reps}

Neural architectures are computational frameworks composed of layers of interconnected nodes, or neurons, designed to process data and learn complex patterns through optimization. Traditionally, neural networks have been viewed as layered structures of nodes that progressively extract hierarchical features from input data. However, a more modern perspective rooted in \textit{functional analysis} frames neural networks as compositions of maps between vector spaces, where each layer applies a learned transformation that projects input representations into increasingly abstract spaces.

\begin{definition}[Neural network]
A neural network is a function \( f: \mathbb{R}^{d_0} \to \mathbb{R}^{d_L} \) that can be expressed as a composition of mappings:
\[
f = f_L \circ f_{L-1} \circ \dots \circ f_1 ,
\]
where each layer \( f_i: \mathbb{R}^{d_{i-1}} \to \mathbb{R}^{d_i} \) is a function, typically of the form:
\[
f_i(\mathbf{x}) = \sigma(W_i \mathbf{x} + \mathbf{b}_i) ,
\]
where \( W_i \in \mathbb{R}^{d_i \times d_{i-1}} \) is a weight matrix, \( \mathbf{b}_i \in \mathbb{R}^{d_i} \) is a bias vector, and \( \sigma \) is a non-linear activation function. 
Here, \( d_0 \) represents the dimensionality of the input space, while \( d_L \) corresponds to the dimensionality of the output space. The network progressively transforms an input \( \mathbf{x} \in \mathbb{R}^{d_0} \) through a sequence of intermediate representations in spaces of dimensions \( d_1, d_2, \dots, d_{L-1} \), ultimately producing an output in \( \mathbb{R}^{d_L} \). 
This formulation highlights how neural networks act as structured function approximators, systematically transforming input data through a sequence of learned linear and non-linear operations.
\end{definition}

The introduction of \textit{neural-based distributed representations} revolutionized NLP by enabling tokens to be represented as dense, low-dimensional vectors in a continuous space. These representations, learned through neural architectures, can leverage \textit{non-linear} transformations to capture complex patterns and are optimized based on the contextual relationships between tokens within large corpora. 
The term \textit{distributed representations} predates modern neural approaches and marks a fundamental shift from symbolic or sparse representations to dense vector-based encodings, where information about a token is distributed across multiple dimensions rather than being tied to a single discrete unit. This contrasts with distributional representations, which rely purely on co-occurrence patterns; distributed representations extend this idea by leveraging neural networks to refine and compress contextual information into compact, expressive vectors. These representations have since become known as \textbf{embeddings}, the standard term for dense vector-based word representations in modern NLP.

\subsection{Word2Vec}

One of the most influential methods for learning distributed representations was Word2Vec, introduced by Mikolov et al.~\cite{mikolov2013distributed}. Word2Vec transformed the way token representations are constructed by introducing a framework that learns dense, low-dimensional embeddings through optimization, rather than relying on co-occurrence counts obtained from a corpus. Its central idea lies in leveraging the distributional properties of tokens in large corpora to train embeddings that capture meaningful relationships based on context.

A key enabler of these advancements is the optimization process itself, grounded in \textbf{gradient descent} and its extension, \textit{backpropagation}. Gradient descent is an iterative algorithm that adjusts the parameters of a neural network to minimize a predefined loss function, which quantifies the error between the model’s predictions and the true outputs. Backpropagation facilitates this process by efficiently computing gradients for all network parameters through the chain rule of calculus, propagating the error backward from the output layer to the input layer. This combination allows neural architectures to fine-tune their parameters to learn meaningful patterns and relationships.

The power of neural distributed representations lies not only in their ability to encode complex relationships but also in the generality of gradient descent-based optimization, which underpins the training of virtually all modern neural language models. By iteratively refining the embeddings through exposure to large corpora, these models learn representations that are both flexible and scalable, enabling breakthroughs across a wide range of NLP tasks.

Word2Vec introduced two neural models for learning embeddings: the \textit{skip-gram} model and the \textit{continuous bag-of-words} model. While both models use neural architectures to learn embeddings, they differ in how they define the prediction task, as detailed below.

\begin{definition}[Skip-Gram]
The skip-gram model focuses on predicting the context words surrounding a target token. Given a target token \( t_i \), the model aims to maximize the probability of its context tokens \( t_{i-k}, \ldots, t_{i-1}, t_{i+1}, \ldots, t_{i+k} \), where \( k \) is the size of the context window. Skip-gram trains embeddings by maximizing the likelihood of context words given the target token:
\[
P(t_{i-k}, \ldots, t_{i-1}, t_{i+1}, \ldots, t_{i+k} | t_i) = \prod_{\substack{-k \leq j \leq k \\ j \neq 0}} P(t_{i+j} | t_i) .
\]
\end{definition}

\begin{figure}[]
    \centering
    \begin{subfigure}{\textwidth}
        \centering
        \includegraphics[width=.6\textwidth]{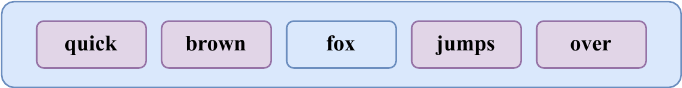}
        \caption[Skip-gram]{Skip-gram: predicting the context words (purple) surrounding the target word ``fox'' (blue) within a given window.}
        \label{fig:skipgram}
    \end{subfigure}
    \hfill
    \begin{subfigure}{\textwidth}
        \centering
        \includegraphics[width=.6\textwidth]{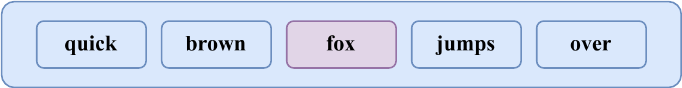}
        \caption{CBOW: predicting the target word ``fox'' (purple) based on the surrounding context words (blue).}
        \label{fig:cbow}
    \end{subfigure}
    \caption[Word2Vec]{Illustration of the Word2Vec training objectives.}
    \label{fig:word2vec}
\end{figure}

By learning to predict the surrounding words, the skip-gram model captures the relationships between tokens based on their co-occurrence patterns. These embeddings encode semantic similarity, as tokens that frequently appear in similar contexts are mapped to close points in the embedding space. Figure \ref{fig:skipgram} illustrates the skip-gram task, where context words (purple) are predicted from a target word (blue).

\begin{definition}[Continuous Bag-of-Words (CBOW)]
In contrast to Skip-Gram, the CBOW model reverses the task, predicting the target token based on its surrounding context tokens. Given context words \( t_{i-k}, \ldots, t_{i-1}, t_{i+1}, \ldots, t_{i+k} \), the CBOW model optimizes the probability of predicting the target token \( t_i \):
\[
P(t_i | t_{i-k}, \ldots, t_{i-1}, t_{i+1}, \ldots, t_{i+k}) .
\]
\end{definition}

CBOW effectively learns embeddings by maximizing the likelihood of a target word given its context. This approach averages the context embeddings, making it computationally efficient and more robust when dealing with noisy data. Figure~\ref{fig:cbow} shows the CBOW task, where the target word (purple) is predicted from context words (blue).

While both skip-gram and CBOW share the same objective of learning word embeddings based on context, they differ in their approach to training. The primary distinction lies in their \textit{prediction direction}: skip-gram predicts surrounding context words given a target word, whereas CBOW predicts the target word based on its surrounding context.  
These models also differ in their focus. Skip-gram is particularly effective for learning representations of rare words and handling smaller datasets because it treats each word-context pair individually, allowing it to capture nuanced relationships. In contrast, CBOW averages the embeddings of context words before making predictions, which makes it more efficient but also more biased toward frequent words. From a computational perspective, CBOW is generally more efficient than skip-gram. By aggregating context word embeddings before prediction, CBOW reduces the number of computations required, making it faster to train. Skip-gram, on the other hand, processes each word-context pair separately, leading to longer training times but often yielding richer embeddings, especially for low-frequency words.

\begin{figure}[]
    \centering
    \includegraphics[width=0.6\textwidth]{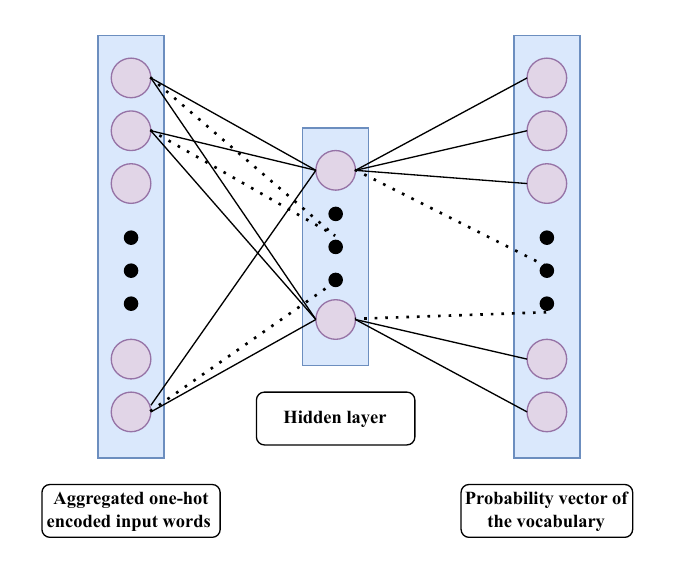}
    \caption[CBOW architecture]{The neural network architecture for CBOW. Context words (one-hot encoded) are projected into dense vectors using a shared weight matrix. These vectors are averaged to create a context representation, which serves as input for predicting the target word.}
    \label{fig:cbow_arch}
\end{figure}

Both models are implemented using a simple, single-layer neural network trained on large corpora. Figure \ref{fig:cbow_arch} illustrates the architecture for CBOW. In this setup, one-hot encoded context words are transformed into dense vectors through a shared weight matrix. These vectors are averaged to form a single context representation, which is then used to predict the target word.

\subsection{GloVe}

GloVe (Global Vectors for Word Representation) \cite{pennington2014glove} extended the idea of distributed representations by incorporating global co-occurrence information directly into the learning process. While Word2Vec relies on predicting word-context pairs, GloVe constructs word embeddings by factorizing a co-occurrence matrix. This method balances the benefits of co-occurrence-based approaches with the efficiency and power of distributed representations.

The GloVe model defines the relationship between token vectors $\mathbf{t}_i$ and $\mathbf{t}_j$ such that their dot product approximates the logarithm of their co-occurrence count:
\[
\mathbf{t}_i^T \mathbf{t}_j \approx \log(M_{ij}) ,
\]
where $M_{ij}$ is the number of times token $t_i$ co-occurs with token $t_j$. By leveraging both local and global co-occurrence information, GloVe embeddings capture both semantic and syntactic relationships between tokens.

GloVe's approach offered a further advancement in distributed representations by directly tying the learned embeddings to observable statistics in the corpus, offering a more explicit connection between token co-occurrences and their geometric representations in the embedding space.

\subsection{Benefits of Distributed Representations}  

The introduction of distributed representations in models such as Word2Vec and GloVe brought several key advantages over earlier word representation approaches. One of the most significant benefits is their ability to create \textbf{dense and low-dimensional representations} directly during training. While methods like LSA also produce lower-dimensional representations, they first require constructing and factorizing a large co-occurrence matrix, which is computationally expensive in terms of both space and time. In contrast, neural-based models like Word2Vec and GloVe learn embeddings efficiently without the need for explicit matrix factorization. These models map words into a significantly lower-dimensional space, typically with only $100$ to $300$ dimensions, reducing computational complexity while preserving essential linguistic information. This direct optimization makes neural embeddings both more scalable and expressive compared to traditional count-based DSMs.

Another key property of word embeddings is that \textit{semantic relationships are encoded in the geometry of the embedding space}. Words that share similar meanings or occur in similar contexts tend to be positioned closer together, while unrelated words are farther apart. This geometric structure naturally reflects linguistic relationships and enables simple vector operations to capture meaningful associations between words. This phenomenon is formalized as the \textit{linear representation hypothesis} \cite{mikolov-etal-2013-efficient}, which suggests that relationships between words can often be expressed through linear transformations in the embedding space.

\begin{example}[Analogy in the Geometric Space]
A well-known example demonstrating the structured nature of word embeddings is the analogy:  
\[
\text{vec}(\text{king}) - \text{vec}(\text{man}) + \text{vec}(\text{woman}) \approx \text{vec}(\text{queen}).
\]  
This example illustrates that word embeddings encode more than mere similarity; they capture relational semantics, allowing analogical reasoning through vector arithmetic. This property makes distributed representations not only useful for measuring similarity but also for uncovering deeper relationships within language.
\end{example}

A further advantage of distributed representations is their role in \textit{efficient pre-training for downstream tasks}. Unlike traditional approaches such as LSA, which require constructing and factorizing a global co-occurrence matrix, neural embeddings are learned directly by optimizing on raw text corpora. Models like Word2Vec and GloVe adjust word vectors to minimize a training objective that reflects meaningful word relationships, allowing embeddings to be produced as a byproduct of an optimization process rather than through explicit matrix decomposition. This enables embeddings to capture general semantic and syntactic properties of language more efficiently. Instead of learning word meanings from scratch for each task, pre-trained embeddings can be transferred to various NLP applications, such as text classification, named entity recognition, and machine translation, often requiring only minimal fine-tuning. This transferability significantly reduces training time while improving model performance, particularly for tasks with limited labeled data.

In summary, distributed representations embed words into vector spaces where semantic relationships are encoded as distances. This capability serves as the foundation for integrating words into machine learning models, enabling them to generalize better, learn faster, and capture intricate linguistic properties. By leveraging these structured representations, modern NLP models can move beyond simple token memorization to a deeper, more contextual understanding of language.

\section{Language Models}

While distributed word representations have significantly advanced NLP, a fundamental limitation of traditional word embeddings is their \textit{static} nature. These embeddings assign a single vector to each word, regardless of its context, which can lead to ambiguities. 
A natural solution to this problem is to \textit{contextualize} word embeddings -- allowing their representations to dynamically adjust based on surrounding words. This shift essentially involves modeling \textit{semantic composition}, where meaning is derived from how words interact within a sequence. The key breakthrough enabling contextualization has been the development of \textbf{language models} (LMs), which have been a cornerstone of NLP since its early days.

At their core, language models are probabilistic models of token sequences. They estimate the likelihood of a sequence of words, enabling models to predict the next word in a sentence or evaluate the plausibility of a given text. Importantly, modern neural LMs naturally produce \textit{contextualized embeddings} as a byproduct, where a word’s representation is dynamically influenced by its surrounding context, overcoming the limitations of static word embeddings.

\begin{definition}[Language Model]
For a sequence of tokens \( \{t_1, t_2, \ldots, t_T\} \), a \textit{language model} assigns a probability to the sequence by modeling the probability of each token given its preceding tokens:
\[
P(t_1, t_2, \ldots, t_T) = \prod_{i=1}^{T} P(t_i | t_1, \ldots, t_{i-1}) .
\]
\end{definition}

The objective of a language model is to approximate this probability distribution as accurately as possible. During training, this is typically achieved by maximizing the likelihood of observed sequences. In the following sections, we will explore the evolution of language modeling techniques, including $n$-grams, recurrent neural networks, and transformers.

Before the advent of neural methods, early statistical approaches to language modeling were dominated by \( n \)-gram models \cite{brown-1992-class}. These models estimate the probability of a word based on a fixed-size window of \( n-1 \) preceding words, simplifying the computation of \( P(t_i | t_1, \ldots, t_{i-1}) \) to:
\[
P(t_i | t_1, \ldots, t_{i-1}) \approx P(t_i | t_{i-(n-1)}, \ldots, t_{i-1}) .
\]

While \( n \)-gram models are simple and computationally efficient, they come with notable limitations that hinder their effectiveness in large-scale NLP applications. One major challenge is \textit{sparsity}. As \( n \) increases, the number of possible \( n \)-grams grows exponentially, making it difficult to estimate probabilities for rare or previously unseen sequences. This issue is particularly problematic in large vocabularies, where many valid sequences may never appear in the training data, leading to poor generalization. Various smoothing techniques have been introduced to mitigate this problem, but they only offer partial solutions.

Another key limitation is the \textit{fixed context window}. Since \( n \)-gram models condition each token on a limited number of preceding tokens, they struggle to capture long-range dependencies and broader contextual information. This constraint makes them less effective in modeling complex linguistic structures, where meaning often depends on relationships spanning entire sentences or even paragraphs.

While \( n \)-gram models played a foundational role in statistical NLP, their limitations prompted the search for more flexible sequence modeling approaches. However, the transition to neural methods was not driven solely by the inadequacies of \( n \)-grams. Instead, it emerged as a consequence of advancements in distributed word representations and neural architectures that made it possible to model sequences in a continuous vector space.

\section{Neural Networks for Sequence Modeling}

A pivotal step in the transition to neural approaches in sequence modeling was the introduction of \textit{distributed word representations} through models like Word2Vec, which learned dense embeddings that captured semantic relationships between words. While these embeddings were highly effective for static word representations, they lacked the ability to encode sequences of words in context. The natural next step was to extend these ideas to dynamic sequence modeling, leading to the integration of neural networks with distributed representations.

\subsection{Recurrent Neural Networks}

Recurrent Neural Networks (RNNs) were among the first neural architectures designed to process sequential data. Initially proposed in the 1980s \cite{rumelhart-etal-1986-learning, jordan-1986-serial}, RNNs introduced a mechanism for maintaining a hidden state that carries information across time steps, allowing models to process variable-length sequences. However, early adoption of RNNs was limited due to computational constraints and issues such as the \textit{vanishing gradient problem}, which hindered their ability to learn long-range dependencies.  

The widespread adoption of RNNs in NLP became feasible only in the 2010s, fueled by key advancements such as the introduction of word embeddings like Word2Vec, improved hardware acceleration (e.g., GPUs), enhanced optimization techniques (e.g., gradient clipping and more effective activation functions), and access to large-scale datasets. These developments allowed deeper architectures to be trained efficiently, making RNNs practical for tasks like machine translation and text generation.

\begin{definition}[RNN]
Formally, at each time step $t$, the hidden state $\mathbf{h}_t$ in an RNN is computed as
\[
\mathbf{h}_t = f(\mathbf{W}_h \mathbf{h}_{t-1} + \mathbf{W}_x \mathbf{x}_t + \mathbf{b}_h),
\]
where:
\begin{itemize}
    \item $t$ represents a time step in the sequence, corresponding to the processing of the token $\mathbf{x}_t$ in an input sequence;
    \item $\mathbf{x}_t$ is the representation of the input token at time step $t$;
    \item $f$ is a non-linear activation function, typically $\tanh$ or $\text{ReLU}$, which introduces non-linearity into the model, enabling it to learn complex patterns in sequential data;
    \item $\mathbf{W}_h$ and $\mathbf{W}_x$ are weight matrices for the hidden state and the input representation $\mathbf{x}_t$, respectively;
    \item $\mathbf{b}_h$ is a bias term.
\end{itemize}
\end{definition}

\begin{figure}[h!]
    \centering
    \includegraphics[width=0.3\textwidth]{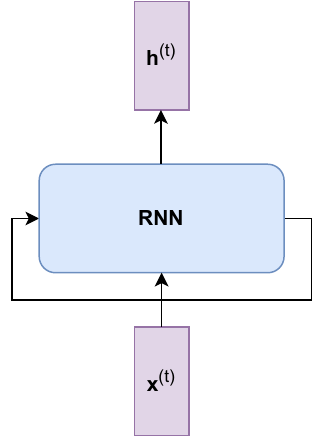}
    \caption[Recurrent structure of an RNN]{Illustration of the recurrent structure of an RNN. At each time step $t$, the hidden state $\mathbf{h}_t$ is computed based on the previous hidden state $\mathbf{h}_{t-1}$ and the input representation $\mathbf{x}_t$. The figure highlights how the recurrence allows contextual information to flow across the sequence, enabling the RNN to process temporal dependencies.}
    \label{fig:rnn}
\end{figure}

The recurrent nature of RNNs, as illustrated in \Cref{fig:rnn}, enables them to process temporal dependencies by leveraging the hidden state $\mathbf{h}_t$, which encapsulates information from both the current input $\mathbf{x}_t$ and the preceding states. This ability to carry context across time steps is what makes RNNs particularly well-suited for sequence-based tasks.

The hidden state $\mathbf{h}_t$ serves as a contextual representation of the sequence up to time $t$. The output probability distribution for the next token is computed as
\[
P(t_{t+1} | t_1, \ldots, t_t) = \softmax(\mathbf{W}_y \mathbf{h}_t + \mathbf{b}_y),
\]
where $\mathbf{W}_y$ and $\mathbf{b}_y$ are the output weight matrix and bias, respectively. Here, the $\softmax$ function is used to convert the raw scores (logits) from $\mathbf{W}_y \mathbf{h}_t + \mathbf{b}_y$ into a probability distribution over the vocabulary. Given a vector of logits $\mathbf{z} = [z_1, z_2, \ldots, z_N]$, the softmax function is defined as:
\[
\softmax(z_i) = \frac{e^{z_i}}{\sum_{j=1}^N e^{z_j}}, \quad \text{for } i = 1, \ldots, N,
\]
where $e^{z_i}$ is the exponential of the $i$-th logit, and the denominator normalizes the values so that they are non-negative and sum to 1, ensuring that the output represents a valid probability distribution. The $\softmax$ function is essential for tasks like language modeling, as it allows the model to predict the probability of each token in the vocabulary based on the given context.

\begin{figure}[h!]
    \centering
    \includegraphics[width=\textwidth]{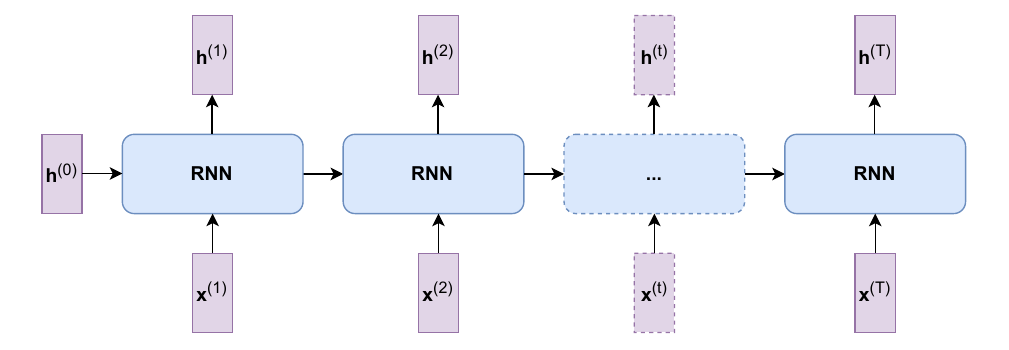}
    \caption[Unrolled RNN]{Unrolled RNN}
    \label{fig:rnn_unrolled}
\end{figure}

\Cref{fig:rnn_unrolled} illustrates the \textit{unrolling} of an RNN across time steps. In its standard form, an RNN is often represented as a compact structure where each hidden state recursively depends on the previous one. However, to better visualize its sequential nature, the network can be unrolled over time, explicitly showing how computations progress step by step. At each step, the hidden state $\mathbf{h}_t$ is updated based on the current input $\mathbf{x}_t$ and the previous hidden state $\mathbf{h}_{t-1}$, enabling the model to retain context from earlier tokens and propagate information through the sequence.

\subsubsection{Advantages and Limitations of RNNs}  

One of the key strengths of RNNs is their ability to process sequences of arbitrary length, making them particularly well-suited for tasks such as time-series analysis and NLP. Unlike traditional feedforward architectures that handle inputs independently, these networks maintain a hidden state that carries information across time steps, allowing them to model dependencies within sequences. Another advantage is their use of \textit{shared parameters} across time steps. Instead of learning separate weights for each input position, the same parameters are reused throughout the sequence. This parameter-sharing mechanism reduces the overall complexity of the model, making it more efficient to train while improving generalization. Additionally, these networks enable \textit{contextual learning} by leveraging hidden states to encode information from previous time steps. This dynamic representation of sequential data marks a significant improvement over static models that treat each token in isolation. By maintaining an evolving internal state, these networks can recognize patterns and relationships across sequences, which is essential for applications such as speech recognition, machine translation, and text generation.  

While these advancements made recurrent architectures powerful tools for sequence modeling, they also introduced new challenges, particularly in maintaining long-range dependencies. One major issue is the \textit{vanishing gradient problem}, where gradients diminish exponentially as they propagate backward through time. This makes it difficult for the network to retain information from distant tokens, leading to weak long-term dependencies. Conversely, these models also suffer from the \textit{exploding gradient problem}, where gradients become excessively large during backpropagation, causing instability in training. While techniques such as gradient clipping help mitigate this issue, it remains a challenge when training deep recurrent architectures. Another fundamental limitation is \textit{limited context retention}. Since each hidden state is updated at every time step, older information is gradually overwritten, making it difficult to retain long-range dependencies. This shortcoming significantly impacts performance on tasks that require understanding relationships between words or events that are far apart in the sequence.  

To address these challenges, more advanced architectures were developed, such as Long Short-Term Memory networks, which introduced gating mechanisms to regulate information flow and improve long-term memory retention.

\subsection{Long Short-Term Memory}

Long Short-Term Memory (LSTM) networks, introduced by Hochreiter and Schmidhuber in 1997 \cite{hochreiter1997long}, were specifically designed to address the shortcomings of vanilla RNNs. While the idea of LSTMs originated in the late 1990s, their practical adoption became widespread only in the 2010s, thanks to improved hardware and training techniques. LSTMs introduced gating mechanisms that effectively regulate the flow of information, enabling the model to retain long-term dependencies and mitigate the vanishing gradient problem.

\begin{definition}[LSTM Cell]
An LSTM cell consists of three gates -- \textit{input}, \textit{forget}, and \textit{output} -- and a cell state $\mathbf{c}_t$, which serves as a memory buffer. At each time step $t$, the following computations are performed:
\[
\begin{aligned}
\mathbf{f}_t &= \sigma(\mathbf{W}_f [\mathbf{h}_{t-1}, \mathbf{x}_t] + \mathbf{b}_f) & \quad & \text{(forget gate)} \\
\mathbf{i}_t &= \sigma(\mathbf{W}_i [\mathbf{h}_{t-1}, \mathbf{x}_t] + \mathbf{b}_i) & \quad & \text{(input gate)} \\
\mathbf{g}_t &= \tanh(\mathbf{W}_c [\mathbf{h}_{t-1}, \mathbf{x}_t] + \mathbf{b}_c) & \quad & \text{(cell state candidate)} \\
\mathbf{c}_t &= \mathbf{f}_t \odot \mathbf{c}_{t-1} + \mathbf{i}_t \odot \mathbf{g}_t & \quad & \text{(cell state update)} \\
\mathbf{o}_t &= \sigma(\mathbf{W}_o [\mathbf{h}_{t-1}, \mathbf{x}_t] + \mathbf{b}_o) & \quad & \text{(output gate)} \\
\mathbf{h}_t &= \mathbf{o}_t \odot \tanh(\mathbf{c}_t) & \quad & \text{(hidden state update)}
\end{aligned}
\]
Here:
\begin{itemize}
    \item $\sigma$ is the sigmoid activation function;
    \item $\odot$ denotes element-wise multiplication;
    \item $\mathbf{W}_f, \mathbf{W}_i, \mathbf{W}_c, \mathbf{W}_o$ and $\mathbf{b}_f, \mathbf{b}_i, \mathbf{b}_c, \mathbf{b}_o$ are weight matrices and bias vectors for the gates and cell updates;
    \item $\mathbf{x}_t$ is the input vector at time step $t$.
\end{itemize}
\end{definition}

\begin{figure}[h!]
    \centering
    \includegraphics[width=0.7\textwidth]{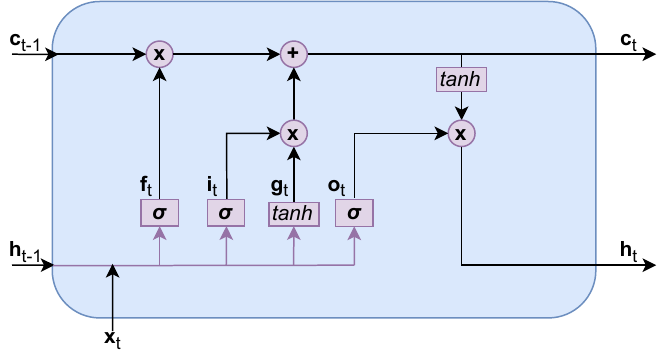}
    \caption[Structure of an LSTM cell]{Structure of an LSTM cell}
    \label{fig:lstm}
\end{figure}

Figure~\ref{fig:lstm} illustrates the internal structure of an LSTM cell. The gating mechanism allows LSTMs to selectively retain relevant information and discard unnecessary data, enabling them to model long-term dependencies more effectively. Specifically, the forget gate $\mathbf{f}_t$ determines which parts of the previous cell state $\mathbf{c}_{t-1}$ to retain. The input gate $\mathbf{i}_t$ and cell candidate $\mathbf{g}_t$ work together to update the cell state $\mathbf{c}_t$ with new information. Finally, the output gate $\mathbf{o}_t$ controls how much of the updated cell state is passed to the hidden state $\mathbf{h}_t$, which serves as the output for the current time step.

\subsubsection{Advantages and Limitations of LSTMs}  

LSTMs introduced several key improvements over standard recurrent architectures, addressing some of the major limitations associated with modeling long-term dependencies. One of their most significant advantages is \textit{improved memory retention}. By incorporating a dedicated cell state, LSTMs provide a direct gradient path, effectively mitigating the vanishing gradient problem and allowing information to persist over extended sequences. Another major benefit of LSTMs is their ability to \textit{selectively control information flow} through gating mechanisms. The input, forget, and output gates enable the model to regulate which information should be stored, updated, or discarded at each time step. This selective filtering allows LSTMs to focus on relevant parts of the sequence while ignoring less important details, improving both efficiency and accuracy in sequential tasks. LSTMs also offer \textit{flexible sequence handling}, making them well-suited for processing variable-length sequences. Their architecture enables them to capture long-range dependencies more effectively than standard recurrent models, making them valuable for applications such as speech recognition, machine translation, and text generation.  

Despite these advantages, LSTMs still face several challenges. One of the main drawbacks is their \textit{high computational cost}. The inclusion of multiple gating mechanisms significantly increases the number of parameters in the model, leading to greater memory requirements and longer training times. Additionally, while LSTMs are better than simple recurrent models at handling long-range dependencies, they may still struggle when sequences span very long contexts. Retaining meaningful information across hundreds of time steps remains a challenge, particularly when the input sequence contains complex dependencies. Another limitation is \textit{sequential processing}. Like traditional recurrent models, LSTMs process input step by step, making them inherently difficult to parallelize. This sequential nature increases computational overhead, making training and inference slower compared to more parallelizable architectures such as transformers.  

To address some of these challenges, the Gated Recurrent Unit (GRU) was introduced by Cho et al.~\cite{cho2014learning} as a simplified alternative to LSTMs. GRUs reduce computational complexity by combining the forget and input gates into a single \textit{update gate} and eliminating the explicit cell state. Instead, they rely solely on a hidden state to manage information. This reduction in complexity results in fewer parameters and faster training times, while still retaining the ability to model long-term dependencies effectively. As a result, GRUs offer a practical alternative in applications where computational efficiency is a priority.

\subsection{Transformers}

The next major breakthrough in language modeling came with the introduction of the \textbf{Transformer architecture} \cite{vaswani2017attention}. Unlike RNNs, which process tokens sequentially, transformers operate on entire sequences in parallel. They achieve this through \textbf{attention}, a mechanism that enables the model to dynamically focus on the most relevant tokens, regardless of their position in the sequence. This architecture not only improves computational efficiency but also enhances the model’s ability to capture both local and long-range dependencies, making it particularly well-suited for large-scale NLP tasks.

\subsubsection{The Attention Mechanism}

\begin{figure}[]
    \centering
    \includegraphics[width=0.3\textwidth]{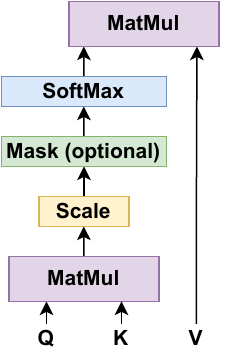}
    \caption[Scaled dot-product attention]{Schematics of the scaled dot-product attention. This diagram illustrates the steps of the scaled dot-product attention mechanism. The inputs consist of three components: queries ($\mathbf{Q}$), keys ($\mathbf{K}$), and values ($\mathbf{V}$). The operation begins with matrix multiplication (\textbf{MatMul}) between $\mathbf{Q}$ and $\mathbf{K}^\top$, producing a similarity score matrix. This score is then scaled to avoid overly large values and passed through an optional \textbf{Mask}, which is used to control visibility in tasks like autoregressive decoding by blocking future tokens. The result is normalized using a \textbf{SoftMax} function to generate attention weights, which are finally applied to $\mathbf{V}$ via another matrix multiplication to produce the output.}
    \label{fig:attention}
\end{figure}

The attention mechanism is a cornerstone of transformer models, enabling their exceptional performance across various NLP tasks. One widely used variant is \textbf{scaled dot-product attention}, which computes the relationship between each token in the sequence and other tokens by generating attention scores that determine how much influence each token has in the final representation. It is formally defined as
\[
\text{Attention}(Q, K, V) = \softmax\left(\frac{QK^T}{\sqrt{d_k}}\right) V ,
\]
where \(Q\), \(K\), and \(V\) represent the queries, keys, and values derived from the input embeddings, and \(d_k\) is the dimensionality of the key vectors. The scaling factor \( \sqrt{d_k} \) is introduced to prevent excessively large dot-product values, which can lead to sharp gradients and instability in training. Figure \ref{fig:attention} illustrates the attention components.

The terminology \emph{queries}, \emph{keys}, and \emph{values} originates from information retrieval systems. In such systems, a \emph{query} represents the search term or request, \emph{keys} represent potential matches in a database, and \emph{values} are the associated information or content retrieved. In the context of attention, each token generates its own query, key, and value representations. The query interacts with all keys in the sequence to compute similarity scores, which are then normalized using the \(\softmax\) function. These scores determine how much weight each value contributes to the final representation. This framework enables the model to capture nuanced relationships between tokens in a structured and interpretable way.

While scaled dot-product attention is the most common formulation in modern transformer architectures, other attention mechanisms exist. For instance, \textbf{additive attention} \cite{bahdanau2014neural} computes attention scores using a feedforward network instead of a dot-product operation, which can be beneficial in certain sequence-to-sequence tasks. Other variations, such as \textbf{local attention} \cite{luong2015effective}, restrict the receptive field to a limited window of tokens rather than attending to the entire sequence, making it computationally more efficient in long documents.

A key variant of scaled dot-product attention used in many transformer architectures is \textbf{self-attention}, where attention is computed entirely within a single sequence -- each token attends only to other tokens in the same input. This enables models to efficiently process entire sequences in parallel while modeling intra-sequence dependencies.

Another important variant is \textbf{cross-attention}, where queries come from one sequence while keys and values are derived from another. Cross-attention is commonly used in sequence-to-sequence architectures, such as BART and T5, where an encoder generates key-value representations that a decoder attends to when producing an output sequence. Similarly, \textit{bi-attention} allows two sequences to attend to each other bidirectionally, a mechanism often used in question-answering and natural language inference tasks.

\subsubsection{The Transformer Architecture}

\begin{figure}[]
    \centering
    \includegraphics[width=0.65\textwidth]{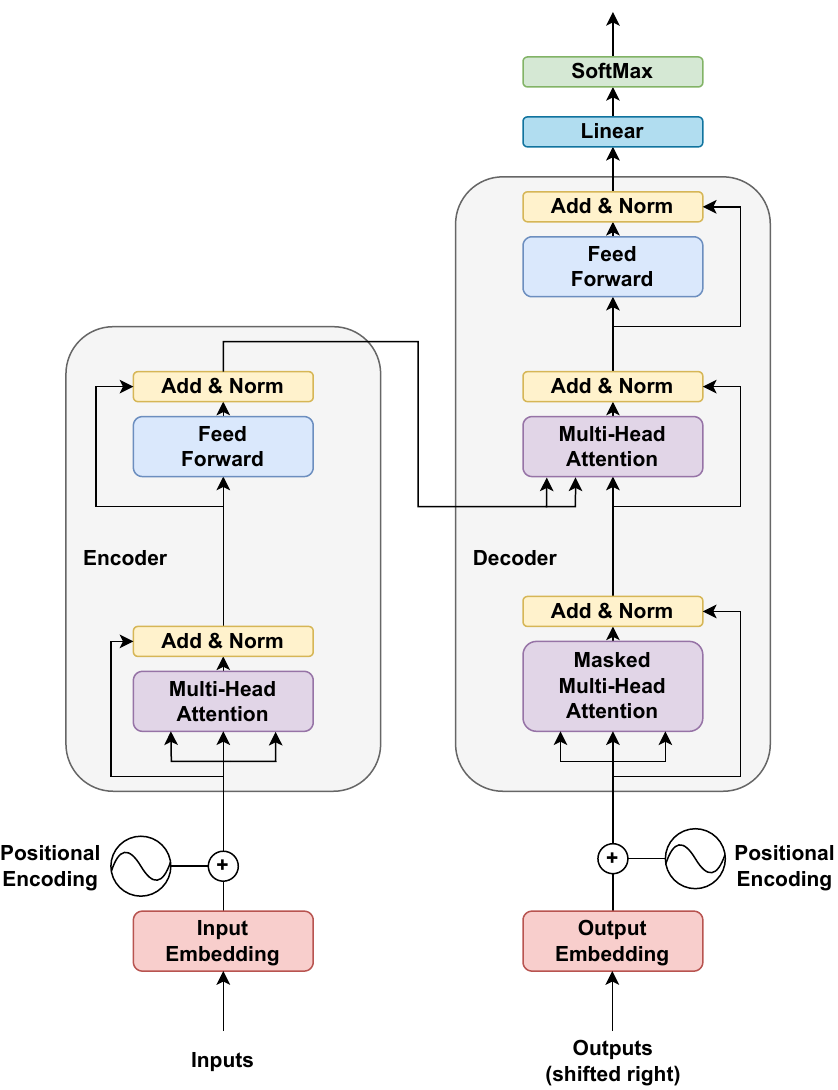}
    \caption[Transformer architecture]{Transformer architecture}
    \label{fig:transformer}
\end{figure}

\begin{figure}[]
    \centering
    \includegraphics[width=0.4\textwidth]{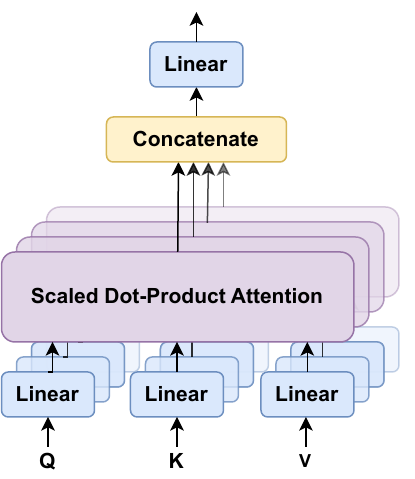}
    \caption[Multi-head attention]{Multi-head attention}
    \label{fig:mha}
\end{figure}

The Transformer architecture (Figure \ref{fig:transformer}) was originally introduced as an encoder-decoder model for machine translation. In this framework, the encoder processes the input sequence in the source language, while the decoder generates the output sequence in the target language. Over time, researchers adapted the transformer architecture for various tasks by using only the encoder or the decoder. Encoders are particularly effective for generating contextualized embeddings, while decoders are well-suited for autoregressive modeling, such as text generation. Both encoder and decoder components consist of the following key features:

\begin{itemize}
    \item \textbf{Multi-head self-attention} (Figure \ref{fig:mha}): Extends the self-attention mechanism by using multiple attention heads, each learning different aspects of token relationships. This is achieved by splitting $Q$, $K$, and $V$ into smaller subspaces and concatenating their outputs. Multi-head attention allows the model to capture diverse types of dependencies between tokens, enhancing its ability to represent complex relationships;
    
    \item \textbf{Feedforward layers}: Position-wise feedforward layers apply non-linear transformations to the attention outputs. These two-layer neural networks operate independently on each token position, enriching the model's expressive power and enabling it to capture complex patterns in the data;
    
    \item \textbf{Positional encoding}: Since transformers process all tokens in parallel, they lack the inherent temporal information present in models like RNNs. To compensate, positional encodings are added to the input embeddings, providing the model with a sense of order. These encodings are often computed using sine and cosine functions at different frequencies, ensuring the positional information is continuous and can be generalized across different input lengths;
    
    \item \textbf{Layer normalization and residual connections}: These components stabilize training by reducing the internal covariate shift and improving the flow of gradients through the network. Residual connections allow the model to maintain information from earlier layers, while layer normalization ensures that inputs to subsequent layers remain well-scaled.
\end{itemize}

In the decoder specifically, several modifications are introduced to enable \textit{autoregressive generation} -- a sequential process in which the model generates one token at a time, using previously generated tokens as context for predicting the next. These modifications ensure that the decoder adheres to a left-to-right generation framework, maintaining coherence while preventing unintended information leakage during training:

\begin{itemize}
    \item \textbf{Masked multi-head self-attention}: In the decoder's first multi-head attention layer, a causal mask is applied to ensure that each token can only attend to itself and the preceding tokens. This prevents the model from ``peeking'' at future tokens during training, maintaining the autoregressive property required for text generation;

    \item \textbf{Shifted outputs}: During training, the target sequence is shifted to the right before being fed into the decoder. This ensures that the decoder predicts the next token based only on previous tokens;
    
    \item \textbf{Cross-attention with encoder outputs}: The decoder's second multi-head attention layer attends to the encoder outputs, enabling the model to integrate contextual information from the source sequence. The encoder outputs act as $K$ and $V$, while the decoder hidden states serve as $Q$;

    \item \textbf{Autoregressive decoding}: During inference, the decoder generates tokens sequentially, predicting one token at a time based on previously generated tokens and encoder outputs (if applicable). This step-by-step process enables the generation of coherent and contextually appropriate sequences.
\end{itemize}

\subsection{Pre-trained Language Models}
\label{sec:plms}

A major breakthrough in the evolution of NLMs was the realization that large-scale \textbf{pre-training} could serve as a foundation for learning general-purpose linguistic representations. Rather than training a model from scratch for each individual task, models first go through a pre-training phase on large corpora using self-supervised learning objectives. This phase equips them with a broad understanding of language.

Pre-training typically follows one of two core self-supervised objectives:  
\begin{itemize}
    \item \textbf{Masked Language Modeling (MLM)}: A bidirectional training approach in which random tokens within a sequence are masked, and the model is trained to predict them based on the surrounding context. This enables the model to learn deep contextual dependencies;
    \item \textbf{Causal Language Modeling (CLM)}: An autoregressive framework in which the model learns to predict the next token in a sequence, given only the preceding tokens.
\end{itemize}

Following pre-training, the model proceeds through \textbf{fine-tuning}, where it is further trained on a smaller, task-specific dataset. This phase allows the model to specialize in various NLP applications. Since the pre-trained model has already learned rich linguistic patterns, fine-tuning requires significantly less labeled data compared to training a model from scratch.

Fine-tuning is accomplished by initializing the pre-trained model weights and optimizing them using supervised learning on task-specific labeled data. The optimization process typically relies on SGD (stochastic gradient descent) \cite{robbins1951stochastic}, a fundamental algorithm that updates model parameters iteratively by computing gradients over mini-batches. The training process is organized into \textbf{epochs}, where an epoch represents one complete pass through the entire training dataset. During each epoch, the model iteratively refines its parameters, gradually improving its ability to generalize to unseen data. However, a key challenge in using SGD is the choice of \textit{learning rate}, which determines the step size for updating model parameters and significantly impacts training stability and convergence speed. Unlike more advanced optimizers, basic SGD does not incorporate \textit{momentum}, a mechanism that smooths gradient updates and accelerates optimization by accumulating past gradients. To address these challenges, more advanced optimizers such as Adam (Adaptive Moment Estimation) \cite{kingma-ba-2015-adam} have been introduced. Adam improves upon SGD by incorporating momentum and adaptive learning rates, allowing for more efficient and stable training. A further refinement, AdamW \cite{loshchilov-and-hutter-2018-decoupled}, extends Adam by introducing weight decay as a separate regularization term, enhancing generalization and training stability. Since the choice of learning rate is critical in fine-tuning, techniques such as learning rate warm-up, which gradually increases the learning rate at the start of training, and gradient clipping, which constrains gradient magnitudes to prevent exploding values, are commonly employed to ensure smooth and stable convergence.

This two-phase paradigm led to the emergence of \textbf{pre-trained language models} (PLMs) \cite{qiu-2020-pre}, which transformed NLP by leveraging vast amounts of unlabeled text for pre-training, followed by targeted fine-tuning with minimal supervision. The success of PLMs demonstrated that this approach significantly outperforms earlier task-specific methods, driving state-of-the-art advancements in modern NLP.

Beyond standard fine-tuning, an additional adaptation step known as \textbf{task-adaptive pre-training (TAPT)} has been shown to further enhance PLM performance on downstream tasks \cite{gururangan-etal-2020-dont}. TAPT involves continuing the pre-training phase using a domain- or task-specific unlabeled corpus before fine-tuning on labeled data. By exposing the model to text more closely aligned with the final task, TAPT helps bridge the gap between general pretraining and task-specific fine-tuning, allowing the model to better capture domain-relevant linguistic structures and terminology.

Unlike standard fine-tuning, which is performed on relatively small labeled datasets, TAPT leverages large amounts of unlabeled data, making it particularly useful in low-resource scenarios or when adapting to specialized domains. This intermediate pre-training step enhances model robustness and reduces the risk of catastrophic forgetting, ultimately leading to improved generalization and performance on the target task.

\subsection{Transformer-Based Language Models}  

The flexibility and scalability of transformers have made them the foundation for modern PLMs. While architectures such as RNNs, LSTMs, and transformers are primarily designed for sequence processing, they become language models when trained on a corpus using objectives derived from language modeling. Training with tasks such as next-word prediction or masked language modeling enables these models to learn statistical patterns and contextual dependencies within the text, allowing them to generate meaningful and coherent outputs.

Decoder-based models are particularly well-suited for language modeling, as they are trained to predict the next word given a sequence of previous words, making their role as language models straightforward. However, the same principle applies to encoder-decoder architectures, which generate text conditioned on an input sequence, and encoder-only models, which capture deep contextual representations through self-supervised objectives such as masked language modeling. By leveraging large-scale pre-training on diverse textual data, PLMs develop general-purpose language representations that can be fine-tuned for a wide range of downstream tasks.

An essential component of modern transformer-based PLMs is \textbf{subword tokenization}, which has become a standard technique for handling large vocabularies and linguistic variations. Instead of representing words as atomic units, subword tokenization splits words into smaller, more manageable units such as prefixes, suffixes, or frequent character sequences.  These character sequences may, but often do not, align with morphemes. For example, the word \textit{unbelievable} might be tokenized into \textit{un}, \textit{believ}, and \textit{able}, preserving semantic integrity while keeping the vocabulary size reasonable. This method addresses the limitations of word-level tokenization, which struggles with out-of-vocabulary words, and character-level tokenization, which may lose semantic coherence. By leveraging subword tokenization, transformer-based PLMs enhance their ability to process diverse linguistic patterns, improving generalization across languages and domains.

A diverse range of transformer-based pre-trained language models (PLMs) have emerged, each designed to optimize different modeling paradigms. These models incorporate architectural modifications and specialized training objectives, allowing them to excel in specific applications. 

A variety of transformer-based PLMs have emerged, each tailored to different modeling paradigms. These models introduce architectural modifications and training objectives that enable them to excel in specific applications. One prominent example is \textbf{BERT} (Bidirectional Encoder Representations from Transformers) \cite{devlin-etal-2019-bert}, an encoder-only transformer model that captures bidirectional context, making it particularly effective for language understanding tasks. BERT is pre-trained using two objectives: MLM, which enables the model to learn deep contextual representations by predicting randomly masked tokens, and next-sentence prediction, which helps the model understand relationships between consecutive sentences.

Building on BERT’s masked language modeling approach, ELECTRA (Efficiently Learning an Encoder that Classifies Token Replacements Accurately) \cite{clark-etal-2020-electra} introduces an alternative pre-training strategy that improves training efficiency. Instead of simply reconstructing masked tokens, ELECTRA employs a discriminative objective, where a small auxiliary generator network corrupts tokens in the input, and the main model is trained to distinguish between real and replaced tokens. This ``replaced token detection'' objective allows the model to learn more efficiently by utilizing all tokens in the sequence, rather than just the masked ones, making pre-training more sample-efficient compared to BERT. The resulting representations retain the strengths of bidirectional modeling while benefiting from a more effective training signal.

While BERT and other encoder-based models have proven highly effective for language understanding tasks, the transformer architecture also enables a fundamentally different approach -- one centered on text generation. Instead of leveraging bidirectional context to enhance comprehension, some models rely on a unidirectional, autoregressive framework that predicts text sequentially. \textbf{GPT} (Generative Pre-trained Transformer) \cite{radford-etal-2019-language} takes this approach by utilizing only the decoder of the transformer and training it with an autoregressive objective. The model learns to predict the next token in a sequence based solely on previous tokens, enabling it to generate coherent and contextually relevant text. This unidirectional training paradigm makes GPT particularly well-suited for generative tasks such as text completion, dialogue systems, and creative writing.  

In the early stages of transformer-based NLP models, encoder-only architectures like BERT dominated the landscape, excelling in language understanding tasks. Encoder-decoder models, such as T5 \cite{raffel2020exploring}, which reframed NLP tasks as text-to-text transformations, also gained traction, particularly in applications like summarization and machine translation. However, in recent years, there has been a growing emphasis on decoder-only models such as GPT. This shift reflects the increasing demand for generative AI, where autoregressive modeling enables applications such as conversational agents and content generation.

\subsection{Contextualized Representations}  

The introduction of recurrent and transformer-based PLMs led to a fundamental shift in how tokens are represented in NLP, giving rise to \textit{contextualized embeddings}. Unlike static embeddings such as Word2Vec or GloVe, which assign a fixed vector to each token regardless of its context, contextualized representations dynamically adjust a token’s embedding based on its surrounding words in a sequence. This advancement significantly improved the model’s ability to capture word meaning variations and resolve ambiguities in natural language.  

Early approaches to contextualized representations emerged naturally from RNN-based language models. Since recurrent networks, including LSTMs and GRUs, maintain a hidden state that is updated at each time step, the representation of each token is implicitly informed by the preceding sequence. This hidden state acts as a form of contextualized embedding, encoding information about the context seen up to that point. However, RNN-based models struggled with capturing long-range dependencies due to the vanishing gradient problem and were computationally inefficient, as their sequential processing limited parallelization. These constraints motivated the transition to transformer-based architectures, which provide a more scalable and effective approach to generating contextualized representations.  

Transformer-based models dramatically improved contextualized embeddings through their self-attention mechanism. In these models, each token’s embedding is computed as a weighted sum of all other tokens in the sequence. The attention mechanism dynamically determines these weights, allowing the model to refine a token’s representation based on its broader context. This results in embeddings that are highly sensitive to meaning and usage. For example, a transformer-based model would assign different embeddings to the word ``bank'' in the following sentences:  
\begin{itemize}
    \item ``I went to the \textit{bank} to deposit money.''
    \item ``The boat is tied to the river \textit{bank}.''
\end{itemize}  
In this case, self-attention enables the model to differentiate between financial institutions and riverbanks by considering surrounding words such as ``deposit money'' and ``river.'' This dynamic adjustment of word representations allows modern language models to achieve a deeper and more nuanced understanding of text.

\section{Large Language Models}

The progression from traditional word embeddings to transformer-based models fundamentally changed how natural language is processed. While contextualized representations, as seen in models like BERT and GPT, significantly improved language understanding and generation, they remained constrained by model size and computational capacity. The natural next step in this evolution was to scale up both the architecture and training data, leading to the development of \textbf{large language models} (LLMs). 

LLMs are typically transformer-based models that leverage massive parameter counts and extensive datasets to enhance language modeling capabilities. Unlike earlier models that were optimized for specific tasks through fine-tuning, LLMs exhibit remarkable generalization abilities, allowing them to perform a wide range of NLP tasks without task-specific retraining. Models such as GPT-3, GPT-4, LLaMA, Phi, and Mistral are trained with billions -- or even hundreds of billions -- of parameters, enabling them to capture nuanced linguistic patterns, reasoning abilities, and broad knowledge across diverse domains. Following empirical scaling laws \cite{kaplan-2020-scaling, hoffmann-2022-chinchilla}, these models are trained on vast corpora, encompassing books, articles, code, and web data, equipping them with broad adaptability. While there is no strict threshold for defining an LLM, models with more than one billion parameters are generally classified as such.

\subsection{Capabilities of LLMs}

A key strength of LLMs is their ability to learn and generalize without explicit fine-tuning. This is made possible through \textbf{in-context learning} \cite{brown-etal-2020-language}, where the model adapts to a given task solely based on the input it receives, without modifying its parameters. Rather than requiring task-specific training, LLMs can infer instructions, recognize patterns, and generate appropriate outputs by conditioning on task descriptions and examples provided in the prompt -- the input text that guides the model’s response.
In-context learning also enables \textbf{few-shot learning}, where the model is presented with a small number of task-specific examples within the prompt, allowing it to generalize to new contexts dynamically. This contrasts with \textbf{zero-shot learning}, in which the model relies entirely on its pre-trained knowledge to perform a task without any explicit examples. The ability to adapt in this way makes LLMs particularly valuable for applications requiring flexibility and rapid deployment without the need for retraining.

Scaling up both model size and training data has been a key driver in the advancement of LLMs. Larger models tend to exhibit better language comprehension, reasoning, and generation abilities. Larger models, such as GPT-3 (175 billion parameters), PaLM (540 billion parameters), and GPT-4 (speculated to exceed a trillion parameters), have shown that increasing parameter counts enhances linguistic generalization and performance across a broad range of NLP tasks, from creative writing and translation to reasoning and knowledge retrieval. While early models like BERT primarily focused on language understanding, recent trends have favored generative architectures, particularly decoder-only models such as GPT. The shift towards generative capabilities aligns with the growing demand for AI-powered applications, including conversational agents, content generation, and interactive AI assistants.

Recent advancements have further refined LLMs to better align with human expectations. A crucial innovation in this area is \textbf{instruction tuning}, where models are fine-tuned to follow explicit human instructions more effectively. This enhances their usability in real-world applications, as users can provide direct prompts to guide the model’s behavior. Additionally, \textbf{reinforcement learning from human feedback (RLHF)} has been employed to further align LLMs with human preferences. This approach involves training the model with reinforcement learning, using human-generated feedback to optimize responses for correctness, coherence, and appropriateness. RLHF has been instrumental in improving AI assistants, reducing undesirable biases, and enhancing response quality.

\subsection{Challenges of LLMs}
\label{sec:llm_challenges}

While LLMs have demonstrated impressive capabilities in terms of linguistic competence, their widespread adoption is hindered by several key challenges. As models grow in size and complexity, they become increasingly difficult to train, deploy, and control, raising concerns related to efficiency, reliability, and interpretability.
One of the most pressing issues is \textbf{computational cost}. Training and deploying LLMs require immense computational power, often relying on high-performance GPUs. The sheer scale of these models leads to high energy consumption and financial costs, making them inaccessible to many research labs and organizations. Additionally, their environmental impact has sparked discussions on developing more sustainable AI training paradigms.

Beyond computational demands, \textbf{data inefficiency} presents another major challenge. Although LLMs are trained on vast text corpora, they do not always utilize data efficiently. Many tokens contribute little to the model’s overall performance, leading to diminishing returns as training data scales up. Compounding these issues is the challenge of \textbf{interpretability and consistency}. LLMs function as black-box models, making it difficult to understand why they produce certain outputs. This lack of transparency poses challenges in sensitive applications where decision-making accountability is crucial. Additionally, model responses can be inconsistent -- subtle changes in input phrasing can lead to vastly different answers. Such unpredictability complicates their use in systems where reliability and stability are essential.

\begin{figure}[]
    \centering
    \includegraphics[width=\textwidth]{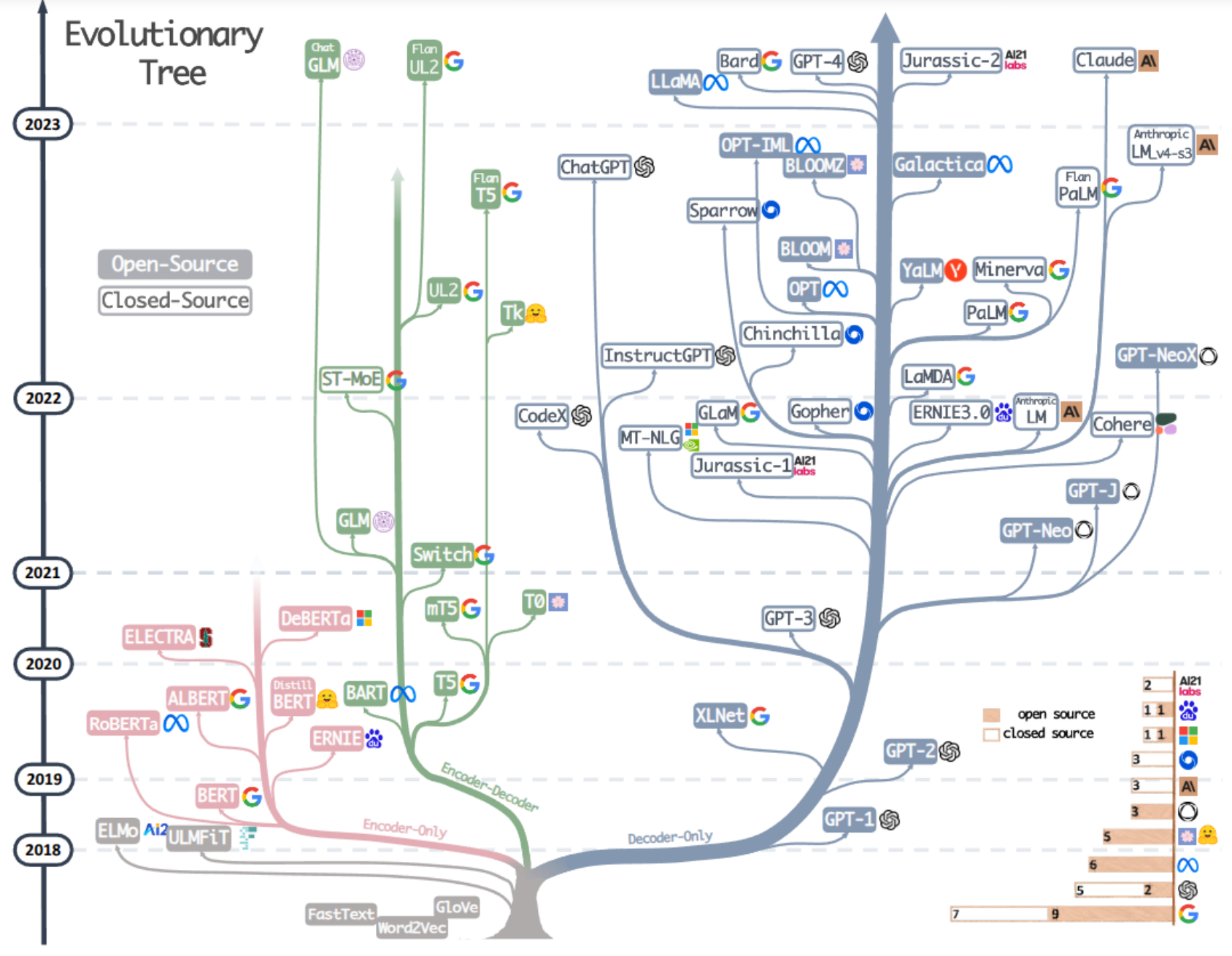}
    \caption[LLM landscape]{LLM landscape. The development of neural language models began with static word embeddings, often combined with RNN architectures. The introduction of transformers led to a shift toward encoder-based architectures, exemplified by BERT and its successors, which excelled in bidirectional contextual understanding. This phase was briefly followed by encoder-decoder architectures like T5, which demonstrated strong performance across diverse NLP tasks. However, the field soon gravitated toward decoder-based transformers -- autoregressive models suited for language generation. This trend has driven the rapid expansion of LLMs, with models growing in size and complexity, as seen in GPT-3, GPT-4, and other large-scale generative models. The figure illustrates this progression, highlighting key milestones in LLM development. Adapted from \cite{yang-2024-harnessing}.} 
    \label{fig:landscape}
\end{figure}

\section{Selection of Pre-trained Language Models}
\label{sec:plm-selection}

The experiments conducted in this thesis utilize a diverse selection of PLMs, classified into two primary groups: \textbf{encoder-based} models and \textbf{decoder-based} models. Each of these approaches leverages the Transformer architecture but is optimized for different types of tasks, influencing how representations are learned and how text is processed.

\subsection{Encoder-Based Models}  

Encoder-based models process input sequences bidirectionally, capturing context from both preceding and succeeding tokens to generate deeply contextualized representations. This makes them particularly well-suited for tasks such as text classification, named entity recognition, and question answering. Our experiments include two widely used encoder-based models: \textbf{BERT} and \textbf{ELECTRA}.

For our experiments, we use the \textit{base} variant of BERT, which consists of 12 transformer layers, a hidden size of 768, and 12 attention heads, amounting to approximately 110 million parameters.
Similar to our selection of BERT, we use the \textit{base} variant of ELECTRA, which also consists of 12 transformer layers, a hidden size of 768, and 12 attention heads, totaling approximately 110 million parameters.

\subsection{Decoder-Based Models}  

Decoder-based models adopt an autoregressive training paradigm, generating text sequentially by predicting the next token based on previously generated tokens. Unlike encoder-based models, which capture bidirectional context, decoder-only models process text unidirectionally, making them particularly well-suited for tasks such as text generation. The decoder-based models included in our experiments span multiple architectures and scales.

\subsubsection{OPT Models}  

The \textbf{Open Pre-trained Transformer (OPT)} models \cite{zhang-etal-2022-opt} are autoregressive transformers trained on large-scale datasets using CLM. Our experiments include three variants of OPT, covering a range of model sizes:
\begin{itemize}
    \item \textbf{OPT-125M} (125 million parameters): 12 layers, hidden size of 768, 12 attention heads;  
    \item \textbf{OPT-1.3B} (1.3 billion parameters): 24 layers, hidden size of 2048, 32 attention heads;  
    \item \textbf{OPT-6.7B} (6.7 billion parameters): 32 layers, hidden size of 4096, 32 attention heads.  
\end{itemize}  
All OPT models are pre-trained on a diverse corpus of web pages, books, and social media text, utilizing memory-efficient optimization techniques to facilitate large-scale training.

\subsubsection{Llama Models}  
Llama 2 \cite{touvron-etal-2023-llama} and Llama 3 \cite{dubey-2024-llama3} are decoder-based LLMs optimized for efficiency and large-scale deployment. While they follow the same autoregressive training framework as OPT, they incorporate architectural improvements to improve inference efficiency. Our experiments include the following Llama model variants:
\begin{itemize}
    \item \textbf{Llama 2-7B} (7 billion parameters): 32 layers, hidden size of 4096, 32 attention heads;
    \item \textbf{Llama 3-8B} (8 billion parameters): 32 layers, hidden size of 4096, optimized for enhanced efficiency over Llama 2.  
\end{itemize}  
Both Llama models are deployed using the \texttt{bfloat16} half-precision format, optimizing memory utilization while maintaining computational efficiency.

\subsubsection{Phi: Lightweight Transformer Models}  

The \textbf{Phi 3 (Mini 4K)} \cite{abdin2024phi3} is a compact yet powerful decoder-based LLM designed to balance computational efficiency with strong language modeling capabilities. Phi models are pre-trained on extensive corpora and fine-tuned for instruction-following and general-purpose text generation. The architecture is streamlined to support efficient inference without sacrificing performance. The Phi 3 (Mini 4K) variant used in our experiments features 3.8 billion parameters, comprising 12 transformer layers, a hidden size of 1536, and 24 attention heads, making it well-suited for resource-constrained applications while maintaining strong modeling capabilities.

\chapter{Representation Properties}
\label{ch:rep-props}

The internal representations of NLMs encapsulate a wealth of information about how these models process, structure, and interpret language. Analyzing these representations provides a systematic approach to understanding the underlying mechanisms that drive model performance, from generalization and stability to robustness against diverse challenges. Through representation analysis, we can probe how effectively models capture patterns in data, adapt to novel inputs, and align their predictions with desired outcomes.

A critical aspect of this analysis involves studying how NLMs, particularly those based on the Transformer architecture, organize linguistic features across their hierarchical layers. Lower layers often emphasize fine-grained linguistic elements like morphology and syntax, while deeper layers progressively refine representations to capture abstract, task-specific semantics \cite{jawahar-2019-does, tenney-etal-2019-bert}. This hierarchical encoding reflects the intricate interplay between syntactic and semantic learning within the model.

By systematically examining these internal structures, representation analysis not only identifies vulnerabilities -- such as sensitivity to small input changes or limited domain transfer -- but also informs the design of targeted interventions, including novel regularization techniques.

\section{Generalization and Robustness}

A fundamental objective in machine learning is achieving strong \textbf{generalization}, ensuring that models not only perform well on their training data but also effectively handle unseen inputs. In NLP, generalization presents unique challenges due to the inherent variability and complexity of human language. Small variations in phrasing, structure, or domain can significantly alter meaning, requiring models to move beyond memorization and instead learn representations that capture underlying linguistic patterns.

To mitigate this, \textbf{regularization} plays a crucial role by constraining the learning process and preventing overfitting. Regularization techniques encourage models to form more stable and generalizable representations rather than fitting to idiosyncrasies in the training data. By promoting smoother, well-structured representation spaces, these techniques help models maintain consistency even under input variations, adversarial perturbations, or shifts in distribution.

PLMs encode words, phrases, and sentences as high-dimensional vectors, mapping semantically similar linguistic units closer together in the representation space. The structure and smoothness of this space directly influence a model’s ability to generalize, allowing it to infer relationships even in previously unseen contexts. However, without proper constraints, learned representations can become overly sensitive to minor perturbations, resulting in brittle predictions and reduced robustness. Regularization not only addresses this by preventing excessive reliance on spurious correlations but also fosters adaptable representations that extend beyond training data.

\subsection{Generalization Enhancers}

At its core, regularization balances model complexity and predictive performance, acting as a generalization enhancer \cite{liu-etal-2023-pac}. Highly flexible models with excessive parameters may fit the training data too precisely, failing to generalize beyond it. By constraining model expressiveness or diversifying training inputs, regularization fosters more stable and adaptable representations.

Regularization has been a cornerstone for improving generalization in PLMs \cite{liu-etal-2023-pac}. Techniques such as weight decay \cite{loshchilov-and-hutter-2018-decoupled} and dropout \cite{srivastava-etal-dropout} have been widely used to mitigate overfitting by reducing model complexity. While effective in controlling model expressiveness, these methods fail to ensure stability under subtle input changes or adversarial perturbations, particularly in the high-dimensional settings of modern PLMs.

To address these gaps, researchers have turned to strategies that leverage unlabeled data and augment existing datasets. For instance, task-adaptive pre-training \cite{gururangan-etal-2020-dont} fine-tunes models on task-specific unlabeled datasets, aligning pre-trained representations more closely with downstream tasks. Additionally, methods like \textit{weak supervision} \cite{yu-etal-2021-fine} and \textit{data augmentation} \cite{okimura-etal-2022-impact, zhou-etal-2021-flipda} introduce controlled modifications to training data, making models more resilient to linguistic variability. While these techniques enhance generalization, they primarily operate at the data level rather than directly refining the representation space.

\subsection{Smoothness in Representation Spaces}
One of the most critical properties that enhance generalization is \textit{smoothness} in representation learning. A well-structured representation space should ensure that small variations in input lead to correspondingly small and predictable changes in learned representations. Ideally, semantically similar inputs should be mapped to nearby points, preserving linguistic regularities in a stable and interpretable manner. Smooth representations not only mitigate overfitting by preventing sharp decision boundaries but also enhance robustness against subtle input perturbations such as paraphrasing, typographical errors, or adversarial attacks.

Despite its advantages, achieving smoothness in high-dimensional neural language models is nontrivial. Without explicit constraints, representations may become highly sensitive to small variations in input, resulting in inconsistent outputs and limited adaptability to new contexts. To address these challenges, recent advances in representation learning \cite{hoffman-etal-2019-robust, mustafa-etal-2020-input} have focused on refining the structure of learned embeddings, ensuring that models remain stable and interpretable across diverse linguistic scenarios.

\subsection{Smoothness-Based Regularization}
\label{sec:smooth-reg}

Inspired by advancements in computer vision, recent work in representation learning has demonstrated that enforcing smoothness can significantly improve both robustness and generalization \cite{czarnecki-etal-2017-sobolev, sokolic-etal-2017-robust}. These techniques focus on controlling how much a model’s predictions change in response to small input perturbations, ensuring that representations remain stable and interpretable. A key approach to achieving this involves regularizing the input-output Jacobian matrix, which captures how input variations affect model outputs. By minimizing the Jacobian norm of the model’s logits with respect to the input \cite{hoffman-etal-2019-robust}, neural models become less sensitive to small perturbations, enhancing robustness against adversarial attacks and minor linguistic variations. Early research in this area \cite{drucker-and-lecun-1992-improving} demonstrated that Jacobian-based constraints could enhance stability, while more advanced techniques, such as \textit{Cross-Hölder regularization} \cite{mustafa-etal-2020-input}, extended this idea by incorporating second-order constraints on the Hessian matrix to regulate representation smoothness further.

Another widely studied concept in smoothness-based regularization is \textit{Lipschitz continuity}, which provides a theoretical framework for bounding the rate of change in model outputs \cite{bartlett-etal-2017-spectrally}. Although computing exact Lipschitz constants in deep networks is computationally intractable, practical approximations using Jacobian matrices have proven effective in reducing sensitivity to perturbations. These constraints implicitly refine the structure of the representation space, making learned embeddings more stable and generalizable across different linguistic contexts. A detailed discussion of Lipschitz continuity and its implications is provided in \Cref{sec:lipschitz}.

\section{Model Calibration}
\label{sec:props-calibration}

\begin{figure}[]
    \centering
    \includegraphics[width=.7\textwidth]{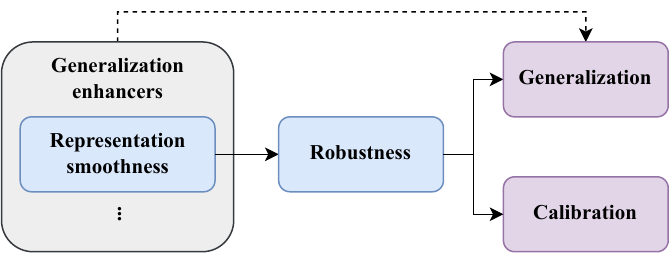}
    \caption[Effects of smoothness-based regularization]{The effects of representation smoothness enforced through regularization. While generalization enhancers are primarily intended to improve performance on unseen data, smoothness-based regularization also promotes robustness -- enhancing a model’s resilience to distribution shifts and adversarial perturbations. This robustness serves as a key mediator, bridging the regularization with improved generalization. Moreover, by fostering robustness, smoothness-based regularization also leads to better calibration, enabling predicted probabilities to more accurately reflect model uncertainty.} 
    \label{fig:gen_cal}
\end{figure}

Machine learning models, including those used in NLP, inevitably make errors. In high-stakes applications -- such as healthcare, finance, and legal decision-making -- these errors can have serious consequences. A model’s ability to express well-calibrated confidence in its predictions is crucial for mitigating risks. \textbf{Calibration} ensures that confidence scores reliably reflect the actual likelihood of correctness \cite{guo-etal-2017-calibration}, enabling models to quantify uncertainty effectively. Well-calibrated models enhance reliability and trustworthiness by identifying uncertain predictions and either rejecting them or deferring decisions to human experts when necessary. However, many PLMs exhibit overconfidence, particularly when faced with out-of-distribution inputs, leading to a misalignment between confidence and correctness. This miscalibration can undermine decision-making, especially in high-stakes settings where model reliability is critical.  

One effective approach to improving calibration is enhancing robustness through representation smoothness (\Cref{fig:gen_cal}). A more robust model maintains stable predictions even under distribution shifts, mitigating overconfidence and improving the alignment between predicted confidence and observed outcomes. In this context, robustness acts as a mediator: by enforcing smoothness in learned representations, models become more resilient to input variations, and this increased robustness ultimately leads to better-calibrated confidence estimates.

By enforcing smoothness in learned representations, models achieve better calibration by ensuring that confidence scores change gradually in response to input variations. This adjustment improves both the reliability and interpretability of model predictions, making them more suitable for real-world deployment. Additionally, smooth representations enhance consistency across different formulations of the same input, reducing variations in predictions caused by minor prompt modifications.

To quantify the impact of smooth representations on model calibration, we consider two key metrics: the Brier score and the expected calibration error. These measures assess how well a model's confidence scores align with actual correctness, providing insight into its reliability in decision-making.

The \textbf{Brier score} \cite{brier1950verification} is a proper scoring rule that quantifies the accuracy of probabilistic predictions.\footnote{A scoring rule is a function that evaluates the quality of probabilistic predictions by assigning a numerical score based on how well the predicted probabilities align with actual outcomes. A proper scoring rule ensures that the expected score is optimized when the predicted probabilities match the true probabilities, encouraging well-calibrated predictions.} It is defined as  
\begin{equation}
\text{Brier Score} = \frac{1}{N} \sum_{i=1}^{N} (p_i - y_i)^2 ,
\end{equation}
where \( p_i \) is the predicted probability for the correct class, \( y_i \) is the true binary label (1 if correct, 0 otherwise), and \( N \) is the number of instances. A lower Brier score indicates better-calibrated predictions. The Brier score can be decomposed into two components: \textit{calibration}, reflecting the alignment of predicted probabilities with observed frequencies, and \textit{refinement}, indicating the spread of confidence scores \cite{degroot-fienberg-1983-comparison}.

\begin{figure}[]
    \centering
    \includegraphics[width=.6\textwidth]{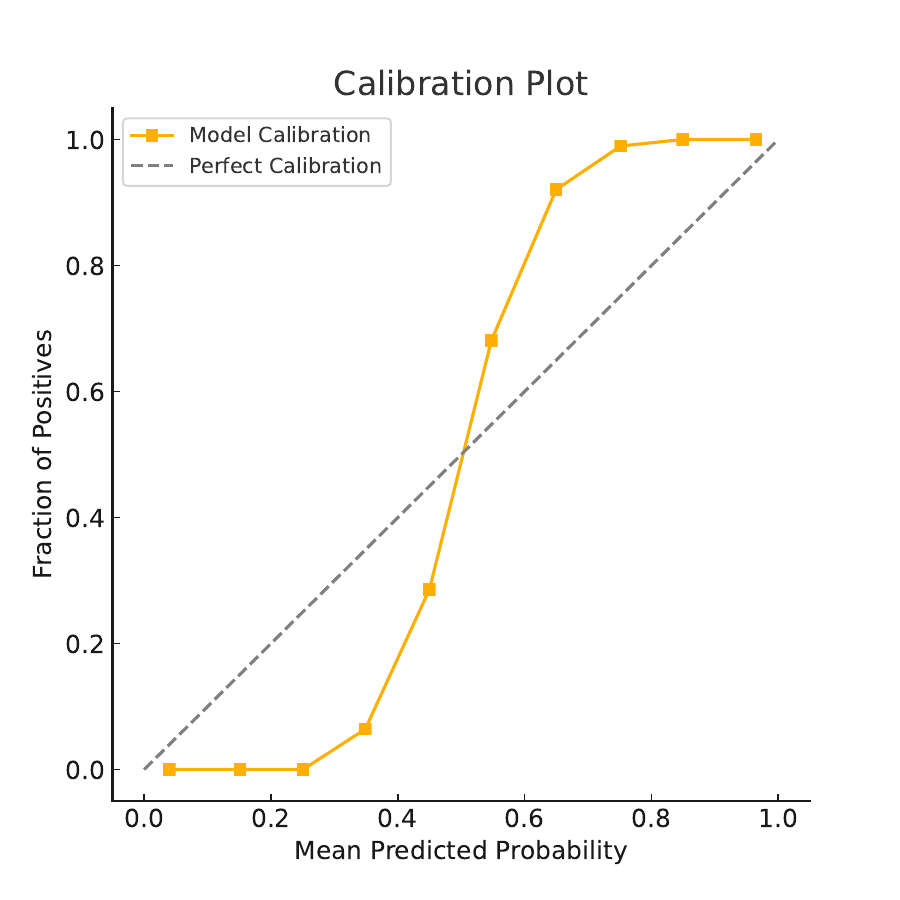}
    \caption[Example of a calibration plot]{Calibration plot illustrating the relationship between predicted probabilities and actual observed frequencies. The dashed diagonal line represents perfect calibration, where predictions match true likelihoods exactly. Deviations from this line indicate miscalibration, where the model is either overconfident or underconfident in its predictions.} 
    \label{fig:cal_plot}
\end{figure}

The \textbf{expected calibration error (ECE)} \cite{naeini2015obtaining} measures the discrepancy between predicted confidence and observed accuracy. It is computed by partitioning predictions into \( M \) bins based on confidence levels and taking a weighted average of the absolute difference between accuracy and confidence in each bin:  
\begin{equation}
\text{ECE} = \sum_{m=1}^{M} \frac{|B_m|}{N} \big| \text{acc}(B_m) - \text{conf}(B_m) \big| ,
\end{equation}
where \( B_m \) is the set of samples in the \( m \)-th bin, \( |B_m| \) is the number of samples in that bin, \( \text{acc}(B_m) \) is the average accuracy, and \( \text{conf}(B_m) \) is the average confidence. A lower ECE indicates better calibration, meaning that the model’s confidence estimates are more aligned with true accuracy. 
\Cref{fig:cal_plot} illustrates a calibration plot, which evaluates how well a model’s predicted probabilities align with actual outcomes. The plot compares the mean predicted probability of positive outcomes with the observed fraction of positives, providing insights into the model’s confidence calibration.

\section{Lipschitz Continuity}
\label{sec:lipschitz}

To gain a mathematical framework for understanding and quantifying representation smoothness, we turn to \textit{Lipschitz continuity}. This concept formalizes the sensitivity of a function’s output to changes in its input, providing a precise apparatus for assessing the stability and robustness of neural networks. By measuring how consistently a network responds to input variations, Lipschitz continuity offers valuable insights into its behavior and reliability \cite{rudin-etal-1964-principles}.

As established in \Cref{sec:distributed_reps}, neural networks are essentially functions mapping inputs from an $n$-dimensional space to an $m$-dimensional space, expressed as $f: \mathbb{R}^n \to \mathbb{R}^m$. The properties of these functions directly influence the reliability and generalization of the network. Lipschitz continuity offers a formal mechanism to assess and constrain these properties, ensuring that small changes in input lead to proportionally small and predictable changes in output \cite{khromov-and-sidak-2024-some}.

\begin{definition}[$L$-Lipschitz Continuity]
A function $f : \mathbb{R}^n \rightarrow \mathbb{R}^m$ is \textit{$L$-Lipschitz continuous} if there exists a constant $L \geq 0$ such that:
\[
\|f(\mathbf{x}) - f(\mathbf{x}')\| \leq L \|\mathbf{x} - \mathbf{x}'\|, \forall \mathbf{x}, \mathbf{x}' \in \mathbb{R}^n .
\]
\end{definition}

When a neural network’s output exhibits large fluctuations due to minor input changes, it becomes prone to overfitting and instability. The Lipschitz constant $L$ provides a bound on these fluctuations, quantifying the extent to which the network’s output can change in response to input variations.

\subsubsection{Lipschitz Continuity and the Jacobian Matrix}
The relationship between the Lipschitz constant and the \textbf{Jacobian} matrix is particularly relevant for understanding the smoothness of neural network functions. Let $\mathbf{J}_f(\mathbf{x}) \in \mathbb{R}^{m \times n}$ denote the Jacobian matrix of $f(\mathbf{x})$, with elements:
\[
\left[\mathbf{J}_f(\mathbf{x})\right]_{i,j} = \frac{\partial}{\partial x_j}f_i(\mathbf{x}).
\]
The spectral norm\footnote{The spectral norm of a matrix $\mathbf{A}$ is defined as $\|\mathbf{A}\|_2 = \max_{\mathbf{x} \neq \mathbf{0}} \frac{\|\mathbf{A}\mathbf{x}\|_2}{\|\mathbf{x}\|_2}$, which corresponds to the largest singular value of $\mathbf{A}$.} of the Jacobian matrix, denoted $\|\mathbf{J}_f(\mathbf{x})\|_2$, provides a lower bound for the Lipschitz constant $L$ \cite{nesterov-2014-convex, dherin-etal-2022-neural}:
\begin{equation}
\|\mathbf{J}_f(\mathbf{x})\|_2 \leq L, \forall \mathbf{x} \in \mathbb{R}^n .
\label{eq:lb}
\end{equation}

\subsubsection{Lipschitz Smoothness and the Hessian Matrix}
\begin{definition}[$L$-Lipschitz Smoothness]
A function $f: \mathbb{R}^n \rightarrow \mathbb{R}$ is \textit{$L$-Lipschitz smooth} if its gradient is $L$-Lipschitz continuous:
\[
\| \nabla f(\mathbf{x}) - \nabla f(\mathbf{x}') \| \leq L\|\mathbf{x} - \mathbf{x}'\|, \forall \mathbf{x}, \mathbf{x}' \in \mathbb{R}^n .
\]
\end{definition}

Lipschitz smoothness provides additional constraints by limiting the variation in a function’s gradient. This ensures that gradients remain bounded, preventing sharp changes in the model’s optimization trajectory. The connection between smoothness and curvature is further captured by the \textbf{Hessian} matrix $\mathbf{H}_f(\mathbf{x})$. For scalar functions, the Hessian matrix, defined as
\[
\left[\mathbf{H}_f(\mathbf{x})\right]_{i,j} = \frac{\partial ^2 f(\mathbf{x})}{\partial x_i \partial x_j},
\]
restricts the curvature of the function by bounding its eigenvalues. For vector-valued functions, the Hessian corresponds to the curvature of individual scalar components. In both cases, $L$-Lipschitz smoothness ensures that the spectral norm of the Hessian matrix does not exceed $L$, further stabilizing the network’s behavior during training.

\subsubsection{Frobenius Norm as a Surrogate for the Lipschitz Constant}
The relationship between matrix norms further refines our understanding of Lipschitz continuity. For any matrix $\mathbf{A}$ of rank $r$, the following inequality holds:
\begin{equation}
\|\mathbf{A}\|_2 \leq \|\mathbf{A}\|_F \leq \sqrt{r} \|\mathbf{A}\|_2 ,
\label{eq:spec_fro}
\end{equation}
where $\|\cdot\|_F$ denotes the Frobenius norm. This implies that reducing the Frobenius norm below its initial spectral norm also reduces the spectral norm, which correlates with a smaller Lipschitz constant. Empirical evidence supports the connection between the Lipschitz constant of neural networks and the lower bound defined by \eqref{eq:lb} \cite{latorre-etal-2020-lipschitz, khromov-and-sidak-2024-some}.

\section{Hutchinson's Estimator}
\label{sec:hutchinson}

Computing the norms of Jacobian and Hessian matrices is computationally expensive, as it requires explicitly materializing these matrices, which is often infeasible for large neural networks. Hutchinson's estimator \cite{hutchinson-1989-stochastic} offers an efficient alternative, enabling the estimation of these norms without fully constructing the matrices, thus significantly reducing computation time. This method provides a practical way to estimate the trace and, consequently, the Frobenius norm of a matrix $\mathbf{A}$, based on the relationship $\trace(\mathbf{A} \mathbf{A}^\top) = \|\mathbf{A}\|^2_F$, where $\trace(\cdot)$ denotes the matrix trace.

The core idea of Hutchinson's estimator is to use random normal vectors to approximate the trace of an arbitrary square matrix $\mathbf{B}$. This approach relies on the expectation of a quadratic form, which provides an unbiased estimate of the trace computed as
\begin{equation}
    \E[\mathbf{v}^\top \mathbf{B} \mathbf{v}] = \trace(\mathbf{B}) ,
\label{eq:hutch_trace}
\end{equation}
where $\mathbf{v} \sim \mathcal{N}(\mathbf{0}, \mathbf{I})$. By setting $\mathbf{B} = \mathbf{J} \mathbf{J}^\top$, where $\mathbf{J}$ represents the Jacobian matrix $\mathbf{J}(\mathbf{x})$, we obtain
\begin{equation}
\begin{aligned}
    \E\left[\mathbf{v}^\top \left(\mathbf{J} \mathbf{J}^\top\right) \mathbf{v}\right] & =
    \trace(\mathbf{J} \mathbf{J}^\top) = \|\mathbf{J}\|^2_F \\
    & = \E\left[ \left( \mathbf{v}^\top \mathbf{J} \right) \left( \mathbf{v}^\top \mathbf{J} \right)^\top \right] \\
    & = \E\left[ \left\| \mathbf{v}^\top \mathbf{J} \right\|^2 \right] ,
\label{eq:fr_trace}
\end{aligned}
\end{equation}
where the vector $2$-norm $\|\cdot\|$ is implied.\footnote{The subscript for the vector $2$-norm is omitted to avoid confusion with the matrix spectral norm.}

Using this relationship, random projections can be employed to compute a Monte Carlo estimate of the Frobenius norm of the Jacobian matrix \cite{varga-etal-2018-gradient, hoffman-etal-2019-robust}. For a network layer, the Frobenius norm of the Jacobian can be approximated as
\begin{equation}
   \|\mathbf{J}\|_F \approx \sqrt{\frac{1}{p} \sum_{i=1}^p \left\| \frac{\partial(\mathbf{v}^{(i)} \mathbf{z})}{\partial \mathbf{x}} \right\|^2} ,
\label{eq:jac}
\end{equation}
where $\mathbf{x}$ is the input, $\mathbf{z} = \mathbf{f}(\mathbf{x})$ is the output of the layer, $\mathbf{v}^{(i)}$ is a random vector sampled from a standard normal distribution for the $i$-th projection, and $p$ is the number of projections.

Similarly, the Frobenius norm of the Hessian matrix for the $j$-th output dimension $z_j$ can be estimated as:
\begin{equation}
    \|\mathbf{H}_j\|_F \approx \sqrt{\frac{1}{p} \sum_{i=1}^p \left\| \frac{\partial\left(\mathbf{v}^{(i)} \frac{\partial z_j}{\partial \mathbf{x}}\right)}{\partial \mathbf{x}} \right\|^2} .
\label{eq:hess}
\end{equation}

Hutchinson's estimator thus provides an efficient way to approximate otherwise intractable computations, making it a practical choice for estimating Jacobian and Hessian norms in large-scale neural networks.

\section{Besov Spaces}
\label{sec:besov-spaces}

Language models generate dynamic representations as inputs progress through successive layers. Each layer refines token embeddings, transitioning from surface-level linguistic features to deeper, task-specific abstractions. This layer-wise refinement can be expressed as:
\[
\mathbf{h}_i^{(l)} = f^{(l)}(\mathbf{h}_i^{(l-1)}, \text{context}),
\]
where $f^{(l)}$ represents the transformation applied at layer $l$, and $\mathbf{h}_i^{(l)}$ is the hidden representation of token $i$ at layer $l$.

Theoretical tools from function space analysis, particularly \textbf{Besov spaces} \cite{triebel1983theory}, provide a powerful framework for analyzing the smoothness and complexity of deep neural network (DNN) representations. Besov spaces generalize Sobolev spaces by accommodating spatially inhomogeneous smoothness, making them well-suited for neural networks where representations vary significantly across layers and dimensions. This allows for the capture of both local and global features, reflecting the hierarchical and complex behavior of learned representations in DNNs \cite{suzuki-2019-adaptivity, suzuki-atsushi-2021-deep}.

\begin{definition}[Smoothness modulus]
For a function $f \in L^p(\Omega)$, $p \in (0, \infty]$, $t \in (0, \infty)$, $h \in \mathbb{R}^d$, and $r \in \mathbb{N}$, the $r$-th \textit{modulus of smoothness} of $f$ is defined as
\[
w_{r,p}(f, t) := \sup_{\|h\|_2 \leq t} \|\Delta^r_h(f)\|_p,
\]
where $\Delta^r_h(f)(x) := \sum_{i=0}^r \binom{r}{i} (-1)^{r-i} f(x + ih)$ is the $r$-th order forward difference operator, defined for $[x, x + rh] \subseteq \Omega$ and zero elsewhere.
\end{definition}

\begin{definition}[Besov space $\mathcal{B}^{\alpha}_{p,q}$]
The \textit{Besov space} $\mathcal{B}^{\alpha}_{p,q}(\Omega)$, where $\Omega \subseteq \mathbb{R}^d$ is the domain, is a function space characterized by three parameters:
\begin{itemize}
    \item $\alpha > 0$: The smoothness parameter, which measures the regularity of the function;
    \item $p > 0$: The integrability index, indicating the norm used for measuring function magnitude;
    \item $q \geq 1$: The summability index, controlling the decay of smoothness across scales.
\end{itemize}

For $f \in L^p(\Omega)$, the Besov seminorm is defined as:
\[
|f|_{\mathcal{B}^{\alpha}_{p,q}} := \left( \int_0^\infty \left( t^{-\alpha} w_{r,p}(f, t) \right)^q \frac{dt}{t} \right)^{1/q} .
\]

The parameter $\alpha$, known as the \textbf{Besov smoothness index}, quantifies the regularity of the function. Higher values of $\alpha$ indicate smoother functions, while lower values suggest greater irregularity or oscillation.
\end{definition}

The Besov seminorm quantifies the regularity of functions in both spatial and frequency domains, making it an effective measure for analyzing representation smoothness in PLMs. When applied to PLM hidden states $\mathbf{H}_l$ at layer $l$, the Besov norm helps identify whether the representations prioritize coarse global structures (low-frequency features) or intricate local patterns (high-frequency features). Smoother representations often correlate with better generalization, while oscillatory ones may signal overfitting.

\paragraph{Approximating Besov smoothness.}
Computing the exact Besov smoothness index is generally intractable. Practical approaches, such as wavelet decomposition, approximate the Besov smoothness of neural network representations. For instance, Elisha and Dekel \cite{elisha-dekel-2017-function} proposed using wavelet decomposition of a random forest (RF) trained on DNN representations. By analyzing the errors of RF estimators with the most important wavelets, they approximated the Besov smoothness index $\alpha_k$ for a specific DNN layer $k$ as
\[
\log(\sigma_m) \sim \log(c_k) - \alpha_k \log(m), \quad m=1,\dots,M,
\]
where $m$ is the number of wavelets, $\sigma_m$ the approximation error, and $\alpha_k$ the Besov smoothness index estimated via least squares regression.

\paragraph{Layer-Wise analysis of smoothness.}
Besov smoothness can provide insights into the geometry of layer-wise representations. Earlier layers in a DNN typically capture generalizable features, while deeper layers tend to focus on memorizing specific patterns \cite{stephenson-etal-2021-geometry, baldock-etal-2021-deep}. Smoother representations in early layers can enhance generalization by mitigating overfitting. This motivates the use of Besov smoothness as a diagnostic and regularization tool for improving PLMs.

\paragraph{Implications for PLMs.}
Leveraging Besov smoothness allows us to analyze and promote smoother representations during training, particularly in early layers. This approach helps align the model's generalization capabilities with its capacity to handle noisy, adversarial, or domain-shifted inputs. By integrating these insights into PLM training regimes, we aim to foster representations that are both robust and adaptable, enhancing the performance of PLMs in diverse real-world scenarios.
\chapter{Representation Regularization Based on Jacobian and Hessian Matrices}
\label{ch:jachess}

Building on recent advancements in representation learning, this chapter introduces a new method for enhancing the generalization and calibration of PLMs by improving their robustness. The method is based on representation-based regularization, which enforces smoothness in intermediate representations to address key shortcomings of conventional approaches. By refining the transition between discrete tokenized inputs and continuous embedding spaces, this approach enhances the stability and adaptability of PLMs across diverse and challenging scenarios. Regularization is applied across multiple layers, enabling the model to better handle noisy, adversarial, or domain-shifted environments.
Our method draws inspiration from techniques originally developed in computer vision and repurposes them to address the unique challenges of NLP. This chapter details the formulation and implementation of our proposed method \cite{jukic-and-snajder-robustness}, followed by an empirical evaluation demonstrating its effectiveness across a range of NLP tasks.

\section{Jacobian and Hessian Regularization}
\label{sec:jachess}

To enforce smoothness in representation learning, we introduce a regularization method designed to enhance robustness by minimizing the norms of the input-output \textbf{Jac}obian and \textbf{Hess}ian matrices, which we call \jachess{}. By leveraging Lipschitz continuity, \jachess{} enforces stability across network layers. Embedded token representations serve as inputs, and subsequent layer outputs are regularized to encourage smooth transitions. To achieve computational efficiency, \jachess{} employs Hutchinson's estimator (Section~\ref{sec:hutchinson}) for norm estimation via (\ref{eq:jac}) and (\ref{eq:hess}). Unlike existing approaches that focus solely on output logits, \jachess{} applies regularization across \textbf{intermediate representations}, targeting the penultimate layer outputs in particular.

\begin{definition}[\jachess{} regularization term]
For a network with $K$ layers, the \jachess{} regularization term is defined as
\begin{equation}
 \sum_{k=1}^K \left( \lambda_1^{(k)} \|\mathbf{J}^{(k)}\|_F + \lambda_2^{(k)} \sum_{d \in D^{(k)}} \|\mathbf{H}^{(k)}_d\|_F \right) ,
\label{eq:jachess}
\end{equation}
where:
\begin{itemize}
    \item $\lambda_1^{(k)}$ and $\lambda_2^{(k)}$ are layer-specific regularization factors,
    \item $D^{(k)}$ is a random subset of output dimensions in the $k$-th layer,
    \item $\|\mathbf{J}^{(k)}\|_F$ and $\|\mathbf{H}^{(k)}_d\|_F$ are the Frobenius norms of the Jacobian and Hessian matrices, respectively.
\end{itemize}
\end{definition}

To handle the high-dimensional nature of intermediate representations, \jachess{} integrates a dimension sampling strategy. While methods such as \cite{mustafa-etal-2020-input} focus on low-dimensional output spaces (e.g., label logits), \jachess{} regularizes the higher-dimensional intermediate spaces within network layers. By sampling dimensions, \jachess{} balances computational feasibility with comprehensive regularization.

The proposed approach incorporates two regularization modes: (1) \textbf{training set regularization} (\jachess{}$_\text{train}$), where the regularization term is added to the original loss function during training, and (2) \textbf{unlabeled data regularization} (\jachess{}$_\text{unlab}$), where the regularization term is minimized independently on a separate unlabeled dataset. These two modes alternate in a \textbf{dual-mode} training strategy, first minimizing the training loss, followed by focused regularization on the unlabeled data.

\subsection{Regularization Factors} 
\label{sec:reg_factors}
The regularization factors $\lambda_1^{(k)}$ and $\lambda_2^{(k)}$ in (\ref{eq:jachess}) are critical for controlling the strength of Jacobian and Hessian regularization across layers. Let $\boldsymbol{\lambda}_i = [\lambda_i^{(1)}, \dots, \lambda_i^{(K)}]$ ($i \in \{1, 2\}$) denote the vectors of these factors across all layers. We explore three strategies for setting these factors:
\begin{enumerate}
    \item \textbf{Uniformly across all layers:} Assigning the same regularization weight to each layer, ensuring equal emphasis across the network;
    \item \textbf{Proportional to the pre-fine-tuning smoothness of the base PLM:} Layers with higher smoothness before fine-tuning are assigned greater regularization weights, reflecting their existing stability;
    \item \textbf{Inversely proportional to the base PLM’s smoothness:} Layers with lower pre-fine-tuning smoothness are assigned greater weights to prioritize their regularization and improve their stability.
\end{enumerate}

To compute layer smoothness, we first estimate the Jacobian Frobenius norms for all layers:
\[
\mathbf{j} = \left[\|\mathbf{J}^{(1)}\|_F, \|\mathbf{J}^{(2)}\|_F, \dots, \|\mathbf{J}^{(K)}\|_F\right] .
\]
Using $\mathbf{j}$, we derive $\boldsymbol{\lambda}_i$ by scaling inversely (or directly) proportional to smoothness. Empirically, scaling proportional to smoothness proves most effective (Section~\ref{sec:jachess_analysis}), with $\softmax$ normalization slightly outperforming standard normalization. For simplicity and consistency, we set $\boldsymbol{\lambda}_1 = \boldsymbol{\lambda}_2 = \boldsymbol{\lambda}$ across all experiments.

\subsection{Application to Transformers}
The definition of \jachess{} introduced earlier primarily assumes a fixed-length multilayer perceptron (MLP) architecture. However, transformer-based models exhibit a more complex computational structure. In particular, while transformers operate on token embeddings of fixed dimensionality, they process token sequences of variable length, introducing additional complexity when applying Jacobian and Hessian regularization.

To extend \jachess{} to transformer-based PLMs, we adapt its formulation to accommodate variable-length token sequences while preserving the computational efficiency of the original approach.  
In this adaptation, each transformer block is treated as a distinct computational layer, with token embeddings as inputs and the block’s transformed representations as outputs. To capture the cumulative effect of the entire sequence, we compute the norms of the Jacobian and Hessian matrices at each layer and aggregate them across all tokens and layers. This aggregation is performed by summing the layer-wise Frobenius norms, ensuring that each layer’s contribution to the overall smoothness of the network is accounted for.

\section{Experimental Setup}
To evaluate the effectiveness of \jachess{}, we conduct an empirical evaluation and compare it with existing methods. Our implementation is built using PyTorch \cite{paszke-etal-2019-pytorch}, leveraging the Hugging Face Transformers library \cite{wolf-etal-2020-transformers} for handling pre-trained models and datasets.
We next outline the models, regularization techniques, datasets, and fine-tuning strategies used in our experiments.

\subsection{Models}

Our experiments include a diverse selection of PLMs to analyze the applicability of \jachess{} across architectures and scales. Specifically, we evaluate decoder-based OPT models \cite{zhang-etal-2022-opt} with 125M, 1.3B, and 6.7B parameters, alongside the 7B parameter \mbox{Llama 2} model \cite{touvron-etal-2023-llama}. To provide a baseline comparison, we also include the encoder-based BERT model \cite{devlin-etal-2019-bert} (cf.~\Cref{sec:plm-selection} for detailed model specifications).

\subsection{Regularization Methods}

We compare \jachess{} against several existing regularization approaches, emphasizing their impact on robustness and generalization. Two key methods from prior work are included (cf.~\Cref{sec:smooth-reg}):
\begin{itemize}
    \item \textbf{Jacobian regularization} \cite{hoffman-etal-2019-robust}: Minimizes the Jacobian norm of the model's logits with respect to the input;
    \item \textbf{Cross-H\"{o}lder regularization} \cite{mustafa-etal-2020-input}: Extends Jacobian regularization by incorporating the input-logit Hessian norm.
\end{itemize}
For both methods, we apply standard regularization on the training data (\texttt{method}$_{\text{\textbf{train}}}$) and adopt the dual-mode approach with separate unlabeled data (\texttt{method}$_{\text{\textbf{unlab}}}$). 

To further contextualize the results, we incorporate several additional techniques that enhance generalization and robustness in PLMs. These include standard regularization approaches as well as more advanced domain-adaptive and optimization-based strategies.
One of the fundamental baselines used in our experiments is \textit{$L_2$ regularization}, which applies a penalty to the magnitude of model parameters to prevent overfitting. By discouraging excessively large weights, $L_2$ regularization helps improve model stability and generalization to unseen data.

Beyond standard regularization, we employ TAPT \cite{gururangan-etal-2020-dont}, an approach designed to refine a pre-trained language model on domain-specific unlabeled data before fine-tuning it for a target task (cf.~\Cref{sec:plms}). Additionally, we evaluate \textit{Sharpness-Aware Minimization} (SAM) \cite{foret-etal-2021-sam}, an advanced optimization technique that enhances model generalization by reducing sensitivity to small perturbations in model parameters. SAM operates by explicitly searching for parameter updates that minimize loss not only at a single point but across a small local neighborhood in the loss landscape. This is achieved by introducing an adversarial perturbation step that seeks sharp minima -- regions where small weight changes cause large fluctuations in loss -- followed by an optimization step that drives the model towards flatter minima, which are associated with better generalization. For our experiments, we set the neighborhood size parameter $\rho$ to 0.05, following best practices established in prior studies.

To establish a no-regularization baseline, models trained without any of these techniques are denoted as \textsc{base}. This provides a reference point for evaluating the impact of different regularization strategies on model performance and robustness.

\subsection{Datasets}
\label{sec:jachess-datasets}

To evaluate the effectiveness of regularization methods, we conduct experiments on the \textbf{GLUE benchmark} \cite{wang-etal-2018-glue}, which consists of diverse NLP tasks spanning different types of classification and regression problems. These tasks serve as a comprehensive testbed for assessing model generalization, robustness, and performance across linguistic challenges.

\subsubsection{Single-Sequence Classification Tasks}  
Some tasks in GLUE require classifying a single input sequence into predefined categories. These tasks assess a model's ability to capture syntactic correctness, sentiment, or semantic meaning within isolated sentences:
\begin{itemize}
    \item \textbf{CoLA (Corpus of Linguistic Acceptability)}: A binary classification task that evaluates whether a given sentence is grammatically acceptable or not;
    \begin{example-item}[]
    \textit{The book was read by the child.} $\rightarrow$ Acceptable \\  
    \textit{Himself enjoyed the movie.} $\rightarrow$ Not acceptable
    \end{example-item}

    \item \textbf{SST-2 (Stanford Sentiment Treebank)}: A sentiment classification task where a model determines whether a sentence expresses positive or negative sentiment;  
    \begin{example-item}[]
    \textit{This movie was absolutely fantastic!} $\rightarrow$ Positive \\  
    \textit{A complete waste of time.} $\rightarrow$ Negative  
    \end{example-item}
    
    \item \textbf{RTE (Recognizing Textual Entailment)}: A natural language inference (NLI) task that involves determining whether a hypothesis logically follows from a given premise.  
    \begin{example-item}[]
    \textbf{Premise}: \textit{A man is playing a guitar.}  \\
    \textbf{Hypothesis}: \textit{A musician is performing.}
    \vspace{5pt}
    \hrule
    \vspace{5pt}
    \textbf{Label}: \textit{Entailment}
    \end{example-item}
    
\end{itemize}

\subsubsection{Sequence-Pair Classification Tasks}  
Other GLUE tasks require reasoning over pairs of sequences, focusing on identifying paraphrases, textual entailment, or relevance between two inputs:
\begin{itemize}
    \item \textbf{MRPC (Microsoft Research Paraphrase Corpus)}: A binary classification task that determines whether two sentences express the same meaning;  
    \begin{example-item}[] 
    \textbf{Sentence 1}: \textit{The company announced a merger.} \\ 
    \textbf{Sentence 2}: \textit{The firm disclosed an acquisition deal.}
    \vspace{5pt}
    \hrule
    \vspace{5pt}
    \textbf{Label}: Paraphrase  
    \end{example-item}

    \item \textbf{QQP (Quora Question Pairs)}: Similar to MRPC, this task identifies whether two questions from Quora are semantically equivalent;  
    \begin{example-item}[]   
    \textbf{Question 1}: \textit{How do I learn Python?} \\
    \textbf{Question 2}: \textit{What are the best ways to master Python?}
    \vspace{5pt}
    \hrule
    \vspace{5pt}
    \textbf{Label}: Paraphrase  
    \end{example-item}

    \item \textbf{QNLI (Question Natural Language Inference)}: Determines whether a given passage contains the answer to a provided question.  
    \begin{example-item}[] 
    \textbf{Question}: \textit{How were the Portuguese expelled from Myanmar?}  \\
    \textbf{Sentence}: \textit{From the 1720s onward, the kingdom was beset with repeated Meithei raids into Upper Myanmar and a nagging rebellion in Lan Na.}
    \vspace{5pt}
    \hrule
    \vspace{5pt}
    \textbf{Label}: Entailment  
    \end{example-item}
\end{itemize}

\subsubsection{Multi-Class Classification Task}  
\begin{itemize}
    \item \textbf{MNLI (Multi-Genre Natural Language Inference)}: A more complex version of RTE, where a model must classify a premise-hypothesis pair into one of three labels: \textit{entailment}, \textit{contradiction}, or \textit{neutral}.  
    \begin{example-item}[]  
    \textbf{Premise}: \textit{You want to punch the button and go.} \\ 
    \textbf{Hypothesis}: \textit{You don't want to push the button lightly, but rather punch it hard.}
    \vspace{5pt}
    \hrule
    \vspace{5pt}
    \textbf{Label}: Neutral  
    \end{example-item}
\end{itemize}

\subsubsection{Regression Task}  
\begin{itemize}
    \item \textbf{STS-B (Semantic Textual Similarity Benchmark)}: A regression task where the model predicts the similarity between two sentences on a continuous scale from 0 to 5.  
    \begin{example-item}[]  
    \textbf{Sentence 1}: \textit{A cat is sleeping on the couch.} \\  
    \textbf{Sentence 2}: \textit{A feline is resting on a sofa.}  
    \vspace{5pt}
    \hrule
    \vspace{5pt}
    \textbf{Similarity Score}: 4.5   
    \end{example-item}
\end{itemize}

Each task in GLUE is evaluated using a task-specific metric, such as Matthew’s correlation for CoLA, $F_1$ score for MRPC and QQP, Spearman’s correlation for STS-B, and accuracy for the remaining tasks. To summarize performance across tasks, we report the average GLUE score, a commonly used metric in benchmarking studies \cite{devlin-etal-2019-bert, houlsby-etal-2019-parameter}.

To evaluate generalization under domain shifts, we pair datasets from the GLUE benchmark with the IMDb sentiment classification dataset \cite{maas-etal-2011-learning}. This setup aligns similar tasks across different domains, allowing for a structured comparison of cross-domain generalization. Specifically, we pair IMDb with SST-2 for sentiment classification, MRPC with QQP for paraphrase detection, and RTE with QNLI for natural language inference. These pairings ensure that the fundamental task objectives remain consistent while assessing the model’s ability to generalize across domains.
For methods that incorporate unlabeled data, we sample $1,000$ unlabeled instances from a separate pool that does not overlap with the test sets, thereby maintaining a clear distinction between training and evaluation data.

\subsection{Fine-Tuning}
All models are fine-tuned on the GLUE tasks with training sets capped at $10,000$ instances for SST-2, MNLI, QNLI, QQP, and IMDb to ensure computational efficiency across various experimental setups. For encoder-based models, we use the \texttt{[CLS]} token's representation, while for decoders, the representation of the last token in the final layer is utilized. A linear head is stacked on top of the final layer, and the entire model is fine-tuned end-to-end.

\section{Enhancing Generalization with Smooth Representations}
\label{sec:jachess_exp}

This section evaluates the efficacy of \jachess{} in improving robustness and its downstream impact on generalization. We explore embedding perturbation, token corruption, in-distribution generalization, and domain shift -- the phenomenon where a model trained on one data distribution encounters a different distribution during testing.

\subsection{Robustness to Embedding Perturbation}

\begin{figure*}[]
\centering

\begin{subfigure}{0.49\linewidth}
  \centering
  \includegraphics[width=\linewidth]{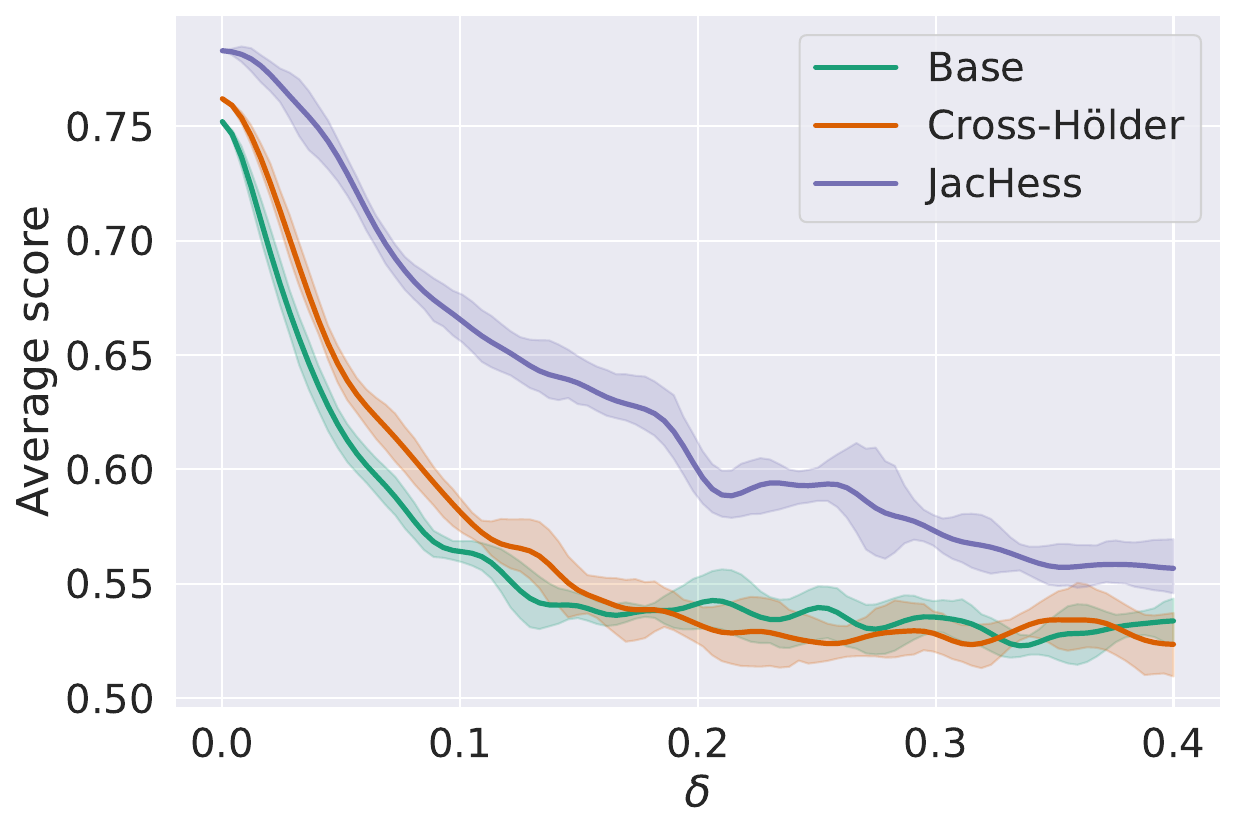}
  \caption{BERT}
\end{subfigure}
\begin{subfigure}{0.49\linewidth}
  \centering
  \includegraphics[width=\linewidth]{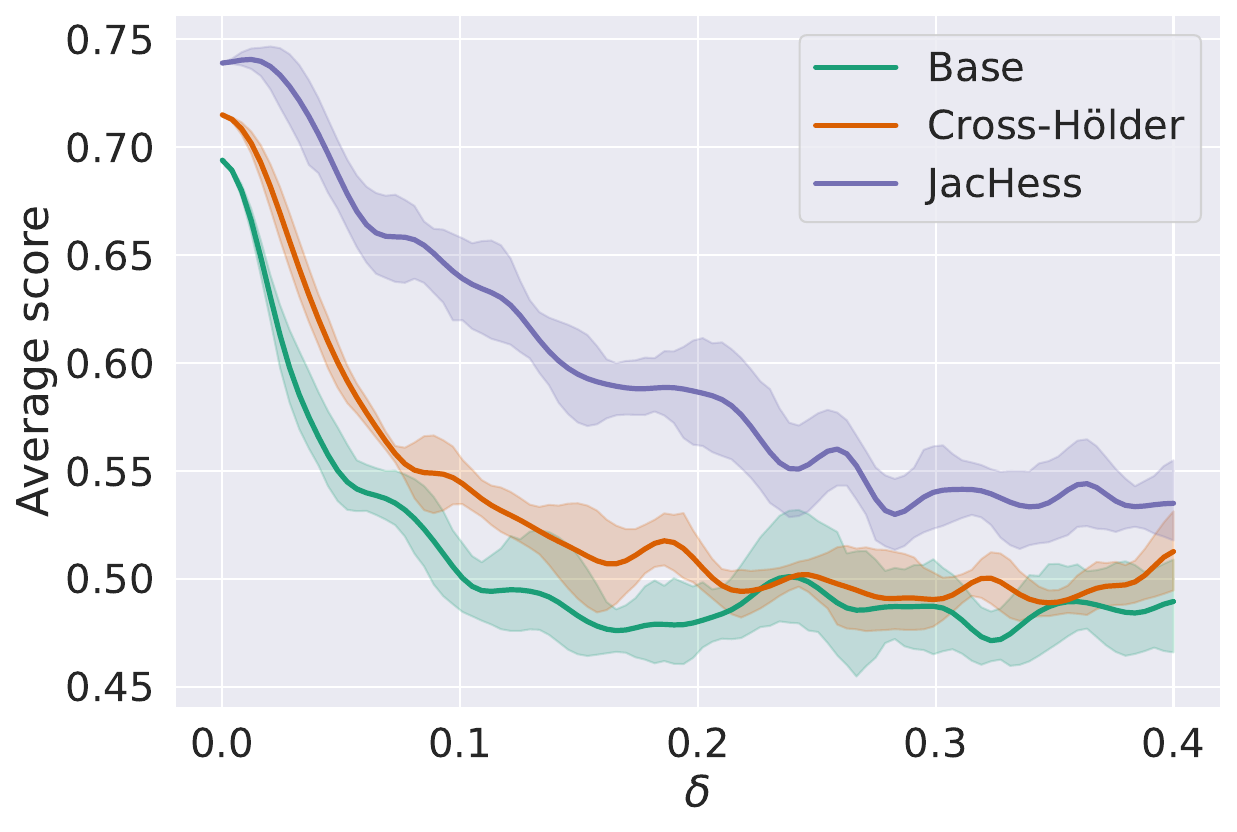}
  \caption{OPT-125M}
\end{subfigure}

\begin{subfigure}{0.49\linewidth}
  \centering
  \includegraphics[width=\linewidth]{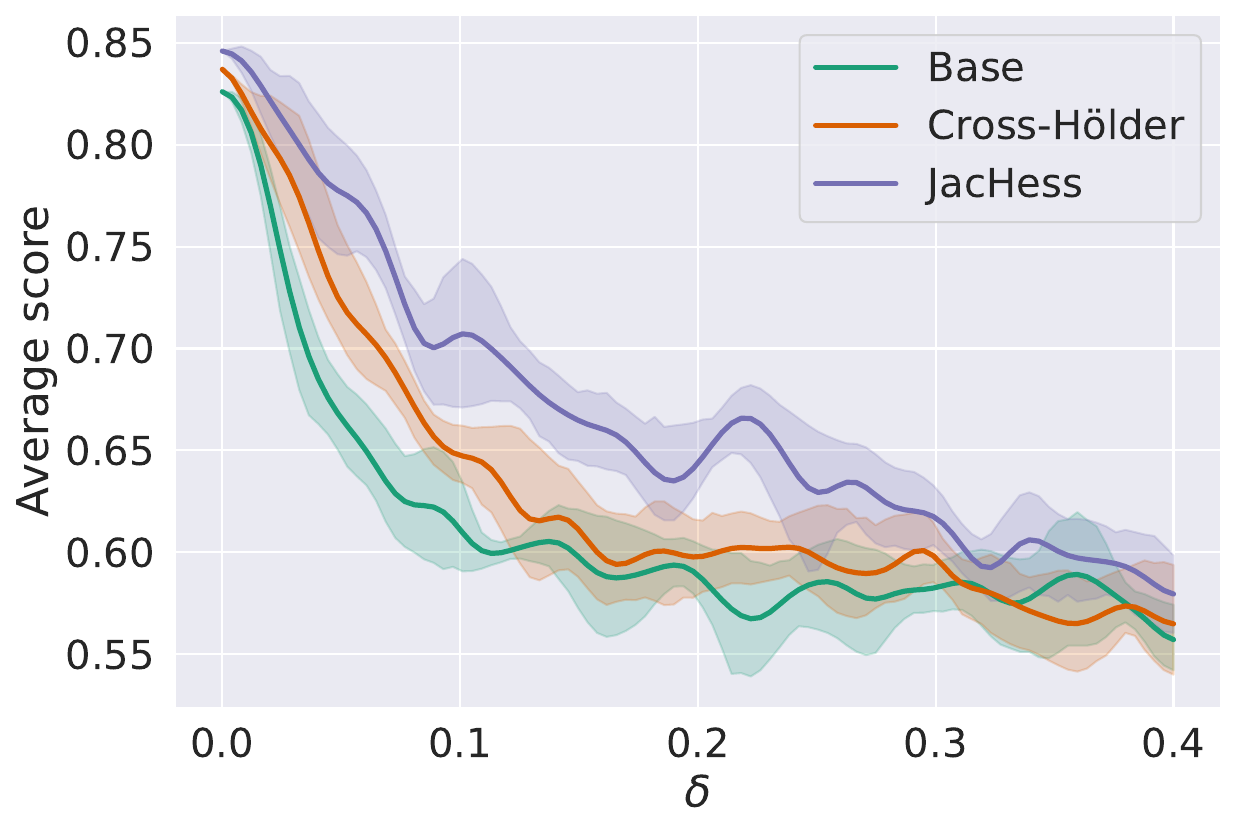}
  \caption{OPT-1.3B}
\end{subfigure}
\begin{subfigure}{0.49\linewidth}
  \centering
  \includegraphics[width=\linewidth]{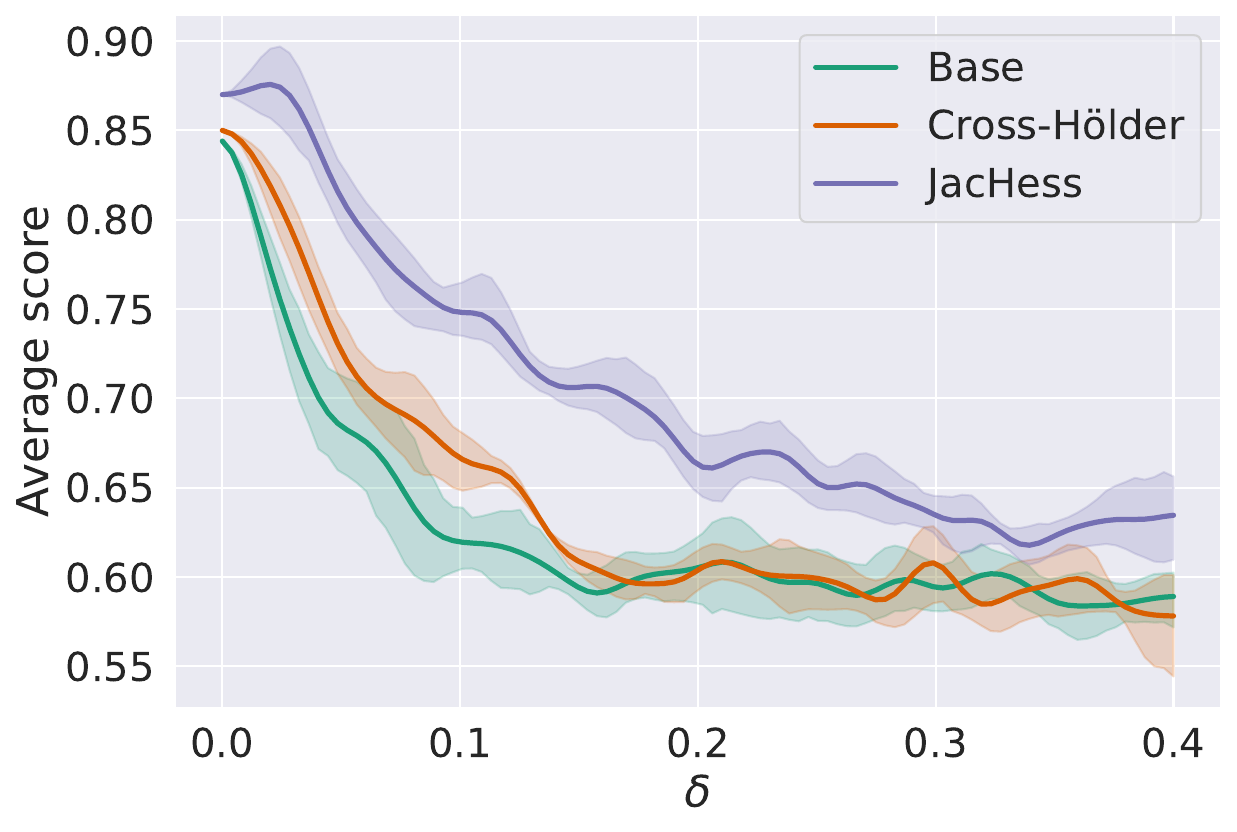}
  \caption{OPT-6.7B}
\end{subfigure}

\begin{subfigure}{0.49\linewidth}
  \centering
  \includegraphics[width=\linewidth]{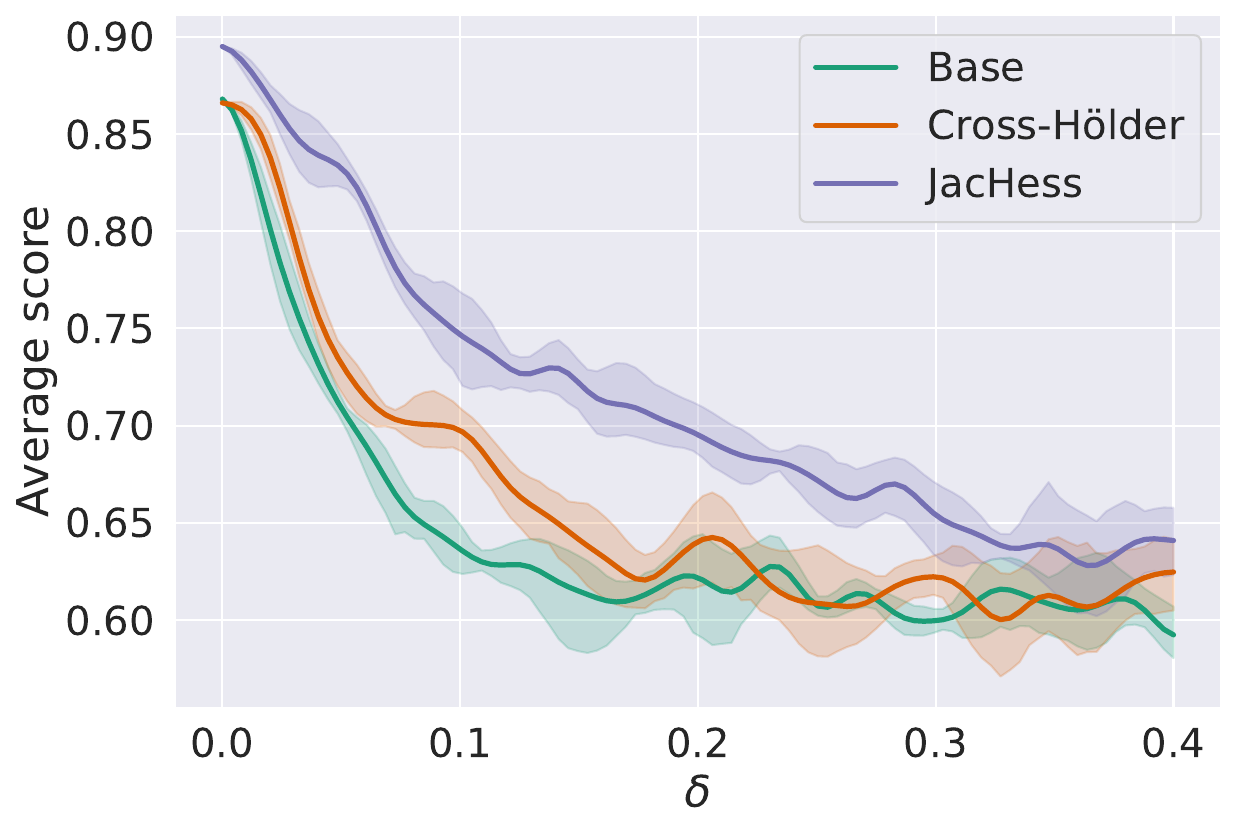}
  \caption{Llama 2}
\end{subfigure}

\caption[\jachess{} -- predictive accuracies under perturbation]{Predictive accuracy under varying perturbation magnitudes $\delta$ averaged across GLUE development datasets. Results are shown for the base model, Cross-H\"{o}lder$_{\text{unlab}}$, and \jachess{}$_{\text{val}}$. Shaded regions indicate standard deviation across multiple runs.}
\label{fig:pert}
\end{figure*}

To evaluate robustness in the continuous embedding space, we introduce controlled perturbations to token embeddings:
\[
\mathbf{e}' = \mathbf{e} + \delta \mathbf{v}, \ \mathbf{v} \sim \mathcal{N}(\mathbf{0},\mathbf{I}),
\]
where $\mathbf{v}$ is a random vector, and $\delta$ controls the perturbation magnitude.

\Cref{fig:pert} illustrates the predictive accuracy of different models under varying perturbation magnitudes $\delta$, averaged across GLUE development datasets. The figure compares the performance of the base model, Cross-H\"{o}lder$_{\text{unlab}}$, and \jachess{}$_{\text{val}}$. As the perturbation magnitude increases, the performance of the base model and Cross-H\"{o}lder$_{\text{unlab}}$ declines more sharply compared to \jachess{}.

These results highlight the effectiveness of \jachess{} in fostering smooth representations that are resilient to input variability. \jachess{} consistently outperforms both Cross-H\"{o}lder$_{\text{unlab}}$ and the base model, maintaining higher predictive accuracy even at larger perturbation magnitudes. This demonstrates its robustness in noisy environments, ensuring stable performance despite significant input perturbations.

\subsection{Robustness to Token Corruption}

\begin{table}[t!]
\caption[Average predictive accuracy with token corruption]{Average predictive accuracy with token corruption. We adjust the percentage of token corruption to $10$\%, $15$\%, and $20$\%. For each dataset, we conduct experiments five times using different seeds and report the average score on the GLUE benchmark. Best scores within the same model and token corruption setup are shown in \textbf{bold}.}
\centering
\small
\scalebox{1}{
\begin{tabular}{llrrrr}
\toprule
& & \multicolumn{4}{c}{Token corruption [\%]} \\
\cmidrule{3-6}
& & $5$ & $10$ & $15$ & $20$ \\
\midrule
\multirow{5}{*}{\rotatebox[origin=c]{90}{BERT}}
& \textsc{base} & $.736$ & $.721$ & $.703$ & $.671$   \\
& $L_2$ & $.730$ & $.724$ & $.715$ & $.677$   \\
& Jacobian$_\text{unlab}$ & $.744$ & $.729$ & $.710$ & $.680$   \\
& Cross-H\"{o}lder$_\text{unlab}$ & $.749$ & $.734$ & $.717$ & $.684$  \\
& \cellcolor{lightgray} \textsc{JacHess}$_\text{train}$ & $.754$ & $.746$ & $.728$ & $.702$ \\
& \cellcolor{lightgray} \textsc{JacHess}$_\text{unlab}$ & $\mathbf{.759}$ & $\mathbf{.751}$ & $\mathbf{.742}$ & $\mathbf{.729}$   \\
\midrule
\multirow{5}{*}{\rotatebox[origin=c]{90}{OPT-125m}}
& \textsc{base} & $.687$ & $.681$ & $.654$ & $.613$   \\
& $L_2$ & $.694$ & $.685$ & $.672$ & $.639$   \\
& Jacobian$_\text{unlab}$ & $.699$ & $.694$ & $.667$ & $.642$   \\
& Cross-H\"{o}lder$_\text{unlab}$ & $.709$ & $.693$ & $.679$ & $.648$  \\
& \cellcolor{lightgray} \textsc{JacHess}$_\text{train}$ & $.719$ & $.710$ & $.702$ & $.689$ \\
& \cellcolor{lightgray} \textsc{JacHess}$_\text{unlab}$ & $\mathbf{.735}$ & $\mathbf{.731}$ & $\mathbf{.723}$ & $\mathbf{.704}$   \\
\midrule
\multirow{5}{*}{\rotatebox[origin=c]{90}{OPT-6.7B}}
& \textsc{base} & $.839$ & $.831$ & $.809$ & $.787$   \\
& $L_2$ & $.846$ & $.830$ & $.805$ & $.792$   \\
& Jacobian$_\text{unlab}$ & $.845$ & $.834$ & $.819$ & $.799$   \\
& Cross-H\"{o}lder$_\text{unlab}$ & $.842$ & $.834$ & $.817$ & $.796$  \\
& \cellcolor{lightgray} \textsc{JacHess}$_\text{train}$ & $.847$ & $.836$ & $.821$ & $.804$ \\
& \cellcolor{lightgray} \textsc{JacHess}$_\text{unlab}$ & $\mathbf{.862}$ & $\mathbf{.851}$ & $\mathbf{.839}$ & $\mathbf{.813}$   \\
\midrule
\multirow{5}{*}{\rotatebox[origin=c]{90}{Llama-2-7B}}
& \textsc{base} & $.852$ & $.831$ & $.813$ & $.786$   \\
& $L_2$ & $.857$ & $.835$ & $.816$ & $.793$   \\
& Jacobian$_\text{unlab}$ & $.848$ & $.837$ & $.822$ & $.804$   \\
& Cross-H\"{o}lder$_\text{unlab}$ & $.854$ & $.832$ & $.811$ & $.796$  \\
& \cellcolor{lightgray} \textsc{JacHess}$_\text{train}$ & $.860$ & $.849$ & $.827$ & $.808$ \\
& \cellcolor{lightgray} \textsc{JacHess}$_\text{unlab}$ & $\mathbf{.883}$ & $\mathbf{.869}$ & $\mathbf{.853}$ & $\mathbf{.829}$   \\
\bottomrule
\end{tabular}
}
\label{tab:corruption}
\end{table}

To assess the robustness of \jachess{} under discrete input perturbations, we simulate token corruption by replacing a percentage of tokens with the \texttt{[UNK]} token. Corruption levels are varied between 10\%, 15\%, and 20\%, and predictive accuracy is measured across the GLUE benchmark.

\Cref{tab:corruption} summarizes the results, showing average predictive accuracy under different levels of token corruption. \jachess{} consistently outperforms other methods across all corruption rates. Notably, \jachess{}$_{\text{unlab}}$ demonstrates greater resilience compared to \jachess{}$_{\text{train}}$ at higher corruption levels, underscoring the benefits of leveraging unlabeled data in mitigating the effects of input perturbations. These results further validate the robustness of \jachess{} in handling noisy or corrupted inputs.

\subsection{In-Distribution Generalization}

\begin{table*}
\caption[ID generalization scores for BERT and OPT-125M with \jachess{}]{In-distribution generalization scores for BERT and OPT-125M on the GLUE development sets averaged across five different seeds for each dataset.  The last column shows the average score across datasets. The highest score for each dataset is in \textbf{bold}, while the second-highest scores are \underline{underlined}. The symbol ``$\dagger$'' indicates significant differences between \jachess{} variants and the \textsc{base} model, as determined by two-sided Mann-Whitney U tests with $p<.05$, corrected for multiple comparisons using the Holm-Bonferroni method.}
\centering
\small
\scalebox{0.8}{
\begin{tabular}{clrrrrrrrrr}
\toprule
& & CoLA & SST-2 & MRPC & STS-B & QQP & MNLI & QNLI & RTE & avg. \\
\midrule
\multirow{10}{*}{\rotatebox[origin=c]{90}{BERT}}
& \textsc{base} & $.466$ & $.894$ & $.852$ & $.855$ & $.810$ & $.700$ & $.832$ & $.610$ & $.752$ \\
& $L_2$ & $.470$ & $.892$ & \underline{$.861$} & $.859$ & $.816$ & $.711$ & \underline{$.843$} & $.608$ & $.758$ \\
& TAPT & $.510$ & $.905$ & $.857$ & $.860$ & $.821$ & \underline{$.714$} & $.835$ & \underline{$.624$} & \underline{$.766$} \\
& SAM & $.483$ & $.889$ & $.849$ & $.851$ & $.809$ & $.708$ & $.822$ & $.619$ & $.754$ \\
& Jacobian$_\text{train}$ & $.471$ & $.883$ & $.850$ & $.847$ & $.806$ & $.704$ & $.821$ & $.607$ & $.749$ \\
& Jacobian$_\text{unlab}$  & $.475$ & $.892$ & $.854$ & $.845$ & $.812$ & $.710$ & $.823$ & $.613$ & $.753$ \\
& Cross-H\"{o}lder$_\text{train}$ & $.498$ & $.901$ & $.842$ & $.839$ & $.824$ & $.707$ & $.830$ & $.605$ & $.756$ \\
& Cross-H\"{o}lder$_\text{unlab}$ & $.504$ & $.905$ & $.852$ & $.836$ & $\mathbf{.829}$ & $.712$ & $.838$ & $.616$ & $.762$ \\
& \cellcolor{lightgray} \textsc{JacHess}$_\text{train}$ & \underline{$.514$}\nospacetext{$^\dagger$} & $\mathbf{.912}$\nospacetext{$^\dagger$} & $.848$ & \underline{$.862$} & $.816$ & $.710$ & $.836$ & $.621$\nospacetext{$^\dagger$} & $.765$ \\
& \cellcolor{lightgray} \textsc{JacHess}$_\text{unlab}$ & $\mathbf{.557}$\nospacetext{$^\dagger$} & \underline{$.906$} & $\mathbf{.864}$\nospacetext{$^\dagger$} & $\mathbf{.891}$\nospacetext{$^\dagger$} & \underline{$.828$}\nospacetext{$^\dagger$} & $\mathbf{.723}$\nospacetext{$^\dagger$} & $\mathbf{.854}$\nospacetext{$^\dagger$} & $\mathbf{.643}$\nospacetext{$^\dagger$} & $\mathbf{.783}$ \\
\midrule
\multirow{10}{*}{\rotatebox[origin=c]{90}{OPT-125M}}
& \textsc{base} & $.452$ & $.883$ & $.760$ & $.824$ & $.693$ & $.614$ & $.742$ & $.582$ & $.694$ \\
& $L_2$ & $.458$ & $.889$ & $.804$ & $.819$ & $.727$ & $.624$ & $.744$ & $.586$ & $.706$ \\
& TAPT & $.461$ & \underline{$.891$} & $.812$ & \underline{$.832$} & $\mathbf{.773}$ & $.653$ & \underline{$.759$} & $.596$ & $.722$ \\
& SAM & $.454$ & $.890$ & $.814$ & $.829$ & $.752$ & $.644$ & $.748$ & $.592$ & $.715$ \\
& Jacobian$_\text{train}$ & $.450$ & $.872$ & $.779$ & $.813$ & $.704$ & $.629$ & $.738$ & $.590$ & $.697$ \\
& Jacobian$_\text{unlab}$ & $.454$ & $.869$ & $.784$ & $.818$ & $.709$ & $.639$ & $.747$ & $.608$ & $.704$ \\
& Cross-H\"{o}lder$_\text{train}$ & $.461$ & $.881$ & $.758$ & $.830$ & $.722$ & $.661$ & $.744$ & $.614$ & $.709$  \\
& Cross-H\"{o}lder$_\text{unlab}$ & \underline{$.470$} & $.880$ & $.771$ & $.839$ & $.731$ & $.657$ & $.751$ & \underline{$.620$} & $.715$  \\
& \cellcolor{lightgray} \textsc{JacHess}$_\text{train}$ & $\mathbf{.474}$\nospacetext{$^\dagger$} & $\mathbf{.896}$\nospacetext{$^\dagger$} & \underline{$.819$}\nospacetext{$^\dagger$} & $.825$ & $.736$\nospacetext{$^\dagger$} & \underline{$.672$}\nospacetext{$^\dagger$} & $.757$\nospacetext{$^\dagger$} & $.610$\nospacetext{$^\dagger$} & \underline{$.724$} \\
& \cellcolor{lightgray} \textsc{JacHess}$_\text{unlab}$  & \underline{$.470$}\nospacetext{$^\dagger$} & $.884$ & $\mathbf{.835}$\nospacetext{$^\dagger$} & $\mathbf{.848}$\nospacetext{$^\dagger$} & \underline{$.768$}\nospacetext{$^\dagger$} & $\mathbf{.691}$\nospacetext{$^\dagger$} & $\mathbf{.787}$\nospacetext{$^\dagger$} & $\mathbf{.628}$\nospacetext{$^\dagger$} & $\mathbf{.739}$ \\
\bottomrule
\end{tabular}
}
\label{tab:robgen1}
\end{table*}

\begin{table*}
\caption[ID generalization scores of \jachess{} for OPT-1.3B, OPT-6.7B, and Llama-2-7B]{In-distribution generalization scores for OPT-1.3B, OPT-6.7B, and Llama-2-7B on the GLUE development sets averaged across five different seeds for each dataset.  The last column shows the average score across datasets. The highest score for each dataset is in \textbf{bold}, while the second-highest scores are \underline{underlined}. The symbol ``$\dagger$'' indicates significant differences between \jachess{} variants and the \textsc{base} model, as determined by two-sided Mann-Whitney U tests with $p<.05$, corrected for multiple comparisons using the Holm-Bonferroni method.}
\centering
\small
\scalebox{0.8}{
\begin{tabular}{clrrrrrrrrr}
\toprule
& & CoLA & SST-2 & MRPC & STS-B & QQP & MNLI & QNLI & RTE & avg. \\
\midrule
\multirow{10}{*}{\rotatebox[origin=c]{90}{OPT-1.3B}}
& \textsc{base} & $.601$ & $.945$ & $.905$ & $.913$ & $.847$ & $.755$ & $.903$ & $.742$ & $.826$ \\
& $L_2$ & $.596$ & $.948$ & $.910$ & $.907$ & $.872$ & $.785$ & $.911$ & $.739$ & $.834$ \\
& TAPT & \underline{$.612$} & $.941$ & $.908$ & $\mathbf{.918}$ & $.874$ & $.801$ & $.914$ & $.749$ & \underline{$.840$} \\
& SAM & $.607$ & $.956$ & $.909$ & $.911$ & $.867$ & $.795$ & $.918$ & $.744$ & $.838$ \\
& Jacobian$_\text{train}$ & $.589$ & $.940$ & $.911$ & $.908$ & $.851$ & $.770$ & $.892$ & $.731$ & $.824$ \\
& Jacobian$_\text{unlab}$ & $.598$ & $.943$ & $.909$ & \underline{$.916$} & $.857$ & $.779$ & $.894$ & $.742$ & $.830$ \\
& Cross-H\"{o}lder$_\text{train}$ & $.612$ & $.939$ & $.910$ & $.902$ & $.879$ & $.773$ & $\mathbf{.924}$ & $.747$ & $.836$ \\
& Cross-H\"{o}lder$_\text{unlab}$ & $.608$ & $.949$ & $.909$ & $.907$ & \underline{$.884$} & $.771$ & \underline{$.921$} & \underline{$.750$} & $.837$ \\
& \cellcolor{lightgray} \textsc{JacHess}$_\text{train}$ & $.610$ & \underline{$.948$} & \underline{$.913$} & $.903$ & $.863$\nospacetext{$^\dagger$} & \underline{$.803$}\nospacetext{$^\dagger$} & $.918$\nospacetext{$^\dagger$} & $.745$ & $.838$ \\
& \cellcolor{lightgray} \textsc{JacHess}$_\text{unlab}$ & $\mathbf{.614}$\nospacetext{$^\dagger$} & $\mathbf{.955}$\nospacetext{$^\dagger$} & $\mathbf{.919}$\nospacetext{$^\dagger$} & $.908$ & $\mathbf{.892}$\nospacetext{$^\dagger$} & $\mathbf{.811}$\nospacetext{$^\dagger$} & \underline{$.921$}\nospacetext{$^\dagger$} & $\mathbf{.751}$\nospacetext{$^\dagger$} & $\mathbf{.846}$ \\
\midrule
\multirow{10}{*}{\rotatebox[origin=c]{90}{OPT-6.7B}}
& \textsc{base} & $.652$ & $.951$ & $.908$ & \underline{$.916$} & $.869$ & $.797$ & $.907$ & $.750$ & $.844$ \\
& $L_2$ & $.650$ & \underline{$.953$} & $.905$ & $.911$ & $.872$ & $.808$ & $.903$ & $.733$ & $.842$ \\
& TAPT & \underline{$.662$} & $.948$ & \underline{$.921$} & $.914$ & \underline{$.903$} & \underline{$.834$} & $\mathbf{.930}$ & \underline{$.754$} & \underline{$.858$} \\
& SAM & $.654$ & $.948$ & $.910$ & $.896$ & $.866$ & $.803$ & $.905$ & $.741$ & $.840$ \\
& Jacobian$_\text{train}$ & $.649$ & $.940$ & $.912$ & $.909$ & $.873$ & $.822$ & $.913$ & $.747$ & $.846$ \\
& Jacobian$_\text{unlab}$ & $.652$ & $.944$ & $.914$ & $.907$ & $.879$ & $.831$ & $.917$ & $.752$ & $.850$ \\
& Cross-H\"{o}lder$_\text{train}$ & $.654$ & $.949$ & $.914$ & $.903$ & $.881$ & $.814$ & $.901$ & $.741$ & $.845$ \\
& Cross-H\"{o}lder$_\text{unlab}$ & $.650$ & $\mathbf{.959}$ & $.920$ & $.907$ & $.887$ & $.821$ & $.907$ & $.745$ & $.850$ \\
& \cellcolor{lightgray} \textsc{JacHess}$_\text{train}$ & $.651$ & \underline{$.953$} & $.919$\nospacetext{$^\dagger$} & $.911$ & $.889$\nospacetext{$^\dagger$} & $.827$\nospacetext{$^\dagger$} & $.918$\nospacetext{$^\dagger$} & $.750$ & $.852$ \\
& \cellcolor{lightgray} \textsc{JacHess}$_\text{unlab}$ & $\mathbf{.688}$\nospacetext{$^\dagger$} & $\mathbf{.959}$ & $\mathbf{.928}$\nospacetext{$^\dagger$} & $\mathbf{.922}$ & $\mathbf{.907}$\nospacetext{$^\dagger$} & $\mathbf{.852}$\nospacetext{$^\dagger$} & \underline{$.929$}\nospacetext{$^\dagger$} & $\mathbf{.776}$\nospacetext{$^\dagger$} & $\mathbf{.870}$ \\
\midrule
\multirow{10}{*}{\rotatebox[origin=c]{90}{Llama-2-7B}}
& \textsc{base} & $.691$ & $.957$ & $.912$ & \underline{$.924$} & $.910$ & $.843$ & $.925$ & $.781$ & $.868$ \\
& $L_2$ & $.686$ & $.950$ & $.904$ & $.915$ & $.908$ & $.846$ & $.923$ & $.792$ & $.866$ \\
& TAPT & \underline{$.722$} & $.953$ & \underline{$.919$} & $.920$ & $.916$ & $.848$ & $.922$ & \underline{$.803$} & $.875$ \\
& SAM & $.682$ & $.961$ & $.914$ & $.920$ & $.911$ & $.846$ & $.925$ & $.794$ & $.869$ \\
& Jacobian$_\text{train}$ & $.681$ & $.940$ & $.893$ & $.903$ & $.882$ & $.837$ & $.912$ & $.764$ & $.852$ \\
& Jacobian$_\text{unlab}$ & $.693$ & $.955$ & $.915$ & $.913$ & $.890$ & $.844$ & $.923$ & $.769$ & $.863$ \\
& Cross-H\"{o}lder$_\text{train}$ & $.688$ & $.951$ & $.909$ & $.915$ & $.914$ & $.832$ & $.927$ & $.759$ & $.862$ \\
& Cross-H\"{o}lder$_\text{unlab}$ & $.691$ & $.949$ & $.913$ & $.917$ & $.909$ & $.838$ & $.931$ & $.779$ & $.866$ \\
& \cellcolor{lightgray} \textsc{JacHess}$_\text{train}$ & $.712$\nospacetext{$^\dagger$} & \underline{$.962$} & $.908$ & $.921$ & \underline{$.919$}\nospacetext{$^\dagger$} & \underline{$.851$}\nospacetext{$^\dagger$} & \underline{$.933$} & $.798$\nospacetext{$^\dagger$} & \underline{$.876$} \\
& \cellcolor{lightgray} \textsc{JacHess}$_\text{unlab}$ & $\mathbf{.746}$\nospacetext{$^\dagger$} & $\mathbf{.973}$\nospacetext{$^\dagger$} & $\mathbf{.951}$\nospacetext{$^\dagger$} & $\mathbf{.934}$\nospacetext{$^\dagger$} & $\mathbf{.929}$\nospacetext{$^\dagger$} & $\mathbf{.872}$\nospacetext{$^\dagger$} & $\mathbf{.940}$\nospacetext{$^\dagger$} & $\mathbf{.813}$\nospacetext{$^\dagger$} & $\mathbf{.895}$ \\
\bottomrule
\end{tabular}
}
\label{tab:robgen2}
\end{table*}

\begin{table*}[h!]
\caption[OOD generalization scores of \jachess{}]{Generalization scores for target datasets under domain shifts for different ``source/target'' dataset pairs. The highest score for each pair is shown in \textbf{bold}.}
\centering
\small
\scalebox{0.9}{
\begin{tabular}{clcccccc}
\toprule
&  & IMDb/SST-2 & SST-2/IMDb & RTE/QNLI & QNLI/RTE & MRPC/QQP & QQP/MRPC \\
\midrule
\multirow{4}{*}{\rotatebox[origin=c]{90}{OPT-6.7B}} 
& \textsc{base} & $.832$ & $.789$ & $.742$ & $.794$ & $.693$ & $.764$ \\
& Jacobian$_\text{unlab}$ & $.865$ & $.798$ & $.772$ & $.804$ & $.707$ & $.772$ \\
& Cross-H\"{o}lder$_\text{unlab}$ & $.853$ & $\mathbf{.809}$ & $.779$ & $.809$ & $.713$ & $.776$ \\
& \textsc{JacHess}$_\text{unlab}$ & $\mathbf{.879}$ & $.804$ & $\mathbf{.795}$ & $\mathbf{.815}$ & $\mathbf{.729}$ & $\mathbf{.810}$ \\
\cmidrule{2-8}
\multirow{4}{*}{\rotatebox[origin=c]{90}{Llama-2}} 
& \textsc{base} & $.892$ & $.824$ & $.768$ & $.832$ & $.741$ & $.790$ \\
& Jacobian$_\text{unlab}$ & $.904$ & $.832$ & $.776$ & $.842$ & $.761$ & $.799$ \\
& Cross-H\"{o}lder$_\text{unlab}$ & $.901$ & $\mathbf{.846}$ & $.780$ & $.848$ & $.754$ & $.812$ \\
& \textsc{JacHess}$_\text{unlab}$ & $\mathbf{.915}$ & $.838$ & $\mathbf{.796}$ & $\mathbf{.852}$ & $\mathbf{.787}$ & $\mathbf{.861}$ \\
\bottomrule
\end{tabular}
}
\label{tab:ood_gen_jachess}
\end{table*}

To evaluate in-distribution generalization,\footnote{In-distribution generalization refers to a model's ability to perform well on test data drawn from the same distribution as the training data. The alternative is out-of-distribution (OOD) generalization, which evaluates a model's robustness to shifts in data distribution, such as domain adaptation or adversarial robustness. OOD generalization is often more challenging and requires techniques beyond standard regularization.} we test \jachess{} on the GLUE benchmark.
\Cref{tab:robgen1,tab:robgen2} together present the performance of different models and regularization methods on the GLUE benchmark tasks. These tables collectively highlight the average scores across all tasks as well as the detailed breakdown for individual tasks. The results show that \jachess{} achieves consistent improvements over baseline models and other regularization techniques. Specifically, \textit{\jachess{} achieves significant improvements in average GLUE scores, ranging from $2\%$ to $4.5\%$ compared to standard fine-tuning and other regularization methods}. These performance gains are particularly evident in larger models, such as OPT-6.7B and Llama 2, underscoring the scalability of \jachess{} to more complex architectures. Furthermore, \jachess{}$_{\text{unlab}}$ consistently outperforms \jachess{}$_{\text{train}}$, emphasizing the utility of incorporating unlabeled data during regularization.

\subsection{Generalization under Domain Shifts}

Domain generalization is critical for assessing the ability of models to transfer knowledge across datasets representing similar tasks in different domains. To evaluate this, we test \jachess{} on paired datasets (e.g., IMDb $\to$ SST-2 for sentiment classification), allowing us to measure its performance under domain shifts.

\Cref{tab:ood_gen_jachess} presents generalization scores across various source/target dataset pairs. The results show that \textit{\jachess{}$_{\text{unlab}}$ consistently outperforms baseline models in cross-domain scenarios}. For example, in paraphrase detection tasks (MRPC $\to$ QQP), \jachess{}$_{\text{unlab}}$ achieves significant accuracy improvements ranging from $4.6\%$ to $7.1\%$. These findings highlight the robust generalization capabilities of \jachess{} and demonstrate its effectiveness in adapting to diverse domains while preserving strong task performance.

\section{Improving Calibration}
\label{sec:calibration}

In this section, we evaluate the impact of \jachess{} on model calibration using the Brier score and ECE (\Cref{sec:props-calibration}), supplemented with calibration plots for a comprehensive evaluation.

\begin{figure*}
\centering
\begin{subfigure}{.49\linewidth}
  \centering
  \includegraphics[width=\linewidth]{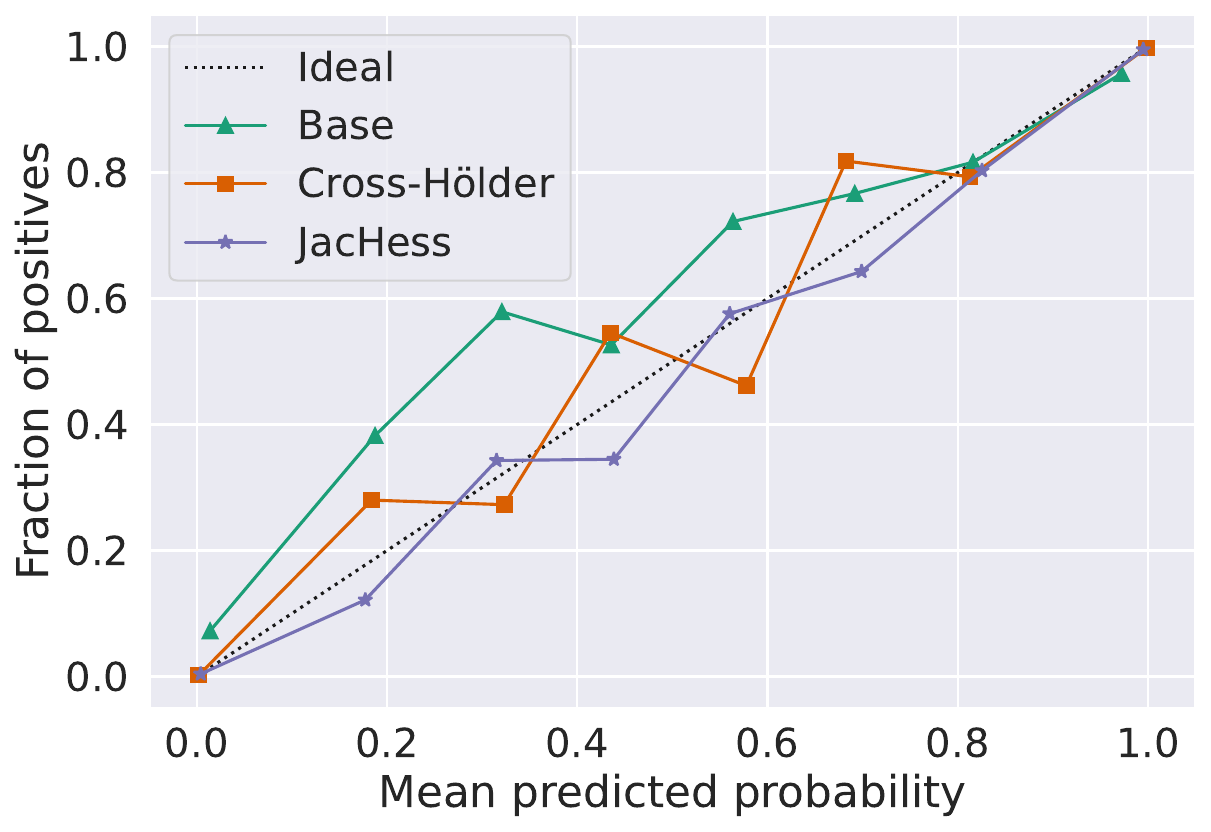}
  \caption{CoLA}
  \label{fig:cal-cola}
\end{subfigure}
\begin{subfigure}{.49\linewidth}
  \centering
  \includegraphics[width=\linewidth]{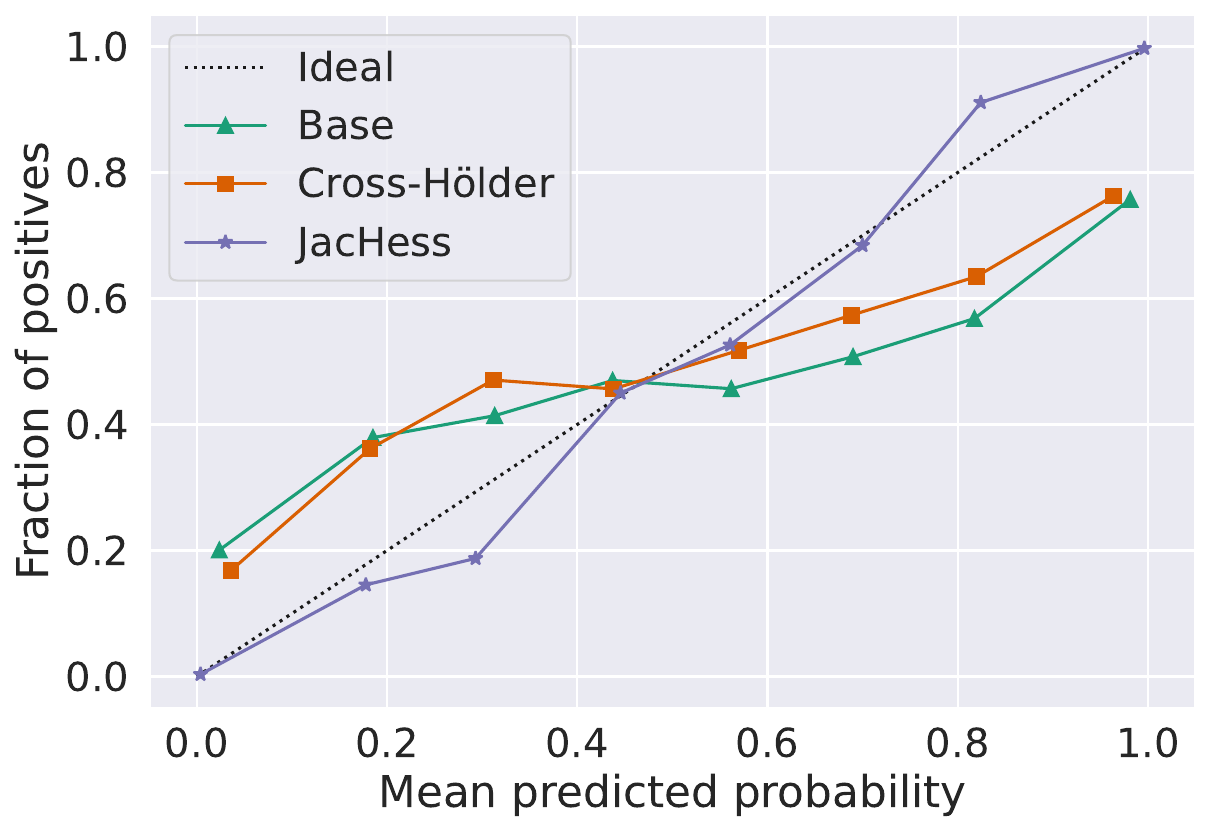}
  \caption{MRPC}
  \label{fig:cal-mrpc}
\end{subfigure}
\begin{subfigure}{.49\linewidth}
  \centering
  \includegraphics[width=\linewidth]{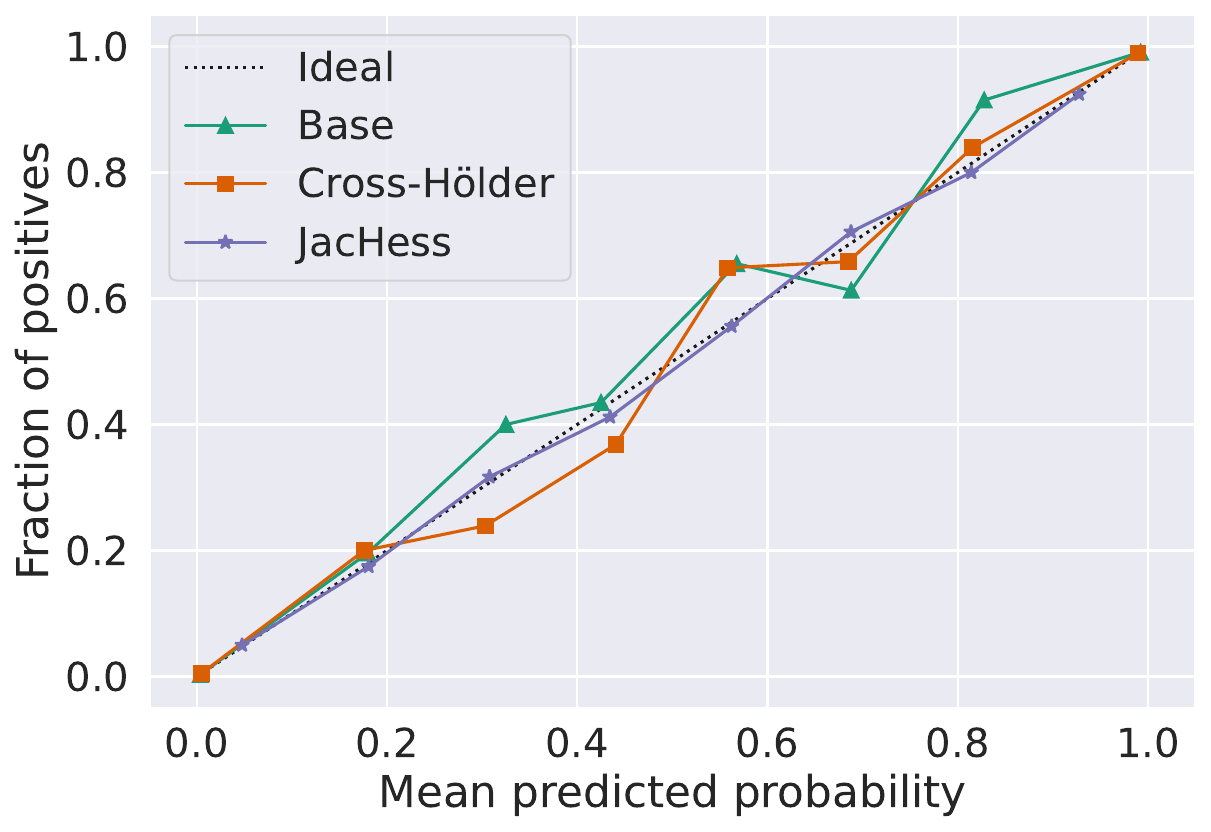}
  \caption{SST-2}
\end{subfigure}
\begin{subfigure}{.49\linewidth}
  \centering
  \includegraphics[width=\linewidth]{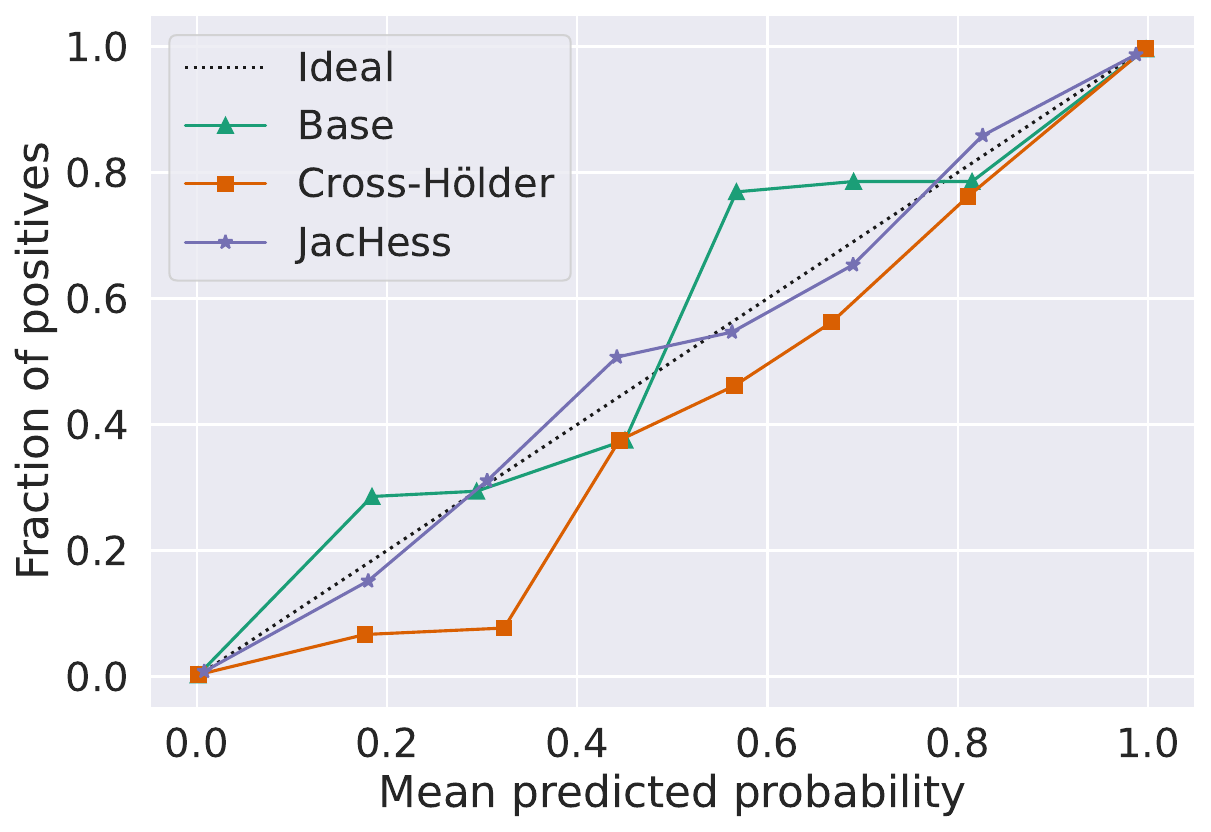}
  \caption{RTE}
\end{subfigure}
\begin{subfigure}{.49\linewidth}
  \centering
  \includegraphics[width=\linewidth]{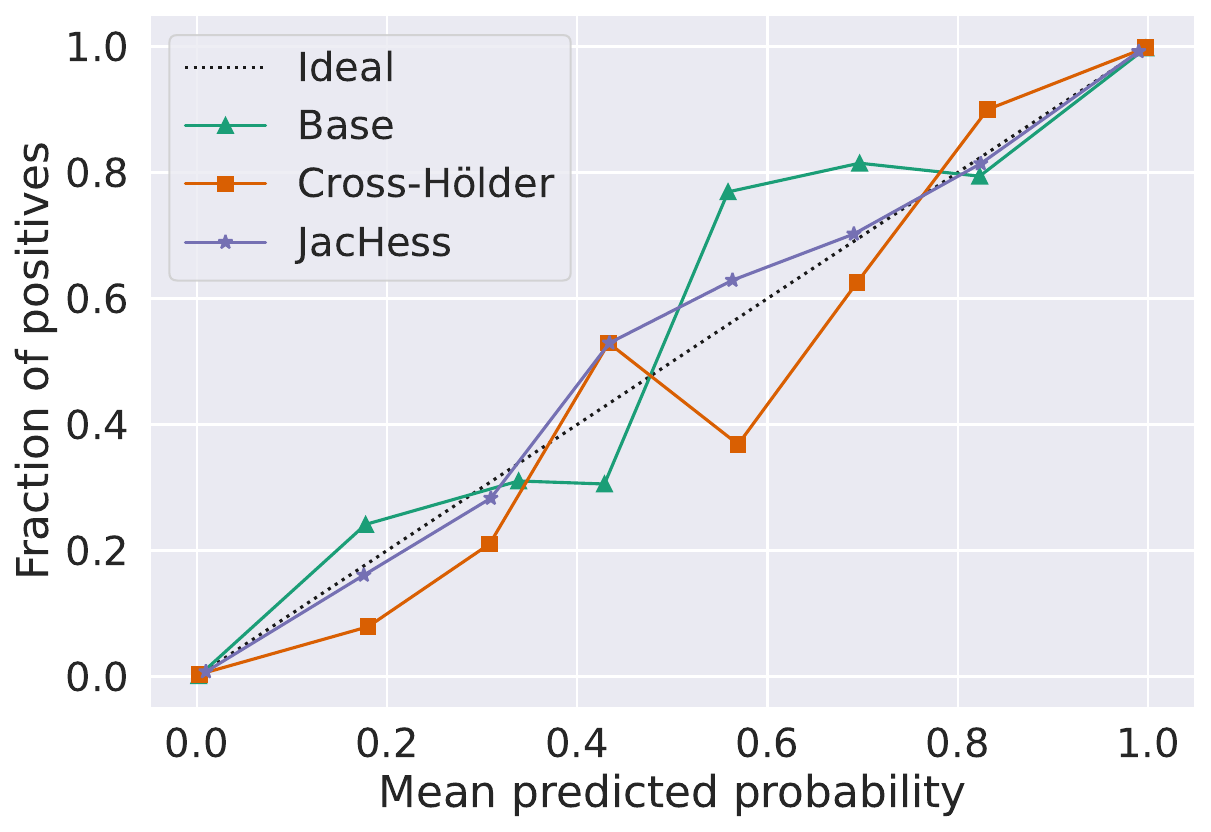}
  \caption{QNLI}
\end{subfigure}
\caption[\jachess{} -- calibration plots]{Calibration plots for binary classification datasets (GLUE) with Llama 2, averaged over five seeds. Eight bins were used to calculate mean predicted probabilities.}
\label{fig:cal}
\end{figure*}

Motivated by prior research linking smoothness in model representations to improved uncertainty estimation \cite{lakshminarayanan-etal-2017-simple, van-etal-2020-uncertainty}, we examine \jachess{}'s ability to promote calibration by leveraging smoothness-enforcing regularization. We utilize the PLM's softmax outputs as confidence scores and compute the Brier score. \Cref{tab:brier} summarizes the Brier scores and ECEs for different models and regularization techniques across a variety of tasks. The data clearly demonstrates that \jachess{} outperforms baseline methods, consistently achieving lower scores. This highlights its effectiveness in improving calibration by aligning predicted probabilities more closely with actual outcomes. These enhancements are observed across all evaluated tasks, confirming \jachess{}'s ability to provide more reliable uncertainty estimates. Additionally, \Cref{fig:cal} provides calibration plots for binary classification tasks from the GLUE benchmark, offering a visual representation of model calibration. The plots reveal that \jachess{} achieves a near-perfect alignment with the ideal calibration line, outperforming both Cross-Hölder regularization and the baseline model. These results further emphasize the effectiveness of \jachess{} in enhancing the trustworthiness and interpretability of PLM predictions.

This evidence highlights the broader significance of representation smoothness, not only for improving generalization and robustness but also for fostering accurate and reliable uncertainty quantification. By leveraging smoothness-enforcing regularization, \jachess{} provides a comprehensive framework for addressing the critical challenge of calibration in NLP models.

\begin{table}[t]
\caption[Average Brier scores and ECE on the GLUE benchmark]{Average Brier scores and ECE on the development sets across binary classification tasks in the GLUE benchmark (CoLA, SST-2, MRPC, RTE, QQP, and QNLI). Regularization methods are applied on a separate, unlabeled set. Lower Brier scores and ECEs indicate better performance. ECE was calculated using eight bins. The best scores are highlighted in \textbf{bold}.}
\centering
\small
\scalebox{1}{
\begin{tabular}{lccccc}
\toprule
& \textsc{base} & Cross-H\"{o}lder & \jachess \\
\cmidrule(lr){2-2} \cmidrule(lr){3-3} \cmidrule(lr){4-4}
Model & \textit{Brier/ECE} & \textit{Brier/ECE} & \textit{Brier/ECE} \\
\midrule
BERT & $.213/.042$ & $.202/.039$ & $\mathbf{.184}/\mathbf{.035}$ \\
OPT-125M & $.256/.057$ & $.233/.045$ & $\mathbf{.204}/\mathbf{.040}$ \\
OPT-1.3B & $.184/.038$ & $.193/.042$ & $\mathbf{.157}/\mathbf{.029}$ \\
OPT-6.7B & $.189/.037$ & $.157/.032$ & $\mathbf{.094}/\mathbf{.024}$ \\
Llama-2-7B & $.167/.034$ & $.163/.031$ & $\mathbf{.089}/\mathbf{.017}$ \\
\bottomrule
\end{tabular}
}
\label{tab:brier}
\end{table}

\section{Analyzing the Design Choices of \jachess{}}
\label{sec:jachess_analysis}

We proceed by analyzing the factors contributing to \jachess{}'s performance, including the scope of regularization, dimension sampling, and the use of unlabeled data.

\begin{table}[]
\caption[Comparison of strategies for 
\jachess{}$_\text{unlab}$ regularization at different application points]{
Comparison of strategies for 
\jachess{}$_\text{unlab}$ regularization at different application points evaluated on the GLUE development sets. Average predictive accuracy for each model is reported across datasets, based on five runs per dataset using different seeds. \textbf{Log} regularizes only the norms corresponding to the Hessian matrices of logits of the penultimate layer. \textbf{Uni} applies regularization uniformly across all layers. \textbf{Inv} sets $\boldsymbol{\lambda}$ inversely proportional to the base model's smoothness, and \textbf{Norm} aligns $\boldsymbol{\lambda}$ directly with the base PLM's smoothness, while \textbf{Soft} applies $\softmax$ instead of standard normalization  (cf.~\Cref{sec:reg_factors}). The highest score for each model is shown in \textbf{bold}.
}
\small
\centering
\scalebox{0.96}{
\begin{tabular}{lccccc}
\toprule
& \multicolumn{5}{c}{Strategy} \\
\cmidrule{2-6}
Model & Logits & Uni & Inv & Norm & Soft \\
\midrule
BERT & $.743$ & $.775$ & $.733$ & $.778$ & $\mathbf{.783}$ \\
OPT-125M & $.709$ & $.726$ & $.692$ & $.735$ & $\mathbf{.739}$ \\
OPT-1.3B & $.832$ & $.845$ & $.803$ & $\mathbf{.848}$ & $.846$ \\
OPT-6.7B & $.851$ & $.868$ & $.811$ & $.865$ & $\mathbf{.870}$ \\
Llama-2-7B & $.874$ & $.883$ & $.819$ & $.892$ & $\mathbf{.895}$ \\
\bottomrule
\end{tabular}
}
\label{tab:ablation}
\end{table}

\paragraph{Scope of regularization.}
We first systematically analyze the \textbf{scope} of regularization and its impact on model performance, focusing on applying \jachess{} at the logits level versus across multiple network layers. Furthermore, we investigate the \textbf{degree} of regularization by varying the regularization factors $\boldsymbol{\lambda}$ applied to different layers.

\Cref{tab:ablation} compares these design choices, previously laid out in \Cref{sec:jachess}. \textit{We observe that initializing the regularization factors in proportion to the base PLM smoothness across layers proves to be the most effective approach.}

\begin{table}[]
\caption[Comparison of predictive accuracies of \jachess{}$_\text{unlab}$ using different numbers of dimensions for the Hessian part of regularization]{Comparison of predictive accuracies of \jachess{}$_\text{unlab}$ using different numbers of dimensions for the Hessian part of regularization. Each number corresponds to the dimensions sampled from a specific layer's output. Average scores for the GLUE benchmark development sets are reported based on five runs per dataset. The highest scores for each model are highlighted in \textbf{bold}.}
\small
\centering
\begin{tabular}{lccccc}
\toprule
& \multicolumn{5}{c}{Number of sampled dimensions} \\
\cmidrule{2-6}
Model & 0 & 5 & 10 & 20 & 50 \\
\midrule
BERT & $.764$ & $.771$ & $\mathbf{.783}$ & $.781$ & $.754$ \\
OPT-125M & $.698$ & $.721$ & $\mathbf{.739}$ & $.737$ & $.713$ \\
OPT-1.3B & $.821$ & $.839$ & $.846$ & $\mathbf{.850}$ & $.831$ \\
OPT-6.7B & $.849$ & $.854$ & $.868$ & $\mathbf{.870}$ & $.857$ \\
Llama-2-7B& $.873$ & $.886$ & $\mathbf{.895}$ & $.892$ & $.880$ \\
\bottomrule
\end{tabular}
\label{tab:hess}
\end{table}

\paragraph{Dimension sampling.}
Regularizing the Frobenius norms of Hessian matrices across sampled dimensions improves computational efficiency. \Cref{tab:hess} indicates that sampling $10$ dimensions yields optimal performance, while higher dimensions result in diminishing returns or degradation due to over-smoothing.

\begin{figure}[]
    \centering
    \includegraphics[width=0.75\linewidth]{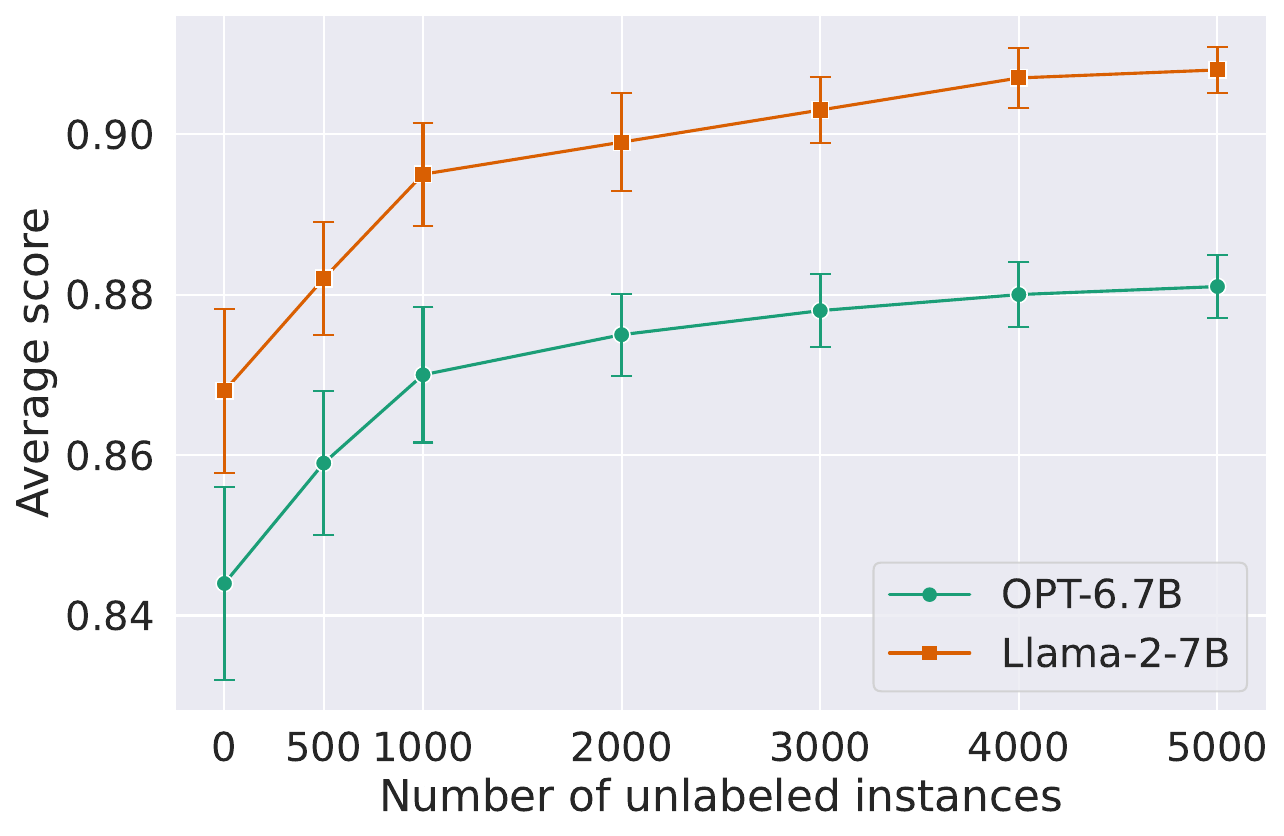}
    \caption[\jachess{} -- performance with increasing numbers of unlabeled examples]{Average GLUE scores for Llama 2 and OPT-6.7B with increasing numbers of unlabeled examples. Error bars represent standard deviations over five runs.}
    \label{fig:unlabeled_data}
\end{figure}

\paragraph{Effect of unlabeled data.}
Finally, we examine the impact of varying the number of unlabeled instances used by \jachess{}$_\text{unlab}$ on model performance. \Cref{fig:unlabeled_data} presents the average scores on the GLUE datasets for both Llama 2 and OPT-6.7B models as the amount of unlabeled data increases. Without any unlabeled data, the setup corresponds to the \textsc{base} model. Incorporating $500$ unlabeled instances for the \jachess{} estimator already yields a substantial improvement in generalization and a slight reduction in standard deviation. 

The generalization scores increase sharply as the number of unlabeled examples rises from $500$ to $1000$, with diminishing returns beyond this point. However, adding more than $1000$ unlabeled instances continues to reduce the standard deviation, indicating enhanced stability across runs. This pattern suggests that while most of the benefits of \jachess{} are realized within the first $1000$ examples, further increases in unlabeled data contribute significantly to the robustness of PLMs on the GLUE datasets.

In terms of runtime, \jachess{} scales linearly with the number of unlabeled instances. Based on averaged runtimes over $10$ runs of Llama 2 on the GLUE datasets, the dual-mode approach increases training time by approximately $3.2\times$ when the unlabeled set matches the size of the training set.\footnote{Runtime estimates are based on $1000$ instances each for the training and unlabeled sets.} In our experiments, we used $1000$ unlabeled instances by default, which is typically one-tenth the size of the training sets in GLUE. This results in a manageable runtime increase of approximately $1.22\times$, balancing the computational cost with the performance gains, making the dual-mode approach both effective and practical for real-world applications. All experiments were conducted on a system equipped with an \textit{AMD Ryzen Threadripper 3970X 32-Core Processor} and $4 \times$ \textit{NVIDIA GeForce RTX 3090} GPUs (each with 24GB of VRAM).

\section{Summary}
This chapter analyzed the relationship between representation properties and their influence on the generalization and uncertainty estimation capabilities of neural language models. By introducing \jachess{}, we demonstrated how promoting smooth representations -- via regularization of Jacobian and Hessian norms -- enhanced the robustness, generalization, and calibration of PLMs. These findings reinforce the critical role of representation properties in improving a model's ability to handle embedding perturbations, token corruption, and domain shifts while providing more reliable uncertainty quantification.

\jachess{} represents a notable advancement in the regularization of PLMs, addressing the dual challenges of generalization and robustness in NLP. By prioritizing the smoothness of intermediate representations, \jachess{} enhances in-distribution generalization while strengthening resilience to adversarial inputs and cross-domain shifts. Furthermore, its contribution to better model calibration highlights its potential for applications in high-stakes domains, where trustworthiness and interpretability are critical requirements.

\part{Active Learning and Parameter Efficiency}
The second part of this thesis focuses on optimizing data and parameter efficiency in the fine-tuning of NLMs. While large-scale PLMs achieve state-of-the-art performance, their dependence on vast labeled datasets and significant computational resources poses substantial challenges. To address these limitations, we explore active learning and parameter-efficient methods as complementary strategies for improving efficiency.

\Cref{ch:al} introduces active learning, highlighting its role in reducing annotation costs by strategically selecting the most informative examples for labeling. The chapter presents theoretical insights into active learning strategies alongside practical considerations in NLP applications. \Cref{ch:beast} builds on this by introducing a smoothness-based early stopping method, which enhances model efficiency without requiring labeled validation data. By leveraging representation smoothness, this method determines optimal stopping points, reducing computational overhead and mitigating overfitting.

\Cref{ch:peft} provides an overview of modular deep learning and parameter-efficient fine-tuning methods, which enable model adaptation without updating the entire model. Finally, \Cref{ch:al-peft} introduces a novel framework that integrates active learning with parameter-efficient fine-tuning, demonstrating how their combination enhances both data efficiency and computational scalability. This framework represents a key contribution of the thesis, offering a principled approach to optimizing NLP model adaptation in low-resource settings.

\chapter{Active Learning}
\label{ch:al}

In machine learning, particularly in supervised learning, the performance of models is largely dependent on the availability of large, labeled datasets. The task of acquiring labeled data is often resource-intensive, involving significant time, labor, and financial costs. In fields such as medical diagnosis, where data labeling requires expert knowledge, or in scenarios where rare events need to be captured, acquiring a large, labeled dataset can be prohibitively expensive.

The dependence of modern machine learning models, especially deep learning models, on large-scale datasets has made data efficiency an essential area of research. While advances in deep learning have led to impressive results across many tasks, they have also highlighted the limitations of models that rely heavily on large, labeled datasets.

To mitigate these challenges, \textit{active learning} has emerged as a data-efficient solution. Active learning (AL)  \cite{settles-2009-active} is a special family of algorithms designed to reduce the cost of labeling by selecting the most informative examples from a pool of unlabeled data and improve \textbf{data efficiency}. The core idea of active learning is to iteratively query an \textit{oracle} -- often a human annotator -- for labels on the data points that are expected to provide the most benefit to the model’s learning process. Through this targeted sampling, active learning aims to achieve high model performance with fewer labeled instances.

\section{Active Learning and Data Efficiency}

One of the primary advantages of AL is its ability to \textit{achieve strong performance with fewer labeled instances}. By focusing on the most uncertain or representative examples, AL enables models to learn more effectively from limited data. This is particularly beneficial in domains where labeling is expensive or time-consuming, such as medical diagnosis, scientific research, or legal document classification.  

Beyond improving data efficiency, AL also \textit{reduces annotation costs} by minimizing the number of labeled samples required for effective model training. Since only the most informative instances are selected for annotation, resources are allocated more efficiently, making it feasible to train high-quality models even in settings with budget constraints. This cost-effectiveness is especially valuable in industries where large-scale labeling is impractical or prohibitively expensive.  

Furthermore, by guiding the model to learn from uncertain or diverse examples, AL naturally \textit{enhances generalization}. Instead of overfitting to redundant or already well-understood patterns, the model is exposed to challenging and underrepresented cases, improving its robustness when encountering unseen data. This results in a more reliable and adaptable system, particularly in dynamic environments where data distributions may shift over time.

At the heart of AL is the ability to \textit{selectively query} data points for labeling, as opposed to the passive learning approach of training on a randomly labeled dataset. The process typically involves the following steps:
\begin{enumerate}
    \item \textbf{Initial model training}: The model is trained on a small, randomly labeled dataset;
    \item \textbf{Query strategy}: The model evaluates the pool of unlabeled data and identifies the data points that are most uncertain or most likely to improve the model’s performance if labeled. Various query strategies are used to determine which instances to select, such as uncertainty sampling, expected model change, or diversity-based sampling;
    \item \textbf{Querying the oracle}: The model queries an oracle (usually a human annotator) for the labels of the selected data points;
    \item \textbf{Model update}: The newly labeled data points are added to the training set, and the model is retrained;
    \item \textbf{Iteration}: This process is repeated iteratively until a stopping criterion is met, such as a labeling budget or a performance threshold.
\end{enumerate}

\section{Theoretical Aspects of Active Learning}

One of the core theoretical aspects of AL revolves around \textit{label complexity} \cite{dasgupta-2011-two}, which refers to the number of labeled examples required for a model to achieve a desired level of performance. In traditional machine learning, the performance of a model is often analyzed in terms of \textit{sample complexity} -- the number of training examples needed for the model to generalize well. In AL, however, the emphasis shifts from sample complexity to label complexity, as the goal is to minimize the number of labeled data points necessary for training.

AL seeks to reduce label complexity by selectively querying the most informative data points from the pool of unlabeled instances. Several theoretical frameworks have been proposed to analyze the efficiency of AL algorithms in terms of label complexity, often providing bounds on the number of labels required compared to passive learning.

\subsection{Version Space}

A common theoretical framework for analyzing AL is the \textit{version space} \cite{mitchell1982generalization}, which represents the set of hypotheses consistent with the labeled data. At each iteration, the AL algorithm queries the oracle to label a data point that reduces the version space as much as possible. The goal is to shrink the version space quickly, leaving only hypotheses that correctly classify the majority of the remaining data.

\begin{figure}[]
\begin{center}
\includegraphics[width=0.6\linewidth]{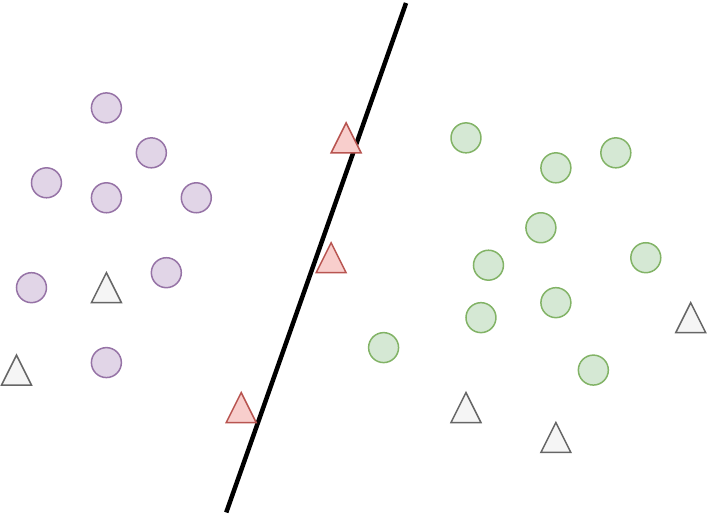}
\caption[Version space in AL]{Illustration of a hypothesis in a version space within AL. The purple circles represent labeled examples from one class, while the green circles belong to another class. Unlabeled instances are shown as triangles, with those highlighted in red representing the most informative instances for query selection. These red-highlighted instances, when labeled, can most efficiently reduce the version space, helping the learning model converge toward the best hypothesis by refining the decision boundary (black line).}
\label{fig:verspace}
\end{center}
\end{figure}

The concept of label complexity is closely tied to the size of the version space. Intuitively, the fewer hypotheses consistent with the labeled data, the fewer additional labels are needed to identify the correct hypothesis. The rate at which the version space shrinks depends on the informativeness of the queried examples, making AL more efficient than passive learning, where data points are selected at random.
As illustrated in \Cref{fig:verspace}, strategically selecting the most informative instances accelerates the reduction of the version space, minimizing ambiguity and guiding the model toward the optimal hypothesis.

\section{Theoretical Insights into Label Complexity in AL}

By adapting to different learning conditions -- whether in ideal, uncertain, or noisy environments -- AL remains a powerful paradigm for reducing annotation costs and improving data efficiency. Theoretical insights into label complexity provide a deeper understanding of when and why AL outperforms passive approaches, guiding the development of more effective learning strategies.

\subsection{Label Complexity Bounds}

AL aims to minimize the number of labeled examples needed to achieve a desired level of performance, a concept often referred to as \textbf{label complexity} \cite{dasgupta-2011-two}. Compared to passive learning, where data points are selected randomly for annotation, AL strategically queries the most informative examples, leading to potentially significant reductions in labeling effort. Several theoretical frameworks have been developed to quantify these benefits under different learning conditions.  

One of the most striking advantages of AL is observed in the \textbf{realizable case}, where there exists a hypothesis in the given hypothesis class that perfectly classifies all data points. In this setting, AL can achieve exponential improvements in label complexity. Specifically, the number of labeled examples required for successful learning can be reduced to a logarithmic function of the hypothesis space size, whereas passive learning typically requires a linear number of labels \cite{freund1997selective}. This efficiency gain highlights the power of actively selecting informative queries in settings where an ideal hypothesis exists.  

Even in the absence of a perfectly fitting hypothesis, AL retains advantages in the \textbf{agnostic case}, where no single hypothesis can classify all data points without error. Here, the efficiency of AL is characterized by the \textit{disagreement coefficient}, a measure of how extensively hypotheses within the version space differ in their predictions on unlabeled data \cite{hanneke2007bound}. A lower disagreement coefficient indicates that fewer labeled examples are needed to refine the hypothesis space, allowing the algorithm to focus on reducing uncertainty more effectively than passive learning. While the improvements in label complexity are not as drastic as in the realizable case, AL still offers meaningful reductions in labeling requirements.  

In \textbf{noisy settings}, real-world challenges, such as mislabeled data and inherent data ambiguity, further influence label complexity. Noise increases the number of labeled examples needed, as AL algorithms must account for erroneous or inconsistent annotations. Theoretical guarantees in these settings depend on specific noise models, such as \textit{Tsybakov noise} \cite{tsybakov2004optimal} or \textit{adversarial noise} \cite{balcan2009agnostic}, which describe the relationship between noise levels and learning efficiency. While robust AL algorithms have been developed to mitigate these effects, the theoretical bounds on label complexity in noisy environments tend to be weaker compared to noiseless cases. Nonetheless, advances in noise-tolerant query strategies continue to make AL a viable option for real-world applications where annotation errors are inevitable.

These theoretical frameworks highlight the diverse factors influencing label complexity in AL and underscore the importance of tailoring algorithms to the specific characteristics of the problem, including hypothesis class properties, data separability, and noise robustness.

\section{Query Strategies in Active Learning}
\label{sec:al-methods}

One of the most critical components of AL is the query strategy, which determines how the model selects data points to label. Broadly, query strategies in AL fall into three main categories: query synthesis, stream-based sampling, and pool-based sampling. Each approach is suited to specific scenarios and impacts label complexity differently.

\textbf{Query synthesis} involves generating synthetic examples for labeling, allowing the active learner to strategically select data points that maximize the reduction of the version space. While theoretically advantageous in minimizing label complexity, this method often produces examples that do not align with real-world data distributions, limiting its generalizability. To mitigate this issue, Wang et al.~\cite{wang2015active} proposed generating synthetic queries near decision boundaries, making the process more realistic and enhancing its effectiveness in reducing label complexity.

\textbf{Stream-based sampling} continuously evaluates a stream of unlabeled data, deciding in real time whether specific instances should be labeled. By selectively choosing only the most informative examples, this method minimizes label complexity while maximizing the value of labeled data. It is particularly beneficial in domains such as natural language processing, where high-utility examples are scarce. Compared to passive learning, stream-based sampling significantly reduces the number of labels required, making it well-suited for dynamic and evolving datasets.

\begin{figure}[]
\begin{center}
\includegraphics[width=0.75\linewidth]{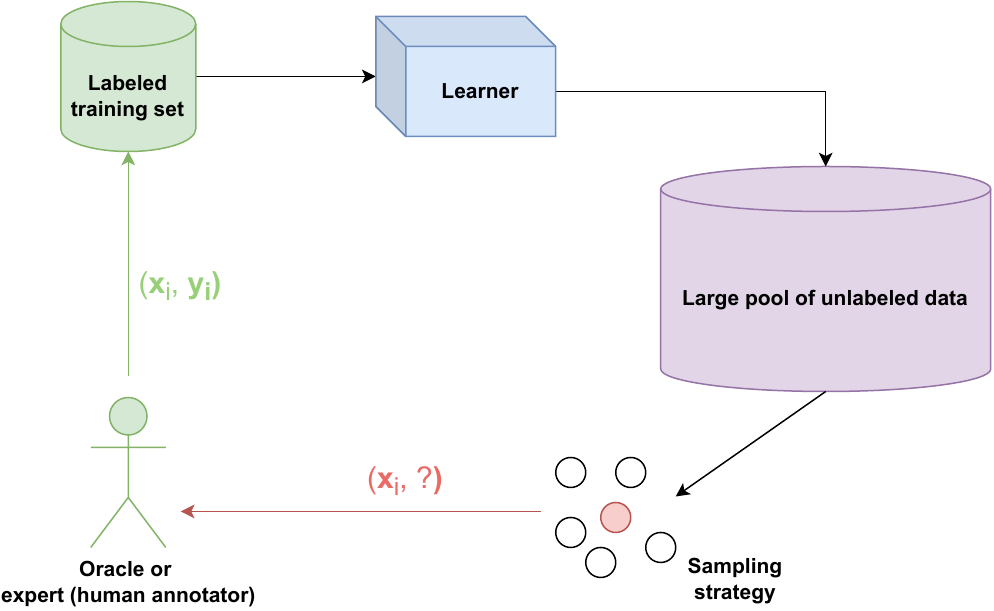}
\caption[AL loop]{Active learning loop in pool-based sampling}
\label{fig:al_loop}
\end{center}
\end{figure}

\textbf{Pool-based sampling}, one of the most widely adopted active learning strategies, operates on a fixed pool of unlabeled data, assessing the entire pool to identify the most informative instances. By querying these examples in batches, this approach ensures that each labeled instance contributes maximally to reducing model uncertainty. Extensively studied in the literature, pool-based sampling has been shown to achieve substantial reductions in label complexity compared to random sampling \cite{tong2001support}. Its structured nature makes it highly versatile and applicable across various real-world tasks. Figure \ref{fig:al_loop} illustrates the iterative process of pool-based AL, highlighting its systematic querying strategy.

Within these frameworks, specific query strategies determine how instances are selected for labeling. These strategies aim to balance exploration, ensuring broad coverage of the input space, and exploitation, focusing on the most uncertain or impactful data points. The most common approaches include:
\begin{itemize}
    \item \textbf{Uncertainty Sampling:} Selects data points where the model has the least confidence in its predictions, focusing on examples near the decision boundary;
    \item \textbf{Disagreement-based sampling:} Uses a \textit{committee} of models to identify data points with the highest disagreement, which often correspond to ambiguous or difficult examples;
    \item \textbf{Expected error minimization:} Selects instances that are expected to reduce the model’s overall error the most, aiming to maximize predictive performance with each newly labeled example;
    \item \textbf{Diversity-based sampling:} Selects a diverse set of examples to ensure broad coverage of the input space and reduce redundancy in the labeled data.
\end{itemize}
The following subsections provide detailed explanations of these strategies, including their mathematical formulations and practical applications.

\subsection{Uncertainty Sampling}

Uncertainty sampling is a widely adopted strategy in AL, built on the premise that the most uncertain instances provide the greatest potential for refining the model's decision boundary. By selectively querying these ambiguous examples, the method seeks to maximize learning efficiency and improve model performance with minimal labeled data. Uncertainty sampling is typically implemented through three main formulations: least confidence, margin sampling, and entropy-based sampling, each utilizing a different metric to quantify uncertainty.

\begin{definition}[Least confidence]
This strategy selects instances for which the model's confidence in its most probable prediction is lowest:
\[
\argmin_{\mathbf{x}} \left[ \max_{y} P_{\boldsymbol{\theta}}\left(y|\mathbf{x}\right) \right] ,
\]
where $P_{\boldsymbol{\theta}}(y|\mathbf{x})$ is the predicted probability of class $y$ for instance $\mathbf{x}$ under model parameters $\boldsymbol{\theta}$. By focusing on examples with the least confident predictions, this method targets instances that are far from being confidently classified.
\end{definition}

\begin{definition}[Margin sampling]
Margin sampling prioritizes instances where the difference in predicted probabilities between the top two classes is smallest:
\[
\argmin_{\mathbf{x}} \left[ P_{\boldsymbol{\theta}}\left(y_1|\mathbf{x}\right) - P_{\boldsymbol{\theta}}\left(y_2|\mathbf{x}\right) \right] ,
\]
where $y_1$ and $y_2$ represent the most probable and second most probable classes, respectively. This approach identifies examples near the decision boundary, which are typically the most ambiguous and informative for learning.
\end{definition}

\begin{definition}[Entropy sampling]
Entropy sampling measures the overall uncertainty in the prediction distribution using entropy:
\[
\argmax_{\mathbf{x}} \left[ - \sum_{y} P_{\boldsymbol{\theta}}(y|\mathbf{x}) \log P_{\boldsymbol{\theta}}(y|\mathbf{x}) \right] ,
\]
where entropy quantifies the dispersion of the probability distribution across all possible classes. Higher entropy indicates greater uncertainty, making this method particularly effective for multi-class problems as it accounts for global uncertainty in predictions.
\end{definition}

While traditional uncertainty sampling methods rely on single-point predictions to estimate uncertainty, they do not account for model uncertainty arising from limited data or inherent ambiguity in learned representations. To address this, Monte Carlo (MC) dropout \cite{gal-ghahramani-2016-dropout} extends entropy-based uncertainty sampling by approximating model uncertainty through Bayesian inference.
\begin{definition}[Monte Carlo dropout with entropy]
Monte Carlo (MC) dropout performs multiple stochastic forward passes with dropout enabled at inference time, generating a distribution over predictions rather than a single deterministic output. Given $T$ stochastic forward passes, the model produces a set of probability distributions:
\[
P_{\boldsymbol{\theta}_t}(y|\mathbf{x}) \quad \text{for } t = 1, \dots, T.
\]
Uncertainty is then estimated using the entropy of the mean predictive distribution across all sampled forward passes:
\[
\argmax_{\mathbf{x}} \left[ - \sum_{y} \bar{P}(y|\mathbf{x}) \log \bar{P}(y|\mathbf{x}) \right] ,
\]
where
\[
\bar{P}(y|\mathbf{x}) = \frac{1}{T} \sum_{t=1}^{T} P_{\boldsymbol{\theta}_t}(y|\mathbf{x}).
\]
By averaging predictions across multiple stochastic forward passes, MC dropout captures epistemic uncertainty. Unlike standard entropy sampling, which relies on a single softmax output, MC dropout accounts for variability in predictions, leading to more reliable uncertainty estimates. However, this method introduces additional computational overhead due to the need for multiple forward passes per query.
\end{definition}

By incorporating Monte Carlo Dropout, uncertainty sampling benefits from a more robust estimation of model uncertainty, allowing for more informed instance selection while mitigating the limitations of single-pass entropy-based methods.

By leveraging these different formulations, uncertainty sampling provides a flexible and powerful framework for actively querying the most informative instances, ensuring efficient exploration of the input space. However, despite its advantages, uncertainty sampling is not without limitations. One notable caveat is its potential to query outliers or mislabeled examples, as such instances can exhibit high uncertainty. These queries may not contribute meaningful information to the learning process and can even degrade model performance. Additionally, uncertainty sampling assumes that the model's uncertainty estimates are reliable. However, poorly calibrated models can misidentify informative examples, leading to suboptimal performance. Addressing these issues often requires incorporating additional heuristics or constraints to ensure that queried instances are both uncertain and representative of the underlying data distribution.

\subsection{Disagreement-Based Sampling}

Disagreement-based sampling, also referred to as \textit{query-by-committee} (QBC), prioritizes instances where there is significant disagreement among a committee of models. The committee may consist of models trained with different initializations, hyperparameters, or subsets of data. The underlying intuition is that regions of high disagreement correspond to areas in the input space that are underrepresented or ambiguous in the current training set, making them informative for model refinement. Below, we outline three prominent formulations of disagreement-based sampling: vote entropy, Kullback-Leibler divergence, and consensus margin.

\begin{definition}[Vote entropy]
Vote entropy quantifies the uncertainty in the class predictions aggregated across the committee. Let $v(y)$ represent the total number of votes for class $y$ across all models in the committee, and $C$ denote the number of models in the committee. The vote entropy is computed as
\[
H_v(\mathbf{x}) = - \sum_{y} \frac{v(y)}{C} \log \frac{v(y)}{C} ,
\]
where higher entropy indicates greater disagreement among the committee members regarding the class label for instance $ \mathbf{x} $.
\end{definition}

\begin{definition}[Kullback-Leibler divergence (KL divergence)]
KL divergence measures the difference between the probability distributions predicted by two models in the committee. Let \( P_i(y|\mathbf{x}) \) and \( P_j(y|\mathbf{x}) \) denote the predicted probability distributions of class \( y \) for instance \( \mathbf{x} \) by two models in the committee. The KL divergence is given by
\[
D_{\text{KL}}(P_i || P_j) = \sum_{y} P_i(y|\mathbf{x}) \log \frac{P_i(y|\mathbf{x})}{P_j(y|\mathbf{x})} .
\]
Instances with high average KL divergence across all pairs of committee members are prioritized for querying, as they indicate substantial disagreement in predicted probabilities.
\end{definition}

\begin{definition}[Consensus margin]
The consensus margin evaluates the difference between the predicted probabilities of the two most agreed-upon classes across the committee. Let \( P_{\text{max}}(\mathbf{x}) \) and \( P_{\text{second}}(\mathbf{x}) \) represent the probabilities of the most and second-most agreed-upon classes, respectively. The margin is defined as
\[
M(\mathbf{x}) = P_{\text{max}}(\mathbf{x}) - P_{\text{second}}(\mathbf{x}) .
\]
Lower margins signify greater disagreement, making such instances valuable for querying.
\end{definition}

Disagreement-based sampling is particularly effective in settings where multiple plausible decision boundaries exist, such as when dealing with sparse or noisy data. By targeting areas of high uncertainty or disagreement, these methods maximize the informativeness of labeled examples, ensuring efficient use of labeling resources and improved learning outcomes.

However, despite its advantages, disagreement-based sampling comes with notable challenges that can limit its practicality in certain scenarios. One major drawback is the computational overhead involved in maintaining and training multiple committee models, which can become particularly burdensome for large-scale datasets or when using complex architectures like deep neural networks. Another issue is the diminishing returns observed as the labeled dataset grows. Over time, disagreements among committee members tend to decrease, reducing the method's effectiveness. Additionally, the performance of disagreement-based sampling is highly sensitive to the design of the committee. If the committee lacks diversity or is poorly constructed, the resulting queries may not be as informative, leading to suboptimal learning outcomes.

\subsection{Expected Error Minimization and Label Complexity}

Another formal approach to reducing label complexity is through \textit{expected error minimization} (EEM).

\begin{definition}[EEM sampling method]
The goal of the EEM sampling method is to select the instance $\mathbf{x}$ from the unlabeled pool $\mathcal{U}$ that minimizes the expected future error of the model. The selection criterion is formalized as
\[
\argmin_{\mathbf{x}} \mathbb{E}_{\mathbf{Y} | \boldsymbol{\theta}, \mathbf{x}} \left[
\sum_{
\mathbf{x}' \in \mathcal{U}} \mathbb{E}_{\mathbf{Y'} | \boldsymbol{\theta}^{+}, \mathbf{x}'}
    \left[ L \left( h\left(\mathbf{x}' \right), y' \right)
    \right]
\right] ,
\]
where $\boldsymbol{\theta}^{+}$ refers to the updated model parameters after incorporating the newly labeled instance. 
\end{definition}

EEM strategy is theoretically appealing because it directly targets reducing the model’s error, but it is computationally expensive and often only feasible for very simple models. The reduction in label complexity is significant in theory but challenging to achieve for large datasets.

\subsection{Diversity-Based Sampling}

Diversity-based sampling is an AL approach that aims to select examples that provide broad coverage of the input space. By prioritizing diversity, these methods reduce redundancy and ensure that the labeled dataset represents the underlying data distribution effectively. Unlike uncertainty sampling, which focuses on selecting the most uncertain examples, diversity-based methods aim to explore the input space comprehensively, enabling the model to generalize better across different regions of the dataset. This approach is particularly valuable for capturing the full variability of the data, improving learning outcomes while minimizing the labeling effort.

Uncertainty sampling often results in selecting points clustered near decision boundaries, leading to redundant queries and potentially neglecting underrepresented regions of the input space. Diversity-based methods address this limitation by selecting examples that span the input space, ensuring that the labeled dataset reflects the variety of patterns in the data. This leads to a more balanced and representative set of training examples, which in turn enhances the model’s performance, particularly in scenarios with limited labeled data. Next, we discuss two key approaches to diversity-based sampling: core-set selection and discriminative AL.

\subsubsection{Core-Set Selection}

A widely used diversity-based sampling strategy is \textbf{core-set selection}, which identifies a subset of data points that effectively approximates the entire dataset. The core set is designed to cover the input space comprehensively, minimizing redundancy while preserving the diversity of the dataset. This makes core-set selection particularly efficient for tasks where labeled data must represent the global structure of the dataset.

\begin{definition}[Core-set strategy]
Formally, let $\mathcal{X} = \{\mathbf{x}_1, \mathbf{x}_2, \ldots, \mathbf{x}_N\}$ represent the unlabeled dataset, and let $\mathcal{S} \subseteq \mathcal{X}$ denote the core-set of size $k$. The objective is to minimize the maximum distance between any point in $\mathcal{X}$ and its closest representative in $\mathcal{S}$. This can be expressed as
\[
\min\limits_{|\mathcal{S}| = k} \, \max\limits_{\mathbf{x} \in \mathcal{X}} \, \min\limits_{\mathbf{s} \in \mathcal{S}} \|\mathbf{x} - \mathbf{s}\|_2 ,
\]
where $\|\mathbf{x} - \mathbf{s}\|_2$ denotes the Euclidean distance between points $\mathbf{x}$ and $\mathbf{s}$. This objective ensures that every point in the dataset is adequately represented by the core-set, thereby reducing the risk of neglecting critical regions of the input space.
\end{definition}

\subsubsection{Discriminative Active Learning}

While traditional AL strategies focus on uncertainty or clustering-based diversity, \textbf{discriminative active learning (DAL)} \cite{gissin-shwartz-2019-discriminative} takes a different approach by directly modeling the distinction between labeled and unlabeled data. Rather than selecting instances based on confidence scores or representative sampling, DAL trains a discriminator to identify examples that are most likely to belong to the unlabeled set. By prioritizing these distinct instances, DAL ensures that newly labeled data introduces novel information, thereby enhancing dataset diversity and improving model generalization.

\begin{definition}[Discriminative AL]
Formally, let $\mathcal{L}$ represent the labeled dataset and $\mathcal{U}$ the pool of unlabeled instances. DAL trains a discriminator $D_{\boldsymbol{\theta}}(\mathbf{x})$ that estimates the probability of an instance $\mathbf{x}$ belonging to the unlabeled set:
\[
D_{\boldsymbol{\theta}}(\mathbf{x}) = P_{\boldsymbol{\theta}}(\mathbf{x} \in \mathcal{U}).
\]
The selection process prioritizes the most confidently unlabeled instances:
\[
\argmax_{\mathbf{x} \in \mathcal{U}} D_{\boldsymbol{\theta}}(\mathbf{x}).
\]
This approach ensures that the selected examples are those that differ the most from the labeled set, effectively expanding the decision boundary and improving model generalization.
\end{definition}

Discriminative selection offers a compelling alternative to traditional uncertainty sampling by directly modeling the distributional discrepancy between labeled and unlabeled data. By focusing on samples that are least similar to the labeled set, DAL encourages diversity and reduces redundancy in data acquisition. However, its effectiveness depends on the robustness of the discriminator, which must generalize well to avoid overfitting to the labeled dataset.

\subsubsection{Caveats and Hybrid Approaches}

While diversity-based sampling ensures broad coverage of the input space, it is not without limitations. Geometric metrics, such as Euclidean distance, may not always align with the true data distribution or task-specific objectives, leading to suboptimal selections. Additionally, diversity methods often involve high computational costs due to pairwise distance calculations or optimization routines, particularly for large datasets. Furthermore, selecting diverse points does not inherently guarantee informativeness, as points that are diverse but redundant or irrelevant to the learning objective may still be chosen.

To address these challenges, hybrid methods that integrate uncertainty and diversity sampling have emerged as effective solutions. By combining the strengths of both approaches, hybrid methods balance the selection of uncertain points near decision boundaries with diverse points that capture the overall data distribution. This synergy improves model performance while maintaining efficient use of labeling resources, making it a practical choice for many real-world AL scenarios.

\section{Practical Considerations in Active Learning}
\label{sec:practical-al}

Despite its conceptual appeal, deploying AL in real-world scenarios comes with several challenges. A fundamental issue is the absence of labeled validation sets, which are typically needed for tasks such as early stopping and hyperparameter tuning. However, constructing a validation set in AL contradicts its core objective of minimizing labeling effort, making traditional model selection strategies impractical \cite{attenberg-provost-2011-inactive, lowell-etal-2019-practical}. 

Another key challenge is the evaluation of AL methods. Comparisons against random sampling baselines often occur under inconsistent training conditions, leading to confounded results that obscure the true benefits of AL. Additionally, deciding when to stop acquiring labels -- known as the AL \textit{stopping criterion} -- remains an open question. Without a reliable stopping rule, models risk incurring unnecessary labeling costs or halting before reaching optimal performance.

The absence of a \textbf{labeled validation set} further complicates model selection and evaluation. Existing solutions often rely on proxies such as model confidence or training error stability \cite{vlachos-2008-stopping, bloodgood-vijay-shanker-2009-method}. However, these approaches frequently require dataset-specific tuning and lack generalizability, limiting their practicality across diverse AL settings.

Beyond these structural challenges, integrating AL with PLMs introduces additional difficulties. Fine-tuning PLMs in low-resource scenarios -- typical for AL -- has been shown to be highly unstable, with performance sensitivity to initialization and data order compounding the problem \cite{mosbach-etal-2021-stability, dodge-etal-2020-fine}. This instability can undermine the benefits of AL, particularly when models are retrained from scratch at each iteration. Furthermore, while traditional AL strategies often rely on confidence-based or uncertainty-driven sampling, they may not fully exploit the rich internal representations learned by PLMs, limiting their effectiveness in modern NLP applications.

The advent of PLMs has significantly shifted how AL is applied in NLP. Earlier AL techniques relied on task-specific models trained from scratch at each AL step \cite{kasai-etal-2019-low, prabhu-etal-2019-sampling}. With PLMs demonstrating superior performance through fine-tuning, they have largely supplanted these models, and researchers have begun exploring how AL can be integrated into PLM-based training regimes \cite{ein-dor-etal-2020-active, schroder-etal-2022-revisiting, yuan-etal-2020-cold}.

In theory, combining the efficiency of AL with the robustness of PLMs should reduce label complexity more effectively than either method in isolation. However, this combination introduces new challenges and remains an active area of investigation. These challenges stem from factors such as the computational cost of iterative fine-tuning, the stability of uncertainty estimates, and the impact of representation shifts over multiple AL cycles. Understanding and addressing these issues is critical to fully realizing the benefits of AL in PLM-driven NLP tasks.
\chapter{Enhancing Active Learning in PLMs through Representation Smoothness}
\label{ch:beast}

The remarkable performance of PLMs across a wide range of NLP tasks has established them as indispensable tools in the field. However, this success comes at the cost of requiring vast amounts of labeled data, which are often expensive and time-consuming to obtain. AL offers a compelling solution by selectively identifying the most informative examples for labeling, thereby reducing the labeling effort while maintaining model performance.

Building on insights from representation analysis and the \jachess{} framework introduced in \Cref{ch:jachess}, this chapter examines how representation smoothness can enhance AL strategies. By leveraging the theory of Besov spaces (\Cref{sec:besov-spaces}), we develop a smoothness-informed early stopping technique \cite{jukic-snajder-2023-smooth} that stabilizes PLM fine-tuning in AL settings, eliminating the reliance on labeled validation sets.  Additionally, we introduce a task-agnostic stopping criterion based on smoothness to further optimize the AL process.

\section{Representation Smoothness Analysis}
\label{sec:beast}

This section introduces a smoothness-based analysis of representation dynamics in PLMs and its application to improving AL. First, we examine the Besov smoothness properties of PLM representations across different training regimes (cf.~\Cref{sec:besov-spaces} for details on Besov smoothness). Building on these insights, we propose an early stopping technique that does not require a labeled validation set. Finally, we introduce a smoothness-based stopping criterion for AL, and evaluate its effectiveness in minimizing label complexity.

\subsection{Experimental Setup}

We outline the experimental setup used to evaluate the proposed method through Besov smoothness analysis, including the datasets, models, AL strategies, evaluation metrics, and stopping criteria. The effectiveness of the proposed approach is empirically assessed by comparing it against baseline methods and alternative AL techniques, focusing on improvements in data efficiency, stability, and generalization in PLMs.  

Our implementation is built on PyTorch \cite{paszke-etal-2019-pytorch} and integrates the Hugging Face Transformers library \cite{wolf-etal-2020-transformers} to facilitate pre-trained model management and dataset processing.

\subsubsection{Datasets}
\label{sec:beast-datasets}

We evaluate AL methods on five datasets, chosen for their diversity in size, complexity, and number of classes. This selection allows us to analyze AL performance across varying task characteristics. To further investigate the impact of reduced complexity, we include binary versions of selected multi-class datasets.

\paragraph{Question type classification (\trec{}).}  
A six-class classification task designed to categorize questions based on their semantic type \cite{li-roth-2002-learning}. The dataset evaluates a model’s ability to recognize different types of information being requested.
\begin{example-item}[]
\textit{Who wrote *Hamlet*?} $\rightarrow$ \textbf{Human} (seeking a person’s name) \\  
\textit{What is the capital of Japan?} $\rightarrow$ \textbf{Location} (asking for a place)  
\end{example-item}  

\paragraph{Binary question type classification (\trecb{}).}  
A simplified binary version of \trec{}, focusing on the two most frequent categories: \textit{Entity} and \textit{Human}. This modification reduces label complexity while retaining the core challenge of identifying question intent.  
\begin{example-item}[]
\textit{Who painted the Mona Lisa?} $\rightarrow$ \textbf{Human} \\  
\textit{What is the largest planet in the solar system?} $\rightarrow$ \textbf{Entity}  
\end{example-item}  

\paragraph{Subjectivity classification (\subj{}).}  
A binary classification task that distinguishes subjective sentences (expressing opinions or personal judgments) from objective statements (factual descriptions) \cite{pang-lee-2004-sentimental}.  
\begin{example-item}[]
\textit{The cinematography in this film is breathtaking.} $\rightarrow$ \textbf{Subjective} \\  
\textit{Water boils at 100 degrees Celsius.} $\rightarrow$ \textbf{Objective}  
\end{example-item}  

\paragraph{AG's news classification (\agn{}).}  
A four-class dataset for news topic classification, covering categories such as \textit{World}, \textit{Sports}, \textit{Business}, and \textit{Science/Technology} \cite{zhang-etal-2015-character}. This dataset evaluates a model’s ability to discern topical differences in news articles.  
\begin{example-item}[]
\textit{Stock markets see a major drop amid economic concerns.} $\rightarrow$ \textbf{Business} \\  
\textit{Scientists discover a new exoplanet in a distant galaxy.} $\rightarrow$ \textbf{Science/Technology}  
\end{example-item}  

\paragraph{Binary AG's news classification (\agnb{}).}  
A binary version of \agn{}, focusing on distinguishing between \textit{World} and \textit{Sports} news articles. This simplification reduces the complexity of the classification task while preserving real-world topical distinctions.  
\begin{example-item}[]
\textit{Germany elects a new chancellor after national elections.} $\rightarrow$ \textbf{World} \\  
\textit{The Lakers secure a last-minute victory in the championship game.} $\rightarrow$ \textbf{Sports}  
\end{example-item}  

By evaluating AL performance across these datasets, we gain insights into how different selection strategies impact learning efficiency across varying task complexities.

\subsubsection{Models and AL Methods}  

As introduced in \Cref{sec:plm-selection}, our experiments utilize \bert{} and \electra{}, two widely adopted encoder-based PLMs. We employ their \textit{base} variants, each with 12 layers, as implemented in the Hugging Face library \cite{wolf-etal-2020-transformers}.

We evaluate five established AL strategies alongside our novel approach. The first five methods, which are described in detail in \Cref{sec:al-methods}, include both uncertainty- and diversity-based techniques. As a baseline, we use \textbf{random selection (\rnd{})}, which selects instances uniformly at random from the unlabeled pool. Among uncertainty-based methods, we employ \textbf{maximum entropy (\ent{})} \cite{lewis-gale-1994-sequential}, which queries instances with the highest prediction entropy, reflecting low model confidence, and \textbf{Monte Carlo dropout (\mc{})} \cite{gal-ghahramani-2016-dropout}, which extends entropy-based sampling by performing multiple stochastic forward passes with dropout enabled to better estimate uncertainty. On the diversity-based side, we use \textbf{core-set selection (\cs{})} \cite{sener-savarese-2018-active}, which selects instances to minimize the maximum distance between unlabeled and labeled examples, ensuring a representative coverage of the input space. Additionally, we employ \textbf{discriminative active learning (\dal{})} \cite{gissin-shwartz-2019-discriminative}, which trains a binary classifier to differentiate labeled from unlabeled data and selects instances most likely to belong to the unlabeled set.

In addition to these established methods, we propose a novel approach called \textbf{representation gradients (\rg{})}, which selects instances based on their impact on the model’s learned representations. Unlike traditional uncertainty-based approaches that rely on prediction confidence, \rg{} ranks data points by the magnitude of their representation gradients. Specifically, it identifies instances that induce the largest changes in the model’s mean representation, selecting data points where the gradient norm is maximized:
\[
\argmax_{\mathbf{x} \in \mathcal{U}} \| \nabla_{\mathbf{x}} \bar{\mathbf{h}} \|_2,
\]
where $\bar{\mathbf{h}}$ denotes the mean representation of the model, and $\mathcal{U}$ represents the unlabeled dataset. 

By prioritizing instances that lead to the most significant representational updates, \rg{} encourages a more effective learning process, ensuring that queried examples provide meaningful contributions to refining decision boundaries. This approach reduces label complexity while maintaining computational efficiency, making it a compelling alternative to traditional AL strategies.

Each AL experiment begins with a warm start of 100 randomly sampled labeled examples. At each iteration, 50 additional examples are selected and labeled. We cap the labeling budget at 1,000 instances for simpler datasets (\trecb{}, \agnb{}, \subj{}) and 2,000 instances for more complex datasets (\trec{}, \agn{}).

\subsubsection{Evaluation Metrics}
\label{sec:al-eval}
To assess the effectiveness of AL strategies, we employ two complementary evaluation metrics that capture both overall performance and labeling efficiency.  

The first metric, \textbf{area under the performance curve (\auc{})}, measures the cumulative classification performance across AL iterations. We use the $F_1$ score to evaluate model performance at each iteration and compute the area under the curve to summarize the overall efficiency of an AL strategy. This provides a holistic view of how well the model improves over successive querying steps, offering insight into the long-term benefits of each method.  

In addition to performance-based evaluation, we also measure labeling efficiency using \textbf{label complexity reduction (\lcr{})}, a metric we introduced in \cite{jukic-snajder-2023-smooth}. \lcr{} quantifies the number of additional labels that random selection would require to match the performance of a given AL strategy at a specific iteration. By comparing the labeling effort needed to achieve the same level of accuracy, \lcr{} provides a practical assessment of how well an AL strategy reduces annotation costs while maintaining model effectiveness.

\subsection{Exploring the Impact of Training Regimes on PLM Smoothness}

To investigate the relationship between representation smoothness and AL performance, we measure the Besov smoothness of PLMs throughout the AL process under three training regimes:
\begin{enumerate}
    \item \textbf{Short training (\st{})}: Models trained for 5 epochs;
    \item \textbf{Extended training (\et{})}: Models trained for 15 epochs;
    \item \textbf{Task-adaptive pre-training (\etar{})}: TAPT (cf.~\Cref{sec:plms}) followed by extended training for 15 epochs.
\end{enumerate}
Smoothness scores are computed for each PLM layer during training, averaged across AL steps. To approximate the Besov smoothness, we use a random forest-based method that leverages wavelet decomposition to estimate smoothness indices \cite{elisha-dekel-2016-wavelet} (described in \Cref{sec:besov-spaces}), providing a practical and computationally feasible alternative to directly computing the Besov seminorm.

\begin{figure}[t]
\centering
\includegraphics[width=\linewidth]{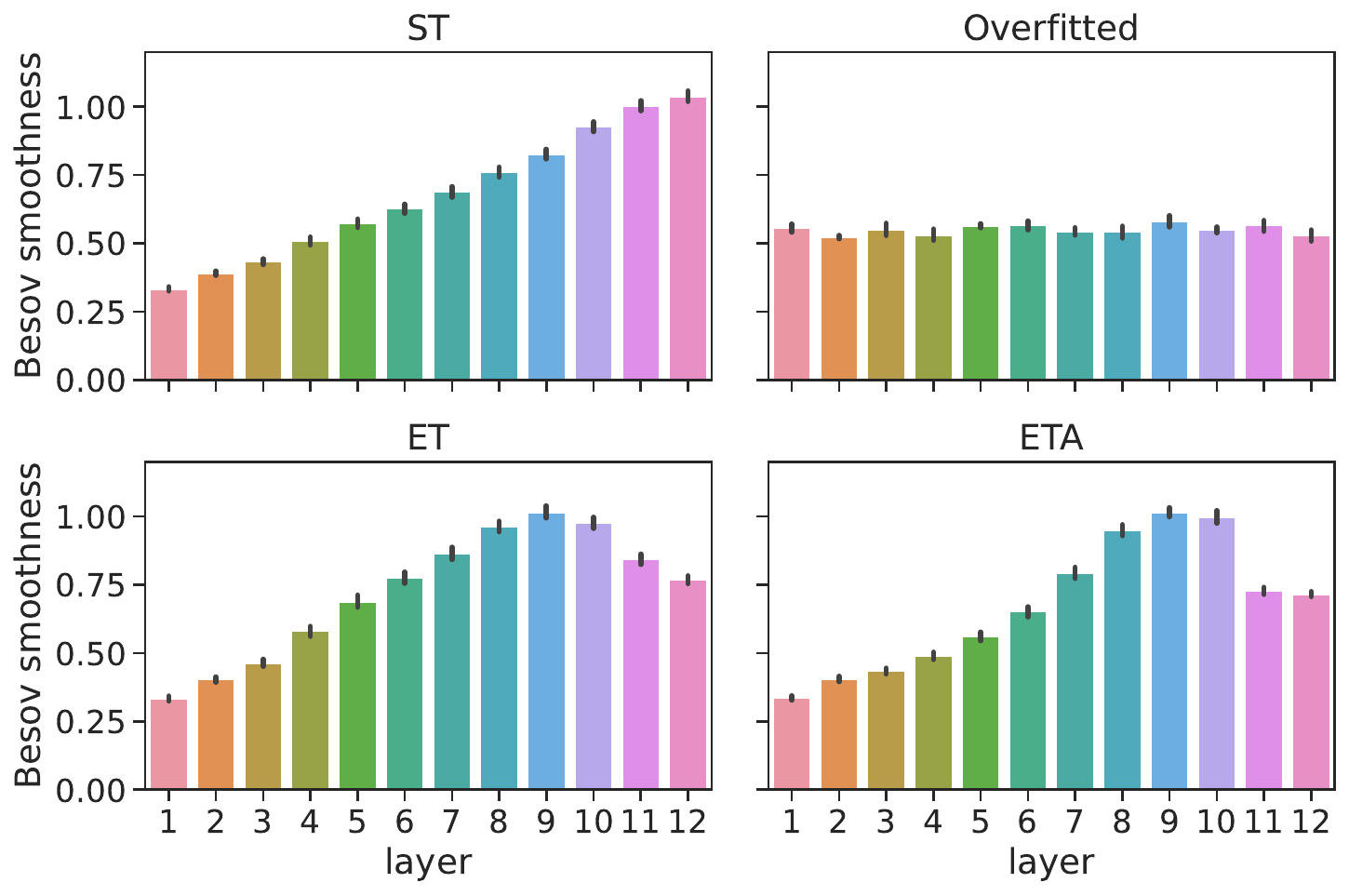}
\caption[\beast{} -- Besov smoothness across layers]{Normalized Besov smoothness of PLM layers across training regimes. Black error bars denote standard deviation, which is small across datasets, models, and AL methods.}
\label{fig:besov-layers}
\end{figure}

Figure~\ref{fig:besov-layers} visualizes the layer-wise Besov smoothness distributions across training regimes, highlighting the impact of training duration and strategies on the representation dynamics of PLMs. The figure includes results for three regimes -- short training (\st{}), extended training (\et{}), and task-adaptive pre-training followed by extended training (\etar{}) -- as well as an overfitting scenario where models are trained for 100 epochs.

Key observations from Figure~\ref{fig:besov-layers} reveal distinct patterns in smoothness progression across training regimes. In the short training regime (\st{}), smoothness increases monotonically through the layers. This suggests that deeper layers compensate for insufficient training by overly smoothing representations. In contrast, extended training (\et{}) and task-adaptive pre-training (\etar{}) produce a shift in the smoothness peak toward earlier layers, reflecting improved representation generalization and more effective layer-wise specialization. Overfitted models, trained for 100 epochs, exhibit flat smoothness distributions, indicating a loss of hierarchical representation dynamics.

In low-resource settings, deeper layers display disproportionately high smoothness, which can be attributed to reliance on spurious correlations -- a phenomenon referred to as heuristic memorization \cite{bansal-etal-2022-measures}. This reliance on shortcuts undermines generalization, leading to poor performance and limiting the efficacy of AL strategies. These observations highlight the importance of balanced training regimes that foster appropriate layer-wise smoothness to enhance both generalization and robustness. Task-adaptive pre-training (\etar{}), in particular, demonstrates a superior ability to achieve this balance, ultimately enabling more effective AL.

\section{Besov Early Stopping}

In AL, the generalization properties of the acquisition model are critical for ensuring effective querying. Poor generalization propagates through AL steps, resulting in suboptimal performance. To mitigate overfitting and preserve generalization, early stopping is a widely used regularization strategy. However, the absence of a labeled validation set in realistic AL scenarios makes traditional early stopping impractical, necessitating an alternative approach.

To address this challenge, we propose \textbf{\beast{}} (\textbf{B}esov \textbf{ea}rly \textbf{st}opping) \cite{jukic-snajder-2023-smooth}, a novel early stopping method based on representation smoothness. Our findings on the layer-wise smoothness distribution of PLMs under different training regimes motivate \beast{}, which determines the stopping point by monitoring the skewness of the smoothness distribution, measured using the Fisher-Pearson coefficient. \beast{} terminates training at the epoch where the skewness stops increasing, indicating that the peak of smoothness no longer shifts toward earlier layers over two consecutive epochs. The model is then reverted to the last epoch where the shift was preserved, effectively preempting overfitting and avoiding the flattened smoothness distributions characteristic of overtrained models.

We evaluate \beast{} in two additional training regimes: \etb{} and \etab{}, which extend \et{} and \etar{} by incorporating our proposed stopping criterion. By leveraging smoothness-based early stopping, \beast{} enhances generalization in AL scenarios without requiring a labeled validation set, making it a practical and effective alternative to traditional early stopping strategies.

\begin{figure}[t]
\centering
\includegraphics[width=0.85\linewidth]{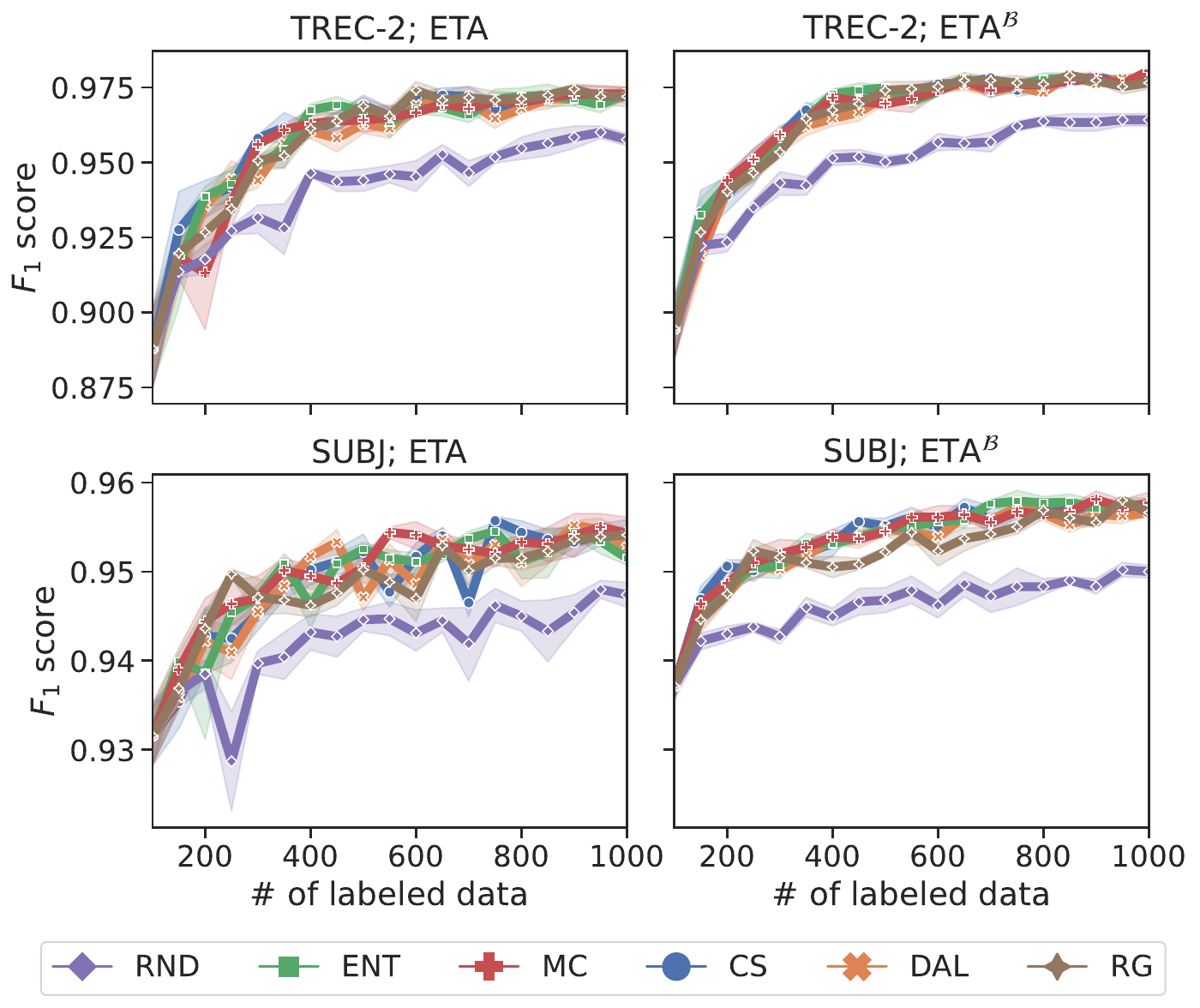}
\caption[\beast{} -- AL performance curves]{Active learning performance curves for \bert{} under \etar{} and \etab{}. Random sampling serves as a baseline. Results are averaged over five runs; confidence intervals indicate standard deviation.}
\label{fig:al-plots}
\end{figure}

\Cref{fig:al-plots} presents AL performance curves for \bert{} under \etar{} and \etab{}, with random sampling as the baseline. Results averaged over five runs show that AL with \st{} performs poorly, often failing to outperform random sampling. Extended training (\et{}) improves performance, with AL occasionally exceeding random sampling. \etar{} and \etab{} deliver consistent gains, surpassing random sampling across all datasets and AL methods.

\Cref{tab:auc,tab:app-auc} provide detailed comparisons using \auc{} as the performance metric. For \bert{}, \etar{} achieves statistically significant improvements over random sampling in 22 of 25 cases, while \etab{} achieves this in all 25 cases. \etb{} and \etab{} outperform their counterparts without \beast{}, reducing result variance (\Cref{tab:std}).

Even in standard fine-tuning without AL, \beast{} enhances performance. When paired with random sampling, it consistently yields higher scores compared to selecting the model from the final epoch.

These findings support the hypothesis that the choice of the AL method is less influential than the training regime. When AL surpasses random sampling, similar results are observed across methods. TAPT consistently enhances AL performance, and \beast{} ensures both practicability and effectiveness, achieving robust improvements across datasets and training setups.

\begin{table}[]
\caption[\auc{} for random sampling and different AL methods across datasets and training regimes for \electra{}]{\auc{} scores for random sampling and different AL methods across datasets and training regimes for \bert{}. The results are averaged over $5$ runs with different seeds. \textbf{Bold} numbers indicate the best \auc{} for each dataset. The ``$\dagger$'' symbol indicates when the mean \auc{} of an AL method is significantly different from random sampling (two-sided Man-Whitney U test with $p<.05$, adjusted for family-wise error rate with the Holm-Bonferroni method).}
\centering
\small
\begin{tabular}{lrrrrrrr}
\toprule
& & \rnd{} & \ent{} & \mc{} & \cs{} & \dal{} & \rg{} \\
\midrule
\multirow{5}{*}{\rotatebox[origin=c]{90}{\trecb{}}}
& \st{} & $.875$ & $.873$ & $.883$ & $.881$ & $.889$ & $.879$ \\
& \et{} & $.912$ & $.932$\nospacetext{$^\dagger$} & $.934$\nospacetext{$^\dagger$} & $.929$ & $.931$ & $.931$\\
& \etbns{} & $.925$ & $.942$\nospacetext{$^\dagger$} & $.942$\nospacetext{$^\dagger$} & $.939$\nospacetext{$^\dagger$} & $.940$\nospacetext{$^\dagger$} & $.938$\\
& \etar{} & $.941$ & $.959$\nospacetext{$^\dagger$} & $.957$\nospacetext{$^\dagger$} & $.960$\nospacetext{$^\dagger$} & $.957$\nospacetext{$^\dagger$} & $.958$\nospacetext{$^\dagger$}\\
& \etabns{}
& $.949$ & $\mathbf{.966}$\nospacetext{$^\dagger$} & $.965$\nospacetext{$^\dagger$} & $.965$\nospacetext{$^\dagger$} & $.964$\nospacetext{$^\dagger$} & $.965$\nospacetext{$^\dagger$}\\
\midrule
\multirow{5}{*}{\rotatebox[origin=c]{90}{\subj{}}}
& \st{} & $.896$ & $.892$ & $.885$ & $.901$ & $.898$ & $.892$\\
& \et{} & $.920$ & $.922$ & $.922$ & $.925$ & $.925$ & $.920$\\
& \etbns{} & $.928$ & $.931$ & $.932$ & $.932$ & $.933$ & $.930$\\
& \etar{} & $.942$ & $.949$\nospacetext{$^\dagger$} & $.950$\nospacetext{$^\dagger$} & $.949$\nospacetext{$^\dagger$} & $.949$\nospacetext{$^\dagger$} & $.948$\\
& \etabns{} & $.946$ & $\mathbf{.954}$\nospacetext{$^\dagger$} & $\mathbf{.954}$\nospacetext{$^\dagger$} & $\mathbf{954}$\nospacetext{$^\dagger$} & $.953$\nospacetext{$^\dagger$} & $.952$\nospacetext{$^\dagger$}\\
\midrule
\multirow{5}{*}{\rotatebox[origin=c]{90}{\agnb{}}}
& \st{} & $.923$ & $.942$\nospacetext{$^\dagger$} & $.941$\nospacetext{$^\dagger$} & $.922$ & $.941$\nospacetext{$^\dagger$} & $.942$\nospacetext{$^\dagger$}\\
& \et{} & $.960$ & $.969$ & $.970$\nospacetext{$^\dagger$} & $.965$ & $.967$ & $.969$\\
& \etbns{} & $.967$ & $.974$\nospacetext{$^\dagger$} & $.975$\nospacetext{$^\dagger$} & $.972$ & $.974$\nospacetext{$^\dagger$} & $.975$\nospacetext{$^\dagger$}\\
& \etar{} & $.974$ & $.981$\nospacetext{$^\dagger$} & $.980$ & $.981$\nospacetext{$^\dagger$} & $.980$ & $.980$\nospacetext{$^\dagger$}\\
& \etabns{} & $.977$ & $\mathbf{.983}$\nospacetext{$^\dagger$} & $\mathbf{.983}$\nospacetext{$^\dagger$} & $\mathbf{.983}$\nospacetext{$^\dagger$} & $.982$\nospacetext{$^\dagger$} & $.982$\nospacetext{$^\dagger$}\\
\midrule
\multirow{5}{*}{\rotatebox[origin=c]{90}{\trec{}}}
& \st{} & $.706$ & $.743$\nospacetext{$^\dagger$} & $.749$\nospacetext{$^\dagger$} & $.666$ & $.689$ & $.693$\\
& \et{} & $.867$ & $.878$ & $.881$\nospacetext{$^\dagger$} & $.867$ & $.878$ & $.867$\\
& \etbns{} & $.873$ & $.885$ & $.890$\nospacetext{$^\dagger$} & $.873$ & $.882$ & $.875$\\
& \etar{} & $.909$ & $.933$\nospacetext{$^\dagger$} & $.931$\nospacetext{$^\dagger$} & $.931$\nospacetext{$^\dagger$} & $.934$\nospacetext{$^\dagger$} & $.930$\nospacetext{$^\dagger$}\\
& \etabns{} & $.925$ & $.939$\nospacetext{$^\dagger$} & $.937$\nospacetext{$^\dagger$} & $.936$\nospacetext{$^\dagger$} & $\mathbf{.940}$\nospacetext{$^\dagger$} & $.935$\nospacetext{$^\dagger$}\\
\midrule
\multirow{5}{*}{\rotatebox[origin=c]{90}{\agn{}}}
& \st{} & $.837$ & $.828$ & $.824$ & $.801$ & $.834$ & $.829$\\
& \et{} & $.869$ & $.869$ & $.871$ & $.871$ & $.880$\nospacetext{$^\dagger$} & $.875$\\
& \etbns{} & $.875$ & $.877$ & $.878$ & $.879$ & $.886$\nospacetext{$^\dagger$} & $.881$\\
& \etar{} & $.891$ & $.905$\nospacetext{$^\dagger$} & $.905$\nospacetext{$^\dagger$} & $.902$\nospacetext{$^\dagger$} & $.906$\nospacetext{$^\dagger$} & $.899$\nospacetext{$^\dagger$}\\
& \etabns{} & $.894$\nospacetext{$^\dagger$} & $.908$\nospacetext{$^\dagger$} & $.908$\nospacetext{$^\dagger$} & $.905$\nospacetext{$^\dagger$} & $\mathbf{.909}$\nospacetext{$^\dagger$} & $.903$\nospacetext{$^\dagger$}\\
\bottomrule
\end{tabular}
\label{tab:auc}
\end{table}

\begin{table}[]
\caption[\auc{} for random sampling and different AL methods across datasets and training regimes for \electra{}]{\auc{} for random sampling and different AL methods across datasets and training regimes for \electra{}. The results are averaged over five runs with different seeds.
}
\centering
\small
\begin{tabular}{lrrrrrrr}
\toprule
& & \rnd{} & \ent{} & \mc{} & \cs{} & \dal{} & \rg{} \\
\midrule
\multirow{5}{*}{\rotatebox[origin=c]{90}{\textsc{trec-2}}}
& \textsc{st} & $.831$ & $.829$ & $.836$ & $.835$ & $.840$ & $.805$\\
& \textsc{et} & $.910$ & $.924$ & $.918$ & $.928$ & $.927$ & $.919$\\
& \textsc{et\nospacetext{$^\mathcal{B}$}} & $.919$ & $.930$ & $.926$ & $.934$ & $.936$ & $.927$\\
& \textsc{eta} & $.932$ & $.953$ & $.953$ & $.953$ & $.951$ & $.949$\\
& \textsc{eta\nospacetext{$^\mathcal{B}$}} & $.939$ & $.959$ & $.958$ & $.959$ & $.956$ & $.956$\\
\midrule
\multirow{5}{*}{\rotatebox[origin=c]{90}{\textsc{subj}}}
& \textsc{st} & $.880$ & $.872$ & $.870$ & $.871$ & $.898$ & $.860$\\
& \textsc{et} & $.926$ & $.927$ & $.925$ & $.935$ & $.936$ & $.935$\\
& \textsc{et\nospacetext{$^\mathcal{B}$}} & $.938$ & $.937$ & $.934$ & $.944$ & $.946$ & $.942$\\
& \textsc{eta} & $.946$ & $.955$ & $.954$ & $.955$ & $.955$ & $.952$\\
& \textsc{eta\nospacetext{$^\mathcal{B}$}} & $.952$ & $.959$ & $.959$ & $.959$ & $.959$ & $.957$\\
\midrule
\multirow{5}{*}{\rotatebox[origin=c]{90}{\textsc{agn-2}}}
& \textsc{st} & $.867$ & $.901$ & $.891$ & $.850$ & $.850$ & $.823$\\
& \textsc{et} & $.963$ & $.963$ & $.954$ & $.963$ & $.966$ & $.965$\\
& \textsc{et\nospacetext{$^\mathcal{B}$}} & $.969$ & $.971$ & $.962$ & $.971$ & $.971$ & $.972$\\
& \textsc{eta} & $.977$ & $.981$ & $.981$ & $.982$ & $.981$ & $.980$\\
& \textsc{eta\nospacetext{$^\mathcal{B}$}} & $.980$ & $.983$ & $.983$ & $.983$ & $.982$ & $.982$\\
\midrule
\multirow{5}{*}{\rotatebox[origin=c]{90}{\textsc{trec-6}}}
& \textsc{st} & $.604$ & $.645$ & $.636$ & $.549$ & $.561$ & $.461$\\
& \textsc{et} & $.837$ & $.848$ & $.839$ & $.817$ & $.811$ & $.814$\\
& \textsc{et\nospacetext{$^\mathcal{B}$}} & $.843$ & $.858$ & $.847$ & $.821$ & $.813$ & $.816$\\
& \textsc{eta} & $.897$ & $.917$ & $.905$ & $.905$ & $.905$ & $.901$\\
& \textsc{eta\nospacetext{$^\mathcal{B}$}} & $.906$ & $.925$ & $.917$ & $.915$ & $.914$ & $.911$\\
\midrule
\multirow{5}{*}{\rotatebox[origin=c]{90}{\textsc{agn-4}}}
& \textsc{st} & $.793$ & $.706$ & $.713$ & $.688$ & $.755$ & $.750$\\
& \textsc{et} & $.857$ & $.844$ & $.824$ & $.845$ & $.866$ & $.857$\\
& \textsc{et\nospacetext{$^\mathcal{B}$}} & $.868$ & $.855$ & $.836$ & $.857$ & $.874$ & $.866$\\
& \textsc{eta} & $.888$ & $.903$ & $.901$ & $.901$ & $.904$ & $.897$\\
& \textsc{eta\nospacetext{$^\mathcal{B}$}} & $.893$ & $.907$ & $.905$ & $.905$ & $.907$ & $.901$\\
\bottomrule
\end{tabular}
\label{tab:app-auc}
\end{table}
\begin{table}[t]
\caption[Average standard deviation for different training regimes with AL]{Average standard deviation for different training regimes. The results are averaged across models and AL methods. \textbf{Bold} numbers indicate regimes with the lowest standard deviation for a particular dataset.}
\small
\centering
\begin{tabular}{lccccc}
\toprule
& \st{} & \et{} & \etar{} & \etb{} & \textsc{eta$^\mathcal{B}$} \\
\midrule
\trecb{} & $.0093$ & $.0053$ & $.0045$ & $.0026$ & $\textbf{.0022}$\\
\subj{} & $.0117$ & $.0045$ & $.0032$ & $.0013$ & $\textbf{.0008}$\\
\agnb{} & $.0100$ & $.0036$ & $.0020$ & $.0009$ & $\textbf{.0005}$\\
\trec{} & $.0134$ & $.0081$ & $.0074$ & $.0032$ & $\textbf{.0027}$\\
\agn{} & $.0118$ & $.0048$ & $.0045$ & $.0022$ & $\textbf{.0014}$\\
\bottomrule
\end{tabular}
\label{tab:std}
\end{table}

\begin{figure*}[]
\centering
\begin{subfigure}{.75\textwidth}
  \centering
  \includegraphics[width=\linewidth]{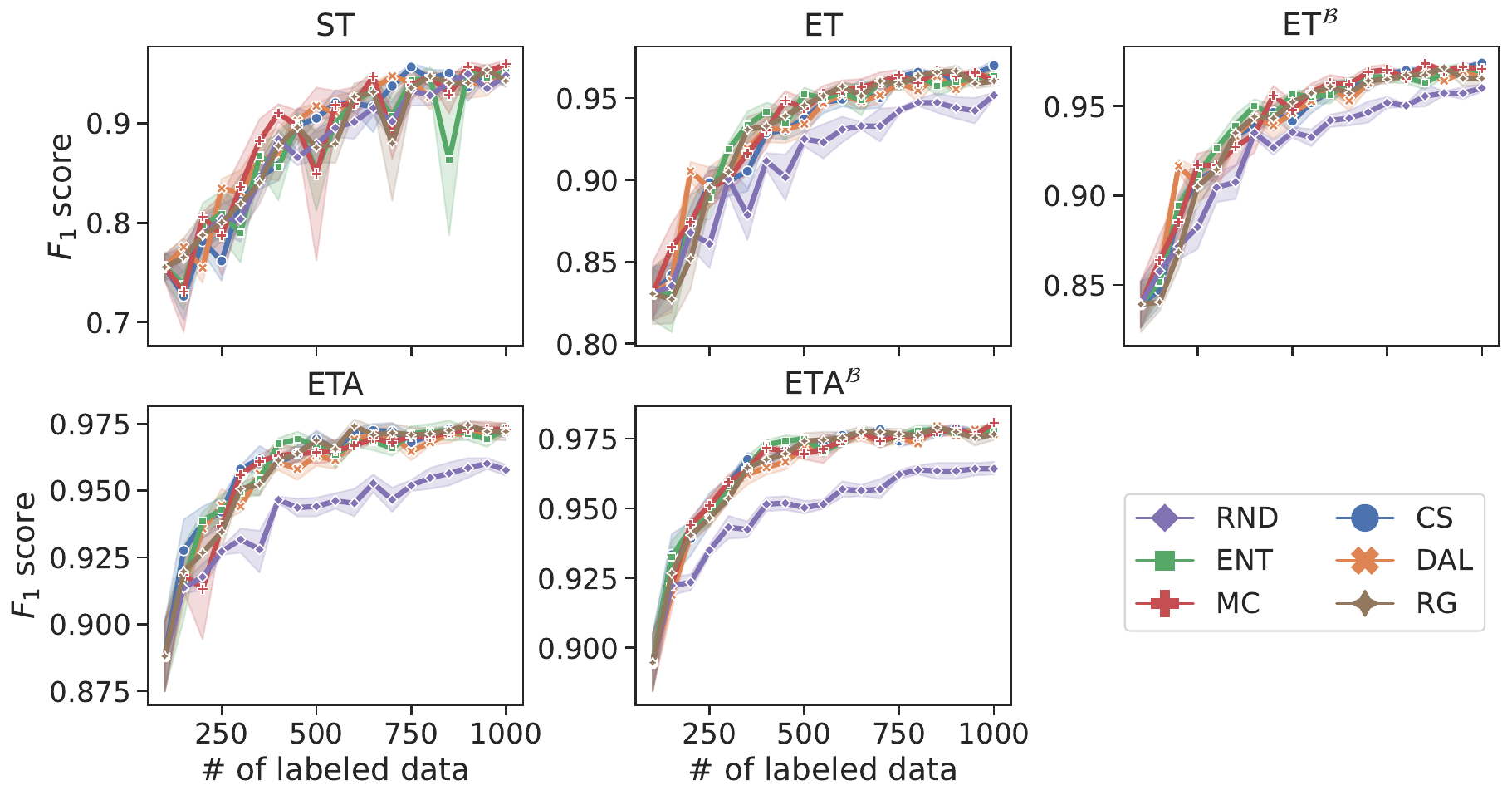}
  \caption{\textsc{trec-2}}
  \label{fig:trec-2-bert}
\end{subfigure}
\begin{subfigure}{.75\textwidth}
  \centering
  \includegraphics[width=\linewidth]{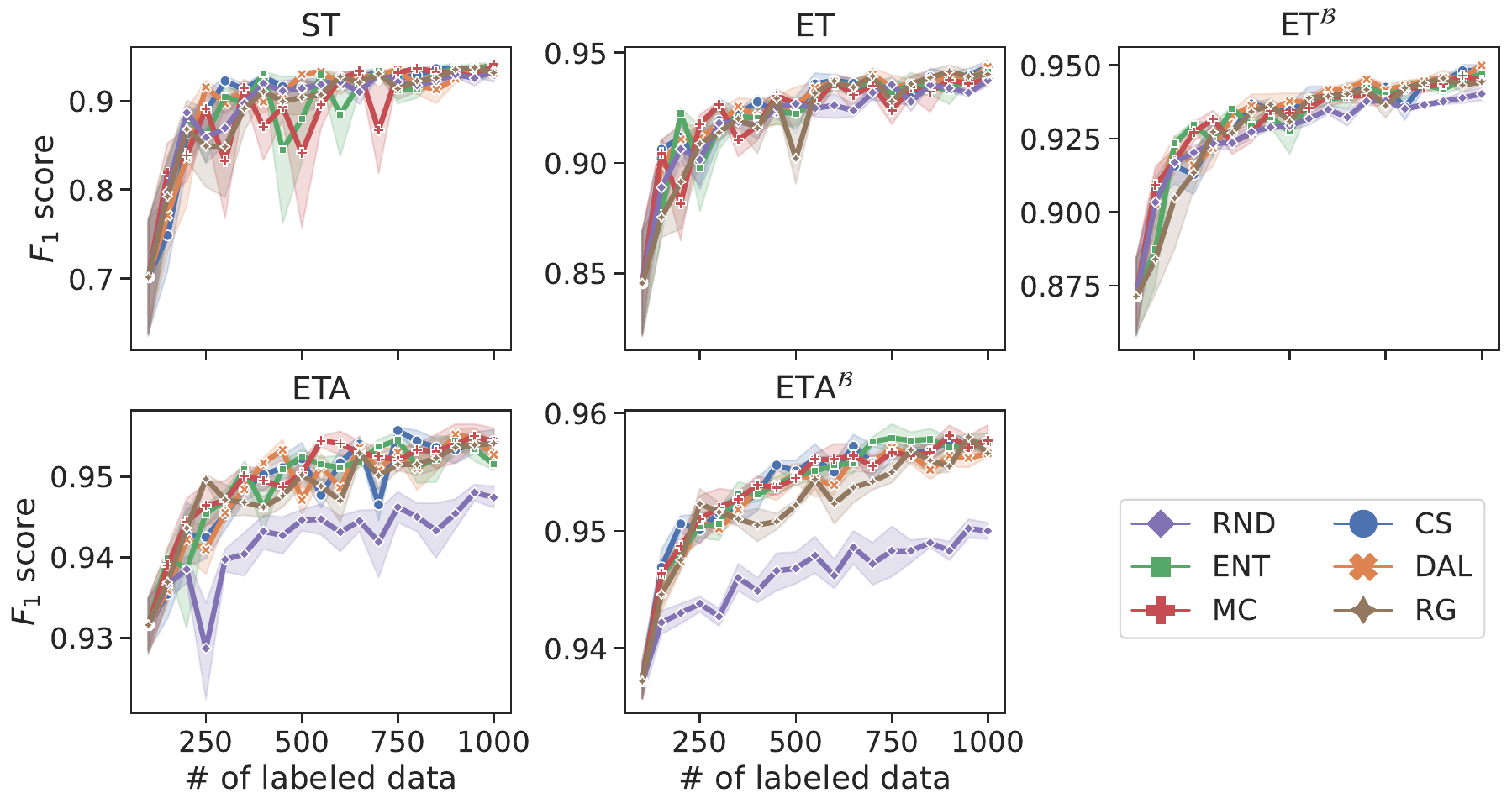}
  \caption{\textsc{subj}}
  \label{fig:subj-bert}
\end{subfigure}
\begin{subfigure}{.75\textwidth}
  \centering
  \includegraphics[width=\linewidth]{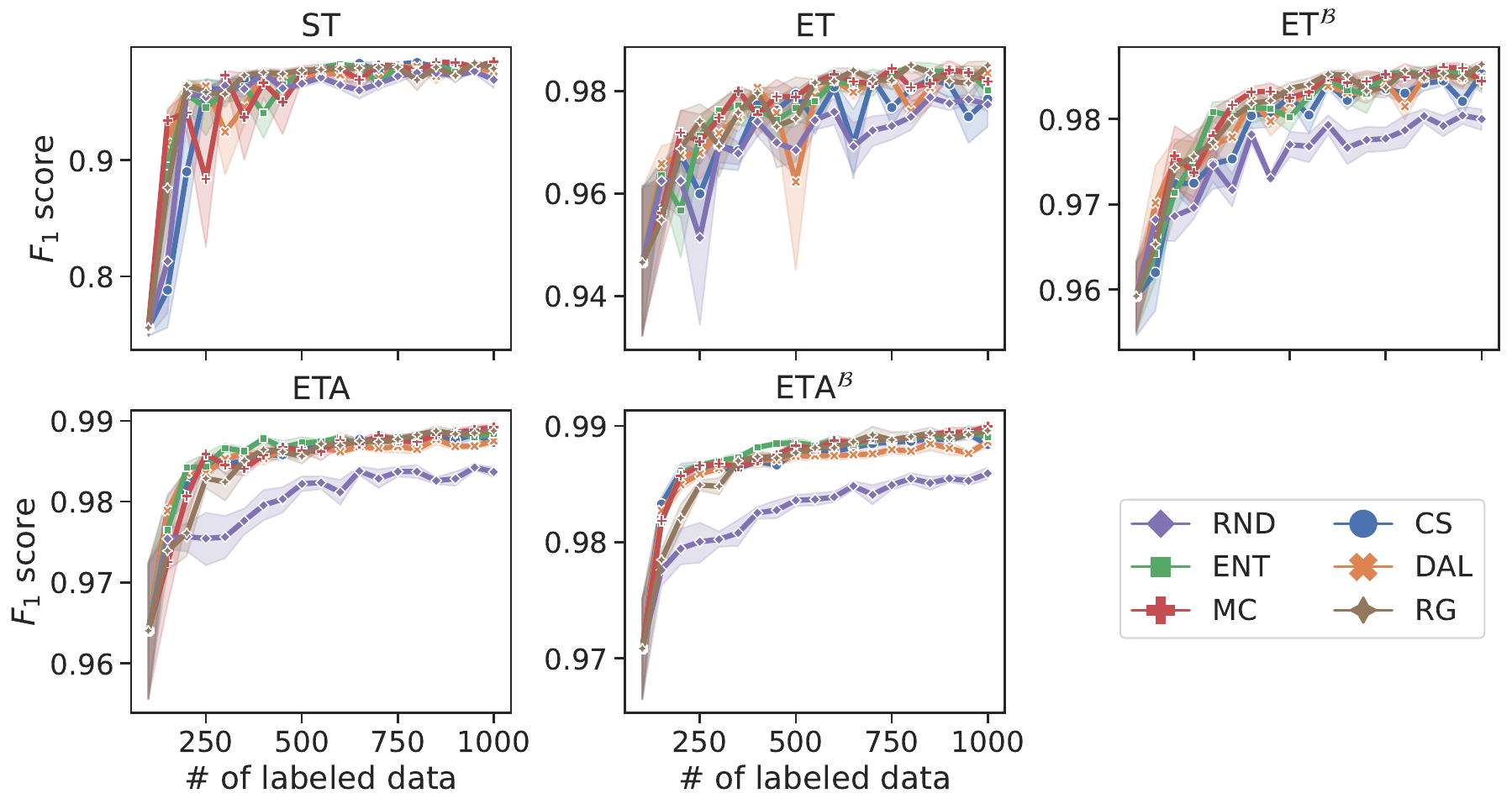}
  \caption{\textsc{agn-2}}
  \label{fig:agn-2-bert}
\end{subfigure}
\caption[\beast{} -- AL performance curves for \bert{}]{AL performance curves for different training regimes across datasets for \bert{}. Random sampling (purple rhombs) serves as a baseline. Best viewed on a computer screen.}
\label{fig:app-curves-bert}
\end{figure*}

\begin{figure*}[]
\centering
\begin{subfigure}{0.75\textwidth}
  \centering
  \includegraphics[width=\linewidth]{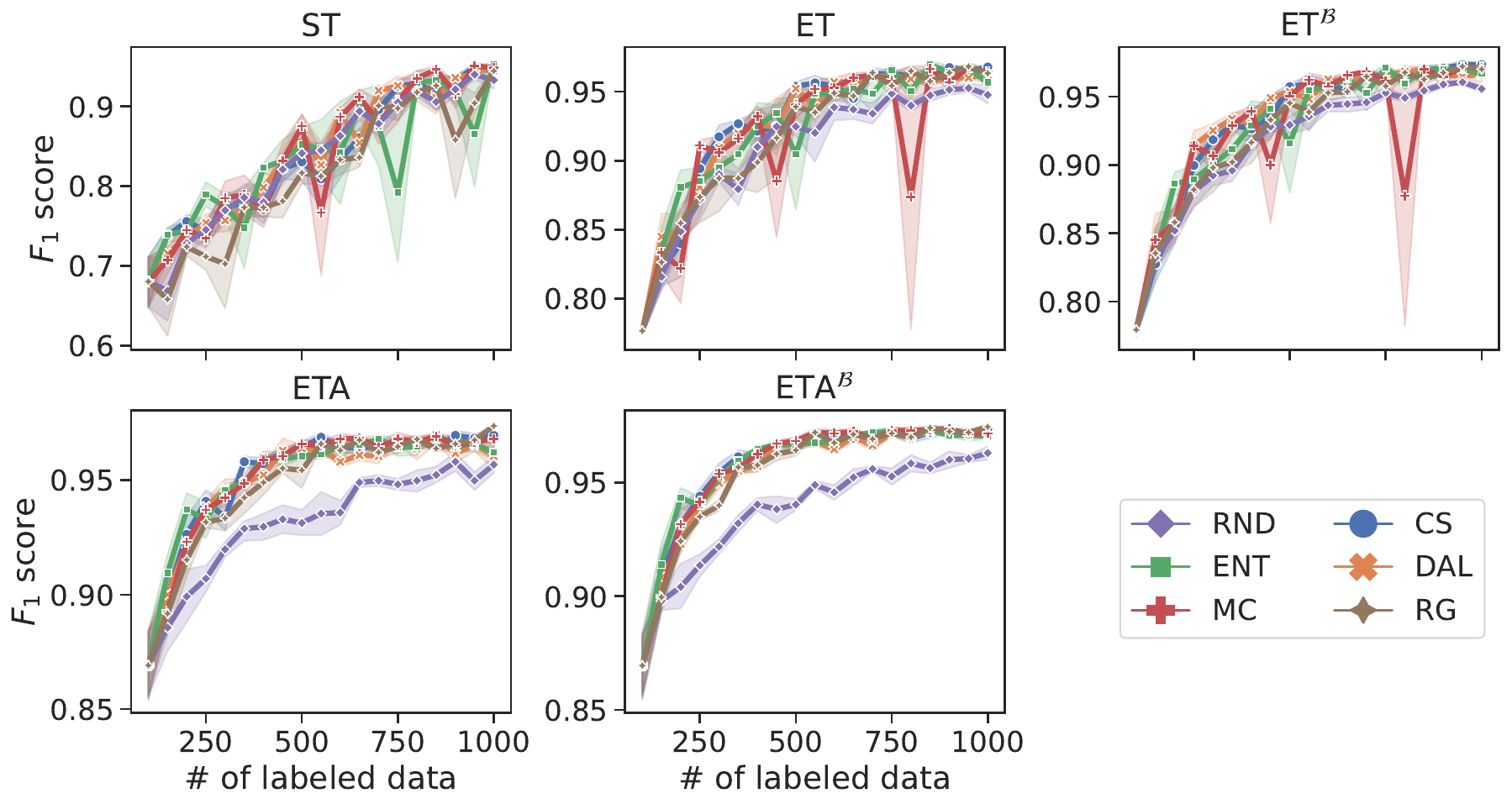}
  \caption{\textsc{trec-2}}
  \label{fig:trec-2-electra}
\end{subfigure}
\begin{subfigure}{0.75\textwidth}
  \centering
  \includegraphics[width=\linewidth]{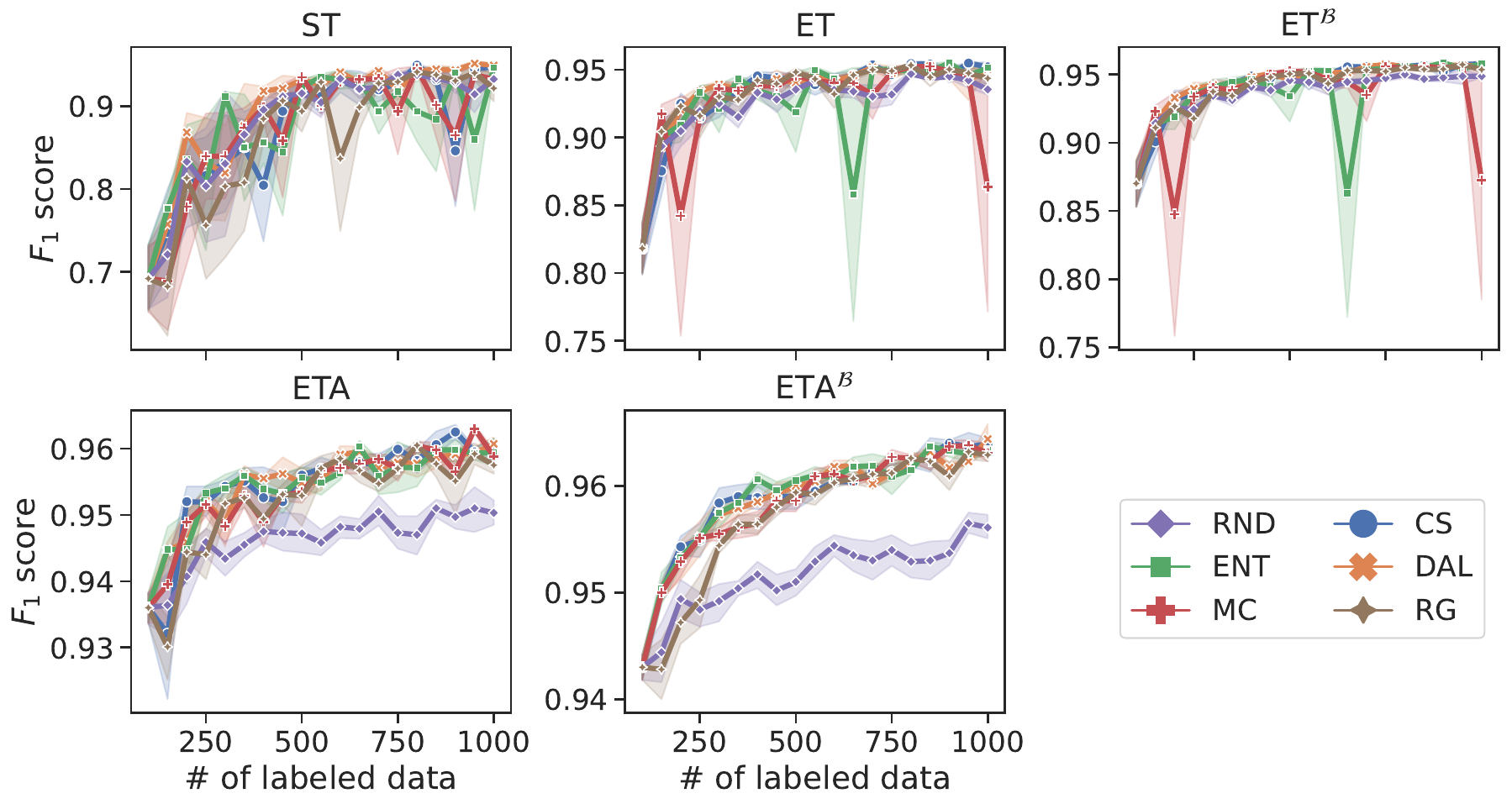}
  \caption{\textsc{subj}}
  \label{fig:subj-electra}
\end{subfigure}
\begin{subfigure}{0.75\textwidth}
  \centering
  \includegraphics[width=\linewidth]{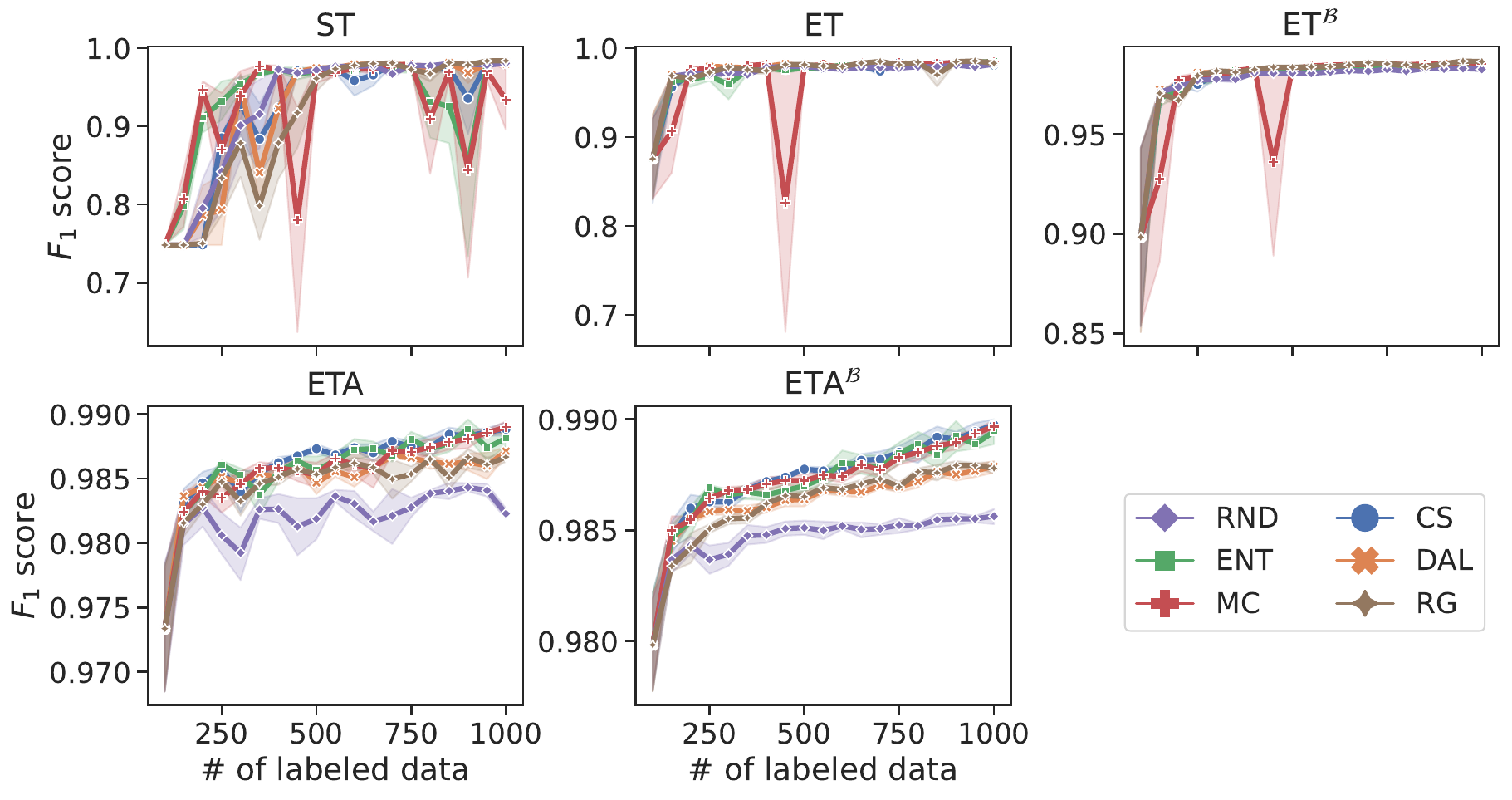}
  \caption{\textsc{agn-2}}
  \label{fig:agn-2-electra}
\end{subfigure}
\caption[\beast{} -- AL performance curves for \electra{}]{AL performance curves for different training regimes across datasets for \electra{}. Random sampling (purple rhombs) serves as a baseline. Best viewed on a computer screen.}
\label{fig:app-curves-electra}
\end{figure*}

\Cref{fig:app-curves-bert,fig:app-curves-electra} show the AL performance curves across the used datasets and for both models (\bert{} and \electra{}). For the \textsc{eta} and \textsc{eta$^\mathcal{B}$} training regimes, we observe a consistent improvement in performance compared to random sampling.

\section{Active Sample Smoothness}
\label{sec:active-smoothness}

To complement our analysis of layer-wise smoothness, we investigate the Besov smoothness of individual active samples -- unlabeled examples selected during AL. Lower smoothness values for active samples suggest higher informativeness, as these examples provide sharper representations that the model can learn from. Conversely, higher smoothness indicates diminishing returns, where the data is already well-represented by the model.

\begin{figure}[]
\small
\centering
\begin{subfigure}{.49\linewidth}
  \centering
  \includegraphics[width=\linewidth]{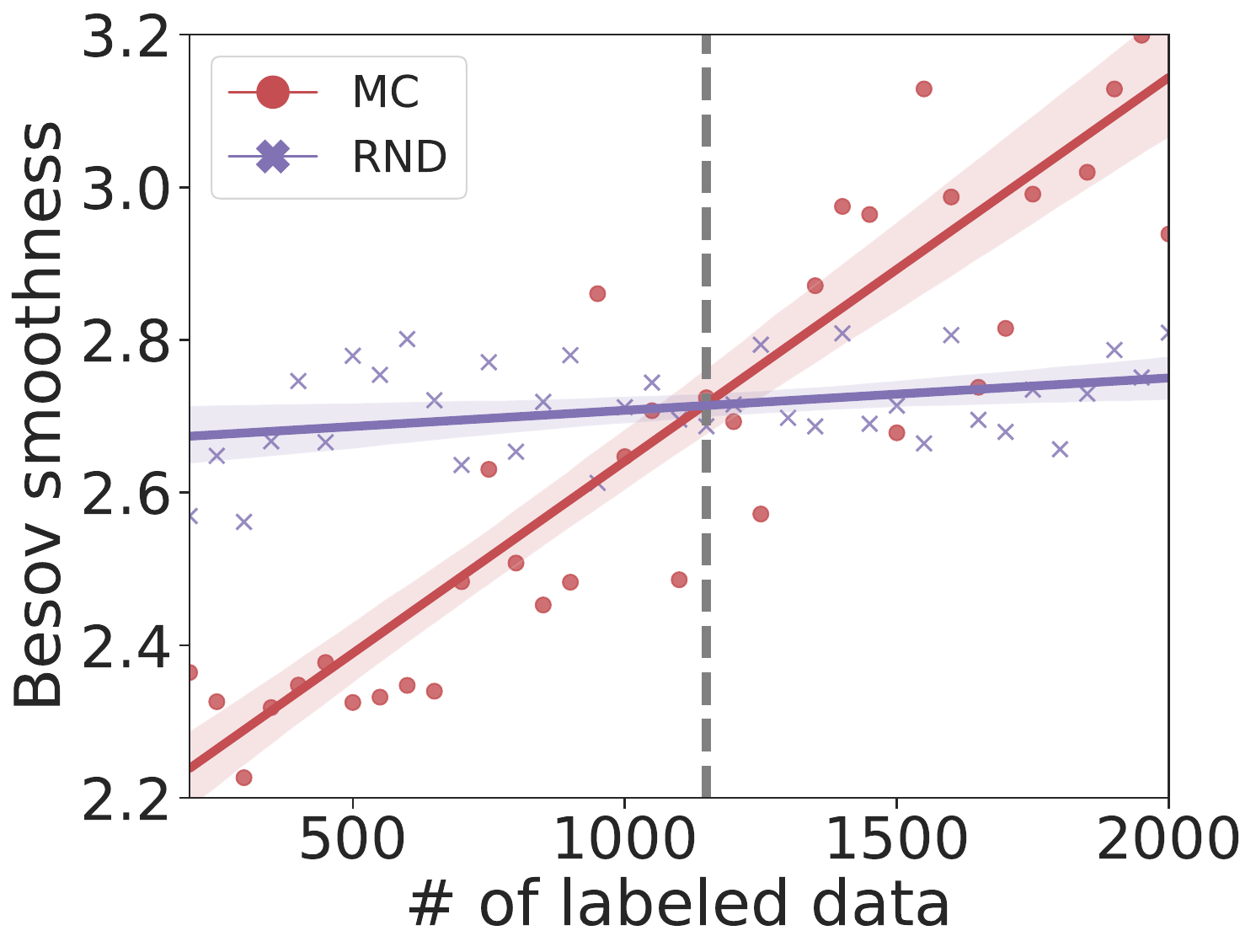}
  \caption{Sample smoothness}
  \label{fig:al-stop-smooth}
\end{subfigure}
\begin{subfigure}{.49\linewidth}
  \centering
  \includegraphics[width=\linewidth]{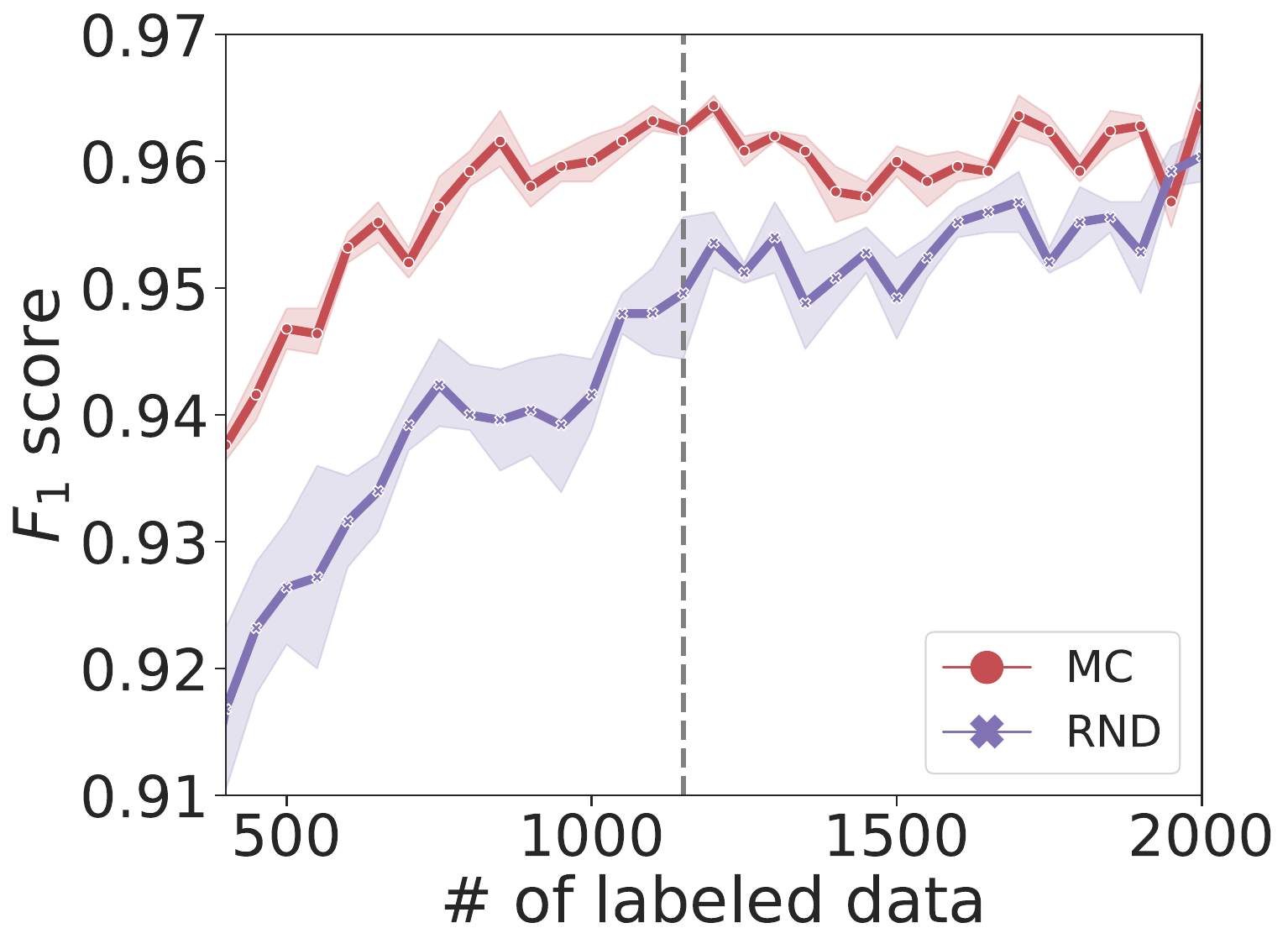}
  \caption{AL performance curve}
  \label{fig:al-stop-curve}
\end{subfigure}
\caption[Comparison of active vs.~random sample smoothness in relation to AL performance]{Relationship between active and random sample smoothness (\Cref{fig:al-stop-smooth}) and corresponding AL performance (\Cref{fig:al-stop-curve}). The gray dotted line indicates the intersection of smoothness trends, signaling diminishing returns. Results are shown for \textsc{bert} in the \etab{} regime on \trec{}; other datasets show similar trends (\Cref{fig:app-active-sample}).}
\label{fig:active-sample}
\end{figure}

\begin{figure}[]
\small
\centering
\begin{subfigure}{.49\linewidth}
  \centering
  \includegraphics[width=\linewidth]{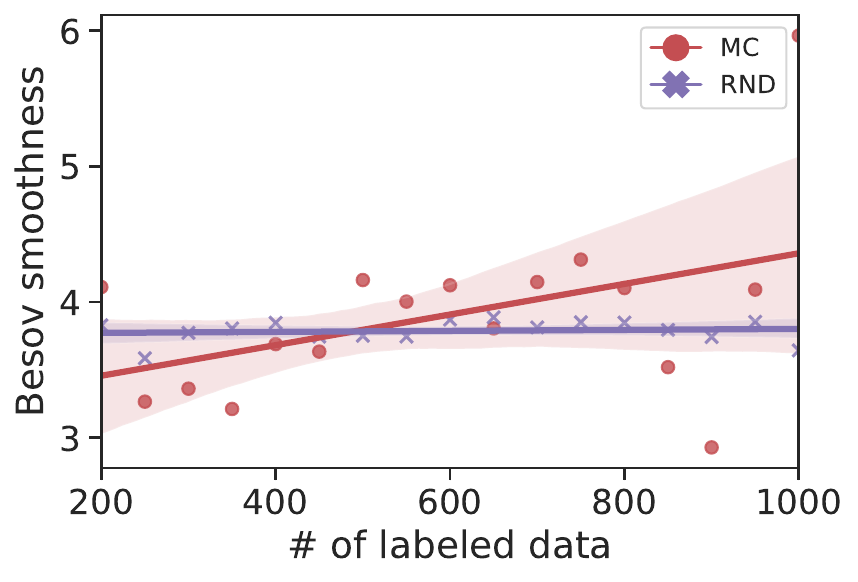}
  \subcaption{\textsc{trec-2}}
  \label{fig:as-trec-2}
\end{subfigure}
\begin{subfigure}{.49\linewidth}
  \centering
  \includegraphics[width=\linewidth]{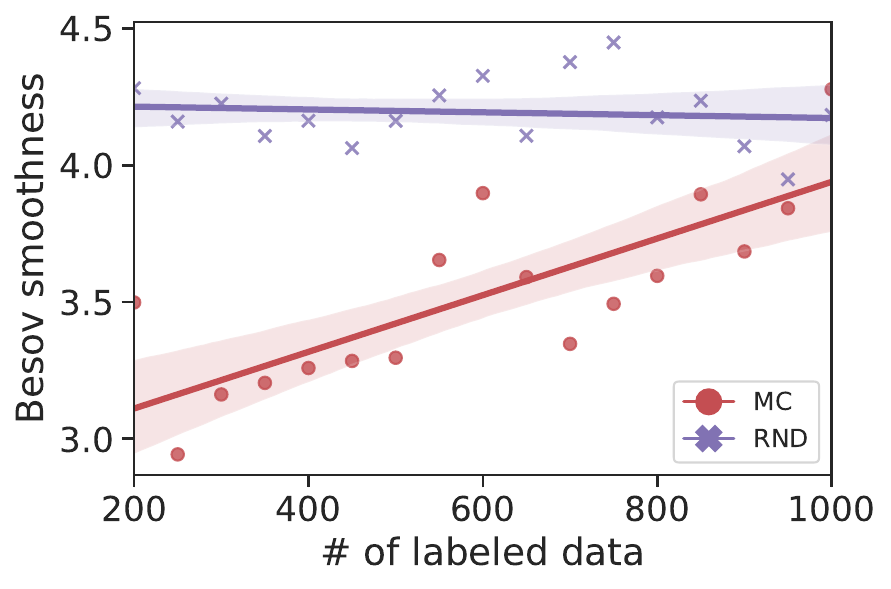}
  \subcaption{\textsc{agn-2}}
  \label{fig:alp-trec-2}
\end{subfigure}
\begin{subfigure}{.49\linewidth}
  \centering
  \includegraphics[width=\linewidth]{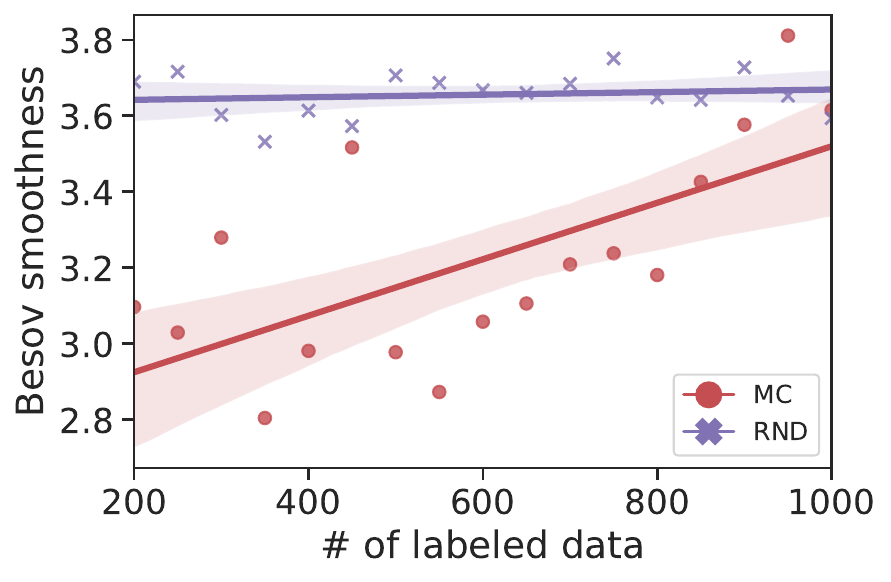}
  \subcaption{\textsc{subj}}
  \label{fig:as-subj}
\end{subfigure}
\begin{subfigure}{.49\linewidth}
  \centering
  \includegraphics[width=\linewidth]{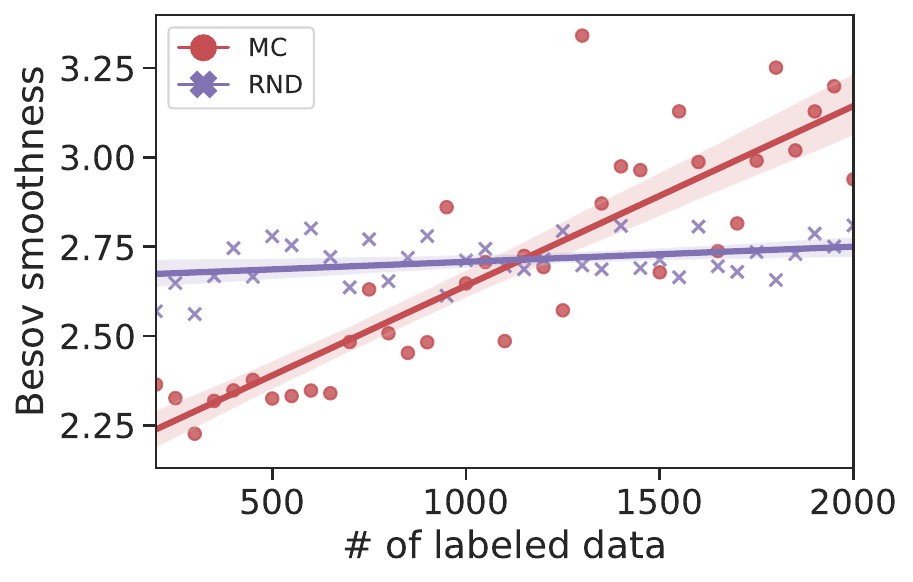}
  \subcaption{\textsc{trec-6}}
  \label{fig:alp-subj}
\end{subfigure}
\begin{subfigure}{.49\linewidth}
  \centering
  \includegraphics[width=\linewidth]{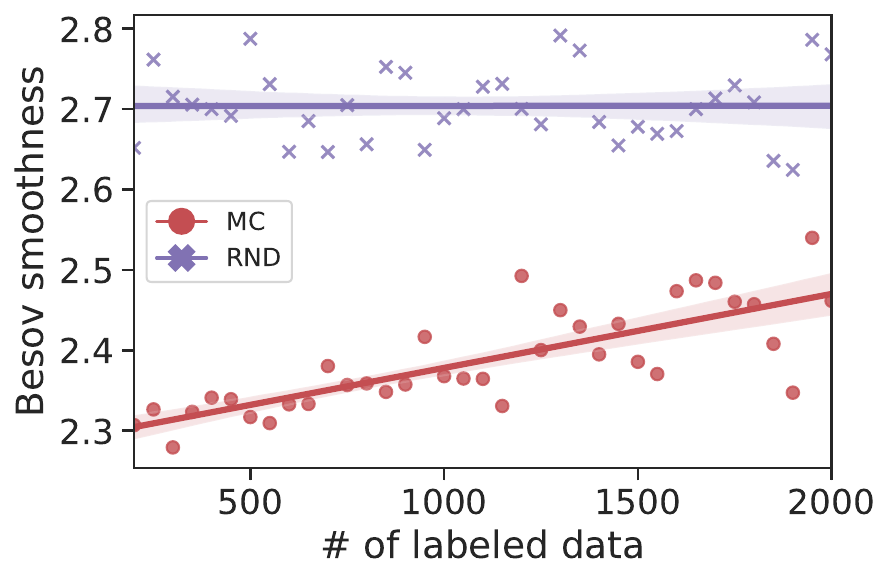}
  \subcaption{\textsc{agn-4}}
  \label{fig:alp-agn-4}
\end{subfigure}
\caption[Active sample smoothness as a stopping criterion]{Besov smoothness of actively acquired samples with \textsc{mc} (red) compared to the smoothness of random samples (purple).}
\label{fig:app-active-sample}
\end{figure}

\Cref{fig:active-sample,fig:app-active-sample} illustrate the relationship between the smoothness of actively acquired and random samples, alongside the corresponding AL performance.
The smoothness trends observed in \Cref{fig:active-sample} reveal distinct behaviors for active and random samples. Random sample smoothness remains relatively stable across AL steps, reflecting the inherent randomness of the selection process. In contrast, the smoothness of active samples starts lower in the initial steps of AL, signifying the informativeness of the selected examples. As the AL process progresses, the smoothness of active samples steadily increases, signaling that the acquisition model is effectively learning from the most informative data. This increase in smoothness eventually aligns with the stable smoothness of random samples, marked by the dotted line, which we interpret as the point of \textit{information depletion}. Beyond this point, actively acquired examples provide minimal additional value, making further labeling inefficient.

Inspired by these observations, we propose \alsbi{} (\textbf{a}ctive \textbf{l}earning \textbf{s}topping by \textbf{B}esov \textbf{i}ndex), a stopping criterion designed to identify the point of information depletion in AL. \alsbi{} terminates the AL process when the average Besov smoothness of actively acquired samples surpasses that of random samples for two consecutive steps. To estimate random sample smoothness in practical settings, we use bootstrapping on warm start examples as a reliable proxy. By halting AL at the point where active samples no longer provide meaningful insights, \alsbi{} ensures that labeling resources are utilized efficiently without compromising performance. The bootstrapped random sample smoothness estimate remains stable across AL steps, with experiments confirming that using $1,000$ bootstrapped samples of size $50$ yields consistent results.

\Cref{tab:label-complexity} quantitatively validates the benefits of \alsbi{}. The table shows the average \lcr{} across datasets and models, reflecting the proportion of the dataset that needs to be labeled for random sampling to achieve the same performance as AL methods. \alsbi{} consistently outperforms the average \lcr{} computed across all AL steps, demonstrating its effectiveness in detecting the optimal stopping point. Among the tested methods, \rg{} achieves the highest \lcr{} gains, likely due to its alignment with \alsbi{}, as both approaches leverage representation smoothness as a guiding principle.

These results affirm the utility of \alsbi{} in identifying the point of diminishing returns in AL, ensuring that resources are allocated efficiently without compromising model performance. By combining theoretical insights with empirical validation, \alsbi{} provides a practical and effective framework for improving the efficiency of AL pipelines.

\begin{table}[tb!]
\caption[Average \lcr{} for different AL methods]{Average \textsc{lcr} across datasets and models. The scores indicate the proportion of the dataset that needs to be labeled for random sampling to match the performance of the corresponding AL method. \alsbi{} is compared to an average \lcr{} throughout the AL steps (avg). The results are averaged over $5$ runs. Numbers in \textbf{bold} indicate the largest \textsc{lcr} for a certain training regime.}
\small
\centering
\begin{tabular}{lrrrrrr}
\toprule
& & \textsc{ent} & \textsc{mc}  & \textsc{cs} & \textsc{dal}  & \textsc{rg}\\
\midrule
\multirow{2}{*}{\rotatebox[origin=c]{0}{\etar{}}}
& avg & $.316$ & $.312$ & $.351$ & $.275$ & $.319$\\
& \alsbi{} & $.447$ & $.401$ & $.511$ & $.282$ & $\textbf{.521}$\\
\midrule
\multirow{2}{*}{\rotatebox[origin=c]{0}{\textsc{eta\nospacetext{$^\mathcal{B}$}}}}
& avg & $.385$ & $.385$  & $.394$ & $.354$ & $.382$ \\
& \alsbi{} & $.586$ & $.532$ & $.488$ & $.447$ & $\textbf{.645}$\\
\bottomrule
\end{tabular}
\label{tab:label-complexity}
\end{table}

\section{Summary}
Our experiments across a diverse set of NLP tasks and datasets demonstrate the substantial benefits of integrating AL with smoothness-based techniques for PLMs. These methods enhance both data efficiency and training stability, effectively addressing key challenges in low-resource settings.

One of the most notable advantages of smoothness-based approaches is their ability to improve data efficiency. By leveraging representation smoothness, PLMs achieve strong performance with fewer labeled examples compared to random selection or traditional AL strategies. This reduction in labeling requirements makes AL more practical in scenarios where annotation costs are a limiting factor.

In addition to improving data efficiency, smoothness-based methods contribute to training stability. The Besov early stopping technique (\beast{}) mitigates overfitting and ensures more consistent model performance across AL iterations. By dynamically determining an appropriate stopping point, \beast{} enhances training reliability and prevents performance degradation due to excessive fine-tuning.

Furthermore, the smoothness-based stopping criterion, \alsbi{}, significantly reduces annotation costs by optimizing the number of labeling iterations required. By identifying the point at which additional labeled data yields diminishing returns, \alsbi{} ensures that annotation resources are utilized efficiently while maintaining high model performance.

These findings highlight the second key contribution of this thesis: the development of a smoothness-driven early stopping algorithm for PLM fine-tuning that operates without labeled validation data. By incorporating representation smoothness into the training process, \beast{} provides a principled and effective method for determining optimal stopping points, further advancing the efficiency and stability of AL pipelines.

\chapter{Parameter-Efficient Learning}
\label{ch:peft}

The rapid advancements in deep learning, coupled with the evolution of language representations discussed in \Cref{ch:background}, have led to the development of increasingly complex models with billions of parameters. While these models achieve state-of-the-art performance across various NLP tasks, their substantial computational and memory demands present significant challenges, particularly in resource-constrained environments. To mitigate these limitations, \textbf{parameter-efficient learning} has emerged as a promising paradigm that optimizes resource utilization by reducing the number of trainable parameters while preserving model performance. This approach enhances the scalability, accessibility, and adaptability of deep learning, making it more feasible for a broader range of applications.

A central aspect of parameter-efficient learning is \textit{modular deep learning} \cite{pfeiffer-etal-2024-modular}, which advocates for decomposing models into smaller, reusable components. Modular architectures not only reduce computational overhead but also enable more flexible training and fine-tuning, allowing models to adapt to new tasks with minimal parameter updates. Rather than retraining entire architectures, modular approaches selectively modify or extend specific components of a model, significantly improving efficiency while maintaining strong performance.

This chapter examines various parameter-efficient learning strategies designed to enhance model adaptation while minimizing computational costs. The techniques covered include \textbf{modular architectures}, which decompose models into reusable components; \textbf{low-rank adaptations}, which constrain parameter updates to lower-dimensional subspaces to reduce redundancy; \textbf{selective fine-tuning}, where only a subset of parameters -- such as specific layers or attention heads -- is optimized; and \textbf{adapter-based methods}, which introduce small, trainable modules into pre-trained models to facilitate efficient specialization.

\section{Modularity in Deep Learning}

Modularity introduces flexibility into model design by allowing components to be independently trained, reused across tasks, or combined to form task-specific configurations. These modular compositions help address challenges in scalability and efficiency while promoting better task-specific adaptability. Below, we outline three prominent types of modular compositions: layer-wise, sequential, and parallel.

\subsection{Layer-Wise Composition}

Layer-wise composition integrates task-specific modules into specific layers of a pre-trained model. This approach is particularly effective when only certain layers require task-specific adjustments, leaving the remaining model parameters unchanged. By modifying select layers, this method balances task adaptability with parameter efficiency.

\begin{definition}[Layer-Wise Composition]
For a pre-trained model $f_{\boldsymbol{\theta}}$ and a task-specific module $g_{\boldsymbol{\phi}}$, the composite model is
\[
  f_{\boldsymbol{\theta}'} = f_{\boldsymbol{\theta}}^{(1:l)} \circ g_{\boldsymbol{\phi}} \circ f_{\boldsymbol{\theta}}^{(l+1:L)},
\]
where
\begin{itemize}
    \item $f_{\boldsymbol{\theta}}^{(1:l)}$ represents the first $l$ layers of the pre-trained model;
    \item $g_{\boldsymbol{\phi}}$ is the task-specific module inserted after layer $l$;
    \item $f_{\boldsymbol{\theta}}^{(l+1:L)}$ denotes the remaining layers of the pre-trained model.
\end{itemize}
Only $g_{\boldsymbol{\phi}}$ is trained during adaptation, ensuring that the rest of the model remains fixed, preserving computational resources.
\end{definition}

This strategy is especially useful for tasks where specific layers encode domain-agnostic features that can be reused across multiple tasks while allowing task-specific layers to adapt flexibly.

\subsection{Sequential Composition}

Sequential composition involves appending a task-specific module to the output of a pre-trained model. This method is simple yet effective, treating the pre-trained model as a feature extractor while delegating task-specific learning to the appended module.

\begin{definition}[Sequential Composition]
For a pre-trained model $f_{\boldsymbol{\theta}}$ and a task-specific module $g_{\boldsymbol{\phi}}$, the composite model is defined as
\[
  h(\mathbf{x}) = g_{\boldsymbol{\phi}}(f_{\boldsymbol{\theta}}(\mathbf{x})),
\]
where
\begin{itemize}
    \item $f_{\boldsymbol{\theta}}$ processes the input $\mathbf{x}$ to extract general-purpose features;
    \item $g_{\boldsymbol{\phi}}$ learns task-specific mappings from the features extracted by $f_{\boldsymbol{\theta}}$.
\end{itemize}
In this approach, $f_{\boldsymbol{\theta}}$ is frozen, and only $g_{\boldsymbol{\phi}}$ is fine-tuned, significantly reducing the number of trainable parameters.
\end{definition}

Sequential composition is well-suited for scenarios where the pre-trained model provides a strong general feature representation, allowing the appended module to specialize in task-specific nuances.

\subsection{Parallel Composition}

Parallel composition enables the pre-trained model and the task-specific module to process the input simultaneously, with their outputs combined in a weighted manner. This approach leverages the complementary strengths of both components, providing a robust and adaptive framework for modular learning.

\begin{definition}[Parallel Composition]
For a pre-trained model $f_{\boldsymbol{\theta}}$ and a task-specific module $g_{\boldsymbol{\phi}}$, the combined output is
\[
  h(\mathbf{x}) = \alpha f_{\boldsymbol{\theta}}(\mathbf{x}) + \beta g_{\boldsymbol{\phi}}(\mathbf{x}),
\]
where
\begin{itemize}
    \item $f_{\boldsymbol{\theta}}$ processes the input $\mathbf{x}$ using general-purpose knowledge;
    \item $g_{\boldsymbol{\phi}}$ processes the input $\mathbf{x}$ with task-specific insights;
    \item $\alpha$ and $\beta$ are learnable weights that balance the contributions of $f_{\boldsymbol{\theta}}$ and $g_{\boldsymbol{\phi}}$.
\end{itemize}
\end{definition}

Parallel composition is particularly effective when task-specific adaptations benefit from general-purpose features while maintaining the flexibility to incorporate additional insights. This configuration ensures parameter efficiency while enhancing task performance.

\section{Parameter-Efficient Fine-Tuning}
\label{sec:peft}

Building on the principles of modular deep learning, \textit{parameter-efficient fine-tuning (PEFT)} offers a practical approach for adapting large pre-trained models to diverse tasks. By updating only a small, task-specific subset of parameters, PEFT minimizes computational and memory overhead while preserving the majority of the pre-trained model's knowledge. This strategy aligns with the modular philosophy of deep learning, enabling efficient task adaptation without the need to retrain or fine-tune entire models.

PEFT integrates seamlessly with the modular design of pre-trained models, treating task-specific adaptation as a localized process. This targeted approach enhances resource efficiency, making it viable to deploy large-scale models in settings with constrained computational resources. In the following sections, we explore some prominent PEFT techniques.

\subsection{Adapters}

Adapters represent a family of PEFT techniques that introduce lightweight modules into a pre-trained model to enable task-specific adaptation with minimal parameter updates \cite{houlsby-etal-2019-parameter}. These modules are typically inserted between the layers of the model, and only the adapter parameters are trained, leaving the rest of the model frozen. This design ensures efficient adaptation to new tasks without disrupting the pre-trained knowledge. 

As a versatile class of PEFT methods, adapters come in various forms, including \textbf{bottleneck adapters}, which project high-dimensional representations into a lower-dimensional space before transforming them back, and \textbf{parallel adapters}, which run alongside the original model layers and merge outputs dynamically. These variations allow adapters to balance efficiency and expressiveness, making them a widely adopted solution for parameter-efficient fine-tuning in NLP.

\begin{definition}[Bottleneck Adapter]
For a hidden representation $\mathbf{h}_{l}$ at layer $l$, a bottleneck adapter is defined as
\[
\mathbf{z}_{l} = W^{\text{up}} \psi(W^{\text{down}} \mathbf{h}_{l}) + \mathbf{h}_{l},
\]
where
\begin{itemize}
    \item $W^{\text{down}} \in \mathbb{R}^{d \times D}$ projects $\mathbf{h}_{l}$ into a lower-dimensional space ($d < D$);
    \item $\psi$ is a non-linear activation function (e.g., ReLU);
    \item $W^{\text{up}} \in \mathbb{R}^{D \times d}$ projects back to the original dimension;
    \item $\mathbf{h}_{l}$ is the hidden state at layer $l$.
\end{itemize}
\end{definition}

The skip connection $\mathbf{h}_{l}$ ensures that the original pre-trained features are preserved. Adapters are typically initialized as identity mappings by setting $W^{\text{down}}$ and $W^{\text{up}}$ such that their combined effect approximates the identity function:

\begin{definition}[Adapter Initialization]
The adapter initialization approximates:
\[
W^{\text{down}} \approx \begin{bmatrix} I_{d} \\ 0 \end{bmatrix}, \quad W^{\text{up}} \approx \begin{bmatrix} I_{d} & 0 \end{bmatrix},
\]
where $I_{d}$ is the $d \times d$ identity matrix, ensuring:
\[
\mathbf{z}_{l} \approx \mathbf{h}_{l}.
\]
\end{definition}

This initialization avoids disrupting the pre-trained model's performance at the start of fine-tuning, allowing the adapters to gradually learn task-specific modifications.

\begin{figure}[]
\begin{center}
\includegraphics[width=0.25\linewidth, height=0.75\linewidth]{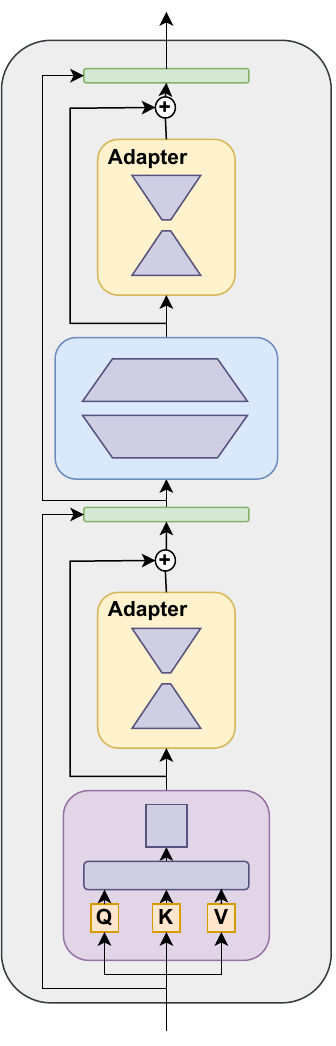}
\caption[Bottleneck adapter]{Bottleneck adapter: A lightweight module inserted between pre-trained layers in a Transformer block to enable parameter-efficient task adaptation.}
\label{fig:bottleneck}
\end{center}
\end{figure}

\subsubsection{Parallel Adapters}

Parallel adapters extend the concept of bottleneck adapters by running a lightweight task-specific module in parallel with the original layer. This approach allows both the pre-trained model and the task-specific adapter to contribute to the output, preserving general-purpose features while incorporating task-specific adjustments.

\begin{definition}[Parallel Adapter]
The output of a layer with a parallel adapter is given by
\[
\mathbf{h}_{l}^{\text{parallel}} = \alpha \cdot f(\mathbf{h}_{l}) + \beta \cdot \mathbf{h}_{l},
\]
where
\begin{itemize}
    \item $f(\mathbf{h}_{l})$ is the task-specific transformation computed by the adapter;
    \item $\alpha$ and $\beta$ are learnable scaling factors balancing the contributions of the adapter and the original layer;
    \item $\mathbf{h}_{l}$ is the original output of the layer.
\end{itemize}
\end{definition}

Parallel adapters provide flexibility by combining pre-trained and task-specific representations, making them particularly suitable for tasks requiring complementary insights from both components.

\begin{figure}[]
\begin{center}
\includegraphics[width=0.4\linewidth]{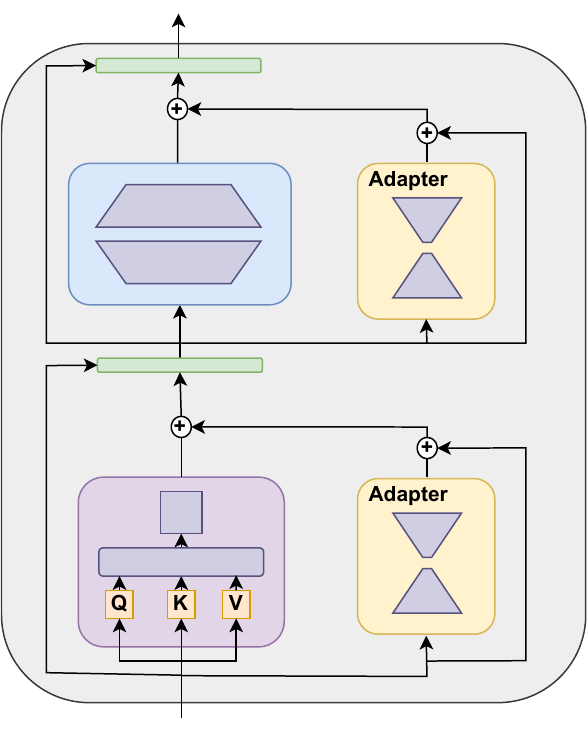}
\caption[Parallel Adapter]{Parallel adapter: A task-specific module operating alongside the pre-trained layer to combine general and task-specific features.}
\label{fig:parallel}
\end{center}
\end{figure}

\subsubsection{Advantages and Limitations of Adapters}

By design, adapters are inherently \textbf{parameter-efficient}, requiring only a small subset of additional parameters to be trained while keeping the majority of the pre-trained model unchanged. This significantly reduces both computational and memory overhead compared to full fine-tuning, making adapters particularly advantageous for large-scale models deployed in resource-constrained environments.

Beyond efficiency, adapters also provide \textbf{modularity}, allowing them to be seamlessly integrated into existing architectures. They can be inserted, removed, or swapped between different layers without modifying the core model, enabling flexible adaptation to various tasks. This modular design also promotes \textbf{reusability}, as trained adapters can be transferred across related tasks, minimizing redundancy and improving knowledge transfer. Furthermore, their \textbf{compatibility} with pre-trained transformers ensures that they can leverage existing learned representations, reducing the need for extensive retraining while maintaining strong performance across tasks.

Despite these advantages, adapters also come with certain limitations that must be considered. A primary concern is their \textbf{capacity constraint}. Due to their bottleneck design, adapters may struggle to capture highly complex task-specific patterns, which can lead to underfitting in cases where extensive adaptation is needed. Additionally, their effectiveness is heavily dependent on \textbf{proper initialization}. If not initialized appropriately, adapters can degrade the quality of learned representations, resulting in unstable training dynamics and suboptimal performance. Another challenge is the \textbf{task-specific overhead} introduced by adapters. While they are computationally lightweight, each task requires a dedicated set of adapter parameters, which can become difficult to manage in multi-task learning scenarios where multiple adapters must be maintained.

Despite these trade-offs, adapters remain a cornerstone of PEFT. Their balance between efficiency, flexibility, and scalability makes them an essential tool for adapting large pre-trained models to a wide range of NLP applications while minimizing computational costs.

\subsection{Low-Rank Adaptation}

Low-Rank Adaptation (LoRA) introduces a parameter-efficient method for adapting pre-trained models by leveraging low-rank updates to their weights \cite{hu-etal-2022-lora}. Unlike adapter-based approaches, which introduce additional trainable modules into a model while keeping the original parameters largely frozen, LoRA directly modifies the weight matrices of the pre-trained model through low-rank decomposition. This means that instead of adding extra layers or modules, LoRA constrains task-specific modifications to a lower-dimensional subspace, significantly reducing the number of trainable parameters while maintaining high task performance. 
This distinction makes LoRA particularly advantageous in scenarios where memory efficiency is critical, as it avoids increasing the model’s forward-pass complexity. While both LoRA and adapters aim to optimize parameter efficiency, LoRA is especially well-suited for tasks where modifying existing weight structures is preferable over introducing new components, offering a complementary yet distinct approach to parameter-efficient learning.

\subsubsection{Core Mechanism}

LoRA assumes that task-specific adaptations lie in a low-dimensional subspace. Instead of updating the entire weight matrix $W$ of a pre-trained model, LoRA introduces an additive update $\Delta W$ parameterized as a low-rank decomposition:

\begin{definition}[LoRA]
For a weight matrix $W \in \mathbb{R}^{D \times D}$, the adapted weight is given by
\[
W_{\text{adapted}} = W + \Delta W,
\]
where
\[
\Delta W = A B, \quad A \in \mathbb{R}^{D \times r}, \, B \in \mathbb{R}^{r \times D}, \, r \ll D.
\]
\end{definition}

In this formulation, $A$ and $B$ represent the low-rank matrices learned during fine-tuning, where $r$ denotes the rank of the adaptation, carefully selected to balance efficiency and expressive power. The original weight matrix $W$ remains frozen, preserving the pre-trained knowledge, while $A$ and $B$ capture task-specific modifications. This design reduces the number of trainable parameters from $D^2$ to $2Dr$, achieving substantial savings in both memory and computational cost.

\subsubsection{Training and Efficiency}

LoRA modifies the forward pass by introducing the low-rank update $\Delta W$ without altering the underlying architecture of the pre-trained model. This modularity allows seamless integration into existing frameworks. The low-rank parameterization ensures efficient use of resources, as only the small matrices $A$ and $B$ are updated during training.

The method's parameter efficiency enables scaling to tasks with limited labeled data or deploying models in scenarios with restricted computational resources, such as edge devices or environments with stringent latency requirements.

\begin{figure}[]
\begin{center}
\includegraphics[width=0.4\linewidth]{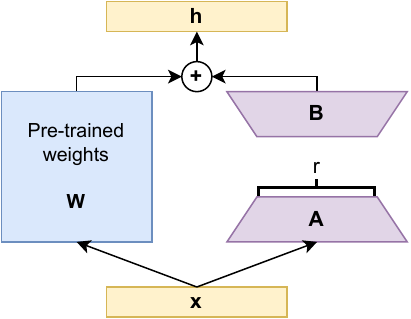}
\caption[LoRA]{LoRA: Task-specific adaptations represented as low-rank updates to the pre-trained weights. The pre-trained weights $W$ remain frozen, while $A$ and $B$ capture task-specific changes.}
\label{fig:lora}
\end{center}
\end{figure}

LoRA provides a flexible and scalable approach to fine-tuning by enabling targeted task-specific adaptations without modifying the entire model. Its modular nature allows low-rank updates to be applied selectively to specific layers, preserving the general pre-trained knowledge while allowing for efficient specialization. This makes LoRA particularly suitable for large-scale models, as it can be deployed on resource-constrained hardware while maintaining strong generalization across diverse tasks, including natural language processing, computer vision, and multimodal applications. Its adaptability has proven effective in various domains, demonstrating robust performance improvements while avoiding the computational burdens of traditional fine-tuning.

Despite its advantages, LoRA introduces certain challenges that must be addressed for optimal performance. One key consideration is the sensitivity to \textbf{rank selection}, as choosing an excessively low rank may lead to underfitting, while a higher rank increases parameter counts and computational costs, partially offsetting the efficiency gains. Additionally, \textbf{layer selection} plays a crucial role in maximizing LoRA's benefits, as identifying the most impactful layers for adaptation often requires careful tuning based on the specific task. Another potential drawback is the slight computational \textbf{overhead} introduced by the additional low-rank matrices during inference, which may be a concern for latency-sensitive applications. Despite these limitations, LoRA remains a cornerstone of parameter-efficient fine-tuning, offering a practical and effective method for adapting large models across a broad range of applications while minimizing computational demands.

\subsection{Prompt Tuning}

Building on the principles of parameter-efficient learning, \textbf{prompt tuning} \cite{lester-etal-2021-power} offers an innovative approach to task adaptation by modifying the input rather than the model's parameters. Similar to adapters, prompt tuning introduces task-specific adjustments, but it achieves this by appending learnable components to the input sequence, akin to how adapters transform intermediate representations.

\begin{definition}[Prompt Tuning]
For a pre-trained model $f_{\theta}$ and an input sequence $\mathbf{x}$, a trainable prompt $\mathbf{p}$ is prepended to the input, resulting in
\[
\mathbf{x}_{\text{prompted}} = [\mathbf{p}; \mathbf{x}],
\]
where
\begin{itemize}
    \item $\mathbf{p}$ is a learnable vector of embeddings;
    \item $\mathbf{x}_{\text{prompted}}$ is processed by the pre-trained model $f_{\theta}$, which remains frozen during training.
\end{itemize}
\end{definition}

Prompt tuning provides a highly efficient mechanism for adapting pre-trained models by learning only a small set of prompt embeddings, drastically reducing the number of trainable parameters. Unlike adapter-based methods or LoRA, which modify the model through additional layers or weight updates, prompt tuning encodes task-specific information directly into the input sequence. This approach allows for rapid adaptation to new tasks while preserving the pre-trained model's weights, ensuring that general-purpose knowledge remains intact. The lightweight nature of prompt tuning makes it particularly scalable, making it well-suited for low-resource settings or applications requiring fast deployment.

Despite its advantages, prompt tuning also introduces certain challenges. The effectiveness of the approach heavily depends on \textbf{prompt design}, which often requires careful task-specific tuning or prior domain knowledge to optimize performance. Additionally, since prompt tuning extends the input sequence with learnable tokens, it increases \textbf{context length}, which can strain the model’s capacity, especially for tasks involving long documents. Furthermore, its success is closely tied to the quality of the underlying pre-trained model, meaning that prompt tuning may be less effective if the base model lacks strong generalization capabilities. Nevertheless, as a flexible and computationally efficient alternative, prompt tuning remains a valuable tool in parameter-efficient learning.

\subsection{Prefix Tuning}

\textbf{Prefix tuning} \cite{li-liang-2021-prefix} extends the principles of prompt tuning by modifying the model’s activations rather than its input tokens. Instead of appending learnable embeddings to the input sequence, prefix tuning optimizes a set of continuous vectors that are inserted as ``prefixes'' into the model’s intermediate layers. These prefixes act as task-specific conditioning mechanisms, guiding the model’s predictions while keeping the pre-trained parameters frozen.

\begin{definition}[Prefix tuning]
For a pre-trained model $f_{\theta}$ and an input sequence $\mathbf{x}$, a trainable prefix $\mathbf{p}$ is inserted into the model’s hidden states, modifying the computation as follows:
\[
\mathbf{h}_{\text{modified}} = [\mathbf{p}; \mathbf{h}_{\text{original}}],
\]
where
\begin{itemize}
    \item $\mathbf{p}$ is a learnable set of prefix parameters injected into the model’s hidden representations;
    \item $\mathbf{h}_{\text{original}}$ denotes the standard hidden states computed by the pre-trained model.
\end{itemize}
\end{definition}

Prefix tuning offers several advantages over traditional prompt tuning. By conditioning the model at the hidden representation level rather than through prepended input tokens, it provides stronger task-specific control without increasing the length of the input sequence. This approach is particularly beneficial for handling long documents, where extending the input with additional tokens -- as in prompt tuning -- can be costly. Moreover, prefix tuning allows for more expressive task adaptation, as the inserted prefix vectors directly influence the transformer’s internal computation rather than being restricted to modifying token embeddings.

However, the effectiveness of prefix tuning depends on selecting appropriate layers for prefix injection, which may require empirical tuning. Additionally, while it remains computationally efficient compared to full fine-tuning, prefix tuning introduces additional memory overhead due to the learnable prefix parameters at each layer.

\subsection{UniPELT: A Unified PEFT Framework}

\textbf{UniPELT} (Unified Parameter-Efficient Language Model Tuning) \cite{mao-etal-2022-unipelt} introduces a framework that integrates multiple PEFT techniques into a single architecture. Recognizing that different PEFT methods -- such as adapters, prefix tuning, and LoRA -- offer complementary advantages depending on the task, UniPELT dynamically selects and combines these techniques to optimize model performance while minimizing parameter overhead.

\begin{definition}[UniPELT]
For a pre-trained model $f_{\theta}$ and an input $\mathbf{x}$, UniPELT introduces multiple PEFT submodules, each parameterized as $\mathbf{m}_i$, and a gating mechanism $\mathbf{g}$ that determines their activation:
\[
\mathbf{h}_{\text{unipelt}} = \sum_{i} g_i(\mathbf{x}) \mathbf{m}_i(\mathbf{h}_{\text{original}}),
\]
where
\begin{itemize}
    \item $\mathbf{h}_{\text{original}}$ is the hidden representation computed by the frozen pre-trained model;
    \item $\mathbf{m}_i$ represents different PEFT methods, such as Adapters, Prefix-Tuning, and LoRA;
    \item $g_i(\mathbf{x})$ is a learned gating function that dynamically activates or deactivates each method based on the task requirements.
\end{itemize}
\end{definition}

UniPELT enhances generalization across diverse tasks by strategically combining multiple PEFT techniques, leveraging their strengths while compensating for their individual limitations. Its adaptive gating mechanism dynamically selects the most effective method -- or a combination thereof -- tailored to the specific task, ensuring both flexibility and efficiency. However, this added adaptability comes at the cost of increased complexity. Learning the gating function and optimizing multiple PEFT modules simultaneously require careful tuning to strike a balance between computational efficiency and performance gains.

\section{Routing in Parameter-Efficient Learning}

As models grow in scale and modularity, efficient routing mechanisms become essential. Routing dictates which modules within a PLM are activated based on the input data or task, allowing the model to dynamically allocate resources where they are most needed. This is particularly important in multi-task or multi-modal settings, where different tasks may require different subsets of modules. By enabling dynamic module selection, routing ensures that parameter-efficient models can adapt to diverse tasks while maintaining scalability and performance \cite{cai-etal-2024-survey}.

\subsection{Fixed Routing}

In fixed routing, the activation of modules is predetermined and remains static during training and inference. Each task is assigned a fixed subset of modules, which are consistently used for processing. Formally, for a task-specific model configuration, the output is given by
\[
f_{\theta}^{\text{fixed}}(\mathbf{x}) = \sum_{i \in \mathcal{S}_{\text{task}}} f_{\theta}^{(i)}(\mathbf{x}),
\]
where $\mathcal{S}_{\text{task}}$ is the predefined set of modules for the task, and $f_{\theta}^{(i)}$ represents the $i$-th module. Fixed routing is computationally efficient and easy to implement, but lacks flexibility, making it less suited for scenarios where tasks or data distributions evolve over time.

\subsection{Learned Routing}

Learned routing allows the model to dynamically select modules based on the input or task. This is achieved using a gating mechanism that outputs a selection vector for the modules. For an input $\mathbf{x}$, the output of the model is given by
\[
f_{\theta}^{\text{selected}}(\mathbf{x}) = \sum_{i=1}^{N} g_i(\mathbf{x}) f_{\theta}^{(i)}(\mathbf{x}),
\]
where
\begin{itemize}
    \item $g_i(\mathbf{x})$ is the gating function that determines the activation weight for module $i$;
    \item $f_{\theta}^{(i)}(\mathbf{x})$ represents the output of module $i$;
    \item $N$ is the total number of modules.
\end{itemize}
This data-driven selection mechanism enhances the model's ability to adapt to new tasks and varying input distributions while maintaining parameter efficiency.

\subsection{Hard vs. Soft Routing}

\begin{figure}[!htb]
\begin{center}
\includegraphics[width=\linewidth]{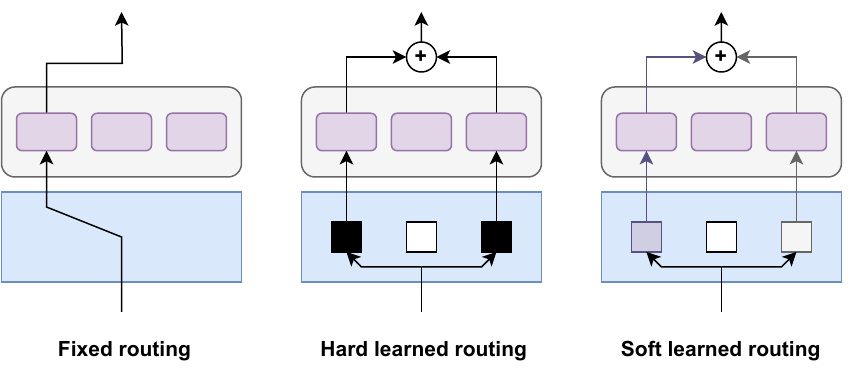}
\caption[Routing mechanisms]{Illustration of routing mechanisms in parameter-efficient learning. Left: Fixed routing assigns predefined modules for each task. Middle: Hard routing activates a subset of modules per input based on binary decisions. Right: Soft routing combines multiple module outputs with learned weights.}
\label{fig:routing}
\end{center}
\end{figure}

Learned routing can be categorized into hard and soft routing, each offering distinct trade-offs between efficiency and flexibility. An overview of these routing strategies is illustrated in \Cref{fig:routing}.

\subsubsection{Hard Routing}
In hard routing, each module is either activated or not based on a binary decision for a given input. The gating function $b_i(\mathbf{x}) \in \{0, 1\}$ indicates whether module $i$ is active:
\[
b_i(\mathbf{x}) = 
\begin{cases} 
1 & \text{if module $i$ is activated}, \\
0 & \text{otherwise.}
\end{cases}
\]
The model's output is then a sum over active modules:
\[
f_{\theta}^{\text{hard}}(\mathbf{x}) = \sum_{i=1}^K b_i(\mathbf{x}) f_{\theta}^{(i)}(\mathbf{x}),
\]
where $K$ is the total number of modules. While this approach is computationally efficient, since inactive modules can be skipped, it poses challenges for optimization due to the discrete, non-differentiable nature of the gating function. To enable training, methods such as reinforcement learning or straight-through estimators are commonly used.

\subsubsection{Soft Routing}
In soft routing, multiple modules are activated, and their outputs are combined in a weighted manner. The model's output is given by
\[
f_{\theta}^{\text{soft}}(\mathbf{x}) = \sum_{i=1}^{N} \alpha_i f_{\theta}^{(i)}(\mathbf{x}),
\]
where
\begin{itemize}
    \item $\alpha_i$ are learned weights representing the contribution of module $i$;
    \item $\sum_{i=1}^{N} \alpha_i = 1$ ensures a normalized mixture of modules.
\end{itemize}
Soft routing is fully differentiable, making it compatible with gradient-based optimization methods. It provides greater flexibility by allowing the model to leverage both general-purpose and task-specific modules.

\chapter{Integrating Active Learning with Parameter-Efficient Fine-Tuning}
\label{ch:al-peft}

PLMs have established themselves as the foundation for state-of-the-art performance across numerous NLP tasks. However, fine-tuning these models in low-resource settings -- where labeled data is scarce -- poses significant challenges. The dual constraints of limited labeled data and the computational cost of fine-tuning necessitate efficient approaches that minimize both label and parameter complexity. 

To address these constraints, two complementary strategies can be leveraged. AL (\Cref{ch:al}) focuses on reducing labeling costs by strategically selecting the most informative examples for annotation, thereby maximizing the utility of scarce labeled data. Simultaneously, PEFT techniques (\Cref{ch:peft}) target computational efficiency by fine-tuning only a small subset of the model's parameters while preserving most of the pre-trained knowledge.

While both AL and PEFT approaches are well-suited to low-resource scenarios, their integration remains unexplored. This chapter bridges this gap by systematically investigating how AL and PEFT can be effectively combined for low-resource text classification. Through extensive experiments (\Cref{sec:exp-adapters}), we assess the impact of this integration on both data efficiency and computational cost, providing empirical insights into the interplay between selective data acquisition and parameter-efficient adaptation. The approach and findings outlined here build on our previous work \cite{jukic-snajder-2023-parameter}, offering a novel perspective on optimizing PLMs in resource-constrained environments.

\section{Experimental Setup}

In this section, we outline the datasets, PEFT methods, AL sampling strategies, experimental setup, and evaluation metrics employed in our study.

\subsection{Datasets}

To comprehensively evaluate the interaction between AL and PEFT, we employ four widely used single-text classification tasks: \subj{} \cite{pang-lee-2004-sentimental}, \trec{} \cite{li-roth-2002-learning}, \sst{} \cite{socher-etal-2013-parsing}, and \agn{} \cite{zhang-etal-2015-character} (cf.~\Cref{sec:jachess-datasets,sec:beast-datasets} for detailed dataset descriptions). These datasets span both binary and multi-class classification scenarios, providing a diverse evaluation framework to assess the impact of AL strategies on PEFT across various linguistic challenges.

\subsection{PEFT and AL Methods}

We evaluate five active learning (AL) methods and four parameter-efficient fine-tuning (PEFT) techniques, covering a diverse range of approaches. For AL, we consider \textbf{random selection} (\rnd{}) as a passive learning baseline, alongside four established strategies: \textbf{entropy sampling} (\ent{}), \textbf{Monte Carlo dropout} (\mc{}), \textbf{core-set selection} (\cs{}), and \textbf{discriminative active learning} (\dal{}) (cf.~\Cref{sec:al-methods} for detailed descriptions).

For PEFT, we assess four widely used techniques: \textbf{bottleneck adapter} (\adapter{}), \textbf{prefix tuning} (\pt{}), \textbf{low-rank adaptation} (\lora{}), and \textbf{UniPELT} (\uni{}) (cf.~\Cref{sec:peft} for details). These methods represent different strategies for reducing trainable parameters while maintaining strong task performance, including modular adaptation (\adapter{}), task conditioning via learnable prefix vectors (\pt{}), low-rank decomposition of weight updates (\lora{}), and a unified framework that dynamically selects between multiple PEFT methods (\uni{}). All PEFT methods are evaluated using BERT \cite{devlin-etal-2019-bert} as the base PLM, with hyperparameters set according to recommendations from their respective original studies.

\subsection{Experiment Pipeline and Evaluation}

Each AL experiment follows a standardized pipeline to ensure consistency and comparability. At each AL step, 50 new examples are sampled and queried for labels, starting with an initial labeled set of 100 randomly selected instances as a warm start. To simulate low-resource settings, the total labeling budget is capped at 1,000 instances. 

For model training, we apply Besov early stopping (\beast{}) \cite{jukic-snajder-2023-smooth}, a stopping criterion introduced in \Cref{ch:beast} that determines the optimal epoch based on layer-wise representation smoothness, eliminating the need for a validation set. In experiments involving TAPT, the base model undergoes masked language modeling on the unlabeled training set before fine-tuning, during which only adapter parameters are updated while the base model remains frozen.

To compare AL methods against passive learning within the same training regime, we introduce the \textbf{relative improvement over passive learning (RIPL)} metric \cite{jukic-snajder-2023-parameter}, defined as  
\begin{equation}
\text{RIPL}(S_\mathrm{AL}, S_\mathrm{PL}) = \frac{\text{\auc}(S_\mathrm{AL}) - \text{\auc}(S_\mathrm{PL})}{1 - \text{\auc}(S_\mathrm{PL})}.
\end{equation}
Here, \( S_\mathrm{AL} \) represents the performance of a model trained using an AL strategy, where instances are selectively queried based on informativeness (e.g., uncertainty-based or diversity-based selection). In contrast, \( S_\mathrm{PL} \) denotes the performance of a model trained using \textit{passive learning}, where training instances are sampled randomly without any active selection. Essentially, RIPL quantifies the proportion of the maximum theoretical improvement an AL method achieves over passive learning.

To ensure robustness, each experiment is repeated five times using different random seeds, and performance is evaluated by reporting the average $F_1$ score at each sampling step. To assess the effectiveness of AL methods, we employ \auc{} and RIPL.

\section{Experiments}
\label{sec:exp-adapters}

In this section, we compare the performance of PEFT methods against full fine-tuning under passive learning. We then evaluate the integration of PEFT methods within AL frameworks.

\subsection{PEFT vs.~FFT}

\textbf{Full fine-tuning} (FFT) is the conventional approach for adapting PLMs to downstream tasks. It involves updating all model parameters during training, allowing the model to fully adapt to task-specific data. While effective, FFT is computationally expensive and requires substantial labeled data to avoid overfitting, making it less practical in low-resource settings.

Previous studies in low-resource settings \cite{li-liang-2021-prefix, mao-etal-2022-unipelt, he-etal-2021-effectiveness} have shown that PEFT methods can match or even surpass FFT. However, these studies often evaluate only a single adapter variant, report performance at a limited number of data points, or rely on full datasets rather than incremental learning curves.

To build upon these findings, we conduct a comprehensive comparison of multiple PEFT methods and FFT by generating detailed learning curves under the passive learning setup. The results, summarized using the \auc{} metric in \Cref{tab:pl}, indicate that \uni{} and \pt{} consistently outperform FFT with statistically significant margins across all datasets. On the other hand, the performance of \adapter{} and \lora{} is generally comparable to FFT, with occasional cases of improvement or underperformance. However, even when \adapter{} and \lora{} surpass FFT, the degree of improvement is notably smaller than that observed with \uni{} and \pt{}.

\begin{figure*}[]
    \centering
    \begin{subfigure}[b]{0.49\linewidth}
        \includegraphics[width=\linewidth]{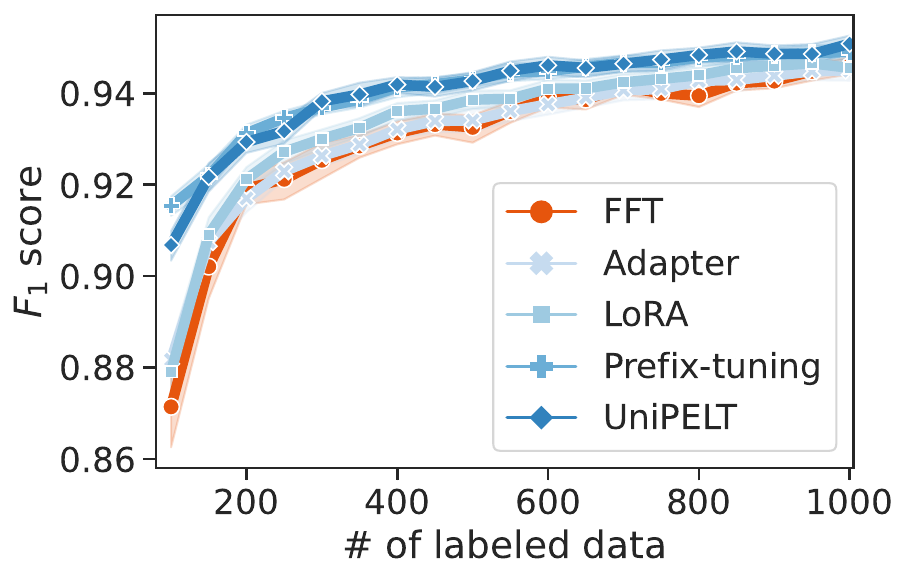}
        \caption{\subj}
        \label{fig:fft-vs-peft-subj}
    \end{subfigure}
    \begin{subfigure}[b]{0.49\linewidth}
        \includegraphics[width=\linewidth]{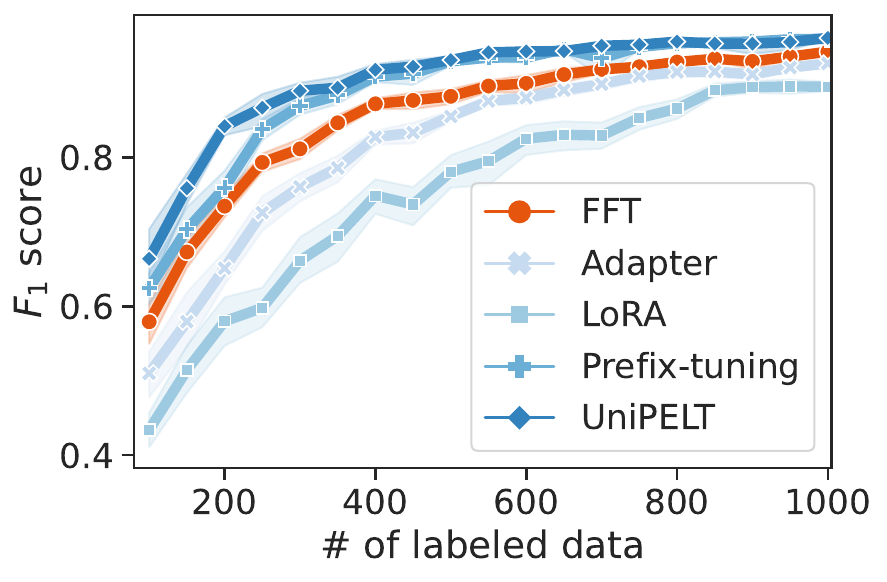}
        \caption{\trec}
        \label{fig:fft-vs-peft-trec}
    \end{subfigure}
    \begin{subfigure}[b]{0.49\linewidth}
        \includegraphics[width=\linewidth]{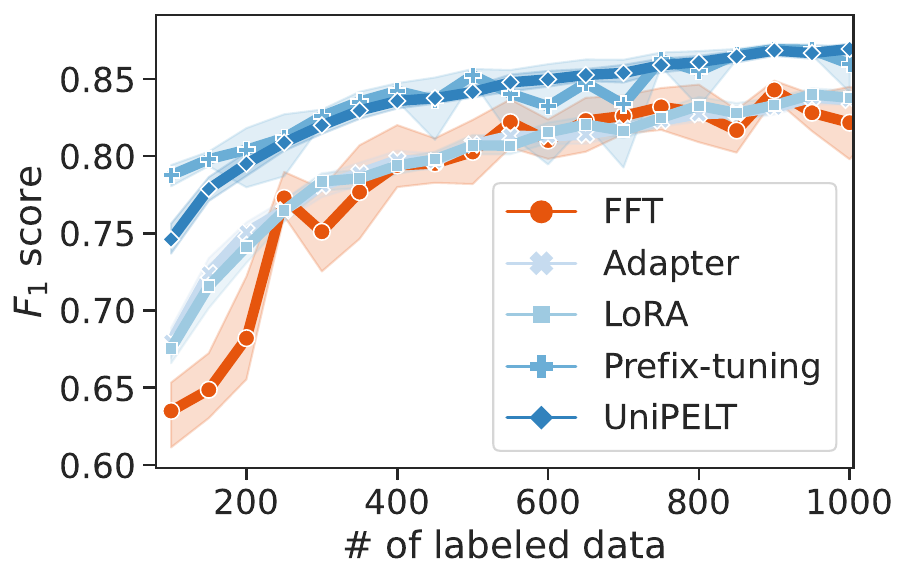}
        \caption{\textsc{sst}}
        \label{fig:fft-vs-peft-sst}
    \end{subfigure}
    \begin{subfigure}[b]{0.49\linewidth}
        \includegraphics[width=\linewidth]{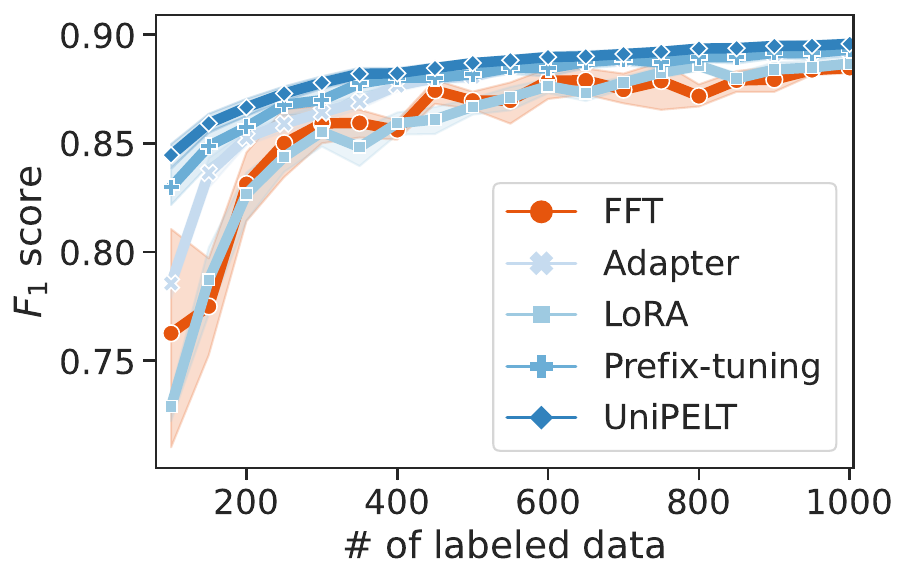}
        \caption{\agn}
        \label{fig:fft-vs-peft-agn}
    \end{subfigure}
\caption[Learning curves under passive learning for PEFT methods and FFT]{Learning curves under passive learning for PEFT methods and FFT. Results are averaged over five runs, with shaded bands indicating standard deviation.}
\label{fig:fft-vs-peft}
\end{figure*}

We further analyze the performance dynamics as the training set size increases. \Cref{fig:fft-vs-peft} illustrates the learning curves for FFT and PEFT. Performance disparities between adapters and FFT are most pronounced under extreme data scarcity, particularly with only $100$--$300$ labeled instances. Notably, the largest performance gap appears in the initial step, where only $100$ labeled examples are available. These results underscore the promise of adapter-based approaches, especially \pt{} and \uni{}, in low-resource settings where labeled data is severely limited.

\begin{table}[]
\caption[\auc{} scores for adapters and FFT with random sampling]{The performance of adapters and FFT in a passive learning setup in terms of the \auc{} metric (based on $F_1$ score) averaged over five runs. Numbers in \textbf{bold} represent the best-performing variant for a particular dataset. The ``$\dagger$'' symbol indicates when the mean \auc{} of an adapter is significantly different from the corresponding mean \auc{} of FFT ($p<.05$ using a two-sided Man-Whitney U test adjusted for family-wise error rate with the Holm-Bonferroni method).}
\centering
\small
\begin{tabular}{llrrrr}
\toprule
& & \textsc{subj} & \textsc{trec} & \textsc{sst} & \textsc{agn} \\
\midrule
\multirow{4}{*}{\rotatebox[origin=c]{90}{adapters}}
& Adapter & $.926$ & $.804$ & $.800$\nospacetext{$^\dagger$} & $.871$\nospacetext{$^\dagger$}\\
& \lora{} & $.929$ & $.750$\nospacetext{$^\dagger$} & $.798$\nospacetext{$^\dagger$} & $.860$ \\
& \pt{} & $\textbf{.936}$\nospacetext{$^\dagger$} & $.847$\nospacetext{$^\dagger$} & $\textbf{.847}$\nospacetext{$^\dagger$} & $\textbf{.875}$\nospacetext{$^\dagger$} \\
& \uni{} & $.934$\nospacetext{$^\dagger$} & $\textbf{.877}$\nospacetext{$^\dagger$} & $.836$\nospacetext{$^\dagger$} & $\textbf{.875}$\nospacetext{$^\dagger$}\\
\midrule
& FFT & $.928$ & $.810$ & $.787$ & $.860$\\
\bottomrule
\end{tabular}
\label{tab:pl}
\end{table}

\subsection{PEFT in AL Settings}

Given PEFT's promising performance in passive learning, we extend our evaluation to AL scenarios. We analyze the effectiveness of PEFT methods when combined with AL strategies and compare these against FFT with and without TAPT. The RIPL metric is used to measure the gains provided by AL methods. \Cref{tab:aucs} summarizes the results across datasets, showing that \textit{PEFT methods consistently achieve higher RIPL scores compared to FFT, particularly for \pt{} and \uni{}}.

\Cref{tab:aucs-app} shows AUC scores for different combinations of AL methods and adapters, complementing the relative improvement scores as AUC represents absolute scores for each configuration. 

\begin{table*}[]
\caption[Improvement over passive learning in terms of the RIPL]{Improvement over passive learning in terms of the RIPL metric for four AL methods considered (\ent{}, \mc{}, \cs{}, and \dal{}) and for all combinations of adapters and datasets considered, shown separately without TAPT and with TAPT. Positive values indicate improvement over passive learning, while negative values indicate performance drops compared to passive learning. Values in \textbf{bold} indicate the best result for a particular dataset across different adapters and AL methods within the same regime (with or without TAPT).}
\centering
\small
\begin{tabular}{lr|rrrr|rrrr}
\toprule
& & 
\multicolumn{4}{c}{without TAPT} & \multicolumn{4}{c}{with TAPT} \\
\cmidrule(lr){3-6} \cmidrule(lr){7-10} 
 & & \ent{} & \mc{} & \cs{} & \dal{} & \ent{} & \mc{} & \cs{} & \dal{} \\
\midrule
\multirow{5}{*}{\rotatebox[origin=c]{90}{\subj}}
& FFT & $.050$ & $.059$ & $.061$ & $.077$ & $.140$ & $.140$ & $.142$ & $.126$\\
& \adapter{} & $.112$ & $.102$ & $.100$ & $.092$ & $.137$ & $.151$ & $.111$ & $.067$\\
& \lora{} & $.127$ & $.115$ & $.091$ & $.081$ & $.165$ & $.160$ & $.122$ & $.100$\\
& \pt & $.095$ & $.110$ & $.106$ & $.111$ & $\textbf{.186}$ & $.181$ & $.170$ & $.151$\\
& \uni{} & $.129$ & $\textbf{.153}$ & $.131$ & $.128$ & $.159$ & $.167$ & $.163$ & $.157$\\
\midrule
\multirow{5}{*}{\rotatebox[origin=c]{90}{\trec}}
& FFT & $.011$ & $.022$ & $.038$ & $.034$ & $.162$ & $.180$ & $.141$ & $.159$\\
& \adapter{} & $.027$ & $.069$ & $.137$ & $.084$ & $.124$ & $.146$ & $.079$ & $.154$\\
& \lora{} &$.098$ & $.065$ & $.048$ & $.007$ & $.254$ & $.237$ & $.243$ & $.074$\\
& \pt{} & $.093$ & $.105$ & $.068$ & $.093$ & $.246$ & $.227$ & $.205$ & $.241$\\
& \uni{} & $.138$ & $.165$ & $.082$ & $\textbf{.200}$ & $.302$ & $\textbf{.334}$ & $.276$ & $.236$\\
\midrule
\multirow{5}{*}{\rotatebox[origin=c]{90}{\sst}}
& FFT & $.002$ & $.011$ & $-.039$ & $.004$ & $.080$ & $.079$ & $.075$ & $.070$\\
& \adapter{} & $.015$ & $.048$ & $.025$ & $.002$ & $.035$ & $.034$ & $.028$ & $.008$\\
& \lora{} & $.001$ & $.007$ & $.064$ & $.031$ & $.036$ & $.022$ & $.032$ & $.014$\\
& \pt{} & $.049$ & $.060$ & $\textbf{.114}$ & $.031$ & $\textbf{.152}$ & $.143$ & $.137$ & $.126$\\
& \uni{} & $.037$ & $.043$ & $.040$ & $.008$ & $.082$ & $.101$ & $.083$ & $.080$\\
\midrule
\multirow{5}{*}{\rotatebox[origin=c]{90}{\agn}}
& FFT & $.014$ & $.032$ & $.007$ & $.092$ & $.134$ & $.021$ & $.089$ & $.017$ \\
& \adapter{} & $.074$ & $.046$ & $.015$ & $.062$ & $.115$ & $.089$ & $.077$ & $.080$\\
& \lora{} & $.020$ & $.025$ & $.067$ & $.016$ & $.028$ & $.102$ & $.071$ & $.023$\\
& \pt{} & $.054$ & $.023$ & $.040$ & $.033$ & $.035$ & $.143$ & $.098$ & $.092$\\
& \uni{} & $.074$ & $\textbf{.096}$ & $.089$ & $.095$ & $\textbf{.185}$ & $.151$ & $.112$ & $.081$  \\
\bottomrule
\end{tabular}
\label{tab:aucs}
\end{table*}

\begin{table*}[]
\caption[\auc{} scores for AL methods with different adapters shown separately with and without TAPT]{\auc{} scores for AL methods with different adapters shown separately without TAPT and with TAPT. We include random sampling for comparison with AL methods. Values in \textbf{bold} indicate the best result for a particular dataset within the same regime (with or without TAPT).}
\centering
\small
\begin{tabular}{lr|rrrrr|rrrrr}
\toprule
& & 
\multicolumn{5}{c}{without TAPT} & \multicolumn{5}{c}{with TAPT} \\
\cmidrule(lr){3-7} \cmidrule(lr){8-12} 
 & & \rnd{} & \ent{} & \mc{} & \cs{} & \dal{} & \rnd{} & \ent{} & \mc{} & \cs{} & \dal{} \\
\midrule
\multirow{5}{*}{\rotatebox[origin=c]{90}{\subj}}
& FFT & $.928$ & $.931$ & $.932$ & $.932$ & $.934$ & $.938$ & $.947$ & $.947$ & $.947$ & $.946$\\
& \adapter{} & $.926$ & $.934$ & $.933$ & $.933$ & $.932$ & $.934$ & $.943$ & $.944$ & $.941$ & $.938$
\\
& \lora{} & $.929$ & $.938$ & $.937$ & $.935$ & $.934$ & $.935$ & $.946$ & $.945$ & $.943$ & $.942$\\
& \pt & $.936$ & $.942$ & $.943$ & $.943$ & $.943$ & $.940$ & $.951$ & $.951$ & $.950$ & $.949$\\
& \uni{} & $.934$ & $.943$ & $\textbf{.944}$ & $.943$ & $.942$ & $.943$ & $.952$ & $\textbf{.953}$ & $.952$ & $.952$\\
\midrule
\multirow{5}{*}{\rotatebox[origin=c]{90}{\trec}}
& FFT & $.810$ & $.812$ & $.814$ & $.817$ & $.816$ & $.818$ & $.847$ & $.851$ & $.844$ & $.847$\\
& \adapter{} & $.804$ & $.809$ & $.818$ & $.831$ & $.820$ & $.820$ & $.842$ & $.846$ & $.834$ & $.848$\\
& \lora{} & $.750$ & $.775$ & $.766$ & $.762$ & $.752$ & $.764$ & $.824$ & $.820$ & $.821$ & $.781$\\
& \pt{} & $.847$ & $.861$ & $.863$ & $.857$ & $.861$ & $.862$ & $.896$ & $.893$ & $.890$ & $.895$\\
& \uni{} & $.877$ & $.894$ & $.897$ & $.887$ & $\textbf{.902}$ & $.896$ & $.927$ & $\textbf{.931}$ & $.925$ & $.921$\\
\midrule
\multirow{5}{*}{\rotatebox[origin=c]{90}{\sst}}
& FFT & $.787$ & $.787$ & $.789$ & $.779$ & $.788$ & $.792$ & $.809$ & $.808$ & $.808$ & $.807$\\
& \adapter{} & $.800$ & $.803$ & $.810$ & $.805$ & $.801$ & $.812$ & $.819$ & $.818$ & $.817$ & $.814$\\
& \lora{} & $.798$ & $.798$ & $.799$ & $.811$ & $.804$ & $.806$ & $.813$ & $.810$ & $.812$ & $.809$\\
& \pt{} & $.847$ & $.854$ & $.856$ & $\textbf{.864}$ & $.852$ & $.868$ & $\textbf{.888}$ & $.887$ & $.886$ & $.885$\\
& \uni{} & $.836$ & $.842$ & $.843$ & $.843$ & $.837$ & $.871$ & $.882$ & $.884$ & $.882$ & $.881$
\\
\midrule
\multirow{5}{*}{\rotatebox[origin=c]{90}{\agn}}
& FFT & $.860$ & $.862$ & $.864$ & $.861$ & $.873$ & $.869$ & $.887$ & $.872$ & $.881$ & $.871$\\
& \adapter{} & $.871$ & $.881$ & $.877$ & $.873$ & $.879$ & $.882$ & $.896$ & $.893$ & $.891$ & $.891$\\
& \lora{}& $.860$ & $.863$ & $.863$ & $.869$ & $.862$ & $.868$ & $.872$ & $.881$ & $.877$ & $.871$\\
& \pt{} & $.875$ & $.882$ & $.878$ & $.880$ & $.879$ & $.886$ & $.890$ & $.902$ & $.897$ & $.896$\\
& \uni{} & $.875$ & $.884$ & $\textbf{.887}$ & $.886$ & $\textbf{.887}$
 & $.887$ & $\textbf{.908}$ & $.904$ & $.900$ & $.896$\\
\bottomrule
\end{tabular}
\label{tab:aucs-app}
\end{table*}

\begin{figure}[]
\centering
\includegraphics[width=0.8\linewidth]{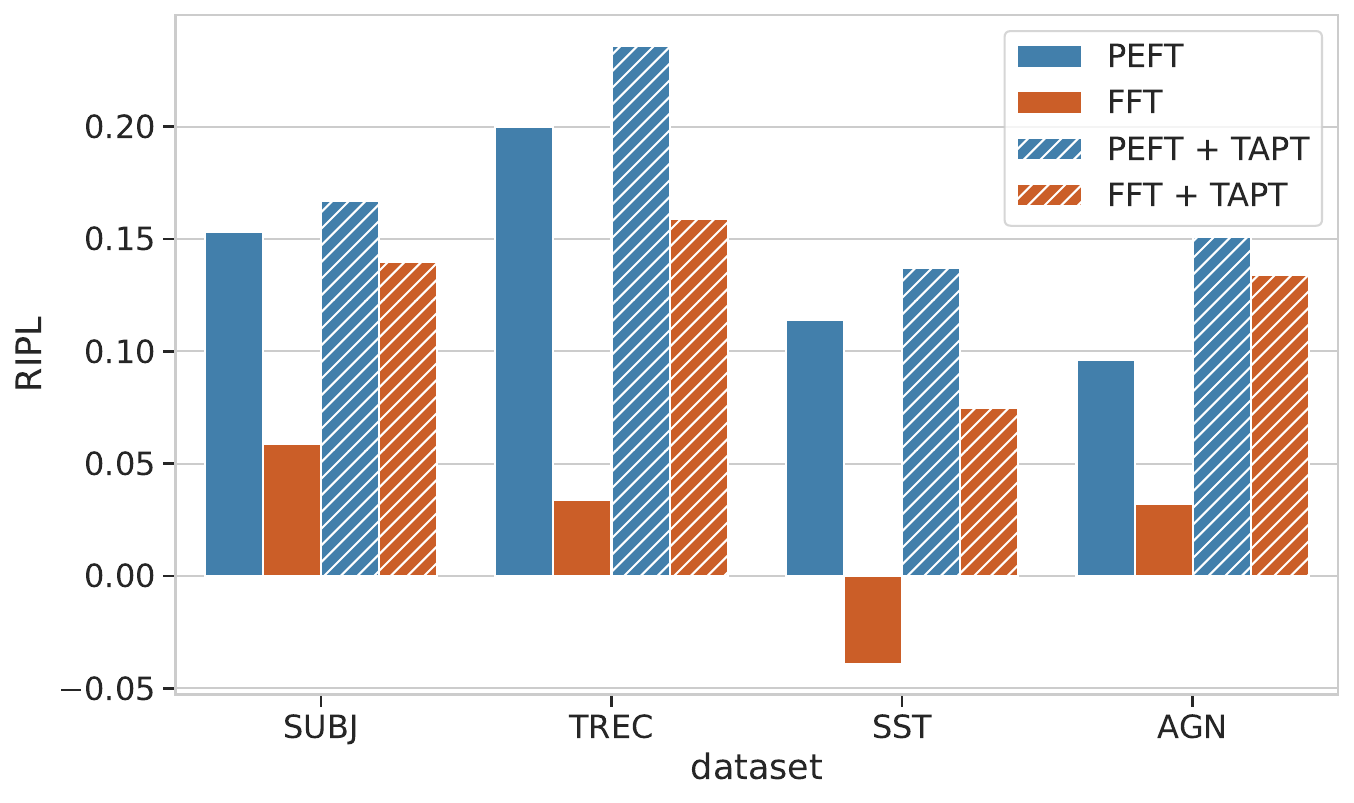}
\caption[Comparison of adapters and FFT with and without TAPT]{Comparison of best-performing adapters and FFT from \Cref{tab:aucs} with and without TAPT.}
\label{fig:tapt-bar}
\end{figure}

\paragraph{Impact of TAPT.}
We observe that TAPT consistently enhances AL performance across all configurations. For FFT without TAPT, \dal{} achieves the highest RIPL scores on two datasets, while \cs{} and \mc{} lead on one dataset each. When TAPT is applied, \ent{} emerges as the top-performing method on three out of four datasets, with \cs{} leading on one. Notably, TAPT narrows the performance gap between FFT and PEFT. However, \pt{} and \uni{} maintain their superiority over FFT, especially in combination with entropy-based AL strategies (\ent{} and \mc{}). \Cref{fig:tapt-bar} highlights the comparison between the best-performing adapters and FFT, with and without TAPT.

\begin{figure*}[]
\centering
\includegraphics[width=\linewidth]{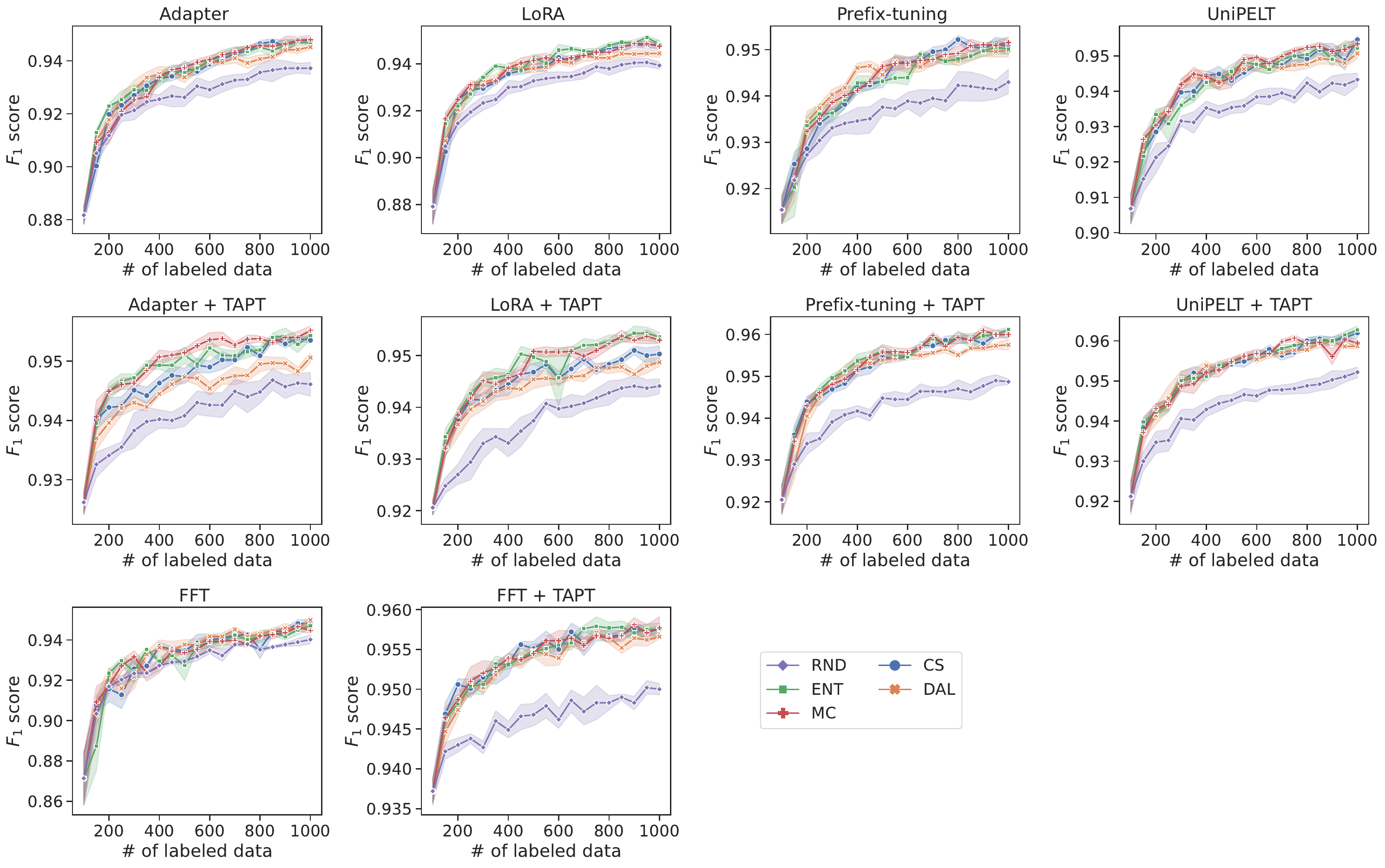}
\caption[Comparison of AL and random sampling on the \subj{} dataset]{AL learning curves compared with random sampling on the \subj{} dataset. Rows 1 and 2 show adapter performance without and with TAPT, respectively. Row 3 shows FFT performance without and with TAPT. Results are averaged over five runs; shaded bands denote standard deviation. Best viewed on a computer screen.}
\label{fig:subj}
\end{figure*}

\paragraph{AL performance across steps.}
To further investigate the behavior of PEFT methods in AL, we analyze learning curves across individual steps. \Cref{fig:subj} illustrates the performance trends on the \subj{} dataset, with similar patterns observed across other datasets. Without TAPT, adapter performance appears largely independent of the specific AL method, with \pt{} and \uni{} outperforming \adapter{} and \lora{} consistently. When TAPT is applied, the gap between AL and random sampling becomes evident as early as the second step (approximately $200$ labeled instances). In contrast, without TAPT, the performance advantage emerges only after $500$ or more labeled examples.

\paragraph{Summary of findings.}
Our experiments reveal several key insights into the interaction between PEFT, TAPT, and AL. PEFT consistently outperforms FFT in low-resource scenarios, with adapters—particularly \pt{} and \uni{} -- achieving superior performance in both passive learning and AL settings under extreme data scarcity. While TAPT narrows the performance gap between FFT and PEFT, the latter retains its advantage, especially in early AL steps. Among AL strategies, entropy-based methods such as \ent{} and \mc{} prove particularly effective, aligning well with TAPT and yielding the best results for both FFT and PEFT models. Furthermore, TAPT enhances AL efficiency by enabling earlier performance gains, reducing the label complexity required to achieve meaningful improvements. These findings reinforce the benefits of integrating PEFT, TAPT, and AL to address data and parameter constraints in low-resource NLP tasks.

\section{Analysis}
\label{sec:analysis}

In \Cref{sec:exp-adapters}, we demonstrated that PEFT methods consistently outperform FFT in AL scenarios for low-resource settings, with similar advantages observed in passive learning. To understand the underlying reasons for PEFT's superior performance, we investigate two key aspects: the influence of TAPT on forgetting dynamics during training, and the differences in representation similarity between PEFT and FFT models relative to their pre-trained counterparts.

\subsection{Forgetting Dynamics}

Understanding how models acquire and retain knowledge throughout training is crucial for evaluating their stability and effectiveness. \textbf{Forgetting dynamics} \cite{toneva-etal-2019-empirical} provides a framework for analyzing this process by tracking how often a model correctly classifies an instance before later misclassifying it. This metric offers insight into model generalization, robustness, and the informativeness of individual training examples. In the context of AL, forgetting dynamics helps identify which instances are most valuable for querying, as frequent forgetting may indicate noise or intrinsic difficulty rather than true informativeness.

\paragraph{Categorization of forgetting events.}
We categorize training examples into three groups based on the frequency of forgetting events during $10$ epochs:
\begin{itemize}
    \item \textbf{Unforgettable}: Instances that are consistently classified correctly throughout training;
    \item \textbf{Moderately forgettable}: Instances that experience one or two forgetting events;
    \item \textbf{Highly forgettable}: Instances with three or more forgetting events, typically harder or noise-prone examples.
\end{itemize}
Moderately forgettable instances are considered the most informative for learning, while highly forgettable instances can negatively impact AL due to their inherent difficulty \cite{karamcheti-etal-2021-mind}.

\begin{figure*}[]
    \centering
    \begin{subfigure}[b]{0.325\linewidth}
        \includegraphics[width=\linewidth, height=3cm]{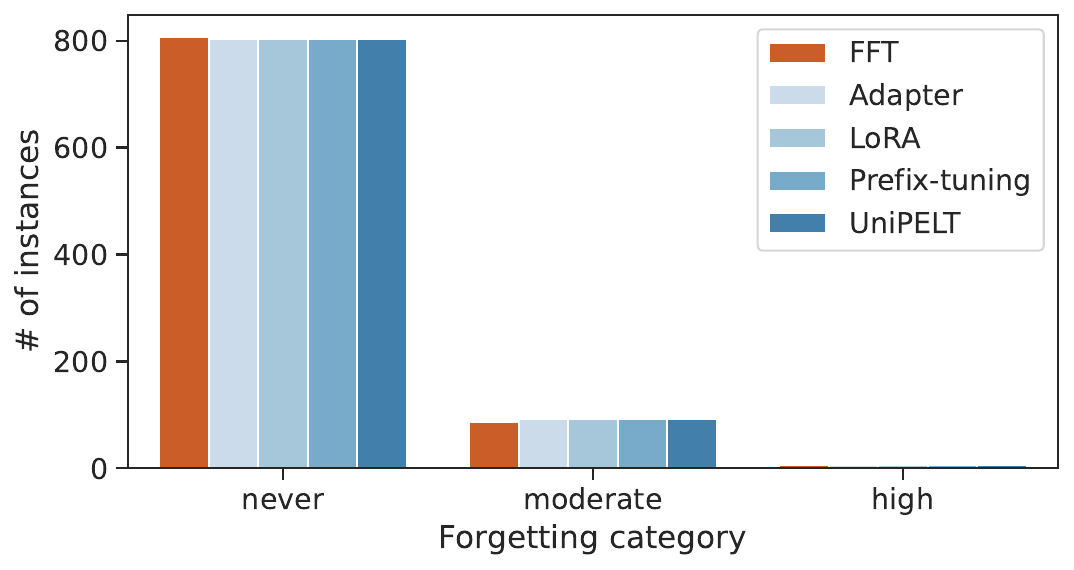}
        \caption{\subj{}; \rnd{}}
        \label{fig:forget_subj_rnd}
    \end{subfigure}
    \begin{subfigure}[b]{0.325\linewidth}
        \includegraphics[width=\linewidth, height=3cm]{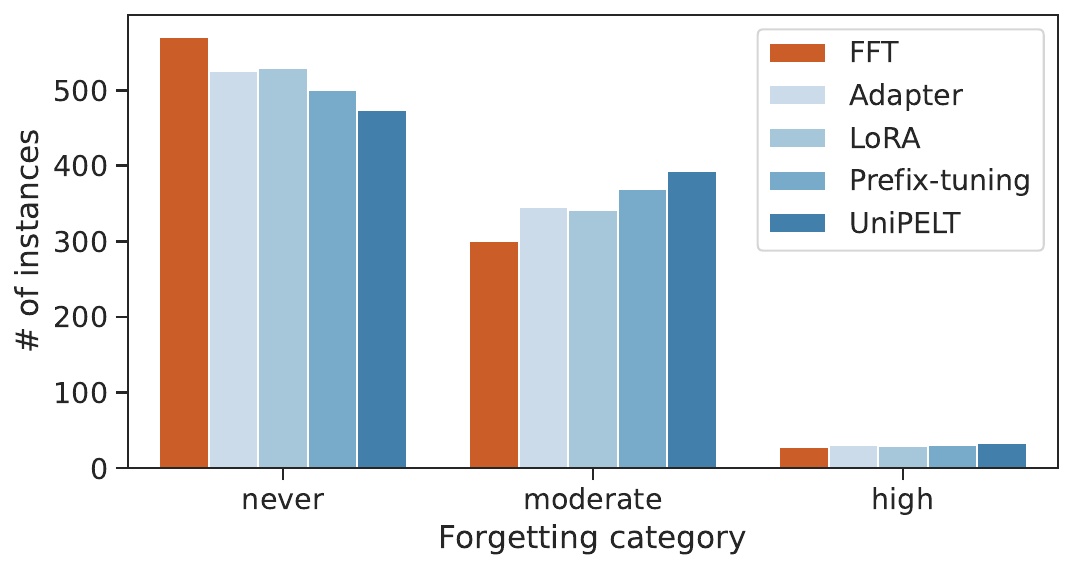}
        \caption{\subj{}; \mc{}}
        \label{fig:forget_subj_mc}
    \end{subfigure}
    \begin{subfigure}[b]{0.325\linewidth}
        \includegraphics[width=\linewidth, height=3cm]{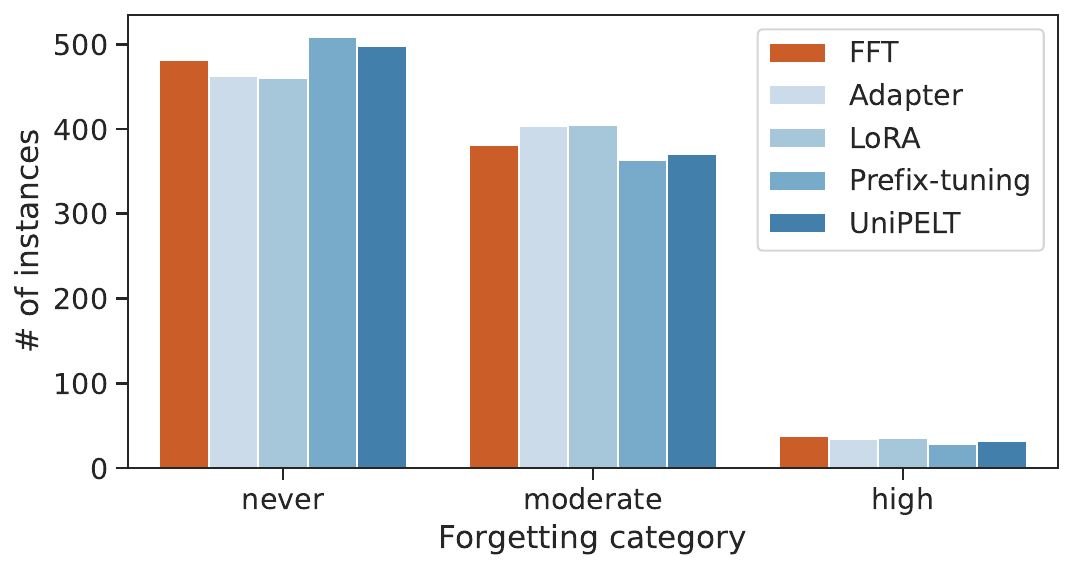}
        \caption{\subj{}; \mc{} + TAPT}
        \label{fig:forget_subj_mc_tapt}
    \end{subfigure}
    \begin{subfigure}[b]{0.325\linewidth}
        \includegraphics[width=\linewidth, height=3cm]{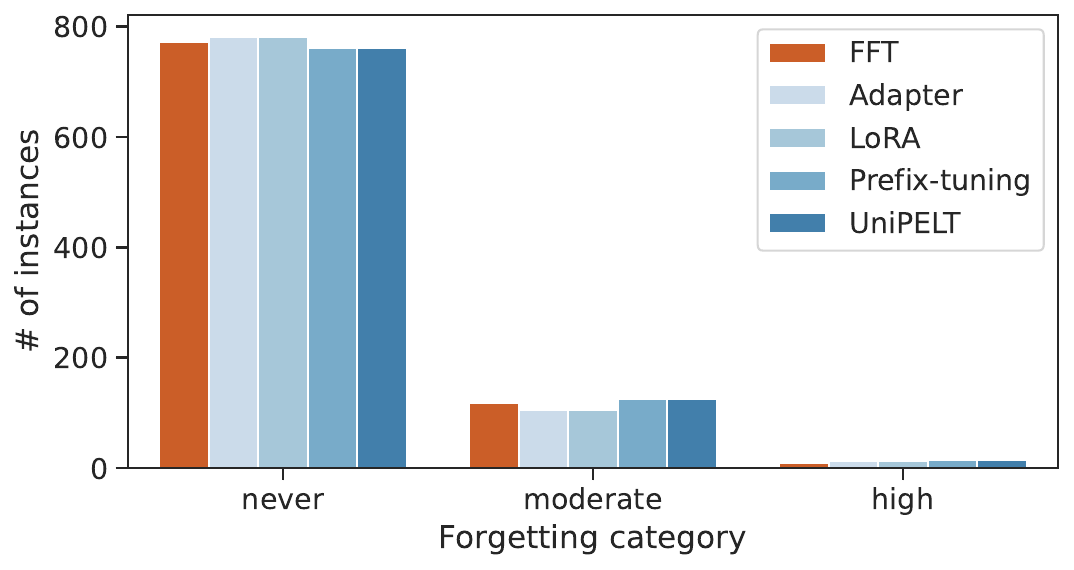}
        \caption{\trec{}; \rnd{}}
        \label{fig:forget_trec_rnd}
    \end{subfigure}
    \begin{subfigure}[b]{0.325\linewidth}
        \includegraphics[width=\linewidth, height=3cm]{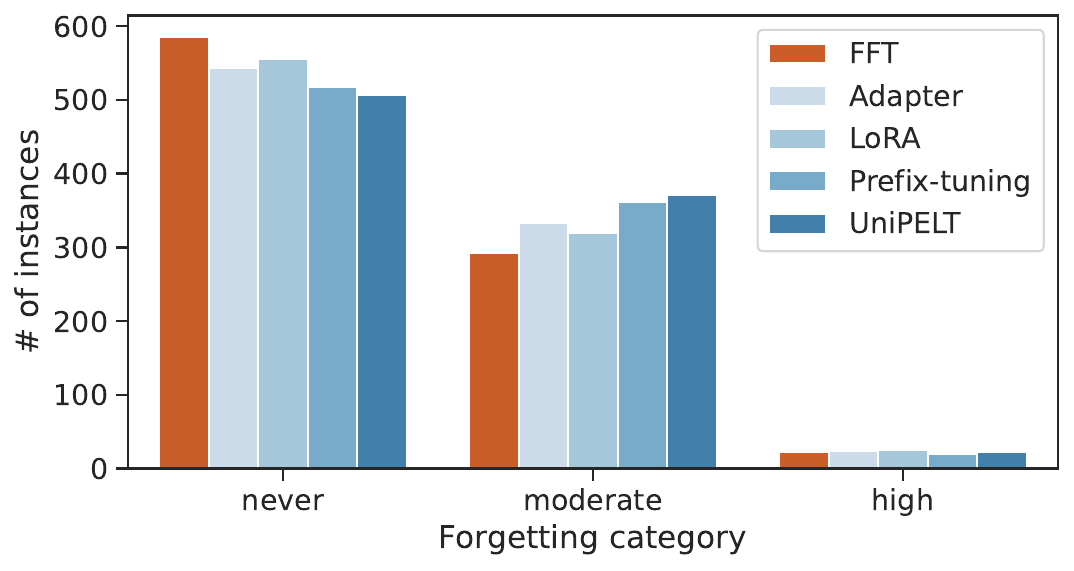}
        \caption{\trec{}; \mc{}}
        \label{fig:forget_trec_mc}
    \end{subfigure}
    \begin{subfigure}[b]{0.325\linewidth}
        \includegraphics[width=\linewidth, height=3cm]{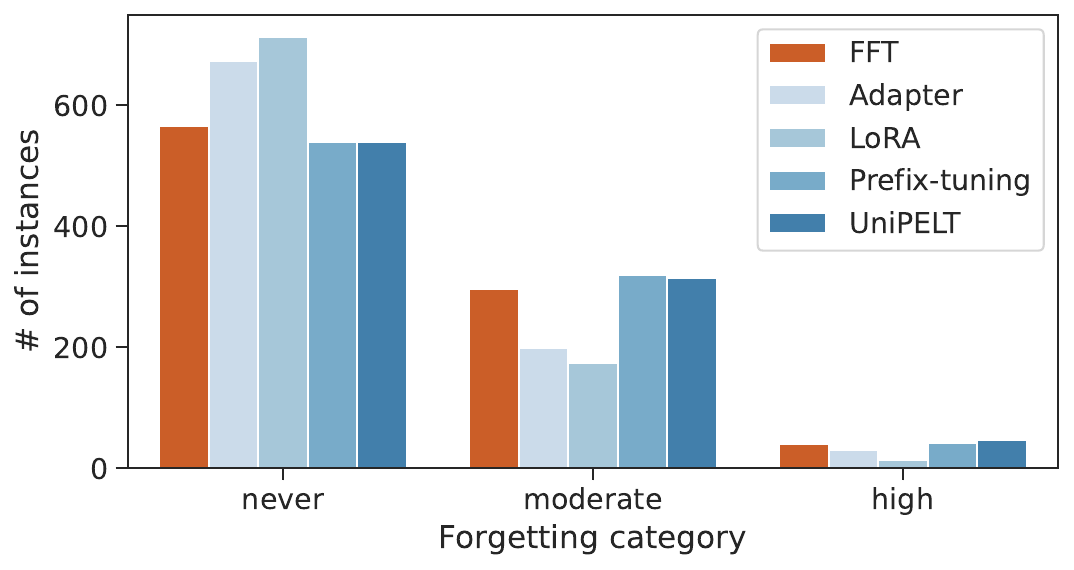}
        \caption{\trec{}; \mc{} + TAPT}
        \label{fig:forget_trec_mc_tapt}
    \end{subfigure}
\caption[Forgetting dynamics for random sampling and AL]{Forgetting dynamics for random sampling (passive learning) and AL with \mc{} without and with TAPT on \subj{} and \trec{}. The x-axis shows the number of instances in each of the forgetting categories: the ``never'' category representing \textbf{unforgettable} instances, \textbf{moderately} forgettable instances, and \textbf{highly} forgettable instances.}
\label{fig:forget}
\end{figure*}

\paragraph{Findings on forgetting profiles.}
\Cref{fig:forget} illustrates the distribution of forgetting events for the \subj{} and \trec{} datasets, comparing FFT and PEFT (\pt{} and \uni{}) under random sampling (\rnd{}) and AL with Monte Carlo dropout (\mc{}). FFT models tend to classify a higher proportion of instances as unforgettable while selecting fewer moderately forgettable examples compared to PEFT. This pattern is consistent across datasets and AL methods. 

Adapters such as \pt{} and \uni{}, which outperform FFT in AL settings, prioritize moderately forgettable instances. This preference aligns with their superior learning stability and performance. Interestingly, when TAPT is applied, the difference in forgetting profiles between FFT and these adapters diminishes, suggesting that TAPT helps FFT emulate the favorable learning behavior of PEFT. Conversely, adapters like \lora{} and \adapter{}, which show smaller performance improvements, retain distinct forgetting profiles even with TAPT.

\subsection{Representation Analysis}

Building on the insights from forgetting dynamics, we analyze the representational behavior (cf.~\Cref{ch:rep-props}) of PEFT and FFT models. Prior studies \cite{he-etal-2021-effectiveness, li-liang-2021-prefix, mao-etal-2022-unipelt} suggest that PEFT methods offer greater representational stability, especially in low-resource scenarios. We focus on how their representations align with those of the pre-trained base model.

Drawing from \cite{stephenson-etal-2021-geometry} and \cite{baldock-etal-2021-deep}, we examine layerwise specialization within models. Early layers generally encode generalized knowledge, while deeper layers focus on task-specific patterns. To quantify representational alignment, we use centered kernel alignment (CKA) \cite{kornblith-etal-2019-similarity} as a similarity metric, which is robust to transformations and effective for high-dimensional representations.

\begin{figure*}[]
    \centering
    \begin{subfigure}[b]{.32\linewidth}
        \includegraphics[width=\linewidth]{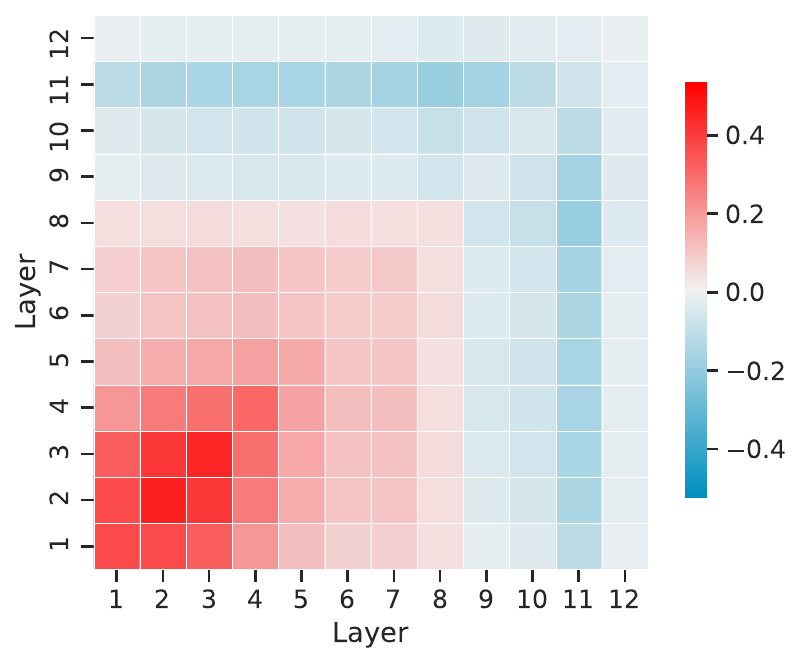}
        \caption{\ent{}}
        \label{fig:repr_ent}
    \end{subfigure}
    \begin{subfigure}[b]{.32\linewidth}
        \includegraphics[width=\linewidth]{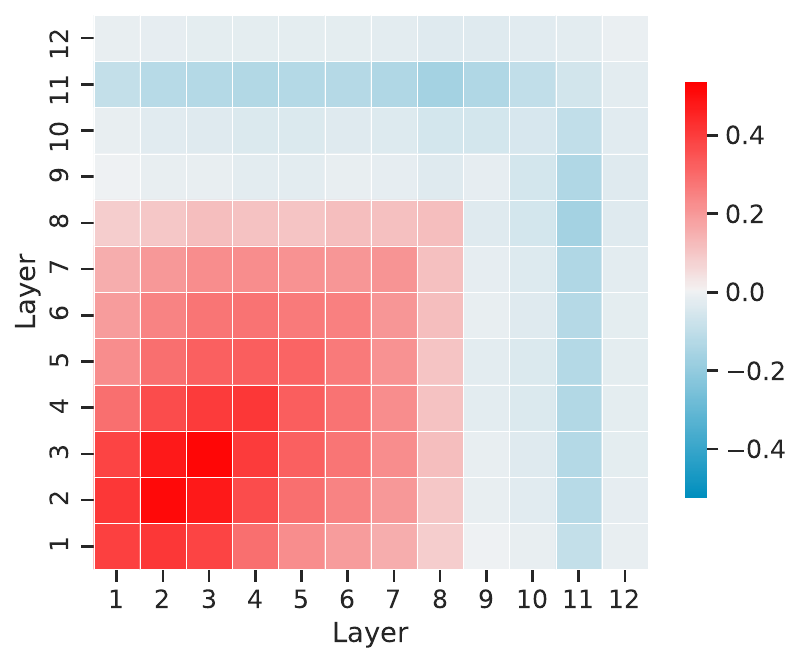}
        \caption{\mc{}}
        \label{fig:repr_mc}
    \end{subfigure}
    \begin{subfigure}[b]{0.32\linewidth}
        \includegraphics[width=\linewidth]{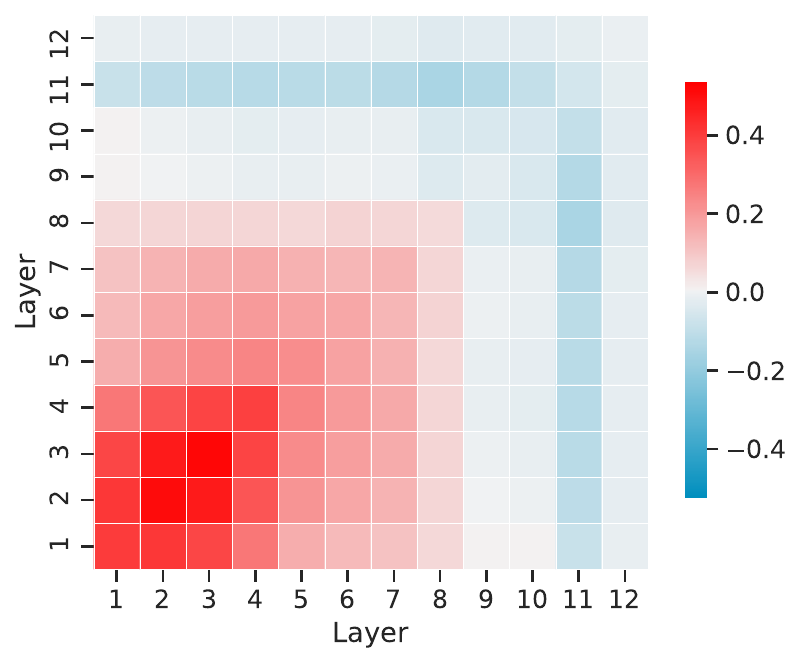}
        \caption{\cs{}}
        \label{fig:repr_cs}
    \end{subfigure}
    \begin{subfigure}[b]{0.32\linewidth}
        \includegraphics[width=\linewidth]{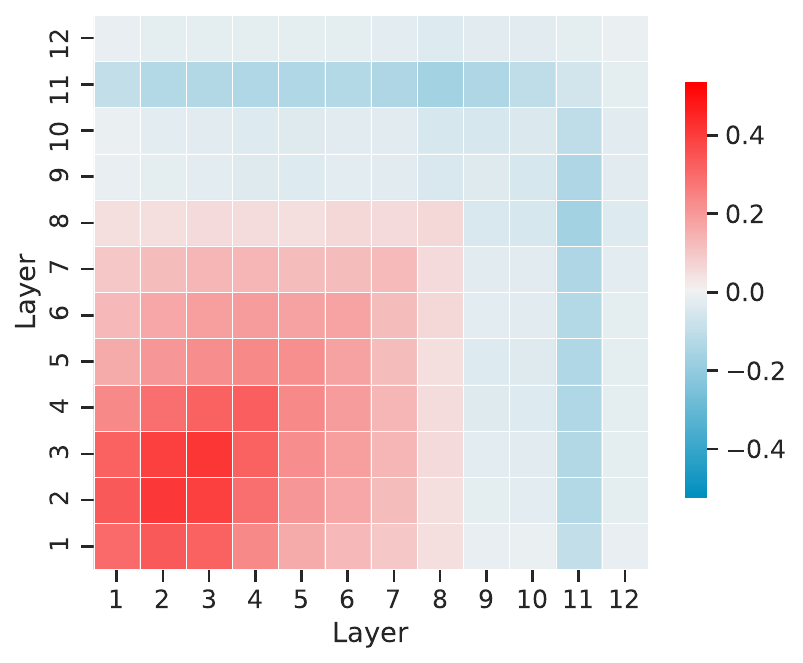}
        \caption{\dal{}}
        \label{fig:repr_dal}
    \end{subfigure}
    \begin{subfigure}[b]{0.32\linewidth}
        \includegraphics[width=\linewidth]{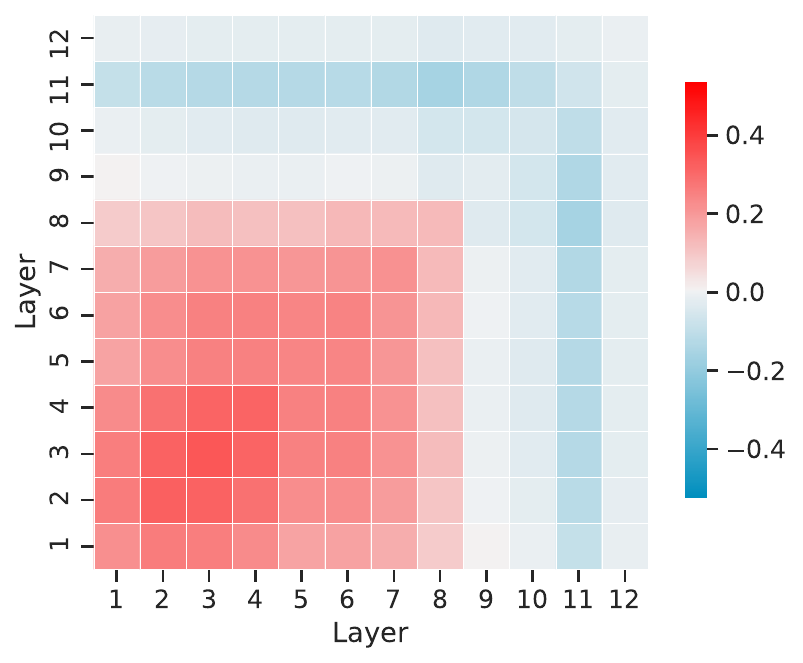}
        \caption{\rnd{}}
        \label{fig:repr_rnd}
    \end{subfigure}
\caption[Representation similarities of \uni{} and FFT (Part I)]{Layerwise difference in representation similarity for the \uni{} adapter and the FFT model on \subj{}. Warm colors (positive values) illustrate layer pairs that demonstrate higher similarity to the base model with the adapter than with FFT. Conversely, cool colors (negative values) represent layer pairs that are more similar to the base model when using the FFT model.}
\label{fig:repr_diff}
\end{figure*}

To compare PEFT and FFT, we calculate the difference in CKA similarity between each model and its base pre-trained counterpart as 
\[ 
\Delta \mathrm{CKA} = \mathrm{CKA}(\textit{adapter}, \textit{base}) - \mathrm{CKA}(\textit{FFT}, \textit{base}).
\] 
\Cref{fig:repr_diff} visualizes these differences across layers for the \subj{} dataset. PEFT methods exhibit higher similarity to the base model in the \textbf{early} and \textbf{middle} layers compared to FFT, regardless of the AL method. This trend is particularly pronounced with \pt{} and \uni{}. In contrast, FFT representations diverge significantly from the base model in these layers, focusing more heavily on task-specific features in deeper layers.

\begin{figure*}[]
    \centering
    \begin{subfigure}[b]{.195\linewidth}
        \includegraphics[width=\linewidth]{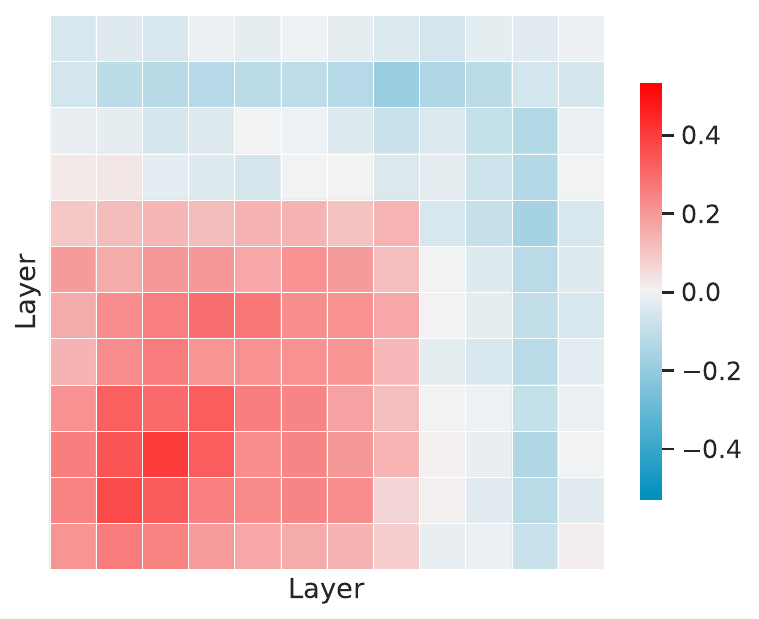}
        \caption{\trec{}; \ent{}}
    \end{subfigure}
    \begin{subfigure}[b]{.195\linewidth}
        \includegraphics[width=\linewidth]{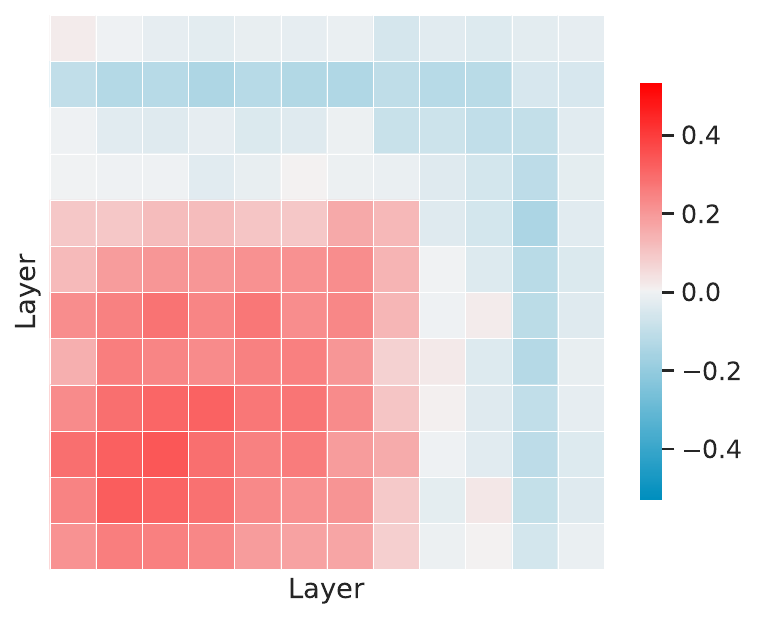}
        \caption{\trec{}; \mc{}}
    \end{subfigure}
    \begin{subfigure}[b]{0.195\linewidth}
        \includegraphics[width=\linewidth]{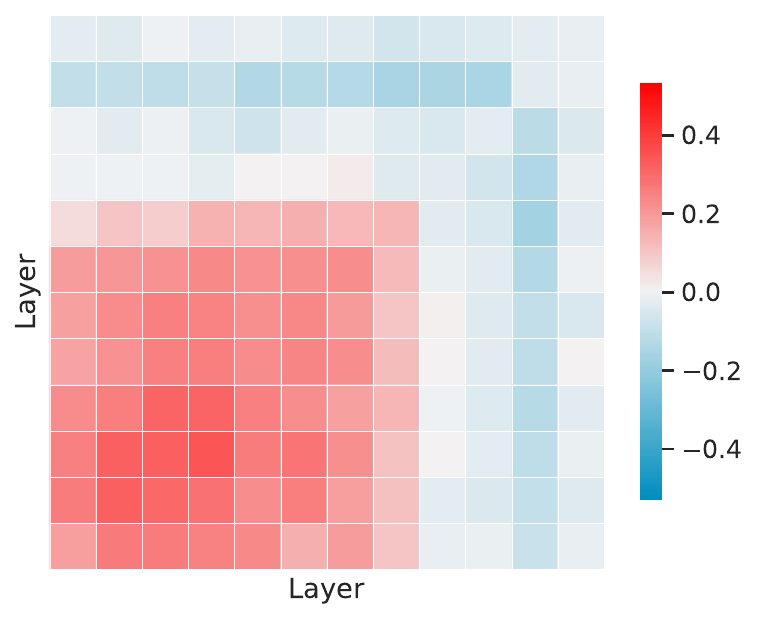}
        \caption{\trec{}; \cs{}}
    \end{subfigure}
    \begin{subfigure}[b]{0.195\linewidth}
        \includegraphics[width=\linewidth]{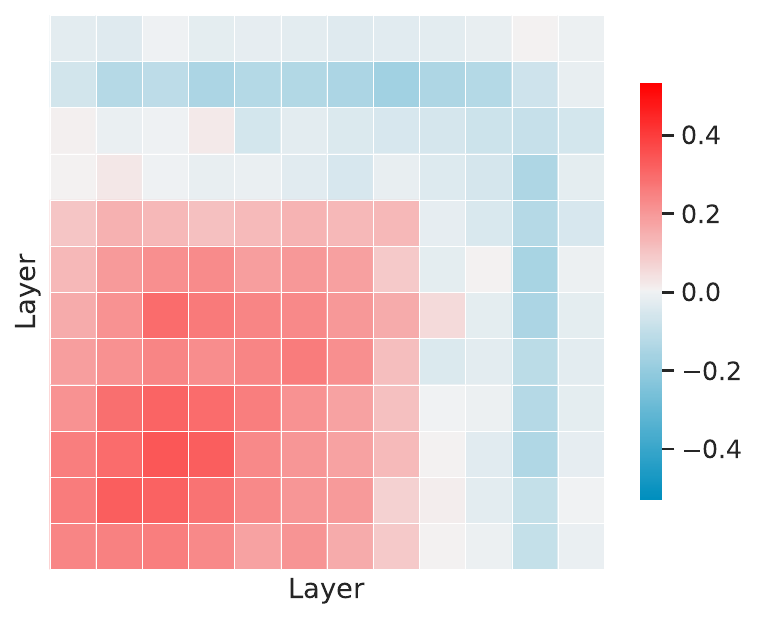}
        \caption{\trec{}; \dal{}}
    \end{subfigure}
    \begin{subfigure}[b]{0.195\linewidth}
        \includegraphics[width=\linewidth]{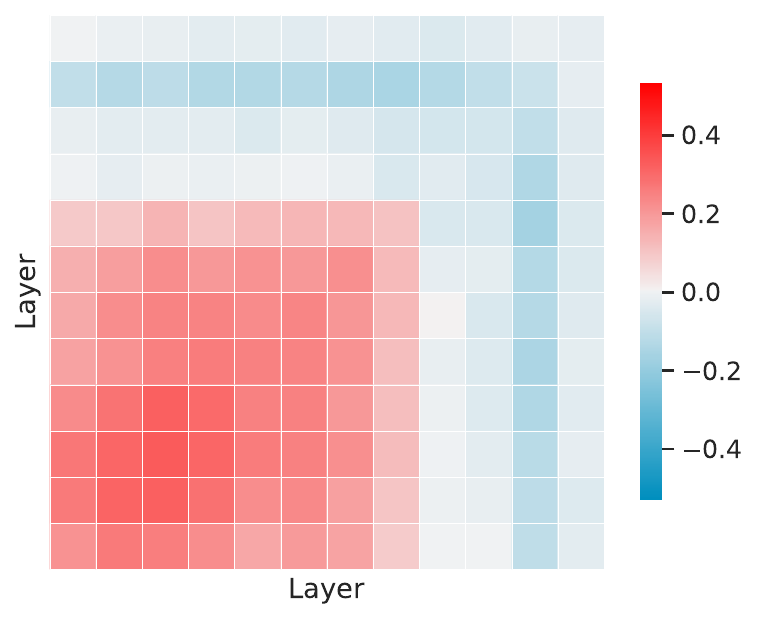}
        \caption{\trec{}; \rnd{}}
    \end{subfigure}

    \begin{subfigure}[b]{.195\linewidth}
        \includegraphics[width=\linewidth]{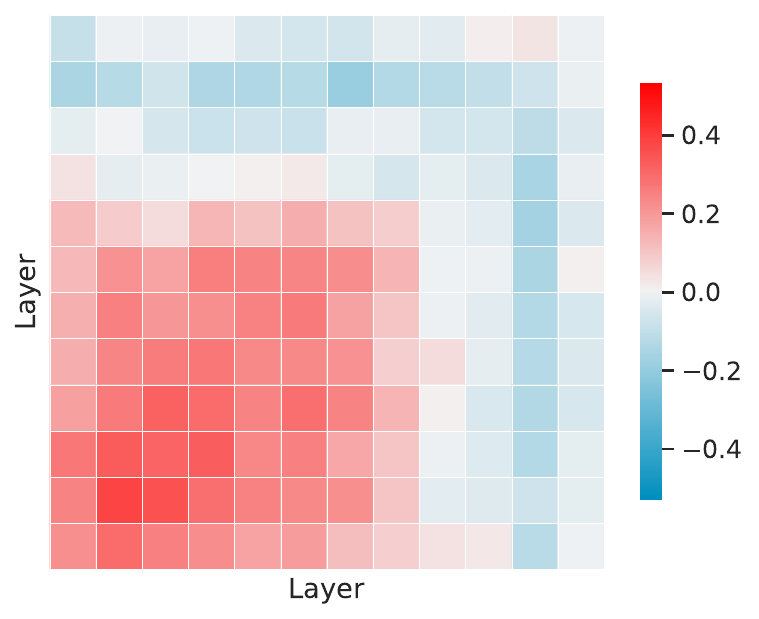}
        \caption{\textsc{sst}; \ent{}}
    \end{subfigure}
    \begin{subfigure}[b]{.195\linewidth}
        \includegraphics[width=\linewidth]{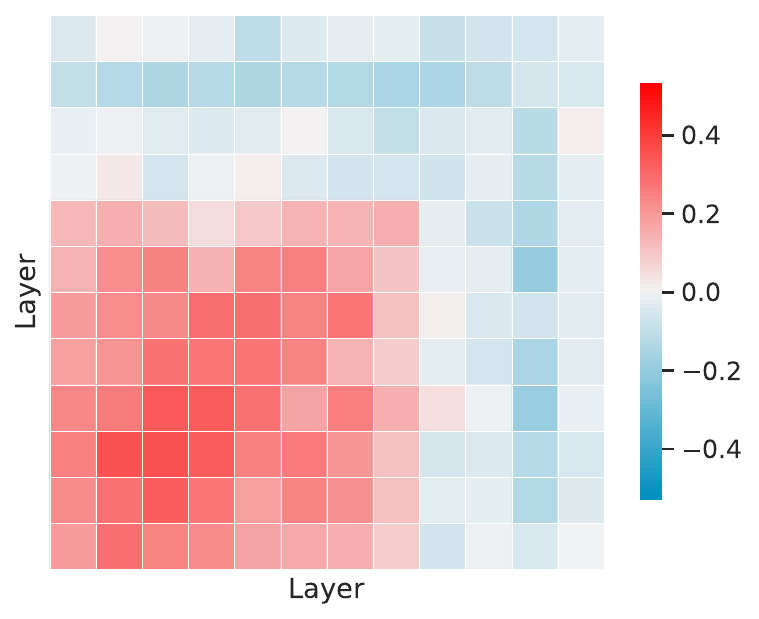}
        \caption{\textsc{sst}; \mc{}}
    \end{subfigure}
    \begin{subfigure}[b]{0.195\linewidth}
        \includegraphics[width=\linewidth]{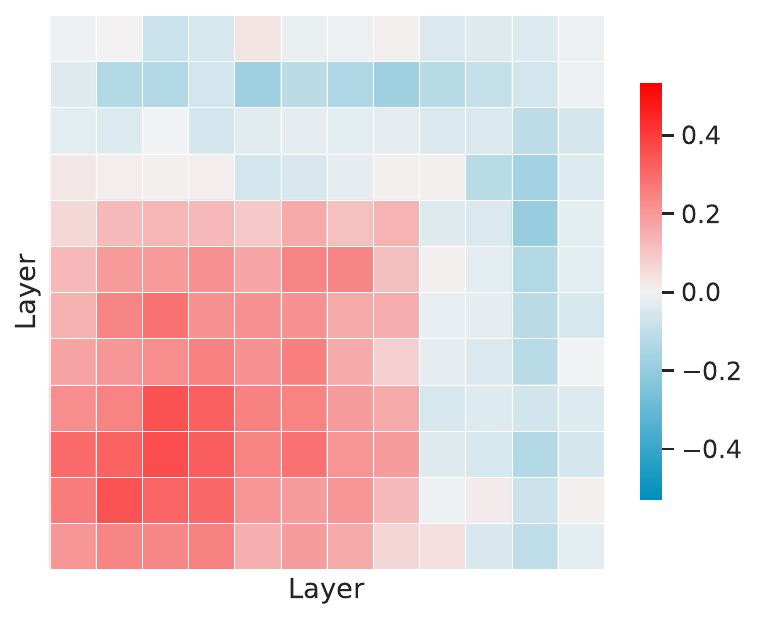}
        \caption{\textsc{sst}; \cs{}}
    \end{subfigure}
    \begin{subfigure}[b]{0.195\linewidth}
        \includegraphics[width=\linewidth]{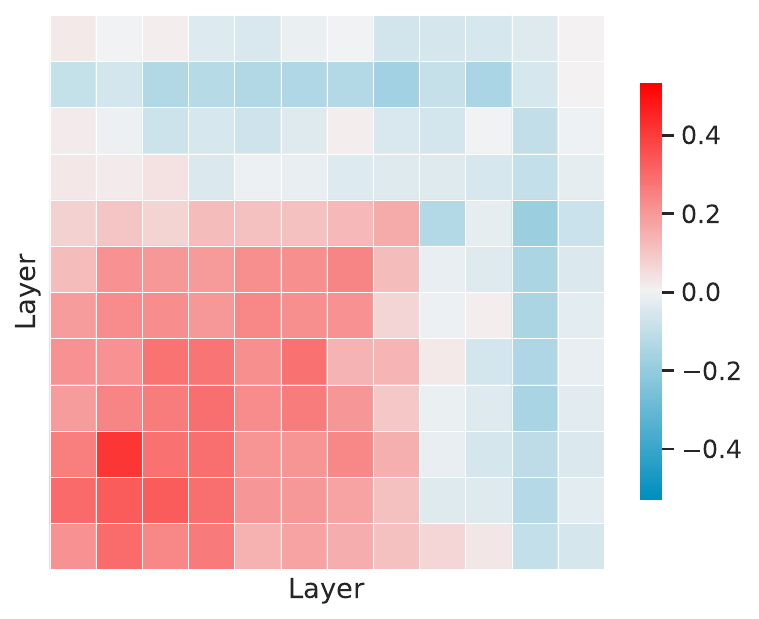}
        \caption{\textsc{sst}; \dal{}}
    \end{subfigure}
    \begin{subfigure}[b]{0.195\linewidth}
        \includegraphics[width=\linewidth]{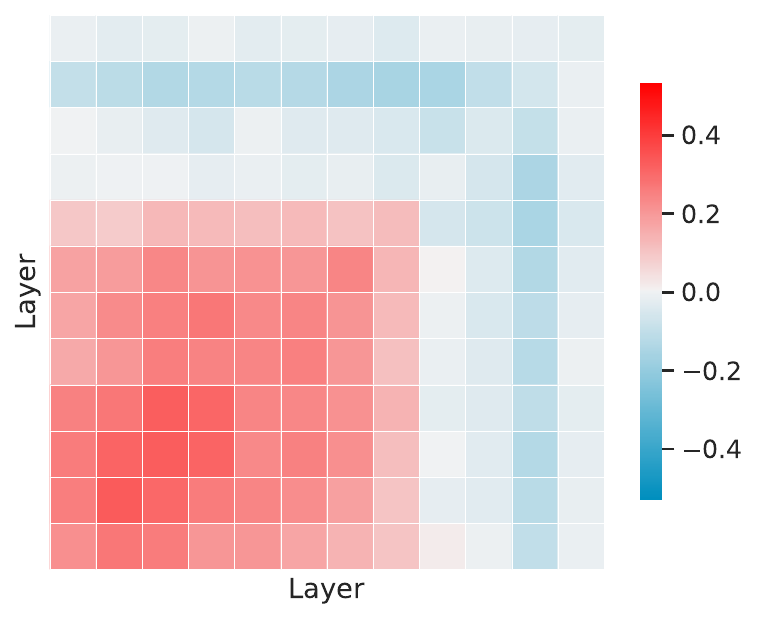}
        \caption{\textsc{sst}; \rnd{}}
    \end{subfigure}

    \begin{subfigure}[b]{.195\linewidth}
        \includegraphics[width=\linewidth]{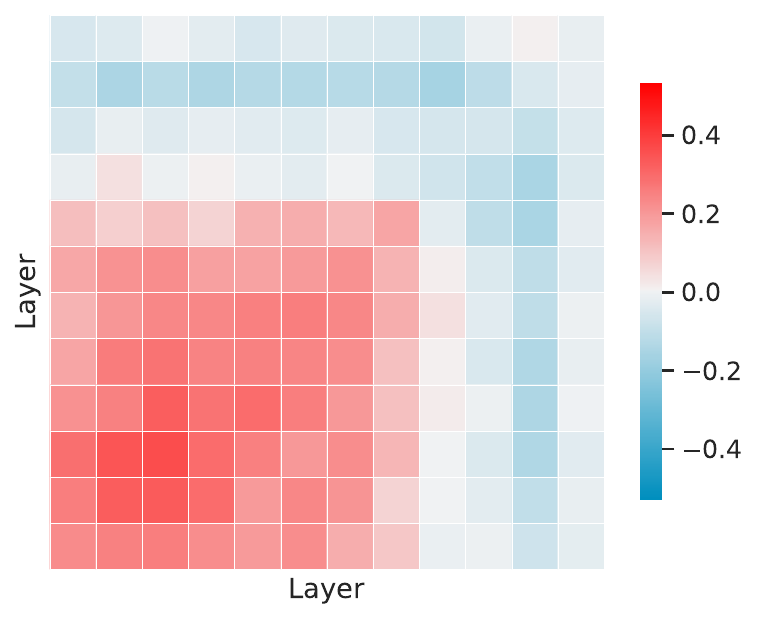}
        \caption{\agn{}; \ent{}}
    \end{subfigure}
    \begin{subfigure}[b]{.195\linewidth}
        \includegraphics[width=\linewidth]{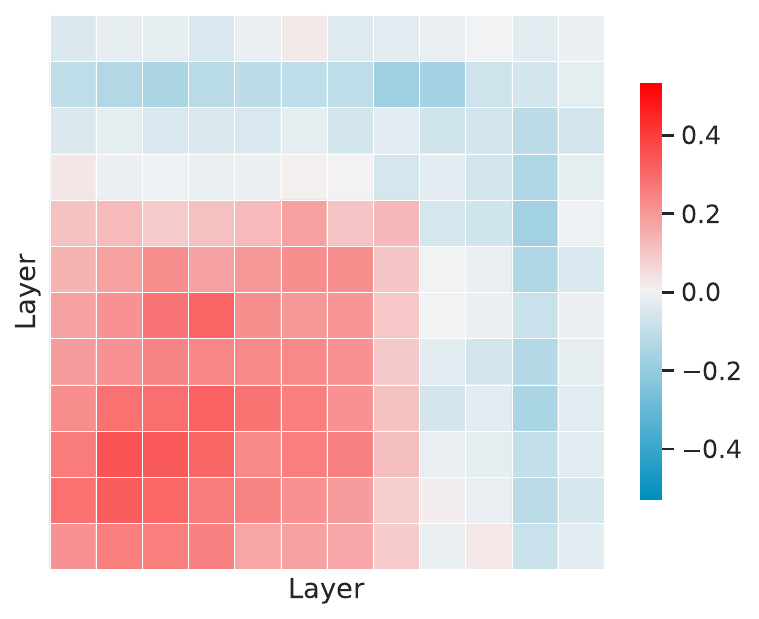}
        \caption{\agn{}; \mc{}}
    \end{subfigure}
    \begin{subfigure}[b]{0.195\linewidth}
        \includegraphics[width=\linewidth]{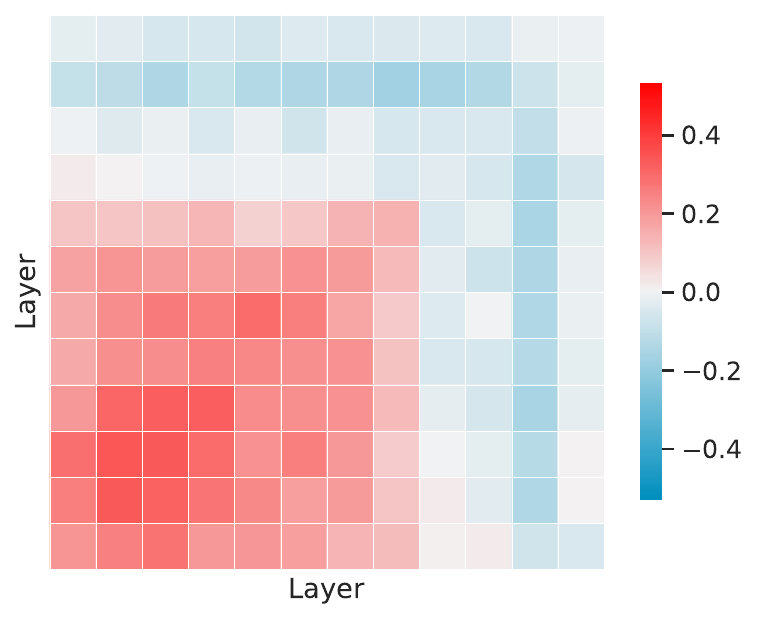}
        \caption{\agn{}; \cs{}}
    \end{subfigure}
    \begin{subfigure}[b]{0.195\linewidth}
        \includegraphics[width=\linewidth]{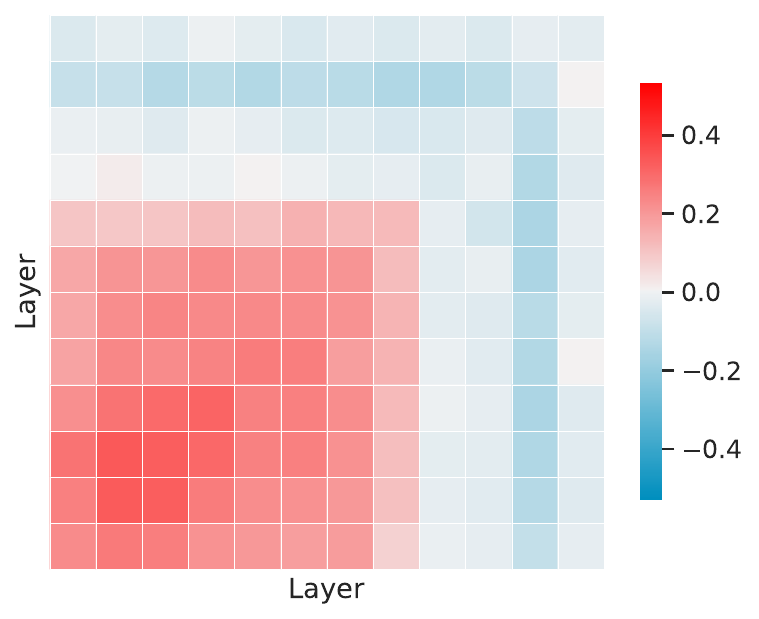}
        \caption{\agn{}; \dal{}}
    \end{subfigure}
    \begin{subfigure}[b]{0.195\linewidth}
        \includegraphics[width=\linewidth]{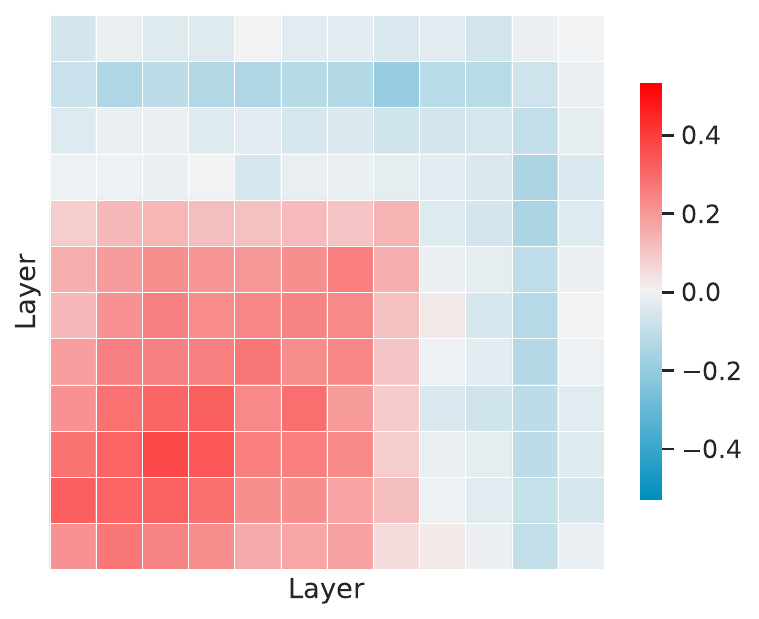}
        \caption{\agn{}; \rnd{}}
    \end{subfigure}
\caption[Representation similarities of \uni{} and FFT (Part II)]{Layerwise difference in representation similarity for the \uni{} adapter and the FFT model on \trec{}, \textsc{sst}, and \agn{}. The differences are computed as $\mathrm{CKA}(\textit{adapter}, \textit{base}) - \mathrm{CKA}(\textit{FFT}, \textit{base})$, where \textit{base} is the corresponding pre-trained \bert{} model. Warm colors (positive values) illustrate layer pairs that demonstrate higher similarity to the base model with the adapter than with FFT. Conversely, cool colors (negative values) represent layer pairs that are more similar to the base model when using the FFT model. Best viewed on a computer screen.}
\label{fig:repr_diff_app}
\end{figure*}

In \Cref{fig:repr_diff_app}, we display the difference in similarities of adapters and FFT compared to their base models on the remaining three datasets.

Our findings suggest that PEFT's stronger alignment with the base model in early layers helps preserve pre-trained knowledge, enhancing generalization. Simultaneously, PEFT introduces task-specific flexibility in deeper layers, achieving a balance between stability and adaptability. This balanced representation enables PEFT to prioritize moderately forgettable instances in AL, driving its superior performance.

\section{Summary}

Our analysis highlights the advantages of PEFT methods in addressing the challenges of low-resource NLP scenarios, both as standalone approaches and when integrated with AL. Specifically, PEFT consistently enhances learning efficiency while reducing computational overhead. 

We found that PEFT methods, particularly \pt{} and \uni{}, outperformed FFT in AL settings by effectively selecting moderately forgettable instances, which contributed to efficient learning. Additionally, TAPT improved the performance of both FFT and PEFT by stabilizing fine-tuning and narrowing the gap between these approaches. Moreover, PEFT methods demonstrated a stronger alignment with the base model’s representations in early and middle layers, ensuring effective task adaptation while preserving pre-trained knowledge.

\part{Weak Supervision with In-Context Learning}
The third part of this thesis investigates the emerging paradigm of in-context learning and introduces a novel weak supervision framework to improve the stability and efficiency of decoder-based LLMs. While ICL enables LLMs to adapt to new tasks without explicit fine-tuning, it remains sensitive to prompt design, demonstration selection, and context window limitations.

\Cref{ch:icl} discusses the principles and limitations of in-context learning, identifying key challenges such as demonstration interference and context window constraints. \Cref{ch:wilda} introduces weak supervision with in-context learning, a framework that employs a teacher-student paradigm to encode contextual knowledge into parameter-efficient modules, reducing reliance on explicit demonstrations during inference. In this approach, the teacher model processes both demonstrations and queries to generate pseudo-labels, which the student model then uses to refine its representations. By iteratively learning from these pseudo-labels, the student internalizes task knowledge more effectively while reducing sensitivity to demonstration variability. 
Designed to minimize computational overhead while preserving strong generalization performance, this framework enables more stable and scalable adaptation to new tasks, making it particularly well-suited for resource-constrained NLP applications.

The contributions in this part of the thesis address critical challenges of in-context learning, offering a structured approach to improving the reliability and efficiency of few-shot adaptation.

\chapter{In-Context Learning}
\label{ch:icl}

Previous research has demonstrated the effectiveness of PEFT and AL in improving data and parameter efficiency for PLMs, particularly in the fine-tuning regime. These methods have been applied mainly to encoder-based PLMs, where direct parameter updates are used to adapt models to downstream tasks while minimizing computational costs.  
However, the emergence of autoregressive decoder-only LLMs has shifted the adaptation paradigm from fine-tuning to prompting and \textbf{in-context learning} (ICL), fundamentally changing how models are leveraged for new tasks. Rather than requiring explicit parameter updates, decoder-only LLMs rely on contextual information within the input to guide their predictions. While this shift has introduced new possibilities, the challenges of data efficiency and parameter constraints persist. These issues warrant alternative strategies that align with the unique properties of LLMs while overcoming their inherent limitations (\Cref{sec:llm_challenges}).

\section{The Foundations of In-Context Learning}

ICL \cite{brown-etal-2020-language} has gained prominence as a transformative paradigm in NLP, particularly with the advent of LLMs. Unlike conventional supervised learning, which involves iterative fine-tuning on labeled datasets, ICL enables models to adapt to new tasks by embedding \textit{demonstrations} and \textit{queries} in a single input prompt. This approach allows LLMs to perform tasks in zero-shot or few-shot settings, making it especially valuable in low-resource scenarios \cite{dong-etal-2024-survey}.

ICL leverages the vast pre-training of LLMs to interpret task demonstrations on the fly, conditioning the model on a few examples within its context window rather than modifying its parameters. This aligns with the growing trend of \textbf{prompt-based learning}, where models are guided via structured prompts rather than explicit gradient-based optimization. Unlike PEFT, which optimizes a subset of model parameters, ICL dynamically utilizes its pre-trained knowledge, adapting representations based on contextual cues rather than explicit updates.

Few-shot ICL extends the model's ability to generalize by providing a limited number of task-specific examples within the prompt. Unlike zero-shot learning, where the model relies solely on pre-trained knowledge and task descriptions, few-shot learning supplies explicit demonstrations, allowing LLMs to infer the intended task structure, label distribution, and input-output mapping. The effectiveness of few-shot ICL is largely influenced by the number, quality, and ordering of demonstrations, as well as the model's ability to recognize patterns from limited context \citep{min-etal-2022-rethinking}.

Few-shot ICL has been widely studied in text classification, machine translation, and reasoning tasks, demonstrating that even with minimal examples, LLMs can approximate supervised learning behavior. This ability is particularly advantageous in low-resource NLP settings, where obtaining large annotated datasets is impractical. However, the success of few-shot ICL depends on factors such as prompt engineering, contextual coherence, and the model’s ability to handle syntactic and semantic variations in the demonstrations.

\begin{example}[Few-shot ICL]
A typical ICL setup consists of demonstrations followed by a query. Consider the task of classifying the sentiment of a movie review:

\begin{quote}
\textbf{Demonstrations:}
\begin{itemize}
    \item \textbf{Review:} ``The cinematography was breathtaking, and the storyline was deeply engaging.''\\ \textbf{Sentiment:} Positive
    \item \textbf{Review:} ``The acting was wooden, and the dialogue felt forced.''\\ \textbf{Sentiment:} Negative
    \item \textbf{Review:} ``The soundtrack perfectly complemented the film's emotional depth.''\\ \textbf{Sentiment:} Positive
\end{itemize}

\textbf{Query:}
\begin{itemize}
    \item \textbf{Review:} ``The pacing was sluggish, and I struggled to stay engaged.''\\ \textbf{Sentiment:} ?
\end{itemize}
\end{quote}

The LLM processes this prompt and generates an answer based on the provided demonstrations. Unlike traditional supervised learning, where the model undergoes weight updates, ICL enables adaptation purely through context.
\end{example}

\section{Theoretical Perspectives on In-Context Learning}

The conceptualization of ICL has shifted from traditional task-learning paradigms toward task-identification frameworks. Its effectiveness has driven extensive empirical and theoretical research into the underlying mechanisms that enable this phenomenon. Research suggests that ICL enables models to recognize latent tasks embedded within their pre-training, allowing for efficient adaptation without explicit parameter updates \cite{wies-etal-2023-the}. This perspective implies that rather than learning new tasks from scratch, LLMs retrieve and align relevant knowledge from pre-training to fit the provided demonstrations. 

Further studies indicate that transformers exhibit structured adaptation mechanisms as they process in-context demonstrations, with models progressing through developmental stages in their ICL capabilities \cite{hoogland-etal-2024-developmental}. The effectiveness of this adaptation depends on the number and quality of examples included in the context, supporting the notion that ICL emerges naturally from deep learning rather than being explicitly programmed.

Empirical analyses have demonstrated that ICL predictions become more robust to input perturbations when longer prompts are used, while exposure to noisy training data can enhance stability \cite{li-etal-2023-transformers}. However, performance improvements are not uniform across different tasks, as the effectiveness of ICL is influenced by the structure of the provided demonstrations and inherent biases from pre-training. These findings highlight the need for further investigation into how LLMs selectively leverage in-context information and whether they can be made more robust to inconsistencies.

\section{Challenges of In-Context Learning}

Despite its advantages, ICL presents several challenges. One major limitation is \textbf{stability}, as LLMs are highly sensitive to prompt variations, including the selection and ordering of demonstrations \cite{li-etal-2024-debiasing, lu-etal-2021-fantastically}. This instability can undermine generalization and make ICL unreliable in practical applications. Additionally, the joint processing of demonstrations and queries within a single prompt introduces constraints related to the model’s context window. As prompts grow longer, issues such as \textbf{primacy and recency biases} arise, where models tend to prioritize information from either the beginning or end of the prompt, leading to inconsistent task adaptation \cite{liu-etal-2024-lost, dong-etal-2024-exploring}.

Another key challenge is \textbf{demonstration selection}, as the placement and ordering of examples significantly impact performance. However, current ICL approaches often rely on heuristic-based selection rather than principled optimization \cite{zhao-etal-2021-calibrate}. Additionally, research suggests that LLMs do not always fully utilize demonstrations, with models sometimes generating outputs based on prior knowledge rather than explicitly leveraging in-context examples \cite{kossen-etal-2024-incontext}. This variability in reliance on provided context underscores the difficulty of ensuring stable and predictable performance across different tasks.

\section{Hybrid Approaches}

While ICL provides a flexible way to adapt LLMs to new tasks without parameter updates, it has inherent limitations related to stability, demonstration selection, and prompt sensitivity. To mitigate these challenges, recent research has explored hybrid learning approaches that often integrate ICL with fine-tuning strategies, leveraging the strengths of both paradigms.

\subsection{Pattern-Based Fine-Tuning}

\textbf{Pattern-based fine-tuning} (PBFT) \citep{schick-schutze-2021-exploiting} is a fine-tuning variant particularly suited for \textbf{decoder-based architectures}. Unlike conventional fine-tuning, which updates model parameters with explicit task-specific objectives, PBFT continues the language modeling objective but on supervised data, effectively reframing traditional fine-tuning as a natural language completion task. This approach is especially beneficial for autoregressive models, as it ensures that task-specific supervision remains aligned with the model’s pre-training distribution.

For example, instead of training on structured classification data in a traditional format, PBFT converts inputs into cloze-style templates.
\begin{example}[Pattern-Based Fine-Tuning]
\textbf{Original classification example:}
\begin{quote}
    \textbf{Input:} ``The movie was visually stunning but lacked emotional depth.'' \\
    \textbf{Label:} Neutral
\end{quote}

\textbf{Pattern-based reformulation:}
\begin{quote}
    \textbf{Prompt:} ``The movie was visually stunning but lacked emotional depth. The overall sentiment of this review is \underline{\hspace{1cm}}.'' \\ 
    \textbf{Expected Output:} ``Neutral''
\end{quote}
\end{example}

By structuring supervised data into a format that resembles the LM’s pre-training distribution, PFT enhances the model’s ability to perform ICL more effectively. Fine-tuned models using this approach exhibit improved generalization in few-shot and zero-shot learning settings, as they maintain the underlying language modeling objective while being exposed to supervised signals.

\subsection{Retrieval-Augmented In-Context Learning}

Another promising hybrid approach is \textbf{retrieval-augmented ICL}, which addresses the challenge of demonstration selection by retrieving the most relevant examples from a knowledge base rather than relying on manually curated prompts \citep{rubin-etal-2022-learning}. In standard ICL, demonstration selection is often heuristic-driven, leading to inconsistencies in performance across different prompts. Retrieval-augmented ICL systematically selects demonstrations that best match the target query, enhancing ICL stability and reducing dependency on pre-training biases.

\begin{example}[Retrieval-Augmented ICL]
\textbf{Standard ICL demonstration:}
\begin{quote}
    \textbf{Task:} Sentiment classification \\
    \textbf{Demonstrations:} (randomly selected)
    \begin{itemize}
        \item ``The cinematography was breathtaking.'' → Positive
        \item ``The acting was subpar.'' → Negative
    \end{itemize}
\end{quote}

\textbf{Retrieval-augmented ICL demonstration:}
\begin{quote}
    \textbf{Query:} ``The soundtrack was mesmerizing.'' \\
    \textbf{Retrieved Demonstrations (semantically similar):}
    \begin{itemize}
        \item ``The visuals were stunning.'' → Positive
        \item ``The sound design was incredible.'' → Positive
    \end{itemize}
\end{quote}
\end{example}

By leveraging retrieval mechanisms, this approach ensures that demonstration selection aligns with the query, reducing variance in ICL performance and improving generalization.

\section{Disentangling Latent Shifts}

Recent studies on ICL's inner mechanisms have illuminated how transformers encode task-specific representations from demonstrations. Transformers have been shown to compress demonstrations into \textbf{in-context vectors} (ICVs), which serve as latent representations of task-specific shifts \citep{hendel-etal-2023-context, liu-etal-2023-context}. These vectors influence how models generate context-appropriate outputs for queries, effectively simulating parameter updates without modifying model weights. Building on these findings, virtual gradients are computed via linear attention to mimic gradient-based learning within the model \citep{dai-etal-2023-gpt}. Similarly, causal mediation analysis has been used to identify the role of specific attention heads in forming robust task representations, introducing the concept of function vectors \citep{todd-etal-2024-function}. These advancements highlight the potential of disentangling latent shifts to improve ICL’s efficiency and reliability.

ICVs offer an alternative view of ICL by representing in-context adaptation as implicit weight updates. Rather than treating ICL as a direct instance of few-shot learning, ICVs frame it as a process where the model dynamically adjusts its intermediate representations to align with task demonstrations. This perspective aligns with research demonstrating that transformers leverage demonstration tokens to form compact, task-specific adjustments \citep{dong-etal-2024-survey}.

Another complementary approach to improving ICL is \textbf{batch in-context learning} (Batch-ICL), which processes multiple ICL instances simultaneously by optimizing batch-wise attention mechanisms. Unlike standard ICL, which treats each prompt separately, Batch-ICL leverages shared attention computations across multiple prompts, enabling more stable and efficient adaptation \citep{zhang-etal-2024-batch}. This method reduces variability in model responses caused by prompt reordering and mitigates inconsistencies in demonstration selection, a major challenge in traditional ICL.

Disentangling in-context knowledge from the query can aid in improving the efficiency and stability of ICL. Current approaches rely on manipulating the outputs of attention heads or hidden states. The motivation behind disentangling lies in previous research \citep{aizerman-1964-theoretical, irie-etal-2022-dual}, demonstrating that linear layers optimized through gradient descent have a dual form of linear attention. To illustrate, consider a neural network's linear layer, where $\mathbf{W}_0, \Delta \mathbf{W} \in \mathbb{R}^{m \times n}$ denote the initial weight matrix and its subsequent updates by backpropagation, respectively. With $\mathbf{x} \in \mathbb{R}^{m}$ as the input representation, a linear transformation $\mathbf{f}: \mathbb{R}^m \to \mathbb{R}^n$ can be expressed as
\begin{equation}
    \mathbf{f}(\mathbf{x}) = (\mathbf{W}_0 + \Delta \mathbf{W}) \mathbf{x} .
\end{equation}
During backpropagation, $\Delta \mathbf{W}$ is computed by accumulating the outer products (denoted by $\otimes$) of $N$ training examples $\{\mathbf{x}_1, \mathbf{x}_2, \dots, \mathbf{x}_N\}$, where $\mathbf{x}_i \in \mathbb{R}^{m}$, and the error signals $\{\mathbf{e}_1, \mathbf{e}_2, \dots, \mathbf{e}_N\}$, where $\mathbf{e}_i \in \mathbb{R}^{n}$, obtained from the gradients of the loss function:
\begin{equation}
    \Delta \mathbf{W} = \sum_{i=1}^N{\mathbf{e}_i \otimes \mathbf{x}_i} .
\end{equation}
The update part of linear layers optimized by gradient descent can be expressed as unnormalized linear dot-product attention \cite{irie-etal-2022-dual}:
\begin{equation}
    \mathbf{f} (\mathbf{x})
    = (\mathbf{W}_0 + \Delta\mathbf{W}) \mathbf{x} 
    = \mathbf{W}_0 \mathbf{x} + \sum_{i=1}^N{(\mathbf{e}_i \otimes \mathbf{x}_i) \mathbf{x}} 
    =  \mathbf{W}_0 \mathbf{x} + \underbrace{\sum_{i=1}^N{\mathbf{e}_i (\mathbf{x}_i^T \mathbf{x}})}_{\text{linear attention}} .\\ 
\label{eq:dual}
\end{equation}
In the context of the attention mechanism, this shows that the latent shift $\Delta \mathbf{W} \mathbf{x}$ corresponds directly to the application of linear attention, with error signals $\mathbf{e}_i$ as values, training examples $\mathbf{x}_i$ as keys, and the current input $\mathbf{x}$ as the attention query.

The concept of disentangling the latent shifts described in (\ref{eq:dual}) can be extended to ICL, albeit only under the approximation of linear attention. Let $\mathbf{W}_V$, $\mathbf{W}_K$, and $\mathbf{W}_Q$ denote the weight matrices for values, keys, and queries, respectively. Let $\mathbf{x}_q^{(t)}$ represent the current query token's embedding at step $t$, and $\mathbf{q}^{(t)} = \mathbf{W}_Q \mathbf{x}_q^{(t)}$ is the corresponding attention query vector. The matrix $\mathbf{X}_q = [\mathbf{x}_q^{(1)}, \mathbf{x}_q^{(2)}, \dots, \mathbf{x}_q^{(t-1)}]$ contains all previous query token representations up to $t-1$, and $\mathbf{X}_d$ is the matrix of demonstration token representations. The concatenation $[\mathbf{X}_d; \mathbf{X}_q]$ along the sequence dimension is used to compute the attention output at step $t$, expressed as:
\begin{equation}
 \mathbf{f}_\text{AH}(\mathbf{x}_q^{(t)}) = \mathbf{W}_V [\mathbf{X}_d; \mathbf{X}_q] \; \softmax \left( \frac{ \left( \mathbf{W}_K [\mathbf{X}_d; \mathbf{X}_q] \right)^\top \mathbf{q}^{(t)} }{ \sqrt{d} } \right),
\end{equation}
where $d$ is the scaling factor (i.e., the dimensionality of the key vectors). By approximating the attention mechanism with linear attention, it becomes possible to disentangle the latent shift of the zero-shot output of an attention head induced by the query from the latent shift induced by the demonstrations \citep{dai-etal-2023-gpt}:
\begin{equation}
\begin{aligned}
    \mathbf{f}_\text{AH}(\mathbf{x}_q^{(t)}) & \approx \mathbf{W}_V [\mathbf{X}_d; \mathbf{X}_q] \left( \mathbf{W}_K [\mathbf{X}_d; \mathbf{X}_q] \right)^\top \mathbf{q}^{(t)} \\
    & = \underbrace{ \mathbf{W}_V \mathbf{X}_q \left( \mathbf{W}_K \mathbf{X}_q \right)^\top }_{\mathbf{W}_{\text{ZS}}} \mathbf{q}^{(t)} + \underbrace{ \mathbf{W}_V \mathbf{X}_d \left( \mathbf{W}_K \mathbf{X}_d \right)^\top }_{\Delta \mathbf{W}_\text{ICL}} \mathbf{q}^{(t)} \\
    & = \left( \mathbf{W}_{\text{ZS}} + \Delta \mathbf{W}_\text{ICL} \right) \mathbf{q}^{(t)} .
\end{aligned}
\label{eq:approx}
\end{equation}
This approximation disentangles the latent shift induced by the demonstrations $\mathbf{X}_d$ from that induced by the query $\mathbf{x}_q^{(t)}$. \textit{The contribution from ICL is captured as a virtual weight update $\Delta \mathbf{W}_\text{ICL}$, corresponding to virtual gradients}, often referred to as ``meta-gradients'' in the literature. The zero-shot latent shift of the query, corresponding to $\mathbf{W}_{\text{ZS}} \mathbf{q}^{(t)}$, reflects the output without demonstrations, providing the initial state. Analogous to $\Delta \mathbf{W} \mathbf{x}$ in (\ref{eq:dual}), the latent shift $\Delta \mathbf{W}_\text{ICL} \mathbf{q}^{(t)}$ reflects the contribution of ICL. Finally, by substituting $\mathbf{h}_\text{ZS} = \mathbf{W}_\text{ZS} \mathbf{q}^{(t)}$ and $\Delta \mathbf{h}_\text{ICL} = \Delta \mathbf{W}_\text{ICL} \mathbf{q}^{(t)}$, we can rewrite the output of an attention head as
\begin{equation}
    \mathbf{f}_\text{AH}(\mathbf{x}_q^{(t)}) \approx \mathbf{h}_\text{ZS} + \Delta \mathbf{h}_\text{ICL} .
    \label{eq:clean}
\end{equation}

To understand (\ref{eq:approx}), we expand on the key intermediate steps for clarity, which were not explicitly covered in the original work. The goal is to decompose the attention head output into separate components corresponding to the demonstrations and the query, thereby disentangling the latent shifts induced by ICL.

\subsubsection{Expanding the Concatenated Matrices}

The key in deriving the expression is in the step of matrix expansion. We can expand the concatenated matrices as follows:
\begin{align}
\mathbf{W}_V [\mathbf{X}_d ; \mathbf{X}_q] &= [\mathbf{W}_V \mathbf{X}_d ; \mathbf{W}_V \mathbf{X}_q] = [\mathbf{V}_d ; \mathbf{V}_q],
\\ \mathbf{W}_K [\mathbf{X}_d ; \mathbf{X}_q] &= [\mathbf{W}_K \mathbf{X}_d ; \mathbf{W}_K \mathbf{X}_q] = [\mathbf{K}_d ; \mathbf{K}_q],
\end{align}
where:
\begin{itemize} \item $\mathbf{V}_d = \mathbf{W}_V \mathbf{X}_d$ is the value matrix for the demonstrations; \item $\mathbf{V}_q = \mathbf{W}_V \mathbf{X}_q$ is the value matrix for the previous queries; \item $\mathbf{K}_d = \mathbf{W}_K \mathbf{X}_d$ is the key matrix for the demonstrations; \item $\mathbf{K}_q = \mathbf{W}_K \mathbf{X}_q$ is the key matrix for the previous queries.
\end{itemize}
The transpose of the concatenated key matrix is:
\begin{equation} \left( \mathbf{W}_K [\mathbf{X}_d ; \mathbf{X}_q] \right)^\top = \begin{bmatrix} \mathbf{K}_d^\top ; \mathbf{K}_q^\top \end{bmatrix}. \end{equation}

\subsubsection{Performing the Matrix Multiplication}

Substituting the expanded forms using rules for block matrix multiplication, we have:
\begin{equation} 
\mathbf{f}_\text{AH}(\mathbf{x}_q^{(t)}) \approx \begin{bmatrix} \mathbf{V}_d ; \mathbf{V}_q \end{bmatrix} \begin{bmatrix} \mathbf{K}_d^\top ; \mathbf{K}_q^\top \end{bmatrix} \mathbf{q}^{(t)} = \left( \mathbf{V}_d \mathbf{K}_d^\top + \mathbf{V}_q \mathbf{K}_q^\top \right) \mathbf{q}^{(t)}. 
\label{eq:expanded_forms}
\end{equation}
This separates the contributions from the demonstrations and the query sequences.

\subsubsection{Defining the Components}

We define:
\begin{align}
\mathbf{W}_{\text{ZS}} &= \mathbf{V}_q \mathbf{K}_q^\top = \mathbf{W}_V \mathbf{X}_q \left( \mathbf{W}_K \mathbf{X}q \right)^\top,
\label{eq:dual_zs}
\\
\Delta \mathbf{W}_\text{ICL} &= \mathbf{V}_d \mathbf{K}_d^\top = \mathbf{W}_V \mathbf{X}_d \left( \mathbf{W}_K \mathbf{X}_d \right)^\top.
\label{eq:dual_icl}
\end{align}
Here: \begin{itemize} \item $\mathbf{W}_{\text{ZS}}$ represents the zero-shot component, capturing the model's behavior based on the query sequence alone; \item $\Delta \mathbf{W}_\text{ICL}$ represents the latent shift induced by the demonstrations, capturing the effect of in-context learning. \end{itemize}

\subsubsection{Final Expression}

Substituting (\ref{eq:dual_zs}) and (\ref{eq:dual_icl}) back into the expression, we obtain:
\begin{equation} 
\mathbf{f}_\text{AH}(\mathbf{x}_q^{(t)}) \approx \left( \mathbf{W}_{\text{ZS}} + \Delta \mathbf{W}_\text{ICL} \right) \mathbf{q}^{(t)} = \mathbf{W}_{\text{ZS}} \mathbf{q}^{(t)} + \Delta \mathbf{W}_\text{ICL} \mathbf{q}^{(t)}.
\end{equation}

\subsubsection{Interpretation}

The decomposition shows that the attention head output can be viewed as the sum of: \begin{enumerate} \item The \textbf{zero-shot component} ($\mathbf{W}_{\text{ZS}} \mathbf{q}^{(t)}$): the model's output when only the query sequence is considered, without any influence from the demonstrations; \item The \textbf{latent shift due to ICL} ($\Delta \mathbf{W}_\text{ICL} \mathbf{q}^{(t)}$): the additional contribution from the demonstrations, representing the knowledge introduced via in-context learning. \end{enumerate}

This separation aligns with the theoretical motivation to disentangle the latent shifts induced by the demonstrations from those induced by the query, allowing for more efficient and stable processing of queries independently of demonstrations.

Although transformer-based LLMs use non-linear attention in practice, many approaches \citep{dai-etal-2023-gpt, zhang-etal-2024-batch, todd-etal-2024-function} rely on the theoretical underpinnings of linear attention. These methods manipulate attention heads or hidden states to disentangle latent shifts despite the inherent non-linearity of the models. Furthermore, this simplification overlooks other crucial components of the transformer architecture, such as the feed-forward layers, activation functions, and residual connections. While approaches based on linear attention have proven effective, they leave room for further improvements in capturing and disentangling the full complexity of how transformers process data. This thesis explores how virtual weight updates can be obtained more directly while preserving the key components of the transformer architecture.

\chapter{Weak Supervision for Disentangling Latent Shifts}
\label{ch:wilda}

A key challenge in ICL is the entanglement of \textbf{latent shifts} -- internal adjustments made by the model in response to demonstrations -- with query processing. This entanglement complicates the extraction of task-relevant information and contributes to instability. Additionally, the limited context window of language models restricts the explicit retention of learned task representations across queries, further exacerbating the issue. To mitigate these challenges, we seek a principled method that can effectively disentangle latent shifts while maintaining computational efficiency. 
Methods leveraging task and function vectors \citep{liu-etal-2023-context, todd-etal-2024-function} have explored storing demonstration-induced knowledge for reuse without reprocessing demonstrations for each query. However, many of these approaches rely on approximations, such as attention manipulation or hidden-state adjustments, leaving room for more principled methods.

To address this gap, this chapter introduces a method for disentangling latent shifts in LLMs through a \textbf{weak supervision} framework. Our approach offers a principled and data-efficient method for refining model representation, thereby eliminating the need for additional supervision.  
In contrast to existing methods that manipulate attention mechanisms or alter hidden states \citep{liu-etal-2023-context, todd-etal-2024-function}, our method encodes ICL-induced knowledge directly into adapter parameters. This enables more stable learning dynamics and improved generalization beyond the initial demonstrations. By avoiding intrusive modifications to model internals, it offers a more structured and reusable integration of task-specific information.  
By embedding in-context knowledge within lightweight adapter modules, this method aligns with broader efforts to develop data- and parameter-efficient strategies for optimizing LLM adaptation in resource-constrained scenarios.  

We begin by outlining the theoretical foundations of our approach, followed by an introduction to our method, which includes a detailed explanation of its underlying mechanism. We then conduct a comprehensive empirical evaluation, assessing the effectiveness, advantages, and limitations of our approach.

\section{Weak-to-Strong Generalization}

Weak supervision is built on the premise that learning signals do not need to be perfect to be effective. Instead of relying on fully annotated datasets, it allows models to learn from noisy, indirect, or incomplete sources of supervision -- such as heuristics, distant supervision, or model-generated signals. When exploited properly, these weak signals can drive meaningful generalization, particularly in low-resource or data-constrained environments.

Our approach is situated within the broader paradigm of \textbf{weak-to-strong} (W2S) generalization \cite{lang-etal-2024-theoretical}, in which a weaker learner improves by leveraging imperfect supervision from a stronger source. This paradigm is often instantiated through a \textit{teacher--student} framework, where a teacher model provides weak supervision, such as noisy or partial labels, that the student model refines over time. Through this process, the student achieves stronger generalization by correcting errors in the supervision signal and expanding its coverage of the representation space. Two key phenomena underlie this transition: \textit{pseudo-label correction} and \textit{coverage expansion}.

\subsection{Pseudo-Label Correction and Coverage Expansion}

\textbf{Pseudo-label correction} refers to the process by which the student model refines the pseudo-labels generated by the teacher, particularly in cases where the initial predictions were incorrect. This correction mechanism is central to W2S generalization, as it enables the model to progressively improve by leveraging high-confidence predictions to correct errors in lower-confidence regions \cite{lang-etal-2024-theoretical}. The ability to refine pseudo-labels is strongly linked to local consistency in the representation space: when the student successfully corrects errors in one region, these corrections propagate to structurally similar instances, fostering \textit{local-to-global consistency} in the learned representations.

\textbf{Coverage expansion} builds on this correction mechanism by extending the effective learning region beyond the initial scope of the teacher model. Typically, the teacher provides pseudo-labels only for regions where it is confident, leaving substantial portions of the input space unexplored. However, when the student corrects pseudo-labels in uncertain regions, it effectively expands coverage, allowing the model to generalize beyond the teacher’s initial decision boundary \cite{lang-etal-2024-theoretical}. This expansion is particularly important for out-of-distribution generalization, as it enables the model to extrapolate knowledge to unseen data while maintaining prediction consistency.

\begin{figure}[]
\centering
\begin{tikzpicture}
    \definecolor{myred}{RGB}{255, 102, 102}
    \definecolor{mygreen}{RGB}{102, 204, 102}
    \definecolor{myorange}{RGB}{255, 178, 102}
    \definecolor{myblue}{RGB}{102, 178, 255}

    \definecolor{myredtext}{RGB}{204, 51, 51}
    \definecolor{mygreentext}{RGB}{0, 153, 0}
    \definecolor{mybluetext}{RGB}{0, 102, 204}

    \fill[myred!30, draw=myred, line width=0.4mm] (0,0) ellipse (2cm and 1.5cm);
    \node at (0.5, -1.1) {$N(X)$};

    \fill[mygreen!30, draw=mygreen, line width=0.4mm] (-0.3,0.3) ellipse (1.4cm and 0.9cm);
    \node at (-1.3, 0.6) {$Y$};

    \foreach \x/\y in {-1.0/0.1, -0.8/0.5, -0.4/-0.1, 0.0/0.4, -0.6/-0.3, 0.5/0.7} {
        \fill[black] (\x,\y) circle (2pt);
    }

    \fill[black] (0.9,-0.7) circle (2pt);

    \fill[myorange!30, draw=myorange, line width=0.4mm] (5,0) ellipse (1.5cm and 1cm);
    \node at (5.5, 0.7) {$X$};

    \foreach \x/\y in {4.6/0.3, 4.9/0.1, 5.2/0.6, 5.4/-0.4, 5.0/-0.6, 5.6/0.0} {
        \fill[black] (\x,\y) circle (2pt);
    }

    \draw[very thin, draw=gray] (0.9,-0.7) to[out=0, in=180] (4.6,0.3);
    \draw[very thin, draw=gray] (-1.0,0.1) to[out=10, in=190] (4.9,0.1);
    \draw[very thin, draw=gray] (-0.8,0.5) to[out=20, in=160] (5.2,0.6);
    \draw[very thin, draw=gray] (-0.4,-0.1) to[out=0, in=200] (5.4,-0.4);
    \draw[very thin, draw=gray] (0.0,0.4) to[out=10, in=220] (5.0,-0.6);
    \draw[very thin, draw=gray] (-0.6,-0.3) to[out=-10, in=190] (5.6,0.0);
    \draw[very thin, draw=gray] (0.5,0.7) to[out=10, in=240] (5.0,-0.6);

    \fill[myblue!30, draw=myblue, line width=0.4mm] (-2.8,0.4) circle (0.6cm);
    \node at (-2.8, 0.6) {$N(Y)$};

    \fill[black] (-2.7,0.2) circle (2pt);

    \draw[very thin, draw=gray] (-2.7,0.2) to[out=10, in=170] (-1.0,0.1);

    \node[anchor=east, text=mybluetext, font=\bfseries] at (-3.4,0.4) {Uncovered space};
    \node[anchor=south, text=mygreentext, font=\bfseries] at (-3.2, 1) {Robust neighborhood};
    \node[anchor=north, text=myredtext, font=\bfseries] at (-2.7, -1) {Neighborhood};

    \node[anchor=north, text=myorange, font=\bfseries] at (5, -1) {Generated pseudo-labels};

\end{tikzpicture}
\caption[Illustration of pseudo-label correction and coverage expansion]{Illustration of pseudo-label correction and coverage expansion. The generated pseudo-labels $X$ (orange) originate from initially labeled instances but may contain errors. If the learner (i.e., student) is locally consistent, pseudo-label correction refines these labels by leveraging high-confidence neighboring points within the robust neighborhood $Y$ (green). Coverage expansion extends learning into the uncovered space $N(Y)$ (blue), allowing the model to incorporate previously unrecognized data points into its decision boundary. The neighborhood $N(X)$ (red) represents the initial region influenced by labeled data, where $N(\cdot)$ denotes the neighborhood function of a given set of points. Gray arrows illustrate how pseudo-labels propagate across the representation space.}
\label{fig:coverage}
\end{figure}
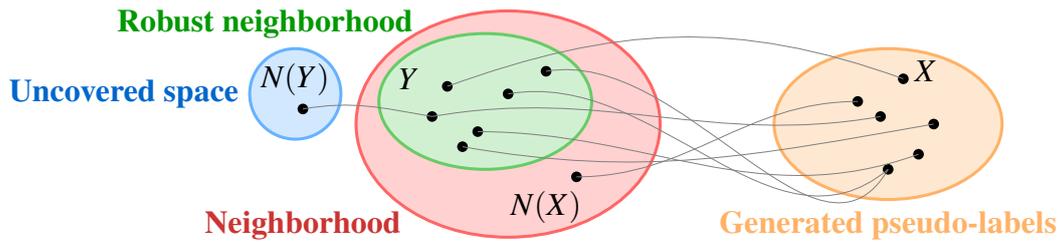

These two mechanisms, pseudo-label correction and coverage expansion, are illustrated in \Cref{fig:coverage}. The figure shows how incorrect pseudo-labels are refined using high-confidence neighboring points and how learning extends beyond the initially labeled regions, incorporating previously unrecognized data points into the model’s decision boundary.

\subsection{Teacher–Student Paradigms in LLMs}

Recent advancements have extended teacher--student learning paradigms to LLMs, demonstrating their potential to enhance reasoning, stability, and generalization. Unlike traditional settings where the teacher provides hard pseudo-labels, modern approaches often rely on more informative supervisory signals, such as rationale-augmented or chain-of-thought responses, to guide the student \cite{huang-etal-2023-large}.
Iterative refinement techniques, such as recursive introspection \cite{qu-etal-2024-recursive} and self-taught evaluators \cite{wang-etal-2024-self}, embody this paradigm by allowing the model to generate, evaluate, and revise its own outputs through weak supervision. In this setting, the model alternates between teacher and student roles across iterations, refining its internal representations and extending its coverage of the task space. These processes directly support the core mechanisms of W2S generalization: \textit{pseudo-label correction} and \textit{coverage expansion}.

Collectively, these studies highlight the significance of teacher--student frameworks for improving model robustness and adaptability. By leveraging pseudo-label correction to refine knowledge and coverage expansion to generalize beyond initial training data, weak supervision provides a compelling framework for achieving W2S generalization.

\section{Weak Supervision with ICL}

Building on the concept of disentangling latent shifts in transformers, we introduce \wilda (\textbf{W}eakly-supervised \textbf{I}n-context \textbf{L}earning \textbf{D}isentanglement via \textbf{A}dapters), an approach that offers a simple yet highly efficient way to internalize ICL knowledge into the parameters of a model. Rather than relying solely on manipulating attention heads, as is common in current methods, \wilda aims to capture the full complexity of the transformer's components -- considering the final output, which depends on all layers, including attention heads, feed-forward layers, and residual connections. By aligning more directly with the actual latent shifts induced by demonstrations, \wilda ensures that the model uses the entirety of its architecture to first embed and later apply in-context knowledge.

\begin{figure}[]
\centering
\begin{tikzpicture}[
    teacherbox/.style={rectangle, rounded corners, draw, minimum height=0.8cm, minimum width=3.5cm, fill=gray!30, align=center, line width=0.5mm},
    adapter/.style={rectangle, rounded corners, draw, minimum height=0.8cm, minimum width=1cm, align=center, line width=0.4mm, fill=orange!30},
    fadedadapter/.style={rectangle, rounded corners, draw, minimum height=0.8cm, minimum width=1cm, align=center, line width=0.2mm, fill=orange!20, opacity=0.6},
    fadedadapter2/.style={rectangle, rounded corners, draw, minimum height=0.8cm, minimum width=1cm, align=center, line width=0.2mm, fill=orange!10, opacity=0.3},
    demobox/.style={rectangle, rounded corners, draw=myblue!70, minimum height=1.4cm, minimum width=2.8cm, align=center, fill=myblue!15, line width=0.6mm},
    querybox/.style={rectangle, rounded corners, draw=myred!70, minimum height=0.4cm, minimum width=2cm, align=center, fill=myred!15, line width=0.6mm},
    arrow_teacher/.style={-{Triangle}, very thick, draw=gray!90},
    arrow_student/.style={-{Triangle}, very thick, draw=black!90},
    concatenation/.style={inner sep=1pt},
    studentbox/.style={rectangle, rounded corners, draw, minimum height=1.6cm, minimum width=5cm, fill=gray!10, align=center, line width=0.5mm},
    dashedbox/.style={rectangle, rounded corners, draw, minimum height=0.8cm, minimum width=2.5cm, align=center, line width=0.4mm, fill=orange!15}
    ]

    \node[teacherbox] (teacherLLM) {\textbf{Teacher LLM}};

    \node[studentbox, below=0.2cm of teacherLLM] (studentLLM) {};
    \node[anchor=north west] at ([xshift=0.1cm, yshift=-0.1cm]studentLLM.north west) {\textbf{Student LLM}};

    \node[dashedbox, minimum width=2.5cm, anchor=west, xshift=0.2cm, yshift=-0.2cm] at (studentLLM.west) (baseLLM) {Base LLM};
    \node[fadedadapter2, minimum width=1.8cm, anchor=west, xshift=0.3cm, yshift=0.2cm] at (baseLLM.east) {};
    \node[fadedadapter, minimum width=1.8cm, anchor=west, xshift=0.2cm, yshift=0.1cm] at (baseLLM.east) {};
    \node[adapter, minimum width=1.8cm, anchor=west, right=0.1cm of baseLLM] (adapter) {Adapter};

    \node[demobox, left=2.5cm of teacherLLM, yshift=-0.3cm] (demonstrations) {Demonstrations:\\[5pt] $\mathbf{X}_d = [\mathbf{x}_1,\, \mathbf{x}_2,\, \ldots,\, \mathbf{x}_n]$};
    \node[querybox, below=0.5cm of demonstrations] (query) {Query: $\mathbf{x}_q$};

    \node[concatenation, right=0.75cm of demonstrations] (concat) {\scalebox{1.5}{$\concat$}};

    \draw [arrow_teacher] (demonstrations.east) -- (concat.west);

    \draw [arrow_teacher] (query.east) to[out=0, in=230] (concat.south west);

    \draw [arrow_teacher] (concat.east) to[out=0, in=180] (teacherLLM.west);

    \node[right=1.25cm of teacherLLM] (yt) {$\mathbf{y}_t$};
    \draw [arrow_teacher] (teacherLLM.east) -- (yt.west);

    \draw [arrow_student] (query.east) to[out=0, in=180] (studentLLM.west);

    \node[right=0.5cm of studentLLM] (ys) {$\mathbf{y}_s$};
    \draw [arrow_student] (studentLLM.east) -- (ys.west);

    \draw [decorate,decoration={brace,amplitude=5,raise=5pt}, thick]
        (yt.north east) -- (ys.south east) node[midway,right=10pt]{$\ell_\text{CE}(\mathbf{y}_t,\, \mathbf{y}_s)$};

\end{tikzpicture}
\caption[Illustration of \wilda]{Illustration of \wilda. The teacher processes a concatenation (denoted by $\concat$) of demonstrations $\mathbf{X}_d$, consisting of $n$ demonstrations $[\mathbf{x}_1, \mathbf{x}_2, \dots, \mathbf{x}_n]$, and the query $\mathbf{x}_q$. The student, using only the query, fine-tunes its adapter weights to produce outputs $\mathbf{y}_s$ aligned with the teacher's pseudo-labels $\mathbf{y}_t$ by minimizing the cross-entropy loss $\ell_\text{CE}$. After fine-tuning, the student can process only queries while still using the knowledge from demonstrations encoded in the adapter.}
\label{fig:wilda}
\end{figure}
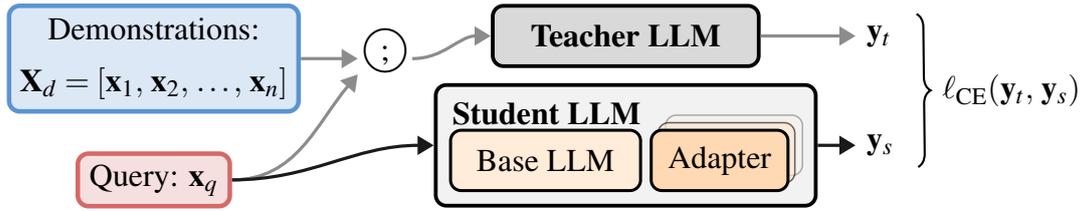

At the core of \wilda is a simple teacher-student framework (\Cref{fig:wilda}): the teacher model, $\mathbf{f}_\text{teacher}$, processes both demonstrations and the query together to generate pseudo-labels without needing additional labeled data. The student model, $\mathbf{f}_\text{student}$, shares the same architecture as the teacher but includes adapter parameters. Unlike the teacher, the student processes only the query, using the adapter to internalize the knowledge from the demonstrations, as illustrated in \Cref{fig:wilda}. Let $\mathbf{x}_q$ denote the query input and $\mathbf{X}_d$ the matrix of demonstration tokens, where each row corresponds to a single demonstration.\footnote{The query $\mathbf{x}_q$ is a vector of token IDs, and $\mathbf{X}_d$ contains token IDs of demonstrations.}
\begin{definition}[\wilda Loss]
The empirical loss, defined using the cross-entropy loss $\ell_{\text{CE}}$, which operates on the teacher's output vector of probabilities for all tokens in the dictionary, is given by:
\begin{equation}
\sum_{\mathbf{x}_q \in \mathcal{D}_\text{unlab}} \ell_{\text{CE}}\left( \mathbf{f}_\text{teacher}\left(\left[  \mathbf{X}^*_d; \mathbf{x}_q\right]\right), \mathbf{f}_\text{student}\left(\mathbf{x}_q\right) \right) ,
\label{eq:loss}
\end{equation}
where $\mathcal{D}_\text{unlab}$ is an unlabeled dataset and $\mathbf{X}^*_d$ is a flattened version of $\mathbf{X}_d$. This approach is grounded in self-training \citep{amini-etal-2022-self}, leveraging the teacher's pseudo-labels to fine-tune the student. 
\end{definition}

\wilda fundamentally differs from existing approaches, which rely on manipulating attention heads or hidden states at query time. Instead, \wilda progressively embeds the knowledge from demonstrations into the adapter parameters, denoted $\mathbf{W}_\text{ICL}$. The base LLM parameters, $\mathbf{W}_\text{ZS}$, capture the zero-shot component, while the total model parameters may be represented as $\mathbf{W}_\text{ZS} \oplus \mathbf{W}_\text{ICL}$, where  $\oplus$ denotes the composition of base and adapter parameters.\footnote{Notably, the number of adapter parameters is significantly smaller compared to the base model parameters.} This setup captures the latent shift introduced by the demonstrations through $\mathbf{W}_\text{ICL}$, extending the disentangling process outlined by (\ref{eq:approx}) across the model's entire architecture. The teacher processes the full input sequence $ \left[\mathbf{X}^*_d; \mathbf{x}_q \right] $, while the student processes only the query, applying $ \mathbf{W}_\text{ICL} $ to integrate demonstration knowledge without explicitly processing the demonstrations. Analogously to (\ref{eq:clean}), the latent shift induced by demonstrations can be recovered by decomposing outputs into zero-shot and ICL components. Let $ \mathbf{h}_\text{LLM}(\mathbf{x}_q \mid \mathbf{W}) $ represent the final latent states of an LLM with parameters $ \mathbf{W} $ when processing the input $ \mathbf{x}_q $. The following decomposition holds:
\begin{equation}
   \mathbf{h}_\text{LLM}(\mathbf{x}_q \mid \mathbf{W}_\text{ZS} \oplus \mathbf{W}_\text{ICL}) = \mathbf{h}_\text{LLM}(\mathbf{x}_q \mid \mathbf{W}_\text{ZS}) + \Delta \mathbf{h}_\text{ICL} ,
\label{eq:decomp}
\end{equation}
where $ \Delta \mathbf{h}_\text{ICL} $ encapsulates the latent shift attributable to the demonstrations. \wilda encodes the latent shift implicitly within the adapter parameters $\mathbf{W}_\text{ICL}$, which is central to our approach.
However, if necessary, the latent shift can also be explicitly calculated owing to the decomposition in (\ref{eq:decomp}).

The stabilizing effect of \wilda extends beyond just handling demonstrations. By iterating over multiple epochs, \wilda leverages the same LLM instance for both the teacher and student roles, transitioning smoothly between them by activating or deactivating the adapter. Demonstrations can be shuffled across epochs to reduce sensitivity to their order, further stabilizing the ICL process. But the true power of \wilda emerges from its parametric nature, which aligns with the optics of W2S generalization \citep{lang-etal-2024-theoretical}. The adapter parameters allow the model to internalize shifts and generalize effectively across both in-domain (ID) and out-of-domain (OOD) data, which we demonstrate empirically in our experiments (\Cref{sec:wilda-gen}).

From the perspective of W2S generalization, the student model is not only expected to match the teacher, but it is also designed to outperform it. \wilda facilitates this by leveraging \textit{pseudo-label correction}, where incorrect labels are refined using high-confidence neighboring examples, and \textit{coverage expansion}, enabling the model to generalize beyond regions initially covered by the teacher and even to near-OOD data.\footnote{Near-OOD data refers to instances that deviate from the training distribution but still share partial structural or semantic similarities with in-domain data. These examples are not fully out-of-distribution but reside in the periphery of the learned representation space, making them a critical test of a model's generalization capacity.} \wilda not only stabilizes ICL but also capitalizes on the parametric regime, where latent shifts can be efficiently encoded, enabling the model to establish implicit local-to-global consistency across the data distribution through extrapolation \citep{wei-etal-2021-theoretical}.

\section{Experimental Setup}
\label{sec:exp}

\subsection{Models}

We utilize a set of decoder-only autoregressive LLMs in our experiments. Specifically, we employ Hugging Face implementations \citep{wolf-etal-2020-transformers} of Llama 3 (8B) \citep{dubey-2024-llama3} and Phi 3 (mini 4k) \citep{abdin2024phi3} as our primary models, with additional comparison results for Llama 2 (7B) \citep{touvron2023llama2}. For all three models -- Llama 3, Llama 2, and Phi 3 -- we utilize the \texttt{bfloat16} half-precision format for parameters. A summary of the models is provided in \Cref{tab:app_models}.

\subsection{Evaluation}

To assess model performance, we evaluate across two key benchmarks: GLUE and MMLU. These benchmarks provide a comprehensive measure of both natural language understanding and domain-specific reasoning.

\paragraph{GLUE Benchmark.} The General Language Understanding Evaluation (GLUE) benchmark \citep{wang-etal-2018-glue} serves as a standard test suite for evaluating natural language understanding capabilities. We select a subset of GLUE datasets, covering both single-sequence and sequence-pair classification tasks. Specifically, we include three binary classification tasks for single sequences (\cola, \sst, \rte) and three binary classification tasks for sequence pairs (\mrpc, \qqp, \qnli). Additionally, we incorporate one multi-class classification task for sequence pairs (\mnli) (cf.~\Cref{sec:jachess-datasets} for detailed dataset descriptions). Following standard evaluation practices, we use the development sets for performance measurement. For generalization assessment, we employ Matthew’s correlation for \cola, $F_1$ scores for \mrpc and \qqp, and accuracy for the remaining datasets.

\paragraph{MMLU Benchmark.} To evaluate the models' ability to perform multiple-choice question answering, we utilize the Massive Multitask Language Understanding (MMLU) benchmark \citep{hendrycks-etal-2021-measuring}. Given the need for robust evaluation, we select two datasets with a sufficient number of instances: ``elementary math'' (\elmath), which tests basic mathematical reasoning skills, and ``miscellaneous'' (\textsc{misc}), which spans diverse subject areas.

\begin{example}[\textsc{math}]
\textbf{Question:} A farmer has 15 apples. He gives away 7 apples to his friends. How many apples does he have left?  \\
\textbf{Options:}  \\
(A) 8  \\
(B) 10  \\
(C) 7  \\
(D) 5  \\

\textbf{Correct Answer:} (C) 7  
\end{example}

\begin{example}[\textsc{misc}]
\textbf{Question:} What is the capital of Canada? \\
\textbf{Options:}  \\
(A) Toronto  \\
(B) Ottawa  \\
(C) Vancouver  \\
(D) Montreal  \\

\textbf{Correct Answer:} (B) Ottawa  
\end{example}

To ensure accurate model evaluation, we compute the first-token probability of task verbalizers. In this context, verbalizers refer to specific tokens that correspond to concrete task labels. For example, in binary classification tasks, verbalizers may map ``Yes'' and ``No'' to class labels, or ``Positive'' and ``Negative'' to sentiment categories. In multiple-choice question answering, verbalizers correspond to the tokens representing the answer choices, such as ``A,'' ``B,'' ``C,'' and ``D.''

The prompt template is carefully designed to guide the model toward generating the answer within the first token while restricting predictions to a predefined subset of verbalizers. This constraint ensures that the model’s responses align with the expected answer format and prevents it from generating free-text completions that deviate from the given answer choices.

\subsection{Baselines and Methods}

We evaluate \wilda by comparing it against three baselines and two ICL disentanglement methods:
\begin{itemize}
    \item \textbf{Zero-Shot ($\mathbf{0}$-shot)}: Predictions made without any demonstrations;
    \item \textbf{Standard ICL ($\mathbf{n}$-shot)}: Utilizes $n$ demonstrations as context during inference;
    \item \textbf{Pattern-Based Fine-Tuning (PBFT)} \cite{schick-schutze-2021-exploiting}: Fine-tunes the model using patterns learned from data, framed as a language modeling task. In our experiments, we fine-tune an adapter module instead of the whole LLM;
    \item \textbf{In-Context Vectors (ICV)} \cite{liu-etal-2023-context}: A forward pass is used on demonstration examples to create in-context vectors from the hidden states of the LLM;
    \item \textbf{Batch-ICL} \cite{zhang-etal-2024-batch}: Utilizes multiple separate one-shot forward computations and aggregates the resulting meta-gradients based on the attention head outputs.
\end{itemize}
In the experiments, we use $n \in \{4, 8, 16, 32\}$ instances for demonstrations and compare methods using a fixed number of demonstrations. Unless stated otherwise, we run each experiment $10$ times with different seeds, which select different demonstrations in each run. In addition to the generalization scores, we report the standard deviation of the runs as an indicator of method stability. We evaluate performance on the GLUE development sets, while for the MMLU datasets, we sample $200$ instances for evaluation.

\subsection{\wilda Variants}

We employ three variants of \wilda, which differ in the variability of demonstrations they use, either in terms of selection or ordering:
\begin{itemize}
    \item \textbf{\wilda-Fixed (\wilda-F)}: Uses a fixed set of demonstrations throughout training;
    \item \textbf{\wilda-Shuffle (\wilda-S)}: Shuffles the order of demonstrations at the start of each epoch;
    \item \textbf{\wilda-Resample (\wilda-R)}: Randomly resamples demonstrations before each epoch.\footnote{Although \wilda-R uses the same number of demonstrations during inference as the other approaches, it requires access to a larger pool of labeled data since it draws new demonstrations in each epoch.}
\end{itemize}
We utilize LoRA (Low-Rank Adaptation) \citep{hu-etal-2022-lora} for the adapter modules (for both PBFT and \wilda), corresponding to $0.1$--$0.3\%$ of the total parameter count, depending on the model (cf.~\Cref{tab:app_models} in the Appendix for adapter sizes per model). For each task, we generate pseudo-labels using the teacher model on unlabeled data. Specifically, we use $100$ unlabeled instances ($\mathcal{D}_\text{unlab}$ in (\ref{eq:loss})) for both the GLUE and MMLU benchmarks. Additionally, for GLUE datasets, we experiment with  $200$ and $500$ instances to assess the impact of the amount of unlabeled data on generalization and stability. We experiment only with $100$ unlabeled instances for MMLU datasets due to their limited size. In all of the experiments, we fine-tune the adapter for $10$ epochs.

\subsection{Hyperparameters}

\paragraph{LoRA adapter configuration.}
\begin{itemize}
    \item $\textbf{r} = 8$  \\
    The rank of the low-rank matrices used to decompose the original weight matrix in LoRA. A smaller $r$ reduces the parameter count while retaining essential information.
    
    \item $\boldsymbol{\alpha} = 32$: \\
    A scaling factor applied to the low-rank updates, balancing the influence of the original weights and the low-rank matrices.

    \item \textbf{Dropout:} 0.1 \\
    The dropout rate applied to the low-rank updates.
    
    \item \textbf{Target modules:} \\ \texttt{q\_proj}, \texttt{k\_proj}, \texttt{v\_proj}, \texttt{o\_proj}, \texttt{gate\_proj}, \texttt{up\_proj}, \texttt{down\_proj}
\end{itemize}

We employ the AdamW optimizer \citep{loshchilov-etal-2017-decoupled} for both PBFT and \wilda-s variants, with a learning rate of $10^{-4}$. For ICV \citep{liu-etal-2023-context} and Batch-ICL \citep{zhang-etal-2024-batch}, we follow the implementations provided in the original papers and adapt them to our codebase, using their default parameters where specified. In the case of Batch-ICL, we utilize attention heads from the last 20 layers ($k=20$) and fine-tune the model for $10$ epochs.

\begin{table}[h]
\caption[\wilda -- summary of models]{Summary of the models used in the experiments, including their Hugging Face IDs, parameter counts, context window sizes, training token volumes, and adapter sizes.}
\centering
\begin{adjustbox}{max width=\textwidth}
\begin{tabular}{lcccccc}
\toprule
\textbf{Model} & \textbf{Hugging Face ID} & \textbf{Parameters} & \textbf{Context window size} & \textbf{Training tokens} & \textbf{Adapter size} \\
\midrule
Llama 3 & Meta-Llama-3-8Bb & 8B & 8k & 15T & 21M \\
Llama 2 & Llama-2-7b & 7B & 4k & 2T & 20M \\
Phi 3 & Phi-3-mini-4k-instruct & 3.8B & 4k & 3.3T & 4.5M \\
\bottomrule
\end{tabular}
\end{adjustbox}
\label{tab:app_models}
\end{table}

\subsection{Prompt Templates}

Prompt templates are a crucial part of the experimental setup, providing the structured inputs needed for the models to effectively leverage in-context learning. These templates are designed to encode task-specific information, balancing clarity and brevity while maximizing the utility of the limited context space. Below, we describe the prompt structures used for the GLUE and MMLU benchmarks, highlighting their tailored design to ensure consistency and task alignment.

\subsubsection{GLUE Prompt Structure}

\begin{center}
\begin{tcolorbox}[colback=gray!5!white, colframe=gray!75!black, width=\textwidth]
\begin{center}
    \textbf{Generic prompt template for GLUE tasks} \\
\end{center}
\hfill \\
\textbf{Demonstrations:}
\begin{verbatim}
{Sentence 1}
{Sentence 2 (if applicable)}
Answer: ({Correct answer})
\end{verbatim}
\textbf{Query:}
\begin{verbatim}
{Sentence 1}
{Sentence 2 (if applicable)}
Question: {Task-specific question}
Answer: (
\end{verbatim}
\end{tcolorbox}
\end{center}

The prompts for GLUE tasks typically consist of two sentences (or one in certain cases) followed by a task-specific question and the corresponding answer. The model is expected to choose from predefined labels like \textit{Yes/No}, \textit{True/False}, or specific class names based on the dataset. The phrasing of the question preceding each answer in the demonstrations is specific to the task. Below is a list of the questions used for each GLUE dataset. To encourage the model to select from predefined labels, we prepend the phrase ``answer with one word'' before each question, and we append clarifying options such as \textit{Yes or No?} to prompt a more targeted response:

\begin{itemize}
\item \textbf{\rte}: \texttt{\{hypothesis\} True or False?}
\item \textbf{\sst}: \texttt{What is the sentiment? Positive or Negative?}
\item \textbf{\qnli}: \texttt{Does the sentence answer the question? Yes or No?}
\item \textbf{\mnli}: \texttt{Is the second sentence an Entailment, Contradiction, or Neutral?}
\item \textbf{\cola}: \texttt{Is this sentence linguistically acceptable? Yes or No?}
\item \textbf{\mrpc}:  \texttt{Do both sentences say the same thing? Yes or No?}
\item \textbf{\qqp}: \texttt{Do both questions ask the same thing? Yes or No?}
\end{itemize}

\subsubsection{MMLU prompt structure}

\begin{center}
\begin{tcolorbox}[colback=gray!5!white, colframe=gray!75!black, width=\textwidth]
\begin{center}
    \textbf{Generic prompt template for MMLU sub-datasets} \\
\end{center}
\hfill \\
\textbf{Demonstrations:}
\begin{verbatim}
Question: {Previous Question 1}
Answer choices:
 (A: {Choice A1}),
 (B: {Choice B1}),
 (C: {Choice C1}),
 (D: {Choice D1})
Answer: (Correct Answer 1)

Question: {Previous Question 2}
Answer choices:
(A: {Choice A2}),
(B: {Choice B2}),
(C: {Choice C2}),
(D: {Choice D2})
Answer: (Correct Answer 2)
...
\end{verbatim}
\textbf{Query:}
\begin{verbatim}
Question: {Current Question}
Answer choices:
(A: {Choice A}),
(B: {Choice B}),
(C: {Choice C}),
(D: {Choice D})
Answer: (
\end{verbatim}
\end{tcolorbox}
\end{center}

\begin{center}
\begin{tcolorbox}[colback=gray!5!white, colframe=gray!75!black, width=\textwidth]
\begin{center}
    \textbf{Example for MMLU \texttt{elementary\_math} (\elmath)} \\
\end{center}
\hfill \\
\textbf{Demonstrations:}
\begin{verbatim}
Question: Ms. Perez drove a total of 40 miles in 5 days.
She drove the same number of miles each day. 
How many miles did Ms. Perez drive each day?
Answer choices: (A: 5), (B: 7), (C: 8), (D: 9)
Answer: (C: 8)

Question: Find the median in the set of data
23, 13, 18, 29, 32, 25.
Answer choices: (A: 18), (B: 24), (C: 25), (D: 29)
Answer: (B: 24)

\end{verbatim}
\textbf{Query:}
\begin{verbatim}
Q: A worker on an assembly line takes 7 hours to produce
22 parts. At that rate how many parts can she produce
in 35 hours?
Answer choices:
(A: 220 parts),
(B: 770 parts),
(C: 4 parts),
(D: 110 parts)
Answer: (
\end{verbatim}
\end{tcolorbox}
\end{center}

\section{Generalization and Stability}
\label{sec:wilda-gen}

\begin{table}[]
\caption[ID generalization scores of \wilda]{ID generalization scores for the $16$-shot setup and $|\mathcal{D}_\text{unlab}| = 100$. The standard deviations of $10$ runs are shown as subscripts. The highest scores and smallest standard deviations are highlighted in \textbf{bold}, while the second-best scores are \underline{underlined}.}
\centering
\begin{adjustbox}{max width=\textwidth}
\begin{tabular}{lllllllllll}
\toprule
& & \multicolumn{7}{c}{\textbf{GLUE}} & \multicolumn{2}{c}{\textbf{MMLU}} \\
\cmidrule(lr){3-9} \cmidrule(lr){10-11}
 \textbf{Model} & \textbf{Method} & \textbf{\rte} & \textbf{\sst} & \textbf{\qnli} & \textbf{\mnli} & \textbf{\cola} & \textbf{\mrpc} & \textbf{\qqp} & \textbf{\elmath} & \textbf{\misc} \\
\midrule
\multirow{8}{*}{\rotatebox{90}{Llama 3 (8B)}} 
& 0-shot & $62.3$ & $79.1$ & $64.3$ & $59.9$ & $44.6$ & $63.6$ & $61.1$ & $31.5$ & $62.5$ \\
& $n$-shot & $75.1_{6.5}$ & $93.5_{2.0}$ & $77.0_{5.5}$ & $68.0_{3.0}$ & $58.5_{4.0}$ & $74.0_{2.5}$ & $70.0_{3.0}$ & $43.5_{3.5}$ & $84.0_{4.0}$ \\
& PBFT & $73.2_{3.8}$ & $93.8_{1.5}$ & $77.8_{6.0}$ & $67.4_{3.5}$ & $56.5_{3.0}$ & $72.0_{2.0}$ & $68.0_{2.5}$ & $44.0_{3.8}$ & $83.5_{4.5}$ \\
& ICV & $72.9_{2.7}$ & $92.2_{1.8}$ & $74.5_{6.3}$ & $67.0_{4.2}$ & $57.3_{3.5}$ & $73.4_{2.3}$ & $69.1_{2.8}$ & $41.5_{4.3}$ & $67.0_{4.2}$ \\
& Batch-ICL & $77.8_{4.7}$ & $94.1_{2.2}$ & $78.0_{6.0}$ & $70.9_{3.5}$ & $59.8_{3.7}$ & $75.2_{2.2}$ & $\underline{72.5}_{2.7}$ & $36.2_{4.0}$ & $81.0_{2.5}$ \\
& \wilda-F & $83.4_{0.3}$ & $95.1_{\textbf{0.6}}$ & $\underline{80.3}_{\textbf{1.4}}$ & $72.1_{2.5}$ & $\underline{63.7}_{\textbf{1.5}}$ & $76.2_{1.8}$ & $71.9_{1.9}$ & $\underline{46.0}_{2.3}$ & $\underline{86.0}_{2.3}$ \\
& \wilda-S & $\underline{86.0}_{\textbf{0.6}}$ & $\textbf{96.1}_{1.2}$ & $\textbf{81.4}_{2.2}$ & $\underline{73.1}_{\textbf{2.0}}$ & $\textbf{64.3}_{2.2}$ & $\textbf{77.7}_{\textbf{1.5}}$ & $\textbf{73.1}_{\textbf{1.8}}$ & $\textbf{49.5}_{\textbf{2.0}}$ & $\textbf{88.0}_{\textbf{2.2}}$ \\
& \wilda-R & $\textbf{86.5}_{3.0}$ & $\underline{95.5}_{0.8}$ & $79.0_{4.3}$ & $\textbf{73.5}_{3.0}$ & $62.5_{2.8}$ & $\underline{76.5}_{1.9}$ & $72.0_{2.2}$ & $44.0_{2.7}$ & $85.5_{3.3}$ \\
\midrule
\multirow{8}{*}{\rotatebox{90}{Phi 3 (mini 4k)}} 
& 0-shot & $60.6$ & $78.3$ & $61.1$ & $58.1$ & $43.7$ & $63.1$ & $57.8$ & $29.5$ & $52.0$ \\
& $n$-shot & $72.1_{5.2}$ & $90.6_{2.1}$ & $75.6_{3.2}$ & $65.3_{3.1}$ & $55.5_{4.1}$ & $71.1_{2.6}$ & $66.2_{3.7}$ & $37.5_{3.6}$ & $75.5_{4.1}$ \\
& PBFT & $70.6_{4.3}$ & $90.9_{1.9}$ & $73.6_{3.4}$ & $63.6_{3.6}$ & $53.6_{3.1}$ & $69.6_{2.3}$ & $64.6_{2.6}$ & $36.5_{4.1}$ & $73.5_{4.6}$ \\
& ICV & $71.5_{3.1}$ & $89.1_{2.1}$ & $74.3_{3.2}$ & $64.1_{4.1}$ & $54.1_{3.6}$ & $70.8_{2.4}$ & $65.4_{2.9}$ & $36.0_{4.6}$ & $74.0_{4.3}$ \\
& Batch-ICL & $75.3_{4.2}$ & $91.2_{2.6}$ & $76.6_{3.1}$ & $67.1_{3.6}$ & $56.1_{4.1}$ & $72.6_{2.6}$ & $67.3_{2.8}$ & $38.0_{3.9}$ & $76.0_{4.1}$ \\
& \wilda-F & $\underline{80.4}_{\textbf{1.2}}$ & $92.1_{\textbf{1.6}}$ & $\underline{78.2}_{\textbf{1.3}}$ & $\underline{69.7}_{2.4}$ & $\underline{59.5}_{2.5}$ & $73.5_{2.1}$ & $\underline{68.6}_{2.2}$ & $\underline{40.5}_{3.2}$ & $\underline{77.5}_{3.6}$ \\
& \wilda-S & $\textbf{82.4}_{1.1}$ & $\textbf{93.2}_{\textbf{1.6}}$ & $79.2_{1.4}$ & $\textbf{70.4}_{\textbf{1.1}}$ & $\textbf{60.7}_{\textbf{2.3}}$ & $\textbf{74.1}_{\textbf{1.4}}$ & $\textbf{69.6}_{\textbf{1.9}}$ & $\textbf{41.5}_{\textbf{2.3}}$ & $\textbf{78.0}_{\textbf{3.3}}$ \\
& \wilda-R & $79.0_{1.9}$ & $\underline{92.6}_{2.0}$ & $\textbf{79.6}_{2.9}$ & $68.6_{3.9}$ & $58.6_{2.9}$ & $\underline{73.6}_{2.0}$ & $68.1_{2.3}$ & $39.5_{3.6}$ & $77.0_{3.7}$ \\
\bottomrule
\end{tabular}
\end{adjustbox}
\label{tab:id_gen}
\end{table}

\begin{table}[]
\caption[ID generalization scores for Llama 2 with \wilda]{ID generalization scores for the $16$-shot scenario and $|\mathcal{D}_\text{unlab}| = 100$ for Llama 2 (7B). The standard deviations of $10$ runs are shown as subscripts.}
\centering
\begin{adjustbox}{max width=\textwidth}
\begin{tabular}{lllllllllll}
\toprule
& & \multicolumn{7}{c}{\textbf{GLUE}} & \multicolumn{2}{c}{\textbf{MMLU}} \\
\cmidrule(lr){3-9} \cmidrule(lr){10-11}
\textbf{Model} & \textbf{Method} & \textbf{\rte} & \textbf{\sst} & \textbf{\qnli} & \textbf{\mnli} & \textbf{\cola} & \textbf{\mrpc} & \textbf{\qqp} & \textbf{\elmath} & \textbf{\misc} \\
\midrule
\multirow{8}{*}{\rotatebox{90}{Llama 2 (7B)}} 
& 0-shot & 57.8 & 75.4 & 59.3 & 55.7 & 40.7 & 59.4 & 58.7 & 29.0 & 59.0 \\
& $n$-shot & 69.2$_{4.3}$ & 89.8$_{2.1}$ & 74.2$_{5.9}$ & 63.3$_{2.8}$ & 54.3$_{3.5}$ & 66.9$_{2.4}$ & 64.7$_{1.5}$ & 37.5$_{4.8}$ & 80.0$_{5.3}$ \\
& PBFT & 69.0$_{2.7}$ & 89.7$_{0.4}$ & 73.3$_{5.0}$ & 64.4$_{4.7}$ & 51.2$_{2.9}$ & 67.9$_{2.0}$ & 64.6$_{1.6}$ & 40.0$_{3.2}$ & 79.5$_{2.1}$ \\
& ICV & 68.0$_{4.6}$ & 87.8$_{2.6}$ & 71.2$_{6.7}$ & 60.9$_{4.0}$ & 53.1$_{2.4}$ & 68.8$_{1.7}$ & 65.0$_{1.9}$ & 39.5$_{2.7}$ & 62.5$_{0.6}$ \\
& Batch-ICL & 75.2$_{0.8}$ & 91.2$_{1.9}$ & 74.0$_{0.8}$ & 66.5$_{3.3}$ & 55.9$_{2.1}$ & 70.3$_{0.8}$ & 69.1$_{1.8}$ & 34.5$_{2.3}$ & 77.0$_{4.1}$ \\
& \wilda-F & 77.2$_{0.7}$ & 90.2$_{0.7}$ & 76.8$_{4.2}$ & 66.5$_{2.4}$ & 60.1$_{1.2}$ & 71.6$_{0.2}$ & 68.8$_{0.8}$ & 43.0$_{1.6}$ & 82.5$_{2.5}$ \\
& \wilda-S & 81.9$_{2.5}$ & 92.1$_{0.3}$ & 77.3$_{0.9}$ & 70.4$_{1.8}$ & 62.8$_{3.4}$ & 72.3$_{2.6}$ & 68.2$_{0.5}$ & 46.5$_{1.5}$ & 82.5$_{1.7}$ \\
& \wilda-R & 81.1$_{1.9}$ & 93.6$_{2.0}$ & 74.7$_{3.6}$ & 69.6$_{2.9}$ & 57.9$_{2.9}$ & 73.1$_{2.0}$ & 66.8$_{2.3}$ & 41.5$_{2.6}$ & 82.0$_{3.7}$ \\
\bottomrule
\end{tabular}
\end{adjustbox}
\label{tab:app_id_gen}
\end{table}

We first evaluate the generalization and stability of \wilda on ID data. \Cref{tab:id_gen,tab:app_id_gen} report the 16-shot ID generalization scores along with standard deviations.
Across all datasets and models, \wilda-S consistently achieves the best generalization scores, outperforming standard ICL, PBFT, and the disentanglement methods ICV and Batch-ICL.
Compared to standard ICL, \wilda-S \textit{shows absolute improvements ranging from $2.6\%$ to $11.9\%$ for Llama 3 and $2.5\%$ to $10.3\%$ for Phi 3}, where the differences in scores are statistically significant across all datasets.\footnote{We assess the statistical significance using a two-tailed Wilcoxon signed-rank test ($p < 0.05$), applying the Holm-Bonferroni method for family-wise error rate correction due to multiple comparisons.} Similar patterns hold for $n \in \{4,8,32\}$, where \wilda-S also surpasses standard ICL (cf.~\Cref{tab:app_nshot} for other $n$-shot setups). Additionally, when a larger set $\mathcal{D}_\text{unlab}$ is used, there is a marginal improvement in scores, while stability improves even further (cf.~\Cref{tab:app_unlab}).
Notably, the improvements in generalization with \wilda-S, compared to standard ICL -- the teacher model in \wilda -- provide strong evidence that the student model is exhibiting W2S generalization; we provide a more detailed analysis of this phenomenon in \Cref{sec:w2s}.
While the \wilda-F and \wilda-R variants also show similar generalization scores as \wilda-S, they generally exhibit higher variance compared to \wilda-S, making \wilda-S the preferred choice due to its higher stability with respect to demonstration selection -- it improves upon standard $n$-shot ICL across all datasets and models. This is supported by the statistically significant differences in standard deviations on all datasets for Llama 3 and on all but \qnli for Phi 3.\footnote{\label{test_footnote} We evaluate significance using a two-tailed Levene's test ($p<0.05$), applying the Holm-Bonferroni method for family-wise error rate correction.}

Having looked at stability with respect to demonstration selection, we now turn to a more focused evaluation of stability with respect to demonstration ordering. \Cref{tab:stab} reports the standard deviations across 50 runs, where the same set of demonstrations is used, but their order is shuffled for each run. Designed to adapt to shuffled demonstrations, \wilda-S \textit{shows the highest stability to demonstration ordering}, as evidenced by the smallest standard deviation. The stability improvements with \wilda-S over standard ICL are statistically significant across all datasets.\footref{test_footnote}

We next assess the capacity of \wilda to perform OOD generalization by fine-tuning an adapter on one dataset and then applying the student model to a different dataset within the same task category, simulating a near-OOD scenario with pairs of closely related datasets.
\Cref{tab:ood_gen,tab:app_ood_gen} show the OOD generalization scores for such pairs of datasets in the GLUE benchmark. The results show that \textit{\wilda-S not only outperforms other methods in OOD generalization but also maintains higher stability when adapting to new domains}.

\begin{table}[]
\caption[Standard deviations of \wilda generalization scores]{Standard deviations of generalization scores across $50$ runs with varied orderings of $16$ demonstrations. The smallest deviations are in \textbf{bold}, and the second-smallest are \underline{underlined}.}
\centering
\small
\begin{adjustbox}{max width=\textwidth}
\begin{tabular}{lllllllllll}
\toprule
& & \multicolumn{7}{c}{\textbf{GLUE}} & \multicolumn{2}{c}{\textbf{MMLU}} \\
\cmidrule(lr){3-9} \cmidrule(lr){10-11}
  \textbf{Model} & \textbf{Method} & \textbf{\rte} & \textbf{\sst} & \textbf{\qnli} & \textbf{\mnli} & \textbf{\cola} & \textbf{\mrpc} & \textbf{\qqp} & \textbf{\elmath} & \textbf{\misc} \\
\midrule
\multirow{7}{*}{\rotatebox{90}{LLama 3 (8B)}} 
& $n$-shot & $4.81$ & $1.62$ & $4.19$ & $2.22$ & $3.04$ & $1.81$ & $2.03$ & $2.52$ & $2.87$ \\
& PBFT & $2.71$ & $1.14$ & $4.53$ & $2.69$ & $2.27$ & $1.57$ & $1.82$ & $2.70$ & $3.22$ \\
& ICV & $2.09$ & $1.23$ & $4.08$ & $2.81$ & $1.95$ & $1.61$ & $2.03$ & $1.96$ & $3.18$ \\
& Batch ICL & $3.04$ & $1.47$ & $2.89$ & $2.24$ & $2.53$ & $\underline{1.42}$ & $1.74$ & $2.51$ & $2.59$ \\
& \wilda-F & $\underline{1.32}$ & $\underline{0.72}$ & $\underline{1.53}$ & $\underline{1.83}$ & $\underline{1.76}$ & $1.54$ & $\underline{1.38}$ & $\underline{1.89}$ & $\underline{2.07}$ \\
& \wilda-S & $\textbf{0.22}$ & $\textbf{0.53}$ & $\textbf{1.04}$ & $\textbf{1.21}$ & $\textbf{1.28}$ & $\textbf{0.73}$ & $\textbf{1.14}$ & $\textbf{1.22}$ & $\textbf{0.97}$ \\
& \wilda-R & $2.04$ & $1.34$ & $2.47$ & $2.05$ & $1.85$ & $1.48$ & $1.64$ & $2.03$ & $2.51$ \\
\bottomrule
\end{tabular}
\end{adjustbox}
\label{tab:stab}
\end{table}
\begin{table}[]
\caption[OOD generalization scores of \wilda]{OOD generalization scores with $16$ shots averaged over 10 runs, with standard deviations shown as subscripts. For each dataset pair, demonstrations are taken from the \textbf{left} dataset, and the model is tested on the \textbf{right} dataset. Columns represent results on the \textbf{right} datasets. The highest scores and lowest standard deviations are in \textbf{bold}, and the second-highest scores are \underline{underlined}. Values in parentheses indicate differences from ID performance for the corresponding target dataset.}
\centering
\small
\begin{adjustbox}{max width=\textwidth}
\begin{tabular}{llllll}
\toprule
\textbf{Model} & \textbf{Method} & \textbf{\qnli $\rightarrow$ \rte} & \textbf{\rte $\rightarrow$ \qnli} & \textbf{\qqp $\rightarrow$ \mrpc} & \textbf{\mrpc $\rightarrow$ \qqp} \\
\midrule
\multirow{7}{*}{\rotatebox{90}{Llama 3 (8B)}}
& $n$-shot & $66.3_{2.4} \ (8.8)$ & $69.6_{1.3} \ (7.4)$ & $66.5_{1.9} \ (7.5)$ & $62.2_{2.3} \ (7.8)$ \\
& PBFT & $66.1_{1.5} \ (7.1)$ & $69.1_{1.6}$ \ (8.7) & $67.2_{1.8} \ (4.8)$ & $62.4_{1.2} \ (5.6)$ \\
& ICV & $65.7_{1.2} \ (7.2)$ & $68.7_{2.3} \ (5.8)$ & $67.5_{1.6} \ (5.9)$ & $63.0_{2.1} \ (6.1)$ \\
& Batch-ICL & $65.3_{1.4} \ (12.5)$ & $66.3_{2.5} \ (11.7)$ & $64.9_{2.3} \ (10.3)$ & $62.1_{2.1} \ (10.4)$ \\
& \wilda-F & $\underline{67.5}_{1.1} \ (15.9)$ & $\underline{70.5}_{1.4} \ (9.8)$ & ${68.5}_{\textbf{1.0}} \ (7.7)$ & ${64.4}_{1.5} \ (7.5)$ \\
& \wilda-S & $\textbf{69.0}_{\textbf{0.5}} \ (17.0)$ & $\textbf{71.3}_{\textbf{0.7}} \ (10.1)$ & $\textbf{69.0}_{2.2} \ (8.7)$ & $\underline{66.4}_{\textbf{1.1}} \ (6.7)$ \\
& \wilda-R & $67.1_{1.7} \ (19.4)$ & $70.0_{1.4} \ (9.0)$ & $68.0_{2.7} \ (8.5)$ & $\textbf{68.3}_{2.0} \ (3.7)$ \\
\bottomrule
\end{tabular}
\end{adjustbox}
\label{tab:ood_gen}
\end{table}

\begin{table}[]
\caption[OOD generalization scores for Phi 3 and Llama 2 with \wilda]{OOD generalization scores for Phi 3 and Llama 2 in a $16$-shot scenario with $\mathcal{D}_\text{unlab} = 100$ over $10$ runs with standard deviations shown as subscripts. In each dataset pair, demonstrations are taken from the left dataset, and the model is tested on the right dataset. The columns correspond to the results on the right datasets.}
\centering
\small
\begin{adjustbox}{max width=\textwidth}
\begin{tabular}{llcccc}
\toprule
\textbf{Model} & \textbf{Method} & \textbf{\qnli $\rightarrow$ \rte} & \textbf{\rte $\rightarrow$ \qnli} & \textbf{\qqp $\rightarrow$ \mrpc} & \textbf{\mrpc $\rightarrow$ \qqp} \\
\midrule
\multirow{3}{*}{{Phi 3 (mini 4k)}}
& $n$-shot & $64.3_{2.5}$ & $67.2_{1.5}$ & $63.7_{2.3}$ & $59.4_{2.2}$ \\
& PBFT & $64.1_{1.8}$ & $66.9_{1.6}$ & $64.7_{2.0}$ & $60.1_{1.4}$ \\
& \wilda-S & $67.4_{0.6}$ & $69.2_{0.9}$ & $66.3_{2.4}$ & $64.4_{1.3}$ \\
\midrule
\multirow{3}{*}{{Llama 2 (7B)}}
& $n$-shot & $62.9_{2.3}$ & $66.3_{1.2}$ & $64.5_{1.9}$ & $61.1_{2.2}$ \\
& PBFT & $62.8_{1.3}$ & $68.1_{1.4}$ & $65.9_{1.8}$ & $61.3_{1.2}$ \\
& \wilda-S & $64.8_{0.4}$ & $70.3_{0.6}$ & $67.8_{2.1}$ & $65.0_{1.1}$ \\
\bottomrule
\end{tabular}
\end{adjustbox}
\label{tab:app_ood_gen}
\end{table}

\begin{table}[]
\caption[ID generalization scores for Llama 3 with \wilda (Part I)]{ID generalization scores for $n$-shot scenarios ($n = 4, 8, 32$, with $\mathcal{D}_\text{unlab} = 100$) for Llama 3 (8B). The standard deviations of $10$ runs are shown as subscripts.}
\centering
\begin{adjustbox}{max width=\textwidth}
\begin{tabular}{llllllllllll}
\toprule
& & & \multicolumn{7}{c}{\textbf{GLUE}} & \multicolumn{2}{c}{\textbf{MMLU}} \\
\cmidrule(lr){4-10} \cmidrule(lr){11-12}
\textbf{Model} & \textbf{$n$} & \textbf{Method} & \textbf{\rte} & \textbf{\sst} & \textbf{\qnli} & \textbf{\mnli} & \textbf{\cola} & \textbf{\mrpc} & \textbf{\qqp} & \textbf{\elmath} & \textbf{\misc} \\
\midrule
\multirow{6}{*}{\rotatebox{90}{Llama 3 (8B)}} 
& \multirow{2}{*}{$4$} & $n$-shot & 71.3$_{5.4}$ & 84.5$_{4.4}$ & 70.1$_{2.9}$ & 62.4$_{2.7}$ & 54.6$_{3.5}$ & 69.2$_{4.1}$ & 62.0$_{2.3}$ & 37.0$_{3.9}$ & 76.5$_{2.5}$ \\
& & \wilda-S & 80.3$_{1.5}$ & 90.9$_{0.9}$ & 76.3$_{1.4}$ & 70.1$_{1.8}$ & 61.4$_{2.0}$ & 72.9$_{1.5}$ & 70.3$_{1.2}$ & 43.0$_{1.3}$ & 77.5$_{1.8}$ \\
\cmidrule(lr){2-12}

& \multirow{2}{*}{$8$} & $n$-shot & 72.7$_{2.1}$ & 89.4$_{2.6}$ & 73.5$_{2.5}$ & 64.7$_{3.1}$ & 55.8$_{2.8}$ & 71.2$_{2.4}$ & 64.3$_{2.9}$ & 37.0$_{1.3}$ & 77.5$_{2.1}$ \\
& & \wilda-S & 82.1$_{1.1}$ & 93.2$_{1.0}$ & 78.3$_{1.3}$ & 72.2$_{1.6}$ & 63.7$_{1.8}$ & 73.9$_{1.3}$ & 72.1$_{0.4}$ & 47.5$_{0.5}$ & 84.0$_{1.4}$ \\
\cmidrule(lr){2-12}

& \multirow{2}{*}{$32$} & $n$-shot & 75.3$_{3.2}$ & 93.2$_{1.9}$ & 77.7$_{2.9}$ & 69.1$_{1.9}$ & 58.3$_{1.5}$ & 76.4$_{2.2}$ & 74.2$_{1.9}$ & 43.0$_{1.5}$ & 84.5$_{2.1}$ \\
& & \wilda-S & 87.9$_{0.6}$ & 97.9$_{0.4}$ & 83.1$_{0.9}$ & 74.0$_{1.1}$ & 64.6$_{1.2}$ & 79.4$_{0.6}$ & 74.8$_{1.5}$ & 56.5$_{0.2}$ & 89.0$_{0.4}$ \\

\bottomrule
\end{tabular}
\end{adjustbox}
\label{tab:app_nshot}
\end{table}

\begin{table}[]
\caption[ID generalization scores for Llama 3 with \wilda (Part II)]{ID generalization scores of \wilda-S for $n=16$ shots and $|\mathcal{D}_\text{unlab}| = 200, 500$ for Llama 3 (8B). Results are shown for GLUE datasets with $n$-shot and \wilda-S methods. The standard deviations of $10$ runs are shown as subscripts.}
\centering
\begin{adjustbox}{max width=\textwidth}
\begin{tabular}{lcllllllll}
\toprule
& & \multicolumn{7}{c}{\textbf{GLUE}} \\
\cmidrule(lr){3-9}
\textbf{Model} & \textbf{$|\mathcal{D}_\text{unlab}|$} &  \textbf{\rte} & \textbf{\sst} & \textbf{\qnli} & \textbf{\mnli} & \textbf{\cola} & \textbf{\mrpc} & \textbf{\qqp} \\
\midrule

\multirow{2}{*}{{Llama 3 (8B)}} 
& 200 & 86.2$_{0.4}$ & 97.2$_{0.4}$ & 81.6$_{1.0}$ & 73.9$_{1.3}$ & 64.7$_{1.1}$ & 78.9$_{0.7}$ & 74.0$_{0.5}$ \\
& 500 & 86.9$_{0.3}$ & 97.1$_{0.5}$ & 81.9$_{0.7}$ & 74.8$_{1.0}$ & 64.6$_{0.8}$ & 81.4$_{0.8}$ & 75.2$_{0.3}$ \\

\bottomrule
\end{tabular}
\end{adjustbox}
\label{tab:app_unlab}
\end{table}

\section{Adapter Arithmetic}

To overcome the limitations of context window sizes and efficiently handle extensive demonstration sets in ICL, we employ \textit{adapter arithmetic} within \wilda. As introduced in \Cref{ch:peft}, adapters serve as lightweight, parameter-efficient modules that enable targeted fine-tuning without modifying the full model.  
Adapter arithmetic refers to the process of combining multiple adapters through parameter-wise operations, such as summation or interpolation, to integrate knowledge from different subsets of data. 

In \wilda, we fine-tune separate adapters for each demonstration subset, with each adapter encoding the latent shift corresponding to its subset. These adapters are then merged by summing their parameters \citep{chitale-etal-2023-task}, resulting in a single adapter that consolidates knowledge from all subsets.  
Partitioning demonstrations into smaller subsets allows for more effective use of long contexts, extending the model’s capacity without exceeding context window limits or requiring modifications to the base LLM architecture. Additionally, distributing the prompt across multiple adapters optimizes GPU utilization, enabling the full prompt to fit within memory during inference while reducing computational overhead.

\Cref{tab:adapters} shows the ID generalization scores of ICV, Batch-ICL, and \wilda in fusing knowledge from multiple demonstration subsets, specifically using $2$, $4$, and $8$ subsets of $16$ demonstrations each.
\wilda-S consistently outperforms baseline methods, demonstrating its ability to fuse knowledge from different subsets. This success parallels broader trends in knowledge fusion within LLMs \cite{wan-etal-2024-knowledge}. Moreover, this form of adapter arithmetic aligns with recent advances in task arithmetic, where merging task-specific parameters promotes generalization across multiple tasks \citep{ilharco-etal-2023-editing, ortiz-jimenez-etal-2023-task}. In our case, \textit{this approach effectively improves generalization and stability when fusing demonstration subsets within the same task.}

\begin{table}[]
\caption[ID generalization scores of knowledge fusion with \wilda for Llama 3]{ID generalization scores of knowledge fusion for Llama 3. The scores are averaged over $10$ runs with standard deviations shown as subscripts. The table compares the effectiveness of knowledge fusion from $2$, $4$, and $8$ subsets of $16$ demonstrations. The highest scores are in \textbf{bold}.}
\centering
\begin{adjustbox}{max width=\textwidth}
\begin{tabular}{cl llllllllll}
\toprule
& & \multicolumn{7}{c}{\textbf{GLUE}} & \multicolumn{2}{c}{\textbf{MMLU}} \\
\cmidrule(lr){3-9} \cmidrule(lr){10-11}
\textbf{Demonstrations} & \textbf{Method} & \textbf{\rte} & \textbf{\sst} & \textbf{\qnli} & \textbf{\mnli} & \textbf{\cola} & \textbf{\mrpc} & \textbf{\qqp} & \textbf{\elmath} & \textbf{\misc} \\
\midrule
\multirow{3}{*}{$\mathbf{2 \times 16}$}
& ICV & $75.2_{4.3}$ & $93.6_{1.9}$ & $77.6_{5.9}$ & $69.2_{3.7}$ & $58.3_{3.5}$ & $74.2_{2.4}$ & $70.6_{2.7}$ & $45.5_{3.7}$ & $72.5_{2.9}$ \\
& Batch-ICL & $80.2_{3.6}$ & $95.3_{1.8}$ & $80.2_{5.8}$ & $72.3_{3.0}$ & $61.2_{3.1}$ & $76.3_{2.0}$ & $72.6_{2.4}$ & $43.5_{2.9}$ & $83.0_{3.6}$ \\
& \wilda-S & $\textbf{87.1}_{1.6}$ & $\textbf{96.4}_{1.3}$ & $\textbf{81.5}_{5.0}$ & $\textbf{75.5}_{2.5}$ & $\textbf{68.4}_{1.8}$ & $\textbf{78.5}_{1.4}$ & $\textbf{74.1}_{1.6}$ & $\textbf{51.5}_{1.6}$ & $\textbf{89.5}_{2.0}$ \\
\midrule
\multirow{3}{*}{$\mathbf{4 \times 16}$}
& ICV & $78.3_{3.6}$ & $94.6_{1.8}$ & $79.3_{5.5}$ & $71.2_{3.1}$ & $60.3_{3.3}$ & $75.6_{2.2}$ & $72.3_{2.4}$ & $47.5_{3.5}$ & $76.5_{3.8}$ \\
& Batch-ICL & $84.4_{3.3}$ & $96.4_{1.5}$ & $82.4_{5.2}$ & $74.3_{2.5}$ & $64.2_{2.8}$ & $78.3_{1.6}$ & $74.3_{2.1}$ & $45.5_{2.6}$ & $84.5_{3.3}$ \\
& \wilda-S & $\textbf{88.4}_{2.3}$ & $\textbf{97.5}_{0.7}$ & $\textbf{83.6}_{4.4}$ & $\textbf{77.3}_{2.2}$ & $\textbf{71.4}_{1.5}$ & $\textbf{79.6}_{0.7}$ & $\textbf{75.2}_{1.3}$ & $\textbf{53.5}_{1.4}$ & $\textbf{91.0}_{1.7}$ \\
\midrule
\multirow{3}{*}{$\mathbf{8 \times 16}$}
& ICV & $81.3_{2.8}$ & $95.6_{1.5}$ & $81.8_{5.0}$ & $73.3_{2.7}$ & $61.3_{2.4}$ & $77.3_{1.7}$ & $73.8_{2.0}$ & $47.5_{2.9}$ & $78.0_{3.5}$ \\
& Batch-ICL & $85.6_{2.5}$ & $96.7_{1.1}$ & $83.8_{4.5}$ & $75.8_{2.1}$ & $65.3_{2.1}$ & $79.8_{1.3}$ & $75.8_{1.8}$ & $45.5_{2.0}$ & $84.0_{2.5}$ \\
& \wilda-S & $\textbf{92.8}_{0.8}$ & $\textbf{98.1}_{0.2}$ & $\textbf{87.9}_{2.5}$ & $\textbf{81.3}_{0.9}$ & $\textbf{74.1}_{0.6}$ & $\textbf{82.8}_{0.4}$ & $\textbf{78.9}_{0.5}$ & $\textbf{57.0}_{0.5}$ & $\textbf{93.0}_{0.7}$ \\
\bottomrule
\end{tabular}
\end{adjustbox}
\label{tab:adapters}
\end{table}

\subsubsection{Few-shot \wilda}
\wilda is primarily designed for $0$-shot operation during the fine-tuning phase, leveraging unlabeled data to encode task-specific information within the adapter. To examine its performance in few-shot setups, we evaluated \wilda-S using Llama 3 (8B) in a $16$-shot configuration, where $16$ additional demonstrations were encoded into the adapter, resulting in a total of $32$ labeled instances. This setup was compared against standard $32$-shot ICL, as well as two \wilda-S variants utilizing $32$ labeled instances in a $0$-shot configuration. Additionally, we included a baseline for a $0$-shot setup with only $16$ encoded demonstrations.

To standardize comparisons, we denote each \wilda variant using the format $n/d$, where $n$ represents the number of shots (n-shot) and $d$ indicates the number of demonstrations encoded in the adapter. The results, averaged over 10 runs, are shown in \Cref{tab:wilda_comparison}.

\begin{table}[]
\caption[\ \ Performance comparison of \wilda-S configurations]{Performance comparison of \wilda-S configurations and standard 32-shot ICL averaged over $10$ runs.}
\centering
\begin{adjustbox}{max width=\textwidth}
\begin{tabular}{lccccccccc}
\toprule
\textbf{Method} & \textbf{\rte} & \textbf{\sst} & \textbf{\qnli} & \textbf{\mnli} & \textbf{\cola} & \textbf{\mrpc} & \textbf{\qqp} & \textbf{\elmath} & \textbf{\misc} \\
\midrule
32-shot ICL & 75.3 & 93.2 & 77.7 & 69.1 & 58.3 & 76.4 & 74.2 & 43.0 & 84.5 \\
\wilda-S (0/32)         & 87.9 & 97.9 & 83.1 & 74.0 & 64.6 & 79.4 & 74.8 & 56.5 & 89.0 \\
\wilda-S (0/16)         & 86.0 & 96.1 & 81.4 & 73.1 & 64.3 & 77.7 & 73.1 & 49.5 & 88.0 \\
\textbf{\wilda-S (16/16)}        & 87.3 & 96.4 & 82.2 & 74.6 & 65.4 & 78.2 & 74.5 & 51.0 & 89.0 \\
\bottomrule
\end{tabular}
\end{adjustbox}
\label{tab:wilda_comparison}
\end{table}

The results demonstrate that \wilda-S in the $16$-shot configuration with $16$ encoded demonstrations ($16/16$) outperforms both standard $32$-shot ICL and \wilda-S ($0/16$) across all datasets, showcasing its ability to utilize additional context during inference. However, it slightly underperforms compared to the $0$-shot \wilda-S variant with 32 encoded demonstrations ($0/32$), likely due to the training process that is exclusive to the 0-shot setup. Nevertheless, the strong performance in n-shot setups ($n > 0$) highlights the flexibility and efficacy of \wilda-S in leveraging additional context provided within the prompt.

\section{Analysis of Weak-to-Strong Generalization}
\label{sec:w2s}

\begin{figure*}[]
    \centering
    \begin{subfigure}{0.49\textwidth}
        \centering
        \includegraphics[width=\linewidth]{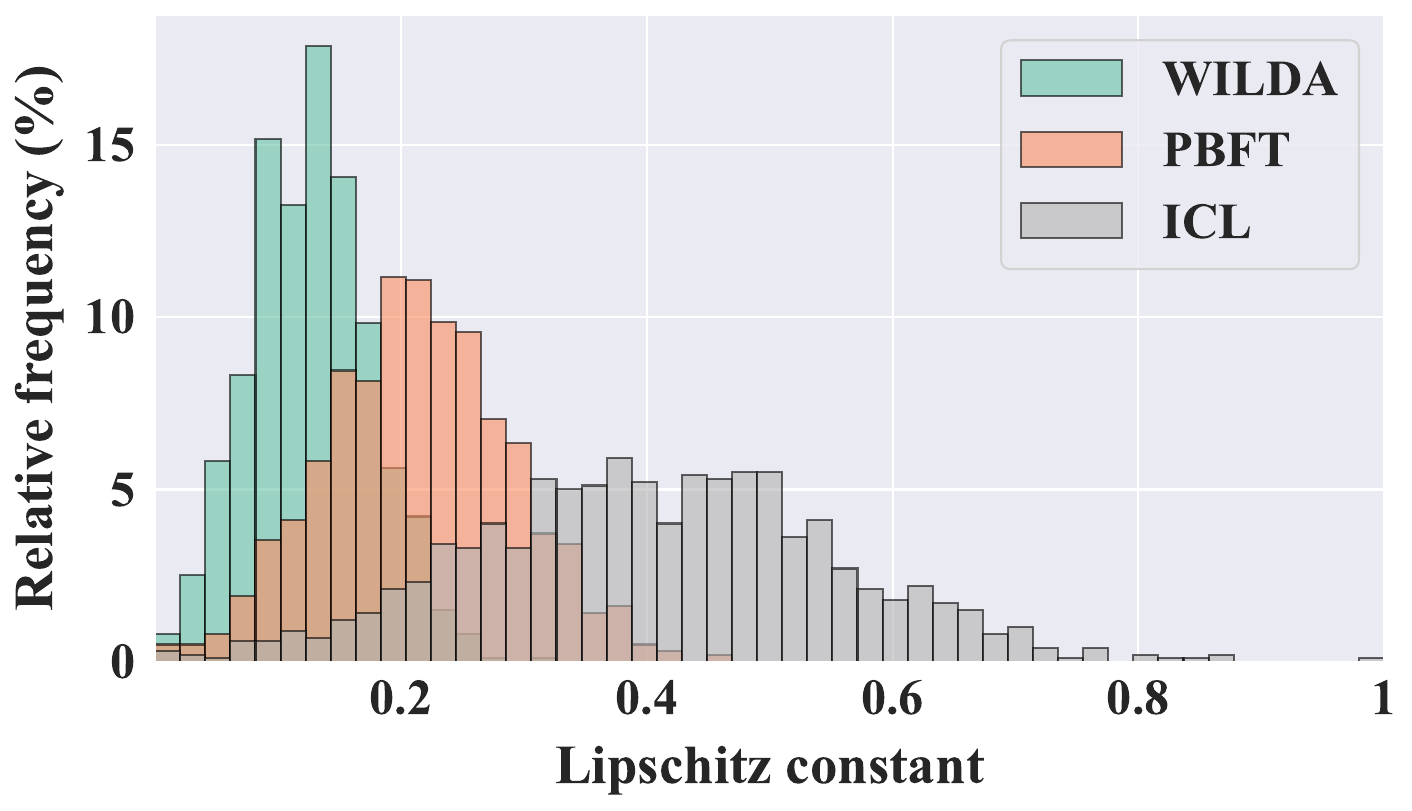}
        \caption{Histogram of approximated Lipschitz constants}
        \label{fig:w2s-a}
    \end{subfigure}
    \hfill
    \begin{subfigure}{0.49\textwidth}
        \centering
        \includegraphics[width=\linewidth]{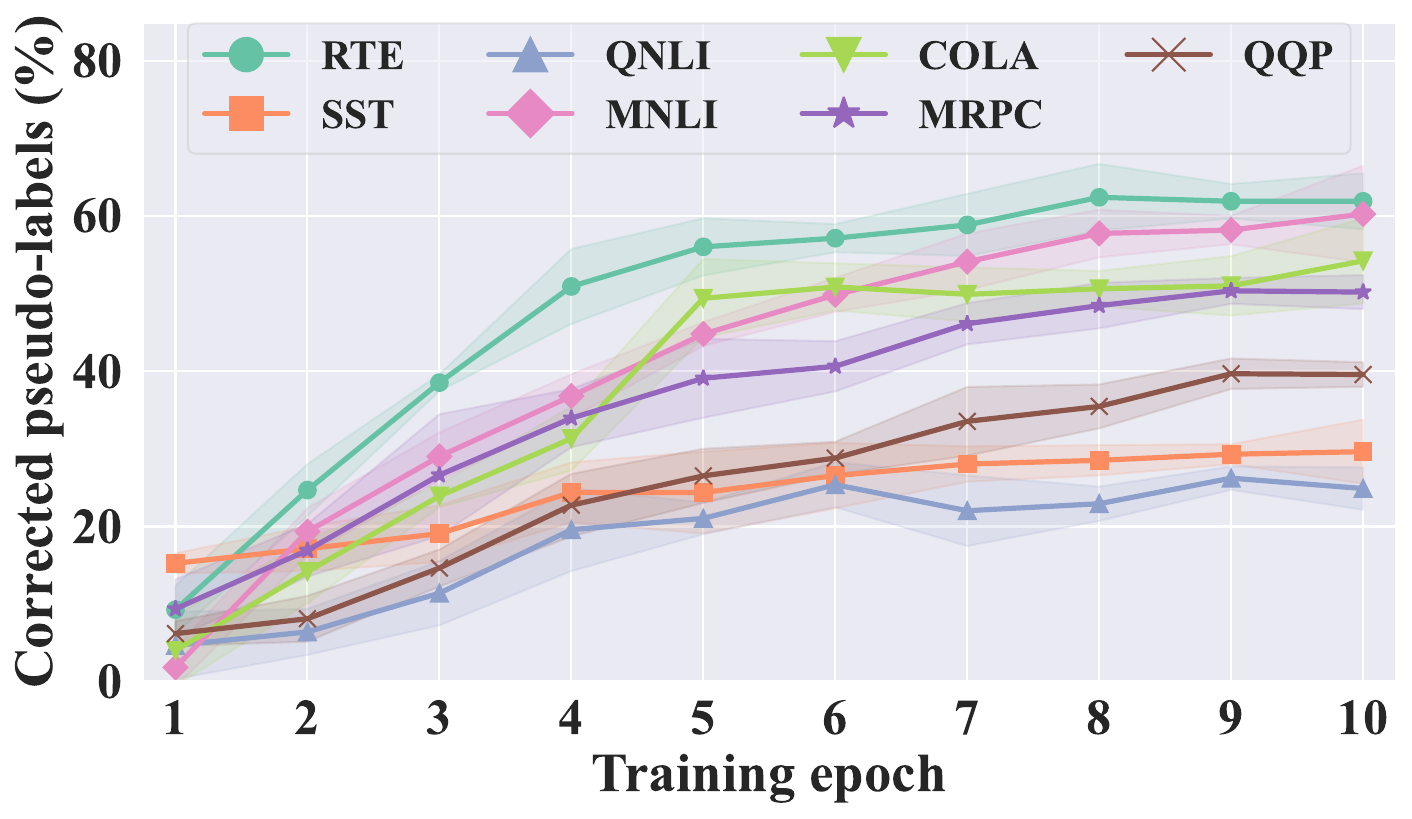}
        \caption{Rate of pseudo-label correction over epochs}
        \label{fig:w2s-b}
    \end{subfigure}
    \hfill
    \begin{subfigure}{0.49\textwidth}
        \centering
        \includegraphics[width=\linewidth]{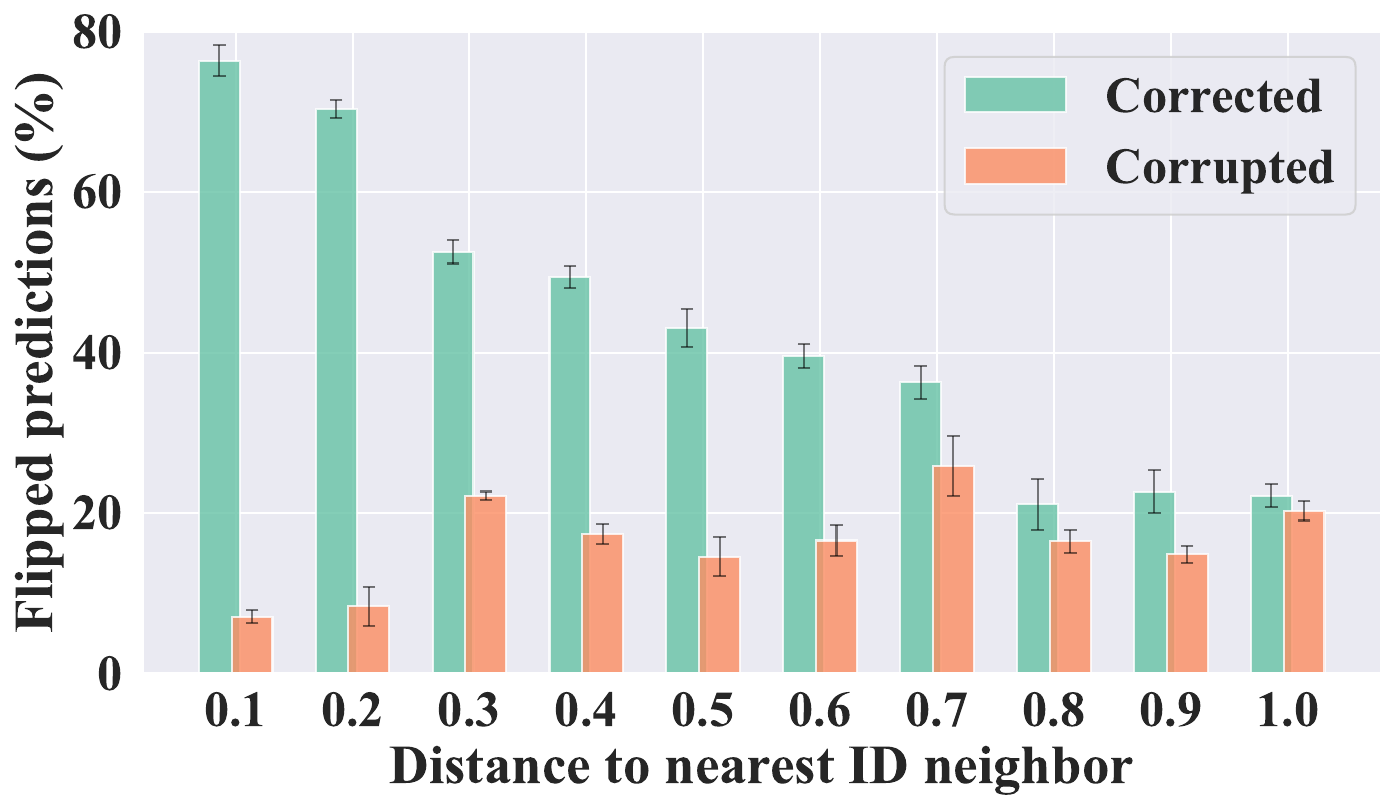}
        \caption{Rate of corrected ID examples}
        \label{fig:w2s-c}
    \end{subfigure}
    \hfill
    \begin{subfigure}{0.49\textwidth}
        \centering
        \includegraphics[width=\linewidth]{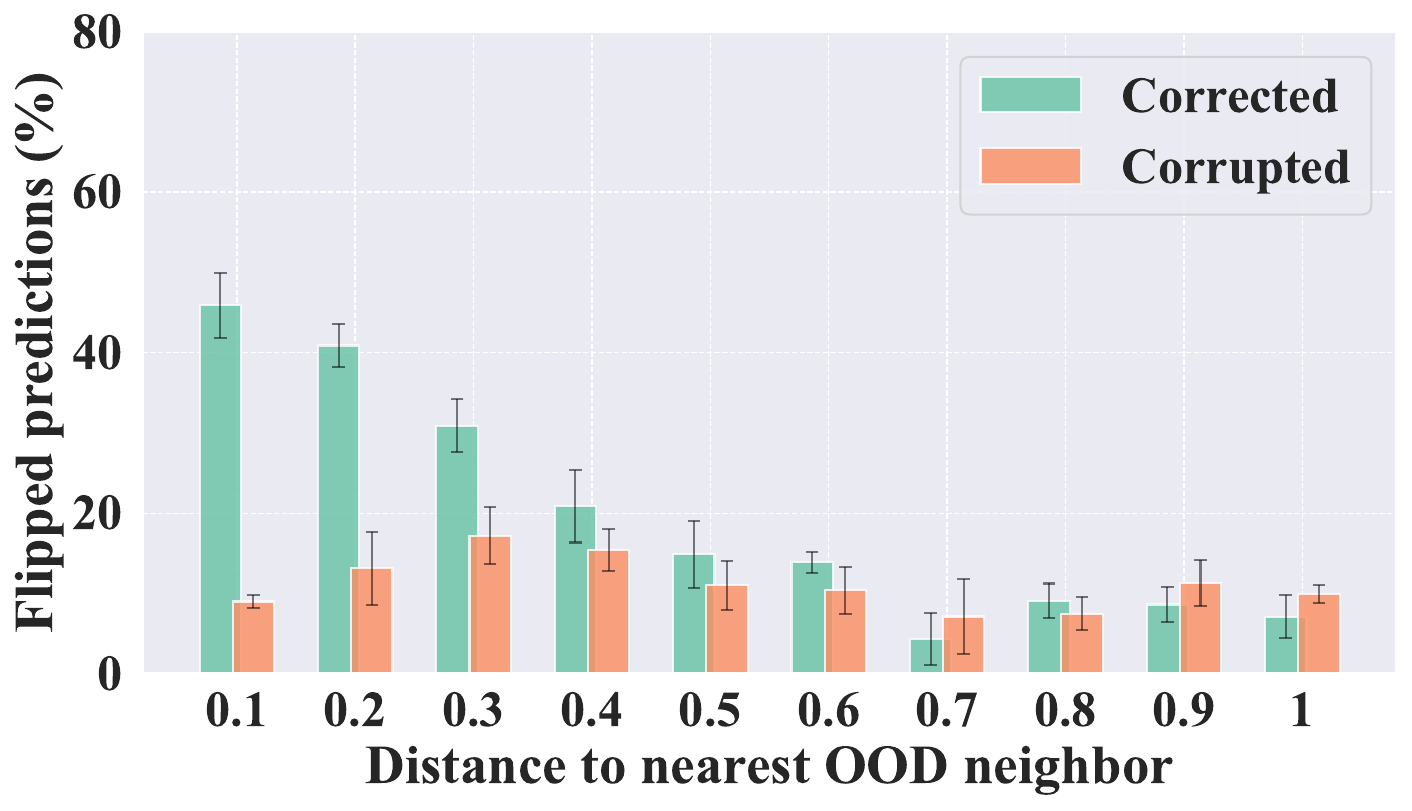}
        \caption{Rate of corrected OOD examples}
        \label{fig:w2s-d}
    \end{subfigure}
    \caption[Empirical analysis of W2S generalization]{
        Empirical analysis of \wilda-S on the aggregated GLUE datasets for Llama 3:
        (a) Histogram of approximated Lipschitz constants across datasets, computed as the Frobenius norm of the input-output Jacobian matrix;
        (b) Rate of pseudo-label correction over training epochs with examples from the unlabeled dataset used for training. Shaded areas indicate the standard deviation over $10$ runs;
        (c) and (d) Corrected and corrupted prediction rates for (c) ID examples and (d) OOD examples, based on the Euclidean distance to the closest correctly pseudo-labeled neighbor (normalized to $[0,1]$). There are $10$ bins ranging from the interval of $[0,0.1]$ to $[0.9,1]$. Error bars denote the standard deviation over $10$ runs.
    }
    \label{fig:w2s}
\end{figure*}

Building on the observation that \wilda consistently outperforms its teacher, standard ICL, we hypothesize that W2S generalization may be driving these improvements, where the model's ability to generalize strengthens progressively from weaker signals. To explore this further, we conduct an empirical analysis of \wilda-S with Llama 3 on aggregated examples from all GLUE datasets, treating them as a single, unified dataset.

\subsection{Local Consistency}

A crucial prerequisite for successful W2S generalization is the student's ability to maintain stable outputs under small perturbations of the input, i.e., robustness to input variations. A low Lipschitz constant serves as a key indicator of this stability, as it bounds the maximum change in the model output for any change in its input \citep{khromov-and-sidak-2024-some}. However, calculating the exact Lipschitz constant for LLMs is intractable. To approximate it, we leverage the relationship between the Lipschitz constant and the input-output Jacobian matrix of a neural network. Specifically, we compute the Frobenius norm of the Jacobian matrix as a tractable proxy, given its relationship to the spectral norm,
which is a known lower bound for the Lipschitz constant \citep{dherin-etal-2022-neural}. We use the embeddings as inputs and the penultimate layer to produce outputs. \Cref{fig:w2s-a} presents the distribution of the approximated Lipschitz constants (normalized to $[0,1]$) for \wilda, PBFT, and ICL, providing a proxy for local consistency. \wilda exhibits a notably lower Lipschitz constant than PBFT and ICL, underscoring its local consistency.

\subsection{Pseudo-Label Correction and Coverage Expansion}

W2S generalization in \wilda emerges from two interrelated phenomena: \textbf{pseudo-label correction}, where erroneous labels are progressively refined, and \textbf{coverage expansion}, where these corrections propagate across the representation space. This section examines how these mechanisms contribute to improved generalization, both within the training domain and on OOD data.

\Cref{fig:w2s-b} shows how the rate of corrected pseudo-labels evolves during training on GLUE datasets. As training progresses, the percentage of corrected pseudo-labels steadily increases, showcasing \wilda's capacity to exhibit W2S generalization. Notably, the rate of pseudo-label correction plateaus faster for simpler datasets like \sst and \qnli, which have lower linguistic variability.

We hypothesize that the core of \wilda's ability to generalize effectively is anchored in coverage expansion, which enables local corrections to propagate globally, creating a ripple effect across the representation space. To understand this dynamic, we analyze which unseen evaluation points are corrected by clustering them based on their proximity to the nearest correctly pseudo-labeled neighbor in $\mathcal{D}_\text{unlab}$. This is quantified by computing the Euclidean distance between the model's representations at the final hidden states, with evaluation points categorized into ten bins based on their normalized distance from the correct neighbor, spanning the range $[0,1]$.

\Cref{fig:w2s-c} illustrates the rate of prediction flips within these bins, where a flip refers to either correcting an incorrect prediction or corrupting a correct one. The rate of corrected predictions shows a strong negative correlation with the distance to the nearest correctly labeled neighbor, as indicated by a Pearson correlation coefficient of $-0.968$, while corrupted predictions are more frequent in regions lacking nearby correct pseudo-labels.

Coverage expansion shows its effects even on OOD data. \Cref{fig:w2s-d}, the counterpart to \Cref{fig:w2s-c}, shows the rate of flipped predictions for OOD data. Although the impact is reduced, a similar correction pattern persists, with a Pearson correlation of $-0.916$. This consistency across domains highlights the model’s ability to propagate accurate predictions not only within the training domain but also across OOD data.

\section{Faithful Encoding and Reconstruction of Demonstrations}

To evaluate whether demonstrations are faithfully encoded and disentangled, we conducted an experiment by encoding a single demonstration into the adapter and assessing the student model's ability to capture this information. Specifically, we utilized $1000$ examples per dataset across the GLUE benchmark using Llama 3 (8B).

For each dataset, the student model was prompted with a simple instruction: ``Repeat the demonstration word for word.'' During the training phase, the teacher model processed input examples using the following template: ``Demonstration: \{\textit{demonstration}\}. Answer: (\{\textit{answer}\}).'' The adapter learned to encode demonstration-specific information indirectly by aligning its outputs with the teacher's responses, without explicitly seeing the demonstration itself. After training, the similarity between the student model's response and the original demonstration was computed.
\Cref{tab:bert_score} shows the average BERTScore similarity \citep{zhang-etal-2020-bertscore} between the original demonstrations and the student's reconstructed response.

\begin{table}[]
\caption[\ \ Average BERTScore of reconstructed demonstrations]{Average BERTScore ($F_1$) similarity across GLUE datasets. Higher scores indicate better fidelity in recalling the encoded demonstration.}
\centering
\begin{tabular}{lccccccc}
\toprule
 & \textbf{\rte} & \textbf{\sst} & \textbf{\qnli} & \textbf{\mnli} & \textbf{\cola} & \textbf{\mrpc} & \textbf{\qqp} \\
\midrule
BERTScore & 0.84 & 0.91 & 0.80 & 0.83 & 0.86 & 0.82 & 0.81 \\
\bottomrule
\end{tabular}
\label{tab:bert_score}
\end{table}

The consistently high BERTScore values across all datasets indicate that the student model can reliably reconstruct the encoded demonstration from the adapter. This suggests that \wilda effectively disentangles and stores task-specific information within the adapter's weights. Notably, when compared to standard ICL, \wilda often produced different outputs for certain queries, particularly in instances where it corrected ``corrupted'' labels provided by the teacher. Despite these differences, the student model maintained a high degree of semantic similarity in reproducing the demonstrations. This suggests that the adapter weights capture not only the demonstration itself but also additional latent information that contributes to improved generalization.

We present below a pair of examples from \sst and \rte, chosen to represent reconstructed demonstrations with similarity scores close to the dataset averages.

\begin{example}[\textbf{\sst: Example 1}]
\textbf{Original:}  
\textit{Proves once again he hasn't lost his touch, delivering a superb performance in an admittedly middling film.} \\
\textbf{Answer:} (Positive) \\[0.5em]
\textbf{Reconstructed:}  
\textit{He demonstrates once more that he hasn't missed a beat, delivering a remarkable performance in what is admittedly an average film.} \\
\textbf{Answer:} (Positive)
\end{example}

\begin{example}[\textbf{\sst: Example 2}]
\textbf{Original:}  
\textit{Though many of the actors spark briefly when they first appear, they can't generate enough heat in this cold vacuum of a comedy to ignite a reaction.} \\
\textbf{Answer:} (Negative) \\[0.5em]
\textbf{Reconstructed:}  
\textit{Although some actors manage to show a hint of energy early on, they fail to create any real warmth or spark within this lifeless and chilly comedy.} \\
\textbf{Answer:} (Negative)
\end{example}

\begin{example}[\textbf{\rte: Example 1}]
\textbf{Original:} \\
Premise: \textit{The source added that the investigation proved that the bases of the genocide crime ``were completed with a series of illegal arrests followed in some cases with assassinations or cases of disappearances and were preceded, according to information attached to the file, by cases of torture.''} \\
Hypothesis: \textit{Investigators discovered that a series of illicit arrests were often followed by disappearances or murders and were preceded by torture.} \\
\textbf{Answer:} (True) \\[0.5em]
\textbf{Reconstructed:} \\
Premise: \textit{The investigation confirmed that genocide involved illegal arrests followed by disappearances or murders, often preceded by torture.} \\
Hypothesis: \textit{Investigators found that unlawful arrests frequently resulted in disappearances or murders, often preceded by acts of torture.} \\
\textbf{Answer:} (True)
\end{example}

\begin{example}[\textbf{\rte: Example 2}]
\textbf{Original:} \\
Premise: \textit{American tobacco companies were showing a profit most quarters due to export sales of cigarettes and diversification of products sold, including food.} \\
Hypothesis: \textit{PM often entered markets with both cigarettes and food.} \\
\textbf{Answer:} (False) \\[0.5em]
\textbf{Reconstructed:} \\
Premise: \textit{Profitability was often maintained by American tobacco companies through diversification into food products and successful cigarette exports.} \\
Hypothesis: \textit{Philip Morris International offered food items and cigarettes.} \\
\textbf{Answer:} (False)
\end{example}

\section{Limitations}

While \wilda offers significant improvements in stability and generalization over standard ICL, it is not without its limitations. These constraints primarily stem from computational requirements, data availability, and scalability concerns when handling large demonstration sets.

\paragraph{Computational cost.}  
\wilda introduces additional computational overhead due to the fine-tuning of adapters. While this fine-tuning is more lightweight compared to full model fine-tuning, it remains more expensive than standard ICL, which avoids weight updates entirely. However, \wilda offsets some of this cost by removing demonstrations from the input during inference. For instance, with Llama 3 (8B) processing $16$ demonstrations from GLUE datasets, inference takes approximately $120$ times longer than a $0$-shot setup (processing only the query). This increased cost scales quadratically with the number of tokens, highlighting the self-attention mechanism as the primary bottleneck when handling $16$ demonstrations. Based on our measurements, fine-tuning with $100$ unlabeled instances and $16$ demonstrations using a single adapter corresponds to a computational cost of approximately $2100$ inferences in a $16$-shot setup. This implies that after about $2100$ inferences, the time spent on fine-tuning is effectively balanced by the reduction in per-inference computational cost.

\paragraph{Applicability.}  
\wilda may be less suitable for scenarios with extremely limited resources, as it relies on access to a supply of unlabeled data. In our experiments with $\{4, 8, 16, 32\}$ demonstrations, we typically used $100$ unlabeled instances, which proved sufficient to achieve strong performance. While unlabeled data is generally easier to acquire than labeled data, there may be scenarios where obtaining even a modest amount of unlabeled data is challenging, potentially limiting the applicability of \wilda.

\paragraph{Large demonstration sets.}  
Although \wilda efficiently encodes demonstrations into adapters to overcome context length limitations, the method has not been extensively tested with very large demonstration sets. From our findings, as the total number of demonstrations increases, using multiple adapters with manageable demonstration sizes tends to be more effective. For instance, we successfully employed $8$ adapters with $16$ demonstrations each (totaling $128$ demonstrations). While this approach theoretically allows for an indefinite increase in the number of demonstrations, its effectiveness with significantly larger sets remains unexplored. Moreover, using additional adapters increases computational costs, introducing a tradeoff between scalability and efficiency.

\section{Summary}

We addressed the challenges of stability and long-context handling that arise when processing multiple demonstrations within ICL using LLMs.
To address these issues, we introduced \wilda, a method that disentangles the latent shifts induced by demonstrations from those of the query, leveraging a teacher-student framework. 
\wilda encodes these latent shifts into an adapter module, enabling the student model to handle queries without requiring demonstrations in the input. Moreover, \wilda allows efficient handling of large demonstration sets by chunking them into manageable subsets, each processed through separate adapter modules. This not only reduces the instability caused by demonstration selection and ordering but also alleviates the context window limitations inherent in transformer-based models. We demonstrated that \wilda exhibits W2S generalization by refining pseudo-labels through progressive corrections, expanding from local consistency to a more comprehensive coverage across the representation space.  Our empirical evaluation of \wilda showed that it consistently outperforms traditional ICL methods, significantly improving generalization and stability across diverse datasets. These findings underscore the effectiveness of weak supervision as a promising strategy for improving ICL performance.

These contributions are integral to the broader narrative of the thesis, which explores the intersection of data- and parameter-efficient strategies to tackle challenges in resource-constrained NLP scenarios. \wilda exemplifies the synergy between PEFT and ICL for efficient data utilization, demonstrating how lightweight, task-specific modules can address complex modeling challenges while preserving computational efficiency. The findings highlight the potential of combining weak supervision, modular design, and targeted parameter updates to create resource-effective solutions for NLP tasks, contributing to a more scalable and sustainable future for neural language models.

\addcontentsline{toc}{part}{}  
\setcounter{part}{0}  
\part*{}

\chapter{Discussion}
\label{ch:discussion}
The remarkable success of NLMs, and more recently LLMs, has come with significant trade-offs, including escalating computational demands, soaring parameter counts, and an increasing dependence on vast amounts of labeled data. The dominant trend of scaling models ever larger has undoubtedly driven performance gains, but it also raises pressing concerns about sustainability, accessibility, and efficiency. As models grow, they rely on finite reserves of high-quality data and require immense computational resources, making brute-force scaling an unsustainable long-term strategy. To address these challenges, research should shift toward optimizing resource utilization, enhancing model interpretability, and ensuring that advancements remain accessible to a broader research community.

Despite the contributions presented in this thesis, several challenges remain, both in terms of methodology and broader implications. The following sections outline key limitations of our work and highlight open questions for future research.

\section{Representation Regularization and \jachess{}}

The introduction of \jachess{} (\Cref{ch:jachess}) provides a novel approach to enhancing representation smoothness through Jacobian and Hessian regularization. By enforcing stability in intermediate representations, this method improves robustness, generalization, and calibration. However, its scope is currently limited to text classification tasks within the GLUE benchmark. The broader applicability of \jachess{} to more complex NLP tasks -- such as structured prediction (e.g., named entity recognition, dependency parsing) and generative modeling (e.g., summarization, machine translation) -- remains an open question. Future research should explore whether smoothness regularization yields similar benefits in tasks where output structures are more intricate and context dependencies are more pronounced.

Beyond task generalization, computational efficiency is another key consideration. While Hutchinson’s estimator reduces the overhead associated with computing Jacobian and Hessian norms, \jachess{} still introduces additional training costs. The dual-mode training strategy (\jachess{}$_\text{train}$ and \jachess{}$_\text{unlab}$) efficiently leverages unlabeled data, but the increase in runtime -- approximately $1.22\times$ with 1000 unlabeled examples -- may be impractical for large-scale applications. Developing more efficient variants, such as adaptive regularization schedules or approximations with lower computational costs, would enhance its viability in real-world scenarios.

A deeper theoretical understanding of how representation smoothness correlates with model generalization is also necessary. While empirical results suggest that regularizing intermediate representations improves performance, a more formal characterization of this relationship could guide future improvements. Investigating layer-wise contributions within transformer architectures could refine \jachess{} and inform more effective regularization strategies. As language models continue to scale, representation-focused research will be crucial for improving interpretability and robustness, ensuring that models remain efficient without sacrificing performance.

\section{Active Learning and Representation Smoothness}

The integration of representation smoothness into AL through \beast{} and \alsbi{} (\Cref{ch:beast}) provides an effective means of improving data efficiency. By leveraging smoothness-based early stopping and sample selection criteria, these methods reduce reliance on labeled validation sets and improve model stability in low-resource settings. However, their effectiveness has been primarily evaluated on classification tasks, raising questions about their applicability to more complex NLP problems. Structured prediction tasks and generative applications introduce unique challenges, such as label ambiguity and long-range dependencies, that may require tailored adaptations of smoothness-based AL strategies.

In addition to task complexity, the fundamental assumption that smoothness correlates with informativeness in AL selection warrants further examination. While empirical results support this assumption in classification tasks, further theoretical work is needed to establish a formal link between smoothness, model uncertainty, and AL efficiency. Additionally, the behavior of smoothness-based AL strategies in highly imbalanced datasets remains unclear. Since these methods prioritize selecting moderately forgettable instances, their effectiveness in cases where informative examples do not exhibit high smoothness needs further exploration. Adapting smoothness-based active learning techniques to the characteristics of specific tasks and datasets may be essential for broader applicability.

\section{Integration of AL and PEFT}

The combination of AL with PEFT in \Cref{ch:al-peft} demonstrates clear advantages in low-resource NLP tasks. By leveraging adapters and other PEFT techniques, this approach reduces computational demands while maintaining strong performance across AL cycles. However, its evaluation has been primarily limited to classification tasks, leaving open questions about its effectiveness in structured prediction and generative modeling. Given the stability benefits of PEFT, future research should investigate its potential for handling more complex task formulations.

One of the key advantages of PEFT in AL settings is its ability to maintain stability across training iterations. However, the selection strategies in AL remain dependent on model-specific uncertainty estimates, which can be unreliable in extremely low-resource settings. This issue is particularly pronounced in entropy-based AL methods, where PEFT models may produce different uncertainty estimates compared to full fine-tuning due to their constrained parameter updates. Developing PEFT-specific uncertainty quantification methods could improve selection strategies and enhance AL effectiveness.

Another important consideration is the role of TAPT in this framework. While TAPT enhances AL efficiency by aligning model representations with task-specific data, its reliance on domain-specific unlabeled data introduces potential constraints. The extent to which TAPT remains effective when in-domain data is scarce is an open question. Additionally, since TAPT is applied uniformly across PEFT methods, different PEFT techniques may require varying levels of adaptation. Future research should investigate whether task-adaptive pre-training strategies should be tailored to specific PEFT configurations.

\section{Weak Supervision and \wilda}

The introduction of \wilda (\Cref{ch:wilda}) provides a novel approach to addressing stability challenges in ICL. By leveraging weak supervision, \wilda internalizes in-context knowledge into adapter modules, improving generalization and efficiency. However, this method comes with trade-offs. Unlike standard ICL, which operates without weight updates, \wilda requires fine-tuning adapter parameters on pseudo-labeled data. While this significantly reduces inference costs by eliminating the need for repeated demonstration processing, the initial fine-tuning step introduces computational overhead that may not be feasible in time-sensitive applications. Future work should explore strategies for optimizing the efficiency of weak supervision, such as adaptive stopping criteria or lightweight fine-tuning regimes.

Another limitation is \wilda’s dependence on a sufficient pool of unlabeled data for generating pseudo-labels. While unlabeled data is generally more accessible than labeled data, certain domains may lack high-quality sources of unlabeled text. Investigating whether smaller unlabeled datasets or synthetic data augmentation techniques can mitigate this constraint remains an open area of research.

Finally, \wilda's ability to generalize across OOD settings depends on the distribution of pseudo-labels. While our experiments show strong generalization across related datasets, the method's robustness in scenarios with significant domain shifts remains to be explored.

\chapter{Conclusion}
\label{ch:conclusion}

Neural language models (NLMs) have transformed natural language processing (NLP), achieving remarkable success in a wide range of tasks, from text classification to machine translation and reasoning. Large language models (LLMs) further advanced the field by demonstrating strong generalization capabilities across diverse domains. However, this progress came with significant costs -- growing parameter sizes, increasing computational demands, and an ever-expanding reliance on large amounts of labeled and unlabeled data. While scaling proved effective in pushing performance boundaries, its reliance on brute-force methods raised concerns about long-term sustainability, accessibility, and efficiency. Addressing these challenges requires a shift toward optimizing resource utilization, improving training stability, and ensuring that state-of-the-art models remain practical and widely accessible.

This thesis tackled these issues by proposing methods that enhanced the efficiency of NLMs, focusing on data efficiency, computational scalability, and model adaptability. The contributions spanned representation learning, active learning (AL), parameter-efficient fine-tuning (PEFT), and weak supervision with in-context learning (ICL), each offering a step toward reducing reliance on large-scale, resource-intensive training while maintaining strong performance across NLP tasks.

A key focus of this research was the role of representation smoothness in improving model generalization and robustness. The introduction of \jachess{} highlighted how Jacobian and Hessian regularization could refine learned representations, making pre-trained language models (PLMs) more resilient to adversarial perturbations and distribution shifts. By enforcing smoothness constraints across intermediate layers, \jachess{} not only enhanced in-domain generalization but also improved cross-domain adaptability. Beyond generalization, this work demonstrated that smoothness played a crucial role in model calibration, helping to produce more reliable confidence estimates. These findings reinforced the importance of structured regularization techniques in building more stable and trustworthy PLMs.

Building on insights from representation smoothness, this thesis explored its integration into active learning. The development of \beast{}, a smoothness-driven early stopping method, and \alsbi{}, a principled stopping criterion for AL, demonstrated how leveraging smoothness could improve data efficiency while reducing the need for labeled validation sets. These methods enhanced training stability in low-resource scenarios, making AL workflows more practical for real-world applications. By ensuring that AL selection strategies prioritized informative yet stable instances, smoothness-based techniques significantly reduced labeling costs while maintaining high model performance. The success of these methods underscored the value of integrating representation analysis with data selection strategies to create more efficient learning pipelines.

Another major contribution of this work was the combination of AL with PEFT, offering a framework that optimized both data and parameter efficiency. While AL strategies traditionally rely on full fine-tuning (FFT), this thesis demonstrated that PEFT techniques not only reduced computational overhead but also improved selection stability when integrated with AL. This synergy was further enhanced by task-adaptive pre-training (TAPT), which helped bridge the performance gap between FFT and PEFT, ensuring that models retained essential pre-trained knowledge while adapting effectively to new tasks. The findings highlighted the advantages of lightweight, modular fine-tuning techniques, particularly in scenarios where computational resources were limited. By combining AL, PEFT, and TAPT, this work provided a comprehensive approach to improving the efficiency of NLP models in low-resource settings.

Beyond fine-tuning strategies, this thesis investigated ICL and its limitations. Standard ICL methods suffer from instability due to their sensitivity to prompt variations and demonstration selection. The introduction of \wilda addressed these challenges by disentangling latent shifts induced by demonstrations and queries, stabilizing the learning process while reducing computational overhead. By leveraging weak supervision, \wilda enables models to internalize in-context knowledge efficiently.

While this thesis made significant strides in improving the efficiency of neural language models, several open directions remain. Future research should further explore the role of representation smoothness in deep transformer architectures, particularly how it evolves across layers and influences different types of NLP tasks. Investigating additional representation properties beyond smoothness could further refine model stability and adaptability.

The synergy between PEFT and AL presented opportunities for refinement. Developing adaptive AL strategies that dynamically adjust based on the characteristics of PEFT models could improve task-specific data selection. Additionally, extending PEFT and AL integration to cross-task transfer learning would enable models to generalize more effectively across domains, further improving sample efficiency.

Weak supervision with ICL offered another promising research avenue. Future work could explore more fine-grained disentanglement of latent shifts, enabling ICL models to better separate task-specific information from prompt-induced biases. The use of adapters and other PEFT techniques in ICL settings could enhance scalability across multiple tasks and domains. Moreover, extending \wilda to handle more complex multi-step reasoning and long-horizon adaptation could further strengthen its applicability.

Addressing the limitations of NLP in low-resource and marginalized languages remains an urgent challenge. Customizing AL and PEFT pipelines for multilingual and low-resource settings could significantly expand language model coverage while mitigating biases. Leveraging structured representation techniques to identify and correct biases in multilingual models would contribute to the development of more equitable and inclusive AI technologies.

Finally, theoretical advancements would be crucial in complementing the empirical findings of this thesis. Formalizing the relationship between smoothness and model generalization could provide deeper insights into why certain methods succeeded. Developing theoretical models to explain the interaction between AL and PEFT would refine their joint application, while studying the convergence properties of AL-PEFT workflows could guide the design of more efficient and reliable training pipelines.

The future of NLMs depends on addressing the challenges posed by scaling, accessibility, and representation. By shifting focus from raw size to efficiency and transparency, the field could ensure that these technologies serve a broader range of societal needs. Achieving this vision would require coordinated efforts across academia, industry, and the open-source community, as well as a commitment to fostering equitable access to the tools and knowledge needed to advance the state of the art.

\backmatter

\bibliographystyle{IEEEtranFER}
\addcontentsline{toc}{chapter}{Bibliography}
\bibliography{bibliography, anthology}



\chapter*{Glossary of Acronyms}
\addcontentsline{toc}{chapter}{Glossary of Acronyms}

\begin{table}[h]
    \centering
    \renewcommand{\arraystretch}{1.2} 
    \begin{tabular}{ll}
        \hline
        \textbf{Abbreviation} & \textbf{Definition} \\
        \hline
        AL    & Active Learning \\
        AUC   & Area Under the Curve \\
        CBOW  & Continuous Bag of Words \\
        CKA   & Centered Kernel Alignment \\
        CLM   & Causal Language Modeling \\
        DAL   & Discriminative Active Learning \\
        DSM   & Distributional Semantic Model \\
        ECE   & Expected Calibration Error \\
        FFT   & Full Fine-Tuning \\
        GRU   & Gated Recurrent Unit \\
        ICL   & In-Context Learning \\
        ID    & In-Distribution \\
        LCR   & Label Complexity Reduction \\
        LLM   & Large Language Model \\
        LSA   & Latent Semantic Analysis \\
        LSTM  & Long Short-Term Memory \\
        MLM   & Masked Language Modeling \\
        NLP   & Natural Language Processing \\
        NLM   & Neural Language Model \\
        OOD   & Out-of-Distribution \\
        PEFT  & Parameter-Efficient Fine-Tuning \\
        PLM   & Pre-trained Language Model \\
        RIPL  & Relative Improvement over Passive Learning \\
        RLHF  & Reinforcement Learning from Human Feedback \\
        RNN   & Recurrent Neural Network \\
        SVD   & Singular Value Decomposition \\
        TAPT  & Task-Adaptive Pre-training \\
        W2S   & Weak-to-Strong \\
        \hline
    \end{tabular}
    \label{tab:abbreviations}
\end{table}

\addcontentsline{toc}{chapter}{List of Figures}
\listoffigures
\cleardoublepage 
\addcontentsline{toc}{chapter}{List of Tables}
\listoftables
\cleardoublepage 

\renewcommand{\leftmark}{Biography}
\chapter*{Biography}
\addcontentsline{toc}{chapter}{Biography}

Josip Jukić received his Bachelor of Computer Science in Computing from the University of Zagreb, Faculty of Electrical Engineering and Computing, in 2018, followed by a Master of Computer Science in Computing from the same institution in 2020. He has received multiple ``Josip Lončar'' Awards in recognition of his academic excellence throughout his studies.

From 2019 to 2021, he was a research associate at the Faculty of Electrical Engineering and Computing, University of Zagreb, focusing on natural language processing. Since 2021, he has been a teaching and research assistant at the same faculty. His research interests encompass natural language processing and representation learning, with a particular focus on large language models.

\subsection*{Journal publications}
\begin{itemize}
    \item Josip Jukić and Jan Šnajder. From Robustness to Improved Generalization and Calibration in Pre-trained Language Models.  
    \textit{Transactions of the Association for Computational Linguistics, 2025}. 
\end{itemize}

\subsection*{Conference publications}
\begin{itemize}
    \item Fran Jelenić, Josip Jukić, Martin Tutek, Mate Puljiz, and Jan Šnajder. Out-of-Distribution Detection by Leveraging Between-Layer Transformation Smoothness.  
    \textit{ICLR 2024, Vienna}.  

    \item Josip Jukić and Jan Šnajder. Parameter-Efficient Language Model Tuning with Active Learning in Low-Resource Settings.  
    \textit{EMNLP 2023, Singapore}.  
    
    \item Josip Jukić and Jan Šnajder. Smooth Sailing: Improving Active Learning for Pre-trained Language Models with Representation Smoothness Analysis.  
    \textit{CLASP: Learning with Small Data 2023, Gothenburg}.  
    
    \item Josip Jukić, Martin Tutek, and Jan Šnajder. Easy to Decide, Hard to Agree: Reducing Disagreements Between Saliency Methods.  
    \textit{Findings of ACL 2023, Toronto}.  
    
    \item Fran Jelenić, Josip Jukić, Nina Drobac, and Jan Šnajder. On Dataset Transferability in Active Learning for Transformers.  
    \textit{Findings of ACL 2023, Toronto}.  

    \item Josip Jukić, Fran Jelenić, Miroslav Bićanić, and Jan Šnajder. ALANNO: An Active Learning Annotation System for Mortals.  
    \textit{EACL 2023, Dubrovnik}.  
    
    \item Josip Jukić, Iva Vukojević, and Jan Šnajder. You Are What You Talk About: Inducing Evaluative Topics for Personality Analysis.  
    \textit{Findings of EMNLP 2022, Abu Dhabi}.  
\end{itemize}

\renewcommand{\leftmark}{Životopis}
\chapter*{Životopis}
\addcontentsline{toc}{chapter}{Životopis}

Josip Jukić stekao je diplomu prvostupnika računarstva na Sveučilištu u Zagrebu, Fakultetu elektrotehnike i računarstva, 2018. godine, a potom i diplomu magistra računarstva na istoj instituciji 2020. godine. Za svoj iznimni akademski uspjeh tijekom studija višestruko je nagrađen nagradom ``Josip Lončar''.

Od 2019. do 2021. godine radio je kao istraživački suradnik na Fakultetu elektrotehnike i računarstva Sveučilišta u Zagrebu, s fokusom na obradu prirodnog jezika. Od 2021. godine radi kao asistent u nastavi i istraživanju na istom fakultetu. Njegovi istraživački interesi obuhvaćaju obradu prirodnog jezika i učenje reprezentacija, s posebnim naglaskom na velike jezične modele.

\end{document}